\documentclass{article}
\usepackage[utf8]{inputenc}
\usepackage{fullpage}

\usepackage{amsthm,mathtools,scalerel}
\usepackage{color,soul,latexsym,amsmath,amssymb,amsfonts,amsthm,dsfont,enumitem,xcolor,bbm,threeparttable,graphicx,float,subcaption}
\usepackage{authblk}
\usepackage[round]{natbib}
\RequirePackage[colorlinks=true,allcolors=blue,hypertexnames=false]{hyperref}%
\usepackage{graphicx,todonotes}
\usepackage{xcolor}
\usepackage{algorithmic} %For Gradient Truncation algorithm

\definecolor{Green}{HTML}{069937}
\newif \ifnotes

\usepackage[most]{tcolorbox}
\colorlet{shadecolor}{gray!15}
\usepackage{outlines}
\usepackage{array}
\usepackage{selectp}
\usepackage[ruled,linesnumbered, vlined, noend]{algorithm2e}
\mathtoolsset{showonlyrefs}
\usepackage{titlesec}

\newtheorem{thm}{Theorem}%[section]

\newtheorem{lemma}{Lemma}
\newtheorem{cor}{Corollary}

\newtheorem*{remark}{Remark}

\DeclareMathOperator*{\argmin}{argmin}

\title{\bf Statistical Inference for Online Algorithms}
\author[1]{Selina Carter}

\author[2]{Arun Kumar Kuchibhotla}
\affil[1]{\texttt{shcarter@andrew.cmu.edu}}
\affil[2]{{\tt arunku@cmu.edu}}
\affil[1,2]{Department of Statistics \& Data Science, Carnegie Mellon University}

\date{}                   
\begin{document}

\maketitle

\begin{abstract}
The construction of confidence intervals and hypothesis tests for functionals is a cornerstone of statistical inference. Traditionally, the most efficient procedures—such as the Wald interval or the Likelihood Ratio Test—require both a point estimator and a consistent estimate of its asymptotic variance. However, when estimators are derived from online or sequential algorithms, computational constraints often preclude multiple passes over the data, complicating variance estimation. In this article, we propose a computationally efficient, rate-optimal wrapper method (HulC) that wraps around any online algorithm to produce asymptotically valid confidence regions bypassing the need for explicit asymptotic variance estimation. The method is provably valid for any online algorithm that yields an asymptotically normal estimator. We evaluate the practical performance of the proposed method primarily using Stochastic Gradient Descent (SGD) with Polyak-Ruppert averaging. Furthermore, we provide extensive numerical simulations comparing the performance of our approach (HulC) when used with other online algorithms, including implicit-SGD and ROOT-SGD.\footnote{We have approached this manuscript from the point of view of a statistician or a data analyst. Although SGD is the most commonly mentioned method in machine learning, our implementation and simulations show that the practical performance of SGD is highly sensitive to the choice of tuning parameters of the algorithm. We could not find a simple remedy that improves performance and also makes the asymptotic properties manageable. This was at least surprising to the authors. It is unclear if this is a well-known observation in the literature on online algorithms, and we hope that our article acts as a word of caution to anyone using online algorithms blindly.} 
\end{abstract}
%Old abstract (1/30/2026): The construction of confidence intervals and hypothesis tests for functionals of asymptotically normal estimators is a cornerstone of statistical inference. Traditionally, the most efficient procedures—such as the Wald interval or the Likelihood Ratio Test—require both a point estimator and a consistent estimate of its asymptotic variance. However, when estimators are derived from online or sequential algorithms, computational constraints often preclude multiple passes over the data, complicating variance estimation. In this article, we propose a computationally efficient, rate-optimal wrapper method (HulC) that wraps around any online algorithm to produce asymptotically valid confidence regions bypassing the need for explicit asymptotic variance estimation. The method is provable valid for any online algorithm that yields an asymptotically normal estimator. We primarily evaluate the practical performance of the proposed method using Stochastic Gradient Descent (SGD) with Polyak-Ruppert averaging. Furthermore, we provide extensive numerical simulations comparing the performance of our approach (HulC) when used with other online algorithms, including implicit-SGD, ROOT-SGD, and other robustified SGD methods.

\maketitle
% \iffalse
\section{Introduction and Motivation}
Suppose we have data $Z_1, \ldots, Z_T$ that are generated independently from a common distribution $P$. The analyst is interested in a summary functional $\theta_{\infty}(P)\in\mathbb{R}^d$ defined by the optimization problem:
\begin{equation}\label{eq:optimization-problem}
    \theta_{\infty} \equiv \theta_{\infty}(P) ~:=~ \argmin_{\theta\in\mathbb{R}^d}\,\mathbb{E}_{P}[\ell(Z; \theta)],
\end{equation}
for some loss function $\ell(\cdot; \cdot)$. Throughout the manuscript, we (implicitly) assume that $\theta_{\infty}$ is uniquely defined. Based on the data, a natural estimator (referred to as the $M$-estimator) of $\theta_{\infty}$ is
\begin{equation}\label{eq:empirical-M-estimator}
    \widehat{\theta}_T := \argmin_{\theta\in\mathbb{R}^d}\,\frac{1}{T}\sum_{i=1}^T \ell(Z_i; \theta).
\end{equation}
This setting includes several commonly used methods such as least squares linear regression, maximum likelihood estimation, generalized linear models, and, in general, any empirical risk minimization. Under certain regularity conditions on the loss function $\ell(\cdot; \cdot)$ including differentiability and invertibility of certain matrix expectations, it is well-known~\citep{Huber1967} that
\begin{equation}\label{eq:asym-normality-M-estimator}
T^{1/2}(\widehat{\theta}_T - \theta_{\infty}) ~\overset{d}{\to}~ N(0, J^{-1}VJ^{-1}),
\end{equation}
where 
\begin{equation}\label{eq:sandwich-variance-components}
    J := \nabla^2\mathbb{E}[\ell(Z; \theta_{\infty})]\quad\mbox{and}\quad V := \mbox{Var}(\nabla\ell(Z_i; \theta_{\infty})).
\end{equation}
Here, $\nabla\ell(Z; \theta_{\infty})$ refers to the derivative of $\theta\mapsto\ell(Z; \theta)$ evaluated at $\theta = \theta_{\infty}$ and $\nabla^2\mathbb{E}[\ell(Z; \theta_{\infty})]$ refers to the second derivative of $\theta\mapsto \mathbb{E}[\ell(Z; \theta)]$ evaluated at $\theta = \theta_{\infty}$. Throughout the manuscript, we implicitly assume that $J$ is a symmetric matrix. 

Assuming~\eqref{eq:asym-normality-M-estimator}, construction of confidence intervals for $e_k^{\top}\theta_{\infty}$, the $k$-th coordinate of $\theta_{\infty}$, is straightforward:
\begin{equation}\label{eq:Wald-interval}
\widehat{\mathrm{CI}}_{T,\alpha}^{(k)} := \left[e_k^{\top}\widehat{\theta}_T \pm \frac{z_{\alpha/2}}{T^{1/2}}(e_k^{\top}\widehat{J}_T^{-1}\widehat{V}_T\widehat{J}_T^{-1}e_k)^{1/2}\right],    
\end{equation}
where
\begin{equation}\label{eq:sandwich-variance-components-estimators}
    \widehat{J}_T := \frac{1}{T}\sum_{i=1}^T \nabla^2\ell(Z_i; \widehat{\theta}_T)\quad\mbox{and}\quad \widehat{V}_T := \frac{1}{T}\sum_{i=1}^T (\nabla \ell(Z_i; \widehat{\theta}_T))(\nabla \ell(Z_i; \widehat{\theta}_T))^{\top}.
\end{equation}
This is the classical Wald interval, and when stitched over all coordinates, this yields a hyper-rectangle confidence region for $\theta_{\infty}$. Although not the best (i.e., the smallest) in the sense of volume, a rectangular confidence region is highly interpretable. 

Computation of the Wald confidence interval~\eqref{eq:Wald-interval} requires unlimited access to the data $Z_1, \ldots, Z_T$, because the exact computation of~\eqref{eq:empirical-M-estimator} requires computing different quantities on the entire data a large (potentially infinite) number of times, depending on the structure of the loss function. For example, if $\ell(\cdot; \cdot)$ is quadratic in the second argument, such as in linear regression, then one pass over the data is enough to compute $\widehat{\theta}_T$ exactly, but if $\ell(\cdot; \cdot)$ is a general convex or non-convex function in the second argument, then an infinite number of passes over the data are necessary. (Non-convex loss functions are substantially more difficult than convex functions, as can be expected.) Even after the calculation of $\widehat{\theta}_T$, one needs at least two passes over the data to calculate $\widehat{J}_T$ and $\widehat{V}_T$ for the estimation of variance. In addition to computational inefficiency, the construction of~\eqref{eq:Wald-interval} is also memory inefficient because the entire data is essentially stored for computation. These problems have long been recognized both in the statistics and machine learning literature and have led to the development of online algorithms for estimating $\theta_{\infty}$ starting from the Robbins-Monro algorithm. 

Since the ground-breaking work~\cite{Robbins-Monro-1951}, several works have proposed modified/refined stochastic approximation algorithms with better finite-sample or asymptotic performance. Stochastic gradient descent with averaging by~\cite{ruppert_1988} and \cite{polyak_1992} is one of the first such refinements. For more recent developments, see~\cite{toulis2017asymptotic},~\cite{Duchi-Ruan-2020-AoS-Riemannian-SGD},~\cite{Li-Mou-Wainwright-Jordon-2020-ROOT-SGD}, and~\cite{Chen-Lai-Li-Zhang-2021-gradient-free-SGD}. All these works propose different online estimators $\widetilde{\theta}_T$ such that 
\[
r_T(\widetilde{\theta}_T - \theta_{\infty}) \overset{d}{\to} N(0, \Gamma),
\]
for some rate of convergence $r_T$ and some covariance matrix $\Gamma$. Under the assumptions needed for~\eqref{eq:asym-normality-M-estimator}, $r_T = T^{1/2}$ and $\Gamma = J^{-1}VJ^{-1}$. In more general settings (e.g., in constrained problems), these quantities can be different.

Although efficient online estimation of $\theta_{\infty}$ has received significant attention in the last 75 years, efficient inference for $\theta_{\infty}$ has garnered research interest relatively recently. Early contributions include~\cite{pelletier2000asymptotic} and~\cite{gahbiche2000estimation} both of which seemingly went unnoticed by the more recent work~\cite{Chen_SGD_2021}. This was followed by a flurry of research on online estimation of the asymptotic variance and its application to the Wald confidence interval using a plug-in approach. A non-exhaustive list of such works includes~\cite{Godichon2019},~\cite{Chen_SGD_2021},~\cite{zhudong_2021},~\cite{lu_song_2021}, and~\cite{li_FL_2022}. In addition, there are some alternatives to this online Wald inference procedure, including bootstrap-based methods, the functional central limit theorem (CLT), or the $t$-statistic; see~\cite{boot_SGD_2018},~\cite{zhong2023online},~\cite{lam2023resampling},~\cite{Lee_random_scaling_2022}, and~\cite{zhu_paralelSGD_2024}. Except for the method of~\cite{zhu_paralelSGD_2024} (based on~\cite{ibragimov_tstat_2010}), all other existing methods require additional computations or memory compared to the original online algorithm. For example, the variance estimators of~\cite{gahbiche2000estimation} and~\cite{Chen_SGD_2021} require storing the intermediate iterations of the SGD, and the bootstrap method of~\cite{boot_SGD_2018} requires running a large number of SGDs parallel to the original SGD. In addition, all existing methods (except~\cite{zhu_paralelSGD_2024}) require some additional structure on the online algorithm in addition to asymptotic normality. As we shall see, even relying on asymptotic normality is restrictive from the point of view of finite-sample performance. In the following section, we propose the application of HulC~\citep{hulc_2023_arun} for computationally efficient, rate-optimal, and asymptotically valid confidence regions for $\theta_{\infty}$.

The remainder of the article is organized as follows. In Section~\ref{sec:HulC-online-alg}, we introduce HulC and its application to online algorithms. In Section~\ref{sec:review-of-asym-SGD}, we review the asymptotics of averaged-SGD and some of the existing inference methods that we use for comparison. In Section~\ref{sec:simulation}, we present simulation results that compare the performance of HulC with three other existing methods in the context of well-specified linear and logistic regression models using averaged-SGD. In Section~\ref{sec:add-methods}, we present additional simulation results comparing HulC with baseline methods using well known other SGD algorithms such as truncated-SGD. Finally, we conclude the article in Section~\ref{sec:conc-remarks} with some remarks and future directions.

\section{HulC and Application to Online Algorithms}\label{sec:HulC-online-alg}
Suppose that we have access to independent and identically distributed random variables $\{Z_t: t \ge 1\}$ sequentially. The goal is to estimate $\theta_{\infty} \equiv \theta_{\infty}(P)$ defined in~\eqref{eq:optimization-problem}. Consider any online algorithm $\mathcal{A}$ starting from $\theta^{(0)}$ of the form
\begin{equation}\label{eq:online-algorithms}
\theta^{(t)} = \mathcal{O}_t(Z_t; \{\theta^{(j)}: 1\le j\le t-1\})\quad\mbox{for}\quad t\ge1.
\end{equation}
In words, the algorithm operates on the new data point $Z_t$ and all (or some of) the past iterates of the algorithm to obtain the current iterate. The way in which the operation $\mathcal{O}_t$ uses $Z_t$ determines whether the algorithm is first-order or second-order, and so on. For example, if $\mathcal{O}_t$ uses only the first derivative of $\ell(Z; \theta)$, then the algorithm is first-order. Similarly, how many past iterates are used in the operation determines the memory efficiency and per-iteration complexity of the algorithm. Let the output of the algorithm at time $T$ be 
\[
\mathcal{A}(Z_T, Z_{T-1}, \ldots, Z_1; \theta^{(0)}) =: \bar{\theta}_T.
\]
obtained from all (or some of) the iterates up to time $T$. Many of the commonly used online algorithms fit into this framework. For a brief overview, we present the following examples:
\begin{enumerate}
    \item Stochastic Gradient Descent (SGD): 
    \[
    \theta^{(t)} = \theta^{(t-1)} - \eta_t\nabla\ell(Z_t; \theta^{(t-1)})
    \]
    and
    \[
    \bar{\theta}_T = \begin{cases}
        \theta^{(T)}, &\mbox{last iterate~\citep{Robbins-Monro-1951}},\\
        T^{-1}\sum_{t=1}^T \theta^{(t)}, &\mbox{average iterate~\citep{ruppert_1988,polyak_1992}},\\
        \theta^{(R)}, &\mbox{randomly stopped~\citep{ghadimi2013stochastic},}\\
        \sum_{t=1}^T w_t\theta^{(t)}, &\mbox{weighted iterate~\citep{rakhlin2012making,wei2023weighted}.}
    \end{cases} 
    \]
    Note that the average iterate output can be iteratively calculated without storing all iterates using $\bar{\theta}_T = (1 - 1/T)\bar{\theta}_{T-1} + \theta^{(T)}/T$. For the randomly stopped iterate, $R$ is a random variable supported on $\{1, 2, \ldots, T\}$.
    \item Implicit-SGD~\citep{toulis2017asymptotic}:
    \[
    \theta^{(t)} = \theta^{(t-1)} - \eta_t\nabla\ell(Z_t; \theta^{(t)})\quad\mbox{and}\quad \bar{\theta}_T = \begin{cases}
        \theta^{(T)}, &\mbox{last-iterate or}\\
        T^{-1}\sum_{t=1}^T \theta^{(t)}, &\mbox{average-iterate.}
    \end{cases}
    \]
    Here $\theta^{(t)}$ is implicitly defined.
    \item Stochastic Heavy-Ball method~\citep{yan2018unified}:
    \[
    \theta^{(t)} = \theta^{(t-1)} - \eta_t\nabla \ell(Z_t; \theta^{(t-1)}) + \beta_t(\theta^{(t-1)} - \theta^{(t-2)})\quad\mbox{and}\quad \bar{\theta}_T = \begin{cases}
        \theta^{(T)}, &\mbox{last-iterate or}\\
        T^{-1}\sum_{t=1}^T \theta^{(t)}, &\mbox{average-iterate.}
    \end{cases}
    \]
    \item Stochastic Nesterov Accelerated Gradient~\citep{yan2018unified}:
    \begin{align*}
    \delta^{(t)} &= \theta^{(t-1)} - \eta_t\nabla \ell(Z_t; \theta^{(t-1)}),\\
    \theta^{(t)} &= \delta^{(t)} + \beta_t(\delta^{(t)} - \delta^{(t-1)})\quad\mbox{and}\quad \bar{\theta}_T = \begin{cases}
        \theta^{(T)}, &\mbox{last-iterate or}\\
        T^{-1}\sum_{t=1}^T \theta^{(t)}, &\mbox{average-iterate.}
    \end{cases}
    \end{align*}
    Here $\delta^{(0)} = \theta^{(0)}$. Other acceleration methods can be found in~\cite{shen2024unified}.
\end{enumerate}
We ignore the stochastic quasi-Newton methods that use the second-order information on the loss function from the list above; for details, see~\cite{guo2023overview} and~\cite{godichon2025adaptive}. In the examples above, $\eta_t, \beta_t$ define step sizes that usually decay to zero as $t\to\infty$. Traditionally, they are chosen to decay polynomially. Because the constants and exponents in such a choice are hard to choose practically, several researchers have proposed data-driven choices. These include, among many others, AdaGrad and ADAM~\citep{duchi2011adaptive,kingma2014adam,zhang2022adam,orvieto2024adaptive}. 

\subsection{Inference with HulC}
Suppose we have an algorithm that takes a stream of data and returns an estimator of $\theta_{\infty}\in\mathbb{R}^d$, and we want to construct a confidence interval for each of the coordinates of $\theta_{\infty}$. The HulC procedure~\citep{hulc_2023_arun} to construct a $(1-\alpha)$-confidence interval works as follows:
\begin{enumerate}[leftmargin=1.4cm]
    \item[\bf Step 1:] For $\alpha\in(0, 1]$, set $B_{\alpha} = \lceil\log_2(2/\alpha)\rceil$. With a standard uniform random variable $U$, define
    \[
    B_{\alpha}^* = \begin{cases}B_{\alpha}, &\mbox{if }U > 2^{B_{\alpha}}(\alpha/2) - 1,\\
    \lfloor\log_2(2/\alpha)\rfloor, &\mbox{otherwise.}
    \end{cases}
    \]
    The choice of $B_{\alpha}$ is so that $\mathbb{E}[2^{1-B_{\alpha}^*}] = 1 - \alpha.$ If $\log_2(2/\alpha)$ is an integer, then $B_{\alpha}^* = \log_2(2/\alpha).$
    \item[\bf Step 2:] Compute
    \[
    \bar{\theta}_T^{(j)} = \mathcal{A}(Z_{j + B_{\alpha}^*\lfloor(T - j)/B_{\alpha}^*\rfloor}, \ldots, Z_j; \theta^{(0,j)}),\quad\mbox{for}\quad 1\le j\le B_{\alpha}^*.
    \]
    This means that $\bar{\theta}_T^{(j)}$ is computed based on the data $Z_j, Z_{B_{\alpha} + j}, Z_{2B_{\alpha} + j}, \ldots$. (Note that this is streaming the data $Z_1, \ldots, Z_T$ into $B_{\alpha}^*$ buckets without the knowledge of the time horizon $T$.) Note that we allow each estimator to be constructed starting from a different initial value $\theta^{(0,j)}$.
    \item[\bf Step 3:] For $1\le k\le d$, construct the confidence interval
    \begin{equation}\label{eq:HulC-CI}
    \widehat{\mathrm{CI}}_{T,\alpha}^{(k)} := \left[\min_{1\le j\le B_{\alpha}^*}\,e_k^{\top}\bar{\theta}_T^{(j)},\, \max_{1\le j\le B_{\alpha}^*}\,e_k^{\top}\bar{\theta}_T^{(j)}\right].
    \end{equation}
\end{enumerate}
We refer to the confidence interval~\eqref{eq:HulC-CI} as the HulC CI. Although we discussed online algorithms of the structure~\eqref{eq:online-algorithms}, the construction of HulC CI does not need such restriction. \cite{hulc_2023_arun} provide a finite-sample validity guarantee for this confidence interval, which we restate in the following result.
\begin{thm}\label{thm:coverage-of-hulc}
    Suppose $Z_1, Z_2, \ldots, Z_T$ are independent random variables. Then for $1\le k\le d,$
    \[
    \mathbb{P}\left(e_k^{\top}\theta_{\infty} \notin \widehat{\mathrm{CI}}_{T,\alpha}^{(k)}\right) \le \alpha\left(1 + 2(B_{\alpha}\Delta_T)^2e^{2B_{\alpha}\Delta_T}\right),
    \]
    where
    \[
    \Delta_T := \max_{1\le j\le B_{\alpha}}\, \left(\frac{1}{2} - \min_{\gamma\in\{\pm 1\}}\;\mathbb{P}(\gamma(e_k^{\top}\bar{\theta}_T^{(j)} - e_k^{\top}\theta_{\infty}) \le 0)\right)_+,
    \]
    represents the maximum median bias of the estimators.
\end{thm}
\begin{proof}
    The result follows from Theorem 2 of~\cite{hulc_2023_arun}.
\end{proof}
Although the maximum median bias seems incomprehensible, it is very interpretable. To elaborate, we note that $\Delta_T = 0$ means that the probabilities of overestimation and underestimation of $e_k^{\top}\theta_{\infty}$ are at least half. In the case where $\mathbb{P}(\bar{\theta}_T^{(j)} = \theta_{\infty}) = 0$, this simplifies the requirement of $\Delta_T = 0$ to mean that overestimation and underestimation are equally likely. In general, $\Delta_T$ measures how different the probabilities of overestimation and underestimation are from half. Some simple bounds on $\Delta_T$ can be informative:
\begin{align*}
    \Delta_T ~&\le~ \sup_{m\in\{T/B_{\alpha}, T/(B_{\alpha} - 1)\}}\,\inf_{F\in\mathcal{F}_{\mathrm{sym}}}\sup_{t\in\mathbb{R}}\left|\mathbb{P}(e_k^{\top}\bar{\theta}_m - e_k^{\top}\theta_{\infty} \le t) - F(t)\right| ~=:~ \Delta_{T}^{\mathrm{Sym}},\\
    ~&\le~ \sup_{m\in\{T/B_{\alpha}, T/(B_{\alpha} - 1)\}}\,\inf_{\sigma \ge 0}\sup_{t\in\mathbb{R}}\left|\mathbb{P}(e_k^{\top}\bar{\theta}_m - e_k^{\top}\theta_{\infty} \le t) - \Phi(t/\sigma)\right| ~=:~ \Delta_T^{\mathrm{CLT}},
\end{align*}
where $\mathcal{F}_{\mathrm{sym}}$ is the collection of all distribution functions of real-valued random variables symmetric around zero (i.e., all non-decreasing functions $F(\cdot)$ satisfying $F(-\infty)=0$ and $F(t) + F(-t) = 1$ for all $t\in\mathbb{R}$), and $\Phi(\cdot)$ denotes the distribution function of the standard normal random variables. The ``distances'' $\Delta_T^{\mathrm{Sym}}$ and $\Delta_T^{\mathrm{CLT}}$ measure the distance of the distribution of $e_k^{\top}(\bar{\theta}_m - \theta_{\infty})$ to the class of all symmetric and (mean-zero) normal distributions, respectively. In particular, if $r_me_k^{\top}(\bar{\theta}_m - \theta_{\infty})$ converges in distribution to a mean zero normal random variables (for some rate of convergence $r_m$), then $\Delta_T^{\mathrm{CLT}}$ converges to zero, and consequently, $\Delta_T \to 0$, implying the asymptotic validity of the HulC CI. Several works~\citep{anastasiou2019normal,shao2022berry,samsonov2024gaussian,sheshukova2025gaussian} have proved bounds for $\Delta_T^{\mathrm{CLT}}$ which implies that $\Delta_T$ converges to zero as $T\to\infty$. A brief review of the proof of asymptotic normality is provided in Section~\ref{sec:review-of-asym-SGD} for the average iterate SGD. 

This discussion yields the following corollary of Theorem~\ref{thm:coverage-of-hulc}.
\begin{cor}
    Suppose the online algorithm based on $m$ independent observations outputs a vector $\bar{\theta}_m\in\mathbb{R}^d$ that satisfies
    \begin{equation}\label{eq:asymp-normality}
    r_m(\bar{\theta}_m - \theta_{\infty}) \overset{d}{\to} N(0, \Gamma),\quad\mbox{as}\quad m\to\infty,
    \end{equation}
    for some rate of convergence $r_m$ and some variance-covariance matrix $\Gamma$, then the HulC confidence interval~\eqref{eq:HulC-CI} satisfies asymptotic validity, i.e.,
    \[
    \mathbb{P}\left(e_k^{\top}\theta_{\infty} \notin \widehat{\mathrm{CI}}_{T,\alpha}^{(k)}\right) \to \alpha,\quad\mbox{as}\quad T\to\infty.    
    \]
\end{cor}
It should be stressed here that the assumption~\eqref{eq:asymp-normality} is only a narrow case of the asymptotic validity of the HulC confidence interval. HulC CI remain valid even if the limiting distribution is not Gaussian but symmetric (e.g., a symmetric stable distribution as in~\cite{wang2021convergence}). However, asymptotic normality is an important case studied in the literature. Several online algorithms have been proven to yield estimators that are asymptotically normal. This includes the online algorithms discussed in~\cite{toulis2017asymptotic},~\cite{Li-Mou-Wainwright-Jordon-2020-ROOT-SGD}, and~\cite{Chen-Lai-Li-Zhang-2021-gradient-free-SGD}, among others. Even in the context of constrained minimization, such asymptotic normality results can hold as shown in~\citet[Section 5]{Duchi-Ruan-2020-AoS-Riemannian-SGD},~\cite{JMLR:v25:22-0832}, and~\cite{davis2024asymptotic}. The application of HulC in these situations is particularly interesting because the asymptotic variance depends on active and inactive constraints, making it difficult to estimate. 

It should be noted that the computation of the HulC confidence interval does not require knowledge of the rate of convergence, unlike the Wald confidence interval. Moreover, the computational cost of the HulC confidence interval is
\[
B_{\alpha}\times\mbox{CompCost}(\mathcal{A}(Z_1, \ldots, Z_{T/B_{\alpha}})),
\]
and, generally, this is less than the computational cost of running $\mathcal{A}(\{Z_1, \ldots, Z_T\})$. For example, if $\mbox{CompCost}(\{Z_1, \ldots, Z_m\}) \asymp m^{-\beta}$ for some $\beta \ge 1$, then the computational cost of HulC is lower by a factor of $B_{\alpha}^{1 - \beta}$. This is much less than the computational cost of existing inference methods.

Finally, given the reliance of HulC on very weak assumptions for validity, some price is paid in terms of width with respect to constants but not the rate. In fact, the HulC CI shrinks at the same rate as the convergence of the online estimator to $\theta_{\infty}$ and is larger than the Wald interval by a factor of $\sqrt{2\log\log_2(1/\alpha)}$. For instance, with $\alpha = 0.05$ (for a 95\% confidence interval), the HulC CI is approximately 50\% wider than the Wald interval.

In the remainder of the article, we focus our efforts on the average iterate SGD, hereafter denoted ASGD, for concreteness. In Section~\ref{sec:review-of-asym-SGD}, we prove the asymptotic normality of the ASGD estimator under some mild conditions on the optimization problem and review some existing methods that we use for comparison. (We do not compare to all existing methods.) In Section~\ref{sec:simulation}, we consider some numerical illustrations on the performance of the ASGD estimator and the performance of HulC along with other existing methods for inference.  To analyze HulC's performance using other common SGD algorithms, in Section~\ref{sec:add-methods} we present additional simulation results comparing HulC with baseline methods using implicit-SGD (both averaged and last-iterate), ROOT-SGD, truncated-SGD, and noisy-truncated-SGD.

\section{SGD and Review of Existing Inference Methods}\label{sec:review-of-asym-SGD}
\subsection{Rate of Convergence of the SGD}
Recall that with streaming data $Z_1, \ldots, Z_T$, the SGD iterates are given by
\[
\theta^{(t)} = \theta^{(t-1)} - \eta_t\nabla\ell(Z_t; \theta^{(t-1)}).
\]
We now provide two results that show an expansion for the ASGD and a rate bound for the SGD iterates. We do not claim any novelty in these results, and we strongly believe they are well-known within the community of online algorithms. An interesting aspect to note about the results below is that we allow the step sizes to be stochastic. Most results on asymptotic normality or rate bounds for non-convex cases only deal with non-stochastic step sizes.

In the following analysis, we allow the step size $\eta_t$ to be a matrix, following~\cite{patel2022global,patel2024gradient}. 
Set $M(\theta) := \mathbb{E}[\ell(Z; \theta)]$. 
Define
\begin{equation}\label{eq:martingale-difference-sequence}
\xi_t ~:=~ \nabla\ell(Z_t; \theta^{(t-1)}) - M(\theta^{(t-1)}).
\end{equation}
The following result provides an expansion for the centered and scaled ASGD estimator. The proof is very standard and can even be isolated from the results of~\cite{polyak_1992}. Consider the following assumptions on maps $\theta\mapsto M(\theta)$ and $\theta\mapsto \ell(Z; \theta)$:
\begin{enumerate}[label=\textbf{(A\arabic*)}]
    \item The step sizes $\eta_t, t\ge 1$ are positive semi-definite and invertible.\label{assump:step-size-invertible}
    \item The function $\theta\mapsto M(\theta)$ is differentiable at $\theta_{\infty}$ and the gradient is zero, i.e., $\nabla M(\theta_{\infty}) = 0.$\label{assump:zero-gradient}
    \item The Jacobian of the function $\theta\mapsto M(\theta)$ is $(L, \mu)$-H{\"o}lder continuous at $\theta_{\infty}$ for some $\mu\in(0, 1]$, i.e., \label{assump:holder-cont}
    \[
    \|J^{-1/2}(\nabla^2 M(\theta) - J)J^{-1/2}\|_{\mathrm{op}} \le L\|\theta - \theta_{\infty}\|_J^{\mu},
    \]
    where, recall from~\eqref{eq:sandwich-variance-components}, $J = \nabla^2M(\theta_{\infty})$, and $\|\theta\|_A := \sqrt{\theta^{\top}A\theta}$ for any symmetric matrix $A$.
\end{enumerate}
Assumption~\ref{assump:holder-cont} is comparable to Assumption 3.2 of~\cite{polyak_1992}. In fact, for the proof of Theorem~\ref{thm:SGD-expansion}, we only need
\[
\|J^{-1/2}(\nabla M(\theta) - \nabla^2M(\theta_{\infty})(\theta - \theta_{\infty}))\|_2 \le L\|\theta - \theta_{\infty}\|_J^{1 + \mu}\quad\mbox{for all}\quad \theta.
\]
This assumption is used to control the remainder terms stemming from Taylor series expansion.
\begin{thm}\label{thm:SGD-expansion}
    Assume~\ref{assump:step-size-invertible}--\ref{assump:holder-cont}. Then
    \[
    \sqrt{T}(\bar{\theta}_T - \theta_{\infty}) ~=~ \frac{1}{\sqrt{T}}\sum_{t=1}^T J^{-1}\xi_t ~+~ \mathrm{Rem}_T,
    \]
    where, setting $\tilde{\eta}_t := J^{1/2}\eta_tJ^{1/2}$,
    \begin{align*}
    \|\mathrm{Rem}_T\|_{J} &\le \frac{1}{\sqrt{T}}\sum_{t=1}^{T-1} \|\tilde{\eta}_t^{-1} - \tilde{\eta}_{t+1}^{-1}\|_{\mathrm{op}}\|\theta^{(t)} - \theta_{\infty}\|_J + \frac{L}{\sqrt{T}}\sum_{t=0}^{T-1} \|\theta^{(t)} - \theta_{\infty}\|_J^{1 + \mu}\\
    &\quad+ T^{-1/2}(\|\tilde{\eta}_T^{-1}\|_{\mathrm{op}} + 1)\|\theta^{(T)} - \theta_{\infty}\|_J + T^{-1/2}(\|\tilde{\eta}_1^{-1}\|_{\mathrm{op}} + 1)\|\theta^{(0)} - \theta_{\infty}\|_J.
    \end{align*}
\end{thm}
If $Z_1, \ldots, Z_T$ are independent random variables, then $\mathbb{E}[\xi_t|Z_1, \ldots, Z_{t-1}] = 0$ and, hence, $\{\xi_t\}_{t\ge1}$ forms a martingale difference sequence. Theorem~\ref{thm:SGD-expansion} shows that the centered and scaled ASGD estimator is approximately a martingale. Hence, applying the martingale central limit theorem~\citep{mcleish1974dependent} implies the asymptotic normality of ASGD. In particular, setting $\Delta_T^{\mathrm{CLT}}$ as
\[
\Delta_T^{\mathrm{CLT}} = \sup_{A\;\mathrm{convex}}\,\left|\mathbb{P}(\sqrt{T}(\bar{\theta}_T - \theta_{\infty}) \in A) - \mathbb{P}(N(0, J^{-1}VJ^{-1})\in A)\right|,
\]
Theorem~\ref{thm:SGD-expansion} implies that
\begin{align*}
\Delta_T^{\mathrm{CLT}} &\le \sup_{A\;\mathrm{convex}}\left|\mathbb{P}\left(\frac{1}{\sqrt{T}}\sum_{t=1}^T J^{-1}\xi_t \in A\right) - \mathbb{P}(N(0,J^{-1}VJ^{-1})\in A)\right|\\ 
&\quad+ \inf_{\delta > 0}\left\{\mathfrak{C}(\mbox{tr}(J^{1/2}V^{-1}J^{1/2}))^{1/4}\delta + \mathbb{P}(\|\mbox{Rem}_T\|_J > \delta)\right\},
\end{align*}
for some universal constant $\mathfrak{C}$.
This follows from a simple truncation argument along with anti-concentration. See, for example, Lemma 2.6 of~\cite{bentkus2003dependence} for the anti-concentration bound for Gaussian distribution over convex sets.  

Theorem~\ref{thm:SGD-expansion} does not require any (weak) convexity assumptions on the loss function $\theta\mapsto\ell(Z; \theta)$. Typically, such assumptions are used to control the error of the iterates, i.e., $\|\theta^{(t)} - \theta_{\infty}\|_J$. We present Theorem~\ref{thm:SGD-expansion} in terms of the norm computed with respect to $J$, $\|\cdot\|_J$, producing unit-free error bounds. This also avoids the appearance of condition numbers of matrices $J$ or $\eta_t$. In particular, such a unit-free property implies that the bounds are scale-invariant. Note that the SGD itself can be rewritten in terms of a unit-free vector: setting $\tilde{\theta}^{(t)} = J^{1/2}\theta^{(t)}$, we get
\[
\tilde{\theta}^{(t)} = \tilde{\theta}^{(t-1)} - \tilde{\eta}_tJ^{-1/2}\nabla \ell(Z_t; J^{-1/2}\tilde{\theta}^{(t-1)}).
\]
Here we note that a similar normalization also appeared in~\citet[Appendix A.1]{LelucProtier23}. 

In particular, if $\eta_t$ is chosen (non-stochastically) so that $\|\tilde{\eta}_t^{-1}\|_{\mathrm{op}} = O(t^{\gamma})$ and $\|\tilde{\eta}_t^{-1} - \tilde{\eta}_{t+1}^{-1}\|_{\mathrm{op}} = O(t^{-\gamma})$, then under strong convexity assumptions, Theorem 1 of~\cite{moulines_2011} implies $\|\theta^{(t)} - \theta_{\infty}\|_J = O_p(t^{-\gamma/2})$ and hence,
\[
\|\mathrm{Rem}_T\|_{J} = O(T^{1/2 - \gamma}) + O(T^{\gamma/2 - 1/2}). 
\]
(Also, see~\citet[Sec. 3.6.3]{bach2016lecture3}.)
Hence, choosing $\gamma\in(1/2, 1)$, the remainder term in Theorem~\ref{thm:SGD-expansion} converges to zero as $T\to\infty$. In particular, the best rate of convergence is attained if $\gamma = 2/3$ in which case the remainder converges to zero at a rate of $T^{-1/6}$. This is in sharp contrast to the convergence of remainder terms in the usual influence function expansion~\citep{he1996general,bellec2019first}, where the remainder converges at a rate of $T^{-1/2}$. The slow convergence to normality is also evident in our simulations. 

It should be noted that there are some rate bounds for $\|\theta^{(t)} - \theta_{\infty}\|_J$ without convexity assumptions on $\theta\mapsto M(\theta)$; see, for example,~\cite{gadat2023optimal},~\cite{liu2023aiming}, and~\cite{mertikopoulos2020almost}. In what follows, we provide a simple result for the rate of convergence of $\theta^{(t)} - \theta_{\infty}$ under the PL condition and a quadratic growth condition. Although various relaxed conditions have been proposed in the literature, we identify the following as the easiest or most straightforward to interpret while still permitting non-convex functions. The following result is obtained by combining the analysis of~\cite{bassily2018exponential},~\cite{khaled2022better},~\cite{liao2024error}, and~\cite{neri2024quantitative}. Consider the following assumptions:
\begin{enumerate}[label=\textbf{(B\arabic*)}]
\item The function $M(\cdot)$ is $\kappa$-smooth for some $\kappa > 0$, i.e., \label{eq:smooth}
\begin{equation}\label{eq:smooth-c}\tag{$\kappa$-Sm}
M(\theta') \le M(\theta) + \nabla M(\theta)^{\top}(\theta' - \theta) + \frac{\kappa}{2}\|\theta' - \theta\|_J^2\quad\mbox{for all}\quad \theta', \theta\in\mathbb{R}^d.
\end{equation}
\item The function $\ell(\cdot; \cdot)$ satisfies the Expected Smoothness (ES) condition, i.e., there exist non-negative constants $A$, $B, C$ such that for all $\theta\in\mathbb{R}^d$,\label{eq:ES}
\begin{equation}\label{eq:ES-c}\tag{ES}
\mathbb{E}[\|\nabla \ell(Z; \theta)\|_{J^{-1}}^2] \le 2A(M(\theta) - M(\theta_{\infty})) + B\|\nabla M(\theta)\|_{J^{-1}}^2 + C,
\end{equation}
where, recall, $J = \nabla^2M(\theta_{\infty})$.
\item The function $M(\cdot)$ satisfies the Polyak-{\L}ojasiewicz (PL) condition, i.e., there exists $\mu_{\mathrm{PL}} > 0$ such that for all $\theta\in\mathbb{R}^d$,\label{eq:PL}
\begin{equation}\label{eq:PL-c}\tag{PL}
\frac{1}{2}\|\nabla M(\theta)\|_{J^{-1}}^2 \ge \mu_{\mathrm{PL}}(M(\theta) - M(\theta_{\infty})).
\end{equation}
%\item The function $M(\cdot)$ satisfies the Quadratic Growth (QG) condition, i.e., there exists $\mu_{\mathrm{QG}} > 0$ and $\nu > 0$ such that for all $\theta\in\mathbb{R}^d$ satisfying $M(\theta) \le M(\theta_{\infty}) + \nu$,\label{eq:QG}
%\begin{equation}\label{eq:QG-c}\tag{QG}
%M(\theta) - M(\theta_{\infty}) \ge \frac{\mu_{\mathrm{QG}}}{2}\|\theta - \theta_{\infty}\|_J^2.
%\end{equation}
\end{enumerate}
These conditions are now standard in the literature on non-convex optimization, albeit with the usual Euclidean norm. We present the modified assumptions using $\|\cdot\|_{J}$ and $\|\cdot\|_{J^{-1}}$ to maintain some scale invariance. In assumption~\ref{eq:ES}, we can, without loss of generality, assume that $B\ge2.$ For a detailed review and generality of these assumptions, see~\cite{liao2024error} and~\cite{khaled2022better}. 
%\cite{liao2024error}, in particular, prove that if $M(\cdot)$ is a weakly convex function (i.e., $\theta\mapsto M(\theta) + \rho\|\theta\|_{J}^2$ is convex), then~\ref{eq:PL} implies~\ref{eq:QG} with $\nu = \infty$ (when the minimizer $\theta_{\infty}$ is uniquely defined). 
It is also well-known that~\ref{eq:PL} is readily satisfied by strongly convex functions. Formally, if $\theta\mapsto M(\theta)$ satisfies
\begin{equation}\label{eq:strong-convex}\tag{SC}
    M(\theta') \ge M(\theta) + \nabla M(\theta)^{\top}(\theta' - \theta) + \frac{\mu^*}{2}\|\theta' - \theta\|_J^2, \quad\mbox{for all}\quad \theta', \theta\in\mathbb{R}^d,
\end{equation}
then minimizing both sides with respect to $\theta'$ yields
\begin{equation}\label{eq:implication-of-SC}
M(\theta_{\infty}) \ge M(\theta) - \frac{1}{2\mu^*}\|\nabla M(\theta)\|_{J^{-1}}^2.
\end{equation}
Note that the right hand side of~\eqref{eq:strong-convex} is minimized at $\theta' = \theta - J^{-1}\nabla M(\theta)/\mu^*$. Note that~\eqref{eq:implication-of-SC} is equivalent to~\eqref{eq:PL-c}. Finally, to indicate the generality of~\eqref{eq:PL-c}, we mention that a class of neural networks trained with the least squares loss (or more generally a strongly convex loss) satisfies the PL condition, i.e., $\theta\mapsto \mathbb{E}[(Y - f(X; \theta))^2]$ for a neural network $f(\cdot; \cdot)$ satisfies the PL condition. We refer the reader to~\citet[Lemma 7.12]{soltanolkotabi2018theoretical},~\citet[Section 4]{pmlr-v80-charles18a}, and~\cite{liu2022loss}.  
Several weaker versions of~\eqref{eq:PL-c} exist in the literature~\citep{lojasiewicz1963propriete,weissmann2024almost,kurdyka1998gradients}. In particular,~\cite{lojasiewicz1963propriete} proposed the alternative $\|\nabla M(\theta)\|^{1/\beta} \ge \mu_{\mathrm{wPL}}(M(\theta) - M(\theta_{\infty}))$ for all $\theta\in\mathbb{R}^d$ for some $\beta\in[1/2, 1]$. Theorem~\ref{thm:rate-bound-SGD} can also be extended to this case, using Lemma 3.1 of~\cite{weissmann2024almost}. Also, see~\cite{karandikar2024convergence} and~\cite{scaman2022convergence} for more details.

The following result proves a Robbins-Sigmund style recursive inequality for $\Delta_t = M(\theta^{(t)}) - M(\theta_{\infty})$ that commonly appears in the literature on stochastic algorithms, and an application of~\eqref{eq:PL-c} implies a rate bound for the SGD iterates. Recall the notation $\tilde{\eta}_t = J^{1/2}\eta_tJ^{1/2}$. Let $\mathcal{F}_{t}$ be the $\sigma$-field generated by $\theta^{(0)}, Z_1, Z_2, \ldots, Z_t$ for $t\ge0$.

\begin{thm}\label{thm:rate-bound-SGD}
    Under assumptions~\ref{eq:smooth} and \ref{eq:ES}, and additionally that $\eta_t$ is $\mathcal{F}_{t-1}$ measurable, for all $t\ge1$, we have 
    \begin{equation}\label{eq:basic-recursion}
        \mathbb{E}[\Delta_t|\mathcal{F}_{t-1}] \le
     \Delta_{t-1}\left(1 + (A + B\kappa)\kappa\|\tilde{\eta}_t\|_{\mathrm{op}}^2\right) - \|\nabla M(\theta^{(t-1)})\|_{\eta_t}^2 + \frac{C\kappa}{2}\|\tilde{\eta}_t\|_{\mathrm{op}}^2.
    \end{equation}
        Consequently, for any $\lambda, \zeta > 0$,
       \begin{equation}\label{eq:tail-bound-sum-of-gradients}
       \mathbb{P}\left(\sum_{t=0}^{\infty} \|\nabla M(\theta^{(t)})\|_{\eta_t}^2 \ge \frac{8(\mathbb{E}[\Delta_0] + \zeta)}{\lambda}\exp\left(\frac{(2A + 2B\kappa)\zeta}{C}\right)\right) \le \lambda + 3\mathbb{P}\left(\frac{C\kappa}{2}\sum_{t=1}^{\infty} \|\tilde{\eta}_t\|_{\mathrm{op}}^2 \ge \zeta\right).
       \end{equation}
    \end{thm}
An inequality very similar to inequality~\eqref{eq:basic-recursion} was effectively used in~\cite{bertsekas2000gradient} and~\citet[Ineq. (53)]{karandikar2024convergence} to imply convergence of SGD to a stationary point.
The basic recursion~\eqref{eq:basic-recursion} can be compared with the setting of the ``almost supermartingale convergence theorem'' of Robbins and Siegmund~\citep{Robbins1971}. Also, see Theorem 40.2 of~\cite{506notes} for a detailed proof. 
%This theorem implies that, under assumptions~\ref{eq:smooth} and~\ref{eq:ES}, if
%\[
%\sum_{t=1}^{\infty} \|\tilde{\eta}_t\|_{\mathrm{op}}^2 < \infty\quad\mbox{almost surely},
%\]
%then, almost surely,
%\[
%\sum_{t=1}^{\infty} \|\nabla M(\theta^{(t)})\|_{J^{-1}}^2 < \infty\quad\mbox{and}\quad \{\Delta_t\}_{t\ge0} \quad\mbox{converges}.
%\]
Some recent works have provided quantitative versions of the Robbins-Siegmund theorem; see, for example,~\cite{liu2022almost} and~\cite{neri2024quantitative}. Assuming~\ref{eq:smooth}--\ref{eq:PL} and also that $\eta_t$ is a scalar multiple of the identity matrix, inequality~\eqref{eq:basic-recursion} can be simplified to
\begin{equation}\label{eq:recursion-Delta_t}
    \mathbb{E}[\Delta_t|\mathcal{F}_{t-1}] \le \Big(1 + (A + B\kappa)\kappa\|\tilde{\eta}_t\|_{\mathrm{op}}^2 - 2\mu_{\mathrm{PL}}\|\tilde{\eta}_t\|_{\mathrm{op}}\Big)\Delta_{t-1} + \frac{C\kappa\|\tilde{\eta}_t\|_{\mathrm{op}}^2}{2},
    \end{equation}
which can be compared with the inequality in Lemma 1 of~\cite{liu2022almost} and inequality (5) of~\cite{weissmann2024almost}. The tail probability inequality~\eqref{eq:tail-bound-sum-of-gradients} implies that
\[
\sum_{t=1}^{\infty} \|\nabla M(\theta^{(t-1)})\|_{\eta_t}^2 = O_p(1),\quad\mbox{whenever}\quad \sum_{t=1}^{\infty} \|\tilde{\eta}_t\|_{\mathrm{op}}^2 = O_p(1).
\]
and consequently, $\|\nabla M(\theta^{(t-1)})\|_{\eta_t} \to 0$ as $t\to\infty$. This does not readily imply that $\nabla M(\theta^{(t)}) \to 0$ as $t\to\infty$ because $\eta_t$ itself converges to zero as $t\to\infty$. (Observe that in the non-stochastic case, if $\eta_t \le \mathfrak{C}t^{-\gamma}$ for all $t$, then $\sum_{t=1}^{\infty} \|\tilde{\eta}_t\|_{\mathrm{op}}^2 = O_p(1)$ is satisfied if and only if $\gamma > 1/2$.) Assuming, in addition, that $\sum_{t=1}^{m} \|\tilde{\eta}_t\|_{\mathrm{op}}^{-1}$ diverges to infinity in probability as $m\to\infty$ implies $\|\nabla M(\theta^{(t)})\|_{J^{-1}} = o_p(1)$. See, for example, the proof of Corollary 40.1 of~\cite{506notes}.   Assuming~\ref{eq:PL}, we can also derive bounds for the iterates.
\begin{cor}\label{cor:under-PL-convergence}
Assume~\ref{eq:smooth} and~\ref{eq:ES}. Then for any $\lambda, \zeta > 0$,
\begin{equation}\label{eq:norm-of-gradients}
\mathbb{P}\left(\sum_{t=0}^{\infty} \frac{\|\nabla M(\theta^{(t)})\|_{J^{-1}}^{2}}{\|\tilde{\eta}_t^{-1}\|_{\mathrm{op}}} \ge  \frac{8(\mathbb{E}[\Delta_0] + \zeta)}{\lambda}\exp\left(\frac{(2A + 2B\kappa)\zeta}{C}\right)\right) \le \lambda + 3\mathbb{P}\left(\frac{C\kappa}{2}\sum_{t=1}^{\infty} \|\tilde{\eta}_t\|_{\mathrm{op}}^2 \ge \zeta\right).
\end{equation}
If, in addition,~\ref{eq:PL} holds, then
\begin{equation}\label{eq:norm-of-iterates}
\mathbb{P}\left(\sum_{t=0}^{\infty} \frac{\|\theta^{(t)} - \theta_{\infty}\|_{J}^{2}}{\|\tilde{\eta}_t^{-1}\|_{\mathrm{op}}} \ge  \frac{8(\mathbb{E}[\Delta_0] + \zeta)}{\mu_{\mathrm{PL}}\lambda}\exp\left(\frac{(2A + 2B\kappa)\zeta}{C}\right)\right) \le \lambda + 3\mathbb{P}\left(\frac{C\kappa}{2}\sum_{t=1}^{\infty} \|\tilde{\eta}_t\|_{\mathrm{op}}^2 \ge \zeta\right).
\end{equation}
\end{cor}
\begin{remark}[Asymptotic Normality of ASGD]
Corollary~\ref{cor:under-PL-convergence} implies that under assumptions~\ref{eq:smooth}--\ref{eq:PL},
\[
\sum_{t=0}^{\infty} \frac{\|\theta^{(t)} - \theta_{\infty}\|_J^2}{\|\tilde{\eta}_t^{-1}\|_{\mathrm{op}}} = O_p(1)\quad\mbox{whenever}\quad \sum_{t=1}^{\infty} \|\tilde{\eta}_t\|_{\mathrm{op}}^2 = O_p(1).
\]
This can be used to control the remainder terms in the asymptotic expansion from Theorem~\ref{thm:SGD-expansion}.
Observe that by H{\"o}lder's inequality,
\begin{align*}
\frac{1}{\sqrt{T}}\sum_{t=1}^{T-1} \|\tilde{\eta}_t^{-1} - \tilde{\eta}_{t+1}^{-1}\|_{\mathrm{op}}\|\theta^{(t)} - \theta_{\infty}\|_J 
%&= \frac{1}{\sqrt{T}}\sum_{t=1}^{T-1} \|\tilde{\eta}_t^{-1} - \tilde{\eta}_{t+1}^{-1}\|_{\mathrm{op}}\|\tilde{\eta}_t^{-1}\|_{\mathrm{op}}^{1/2}\frac{\|\theta^{(t)} - \theta_{\infty}\|_J}{\|\tilde{\eta}_t^{-1}\|_{\mathrm{op}}^{1/2}}\\
&\le \frac{1}{\sqrt{T}}\left(\sum_{t=1}^{T} \|\tilde{\eta}_t^{-1} - \tilde{\eta}_{t+1}^{-1}\|_{\mathrm{op}}^2\|\tilde{\eta}_t^{-1}\|_{\mathrm{op}}\right)^{1/2}\left(\sum_{t=1}^T \frac{\|\theta^{(t)} - \theta_{\infty}\|_{J}^{2}}{\|\tilde{\eta}_t^{-1}\|_{\mathrm{op}}}\right)^{1/2},\\
\frac{L}{\sqrt{T}}\sum_{t=0}^{T-1} \|\theta^{(t)} - \theta_{\infty}\|_J^{1 + \mu} 
%&= \frac{L}{\sqrt{T}}\sum_{t=0}^{T-1} \left(\frac{\|\theta^{(t)} - \theta_{\infty}\|_J^2}{\|\tilde{\eta}_t^{-1}\|_{\mathrm{op}}}\right)^{(1 + \mu)/2} \|\tilde{\eta}_t^{-1}\|_{\mathrm{op}}^{(1 + \mu)/2}\\
&\le \frac{L}{\sqrt{T}}\left(\sum_{t=0}^{T} \frac{\|\theta^{(t)} - \theta_{\infty}\|_{J}^{2}}{\|\tilde{\eta}_t^{-1}\|_{\mathrm{op}}}\right)^{(1 + \mu)/2}\left(\sum_{t=0}^{T} \|\tilde{\eta}_t^{-1}\|_{\mathrm{op}}^{(1 + \mu)/(1-\mu)}\right)^{(1-\mu)/2}.
\end{align*}
Therefore, asymptotic normality of ASGD holds under assumptions~\ref{assump:step-size-invertible}--\ref{assump:holder-cont} and~\ref{eq:smooth}--\ref{eq:PL} whenever the following conditions hold on the step size sequence:
\begin{equation}
\|\tilde{\eta}_1^{-1}\|_{\mathrm{op}} = O_p(1),\quad
\|\tilde{\eta}_T^{-1}\|_{\mathrm{op}} = O_p(T^{1/2}),\quad
\sum_{t=1}^{\infty} \|\tilde{\eta}_t\|_{\mathrm{op}}^2 = O_p(1),
\end{equation}
and
\begin{align*}
\sum_{t=1}^{T} \|\tilde{\eta}_t^{-1} - \tilde{\eta}_{t+1}^{-1}\|_{\mathrm{op}}^2\|\tilde{\eta}_t^{-1}\|_{\mathrm{op}} = o_p(T),\quad\mbox{and}\quad \left(\sum_{t=0}^{T} \|\tilde{\eta}_t^{-1}\|_{\mathrm{op}}^{(1 + \mu)/(1-\mu)}\right)^{1-\mu} = o_p(T).
\end{align*}
If $\mu = 1$ in assumption~\ref{assump:holder-cont}, then the last condition is interpreted as $\max_{1\le t\le T}\|\tilde{\eta}_t^{-1}\|_{\mathrm{op}} = o_p(T^{-1/2})$.
\end{remark}

Notably, our results require smoothness only for the expected loss function, including the existence of higher-order derivatives. The loss function itself needs only the existence of a derivative. This distinction is particularly advantageous when dealing with non-smooth, ``irregular" losses like the pinball~\citep{chen2023recursive} or hinge loss. Furthermore, even the requirement for the loss function's derivative can be relaxed, provided an unbiased estimate of the expected loss function's derivative is obtainable.
\subsection{Existing Inference Methods}
All the existing inference methods for ASGD rely on the expansion and asymptotic normality of $\bar{\theta}_T$. In the following, we briefly summarize the existing methods that are used for comparison with the HulC confidence intervals in Section~\ref{sec:simulation}. Our comparison is not exhaustive, given the numerous methods in existence.
\paragraph{Wald interval} (offline method --- baseline): We use the Wald interval as a baseline method. This uses the global minimizer of the empirical loss function as defined in~\eqref{eq:empirical-M-estimator}. The Wald interval is
\begin{equation}\label{eq:Wald-interval-2}
\widehat{\mathrm{CI}}_{T,\alpha}^{(k)} := \left[e_k^{\top}\widehat{\theta}_T \pm \frac{z_{\alpha/2}}{T^{1/2}}(e_k^{\top}\widehat{J}_T^{-1}\widehat{V}_T\widehat{J}_T^{-1}e_k)^{1/2}\right],    
\end{equation}
where
\begin{equation}\label{eq:sandwich-variance-components-estimators-2}
    \widehat{J}_T := \frac{1}{T}\sum_{i=1}^T \nabla^2\ell(Z_i; \widehat{\theta}_T)\quad\mbox{and}\quad \widehat{V}_T := \frac{1}{T}\sum_{i=1}^T (\nabla \ell(Z_i; \widehat{\theta}_T))(\nabla \ell(Z_i; \widehat{\theta}_T))^{\top}.
\end{equation}
\paragraph{ASGD Plug-in} \citep{Chen_SGD_2021}: This is possibly the first general inference method using ASGD. The confidence interval is given by
\begin{equation}\label{eq:ASGD-plug-in}
    \widehat{\mathrm{CI}}_{T,\alpha}^{(k)} := \left[e_k^{\top}\bar{\theta}_T \pm \frac{z_{\alpha/2}}{T^{1/2}}(e_k^{\top}\widetilde{J}_T^{-1}\widetilde{V}_T\widetilde{J}_T^{-1}e_k)^{1/2}\right],
\end{equation}
where 
\begin{equation}\label{eq:sandwich-variance-components-estimators-2}
    \widetilde{J}_T := \frac{1}{T}\sum_{t=1}^T \nabla^2\ell(Z_t; \theta^{(t-1)})\quad\mbox{and}\quad \widetilde{V}_T := \frac{1}{T}\sum_{t=1}^T (\nabla \ell(Z_t; \theta^{(t-1)}))(\nabla \ell(Z_t; \theta^{(t-1)}))^{\top}.
\end{equation}
In contrast to our asymptotic normality result, the construction of this confidence interval requires the loss function to be twice differentiable. Additionally, the results of~\cite{Chen_SGD_2021} imply a slow rate of convergence of $\widetilde{J}_T^{-1}\widetilde{V}_T\widetilde{J}_T^{-1}$ compared to the offline estimator $\widehat{J}_T^{-1}\widehat{V}_T\widehat{J}_T^{-1}$. This directly impacts the coverage of the ASGD plug-in interval; see~\citet[Theorem 1]{kauermann2001note}. More recent studies offer alternative online variance estimators without matrix inversion, for instance,~\cite{zhu2023online}.
\paragraph{ASGD $t$-stat} \citep{ibragimov_tstat_2010,zhu_paralelSGD_2024}: The $t$-statistic for ASGD, proposed by~\cite{zhu_paralelSGD_2024} and building on the general method envisioned by~\cite{ibragimov_tstat_2010}, also relies on bucketing the data into B buckets, similar to HulC. The method is as follows: Fix any $B\ge 2$, and compute  
\[
\bar{\theta}_T^{(j)} = \mathcal{A}(Z_{j+ B\lfloor (T - j)/B\rfloor}, \ldots, Z_j; \theta^{(0,j)}),\quad\mbox{for}\quad 1\le j\le B.
\]
Compute
\[
\widetilde{\theta}_T = \frac{1}{B}\sum_{j=1}^B \bar{\theta}_T^{(j)}\quad\mbox{and}\quad \widetilde{\sigma}_{k,T}^2 = \frac{1}{B-1}\sum_{j=1}^B (e_k^{\top}\bar{\theta}_T^{(j)} - e_k^{\top}\widetilde{\theta}_T)^2.
\]
Report
\[
\widehat{\mathrm{CI}}_{T,\alpha}^{(k)} := \left[e_k^{\top}\widetilde{\theta}_T \pm t_{B-1,\alpha/2}\widetilde{\sigma}_{k,T}\right],
\]
where $t_{B-1,\alpha/2}$ is the $(1-\alpha/2)$-th quantile of $t$ distribution with $B-1$ degrees of freedom. The results of~\cite{ibragimov_tstat_2010} and~\cite{zhu_paralelSGD_2024} imply that this is an asymptotically valid $(1-\alpha)$ confidence interval for any $B \ge 2$. Because there is no straightforward method to choose $B$, we choose $B = B^*$ from the HulC confidence interval. This means that for a 95\% confidence interval, we use approximately 5 buckets.

\section{Numerical Illustrations}\label{sec:simulation}
In a simulation study, we assess the utility of HulC by comparing confidence regions for ${\theta}_{\infty} \in \mathbb{R}^d$ on two simple cases: linear regression and logistic regression. Mimicking the simulation settings from~\cite{Chen_SGD_2021}, we generate $T$ iid samples $X_i = (1, Z_i^{\top})^{\top}\in \mathbb{R}^d$, $1\le i\le T$, where $Z_i\sim N(0, \Sigma)$. We consider three different types of covariance matrices: (i) Identity, $\Sigma_{i,j} = \mathbf{1}\{i= j\}$; (ii) Toeplitz matrix ($\Sigma_{i,j} = 0.5^{|i-j|}$); and (iii) an equicorrelation matrix ($\Sigma_{i,j}=(0.2)\mathbf{1}\{i\ne j\} + \mathbf{1}\{i= j\}$). 
The responses $Y_i, 1\le i\le T$ for linear and logistic cases are generated as follows:
\[
Y_i|X_i \sim 
\begin{cases}
N(\theta_{\infty}^{\top}X_i,\, 1),&\mbox{for linear regression},\\
\mbox{Bernoulli}(\exp(\theta_{\infty}^{\top}X_i)/(1 + \exp(\theta_{\infty}^{\top}X_i))), &\mbox{for logistic regression.}
\end{cases}
\]
In all cases, the coordinates of the parameter vector $\theta_{\infty}$ are obtained by linearly spacing 0 and 1. For example, if $d=5$, then $\theta_{\infty} = [0, 0.25, 0.5, 0.75, 1]^\top$. We consider three different dimension sizes $d \in \{5, 20, 100\}$ corresponding to low, moderate, and high dimensions. We also consider four sample sizes $T\in\{10^3, 10^4, 10^5/2, 10^5\}$, again corresponding to low, moderate, and high sample sizes. Throughout, we aim to cover $e_k^{\top}\theta_{\infty}$ with a 95\% confidence for each coordinate $k\in\{1, 2, \ldots, d\}$.

For simplicity, we take non-random scalar step sizes of the form $\eta_t = ct^{-0.505}$. The choice of the hyperparameter $c$ is very crucial for the performance of ASGD and the resulting confidence intervals. Clearly, if the performance of ASGD is poor, then all inference methods based on the ASGD estimator are poor. Due to the lack of enough resources on the ``right'' way to choose $c$, we present our results over a range of $c$ values (see Table~\ref{tab:c-ASGD} in Supplement~\ref{sec:annex_plots} for the grid of $c$ values we tested). For each choice of $c$, $d$, $T$, and covariance structure, we generate 200 datasets and construct the confidence intervals from each dataset to estimate the coverage and width of the intervals. 
Formally, in each run of $200$ independent experiments, we first generate the data. Given the data, we fix $c$ from a grid of values and record the coverage and width ratios for each inference technique --- specifically, we check whether the $k$th coordinate of the parameter, $e_k^{\top}\theta_{\infty}$, falls within the corresponding confidence interval $\widehat{\mathrm{CI}}_{T,\alpha}^{(k)}$, assigning a value of $1$ if $e_k^{\top}\theta_{\infty} \in \widehat{\mathrm{CI}}_{T,\alpha}^{(k)}$ and $0$ otherwise. The estimated coverage is then calculated as the proportion of $200$ replications in which the parameter was covered, with a target of approximately $95\%$ (equivalent to $190$ out of $200$ independent instances of coverage).\footnote{The code for all simulations is available at \url{https://github.com/Arun-Kuchibhotla/HulC/tree/main/Online_Algorithms_HulC_simulations}. An interactive graph for all simulations is available at \url{https://public.tableau.com/app/profile/selina.carter6629/viz/OnlineinferencesimulationsOLSandlogisticregression/Coverageandwidthratio_paper}. }

Our findings at a high level are as follows:
\begin{itemize}
    \item \textbf{The offline Wald interval achieves the desired $95\%$ coverage of $\theta_{\infty}$, making it a reliable baseline method}: Regardless of model type, sample size $T$, dimension $d$, and covariance type, the Wald interval usually achieves approximate $95\%$ coverage across all coordinates of $\theta_{\infty}$, making it a useful baseline in both linear and logistic regression. 
    % The only exception arises in high-dimensional  ($d=100$) logistic regression with a small sample size ($T=10^3$), where the Wald interval either cannot be computed due to the singularity of the data matrix or $X^\top X$ fails to achieve adequate coverage.
    
    \item \textbf{ASGD is highly sensitive to hyperparameter $c$}: There is a ``basin of attraction'' of $c$ (depending on the model parameters) in which the ASGD estimator $\bar{\theta}_T$ converges to $\theta_{\infty}$. This basin of attraction becomes narrower as the dimension $d$ increases.  The sample size $T$ and the covariance scheme also shift the correct basin of $c$. %In turn, the ASGD plug-in confidence interval, HulC, and $t$-stat confidence intervals are all sensitive to $c$.
    \item \textbf{The ASGD plug-in confidence interval undercovers $\theta_{\infty}$}: The ASGD plug-in interval typically only reaches 90\% coverage, well below the target 95\%. The coverage decreases as $d$ increases. This undercoverage might stem from two sources: (i) slow rate of convergence to asymptotic normality; and (ii) slow rate of consistency of the ASGD plug-in variance estimation. 
    \item \textbf{Both the $t$-stat and HulC intervals achieve the correct coverage for an appropriately chosen $c$}: In low and medium dimensions ($d=5, 20$), there always exists a $c$ such that the $t$-stat  and HulC intervals achieve correct coverage for both linear and logistic settings. In high-dimensions ($d = 100$), the empirical evidence is more favorable for the linear regression setting, and favors lower values of $c$ for all sample sizes. For logistic regression, the choice of grid of values for $c$ needs to be tuned carefully to achieve correct coverage.
    \item \textbf{HulC confidence intervals are ``reasonable'' in width}: Typically, HulC confidence intervals are wider than all other methods and this is in line with the theoretical width analysis of~\cite{hulc_2023_arun}. Our numerical study suggests that the HulC intervals are only slightly wider than those of the $t$-stat: when correct coverage is achieved, HulC intervals are $10\%$ higher, on median, than $t$-stat intervals. For larger sample sizes ($T\ge 5\cdot 10^4$), HulC intervals are $2.3$ times the width of the Wald intervals, on median.  %For well-chosen $c$ (i.e., where coverage is between 92.5-97.5\% and HulC width ratios are less than 10), the median HulC width ratio is 2.1 across all coordinates of $\theta$, dimensions, covariance schemes, sample sizes in the case of linear regression. For logistic regression, the median ratio is 1.9 (but we must exclude $T=1,000$ and $d=100$ for Toeplitz and equicorrelation covariance schemes, in which the Wald estimator does not converge due to non-singularity).
    
\end{itemize}

\cite{zhu_paralelSGD_2024} also conducted a numerical study for the $t$-stat method (which they call ``parallel stochastic optimization''). Unlike their simulations, we do not compare the running times of each method, as our SGD code is not optimized for efficiency using C++ or other compiled languages. Whereas \cite{zhu_paralelSGD_2024} examine the effect of the number of batches $B$, we do not, since we fix $B$ between $5$ and $6$ to be comparable to HulC. These choices of $B$ fall within the ``reasonable range'' for $B$ identified by \cite{zhu_paralelSGD_2024}, and also show that $B=6$ is a good choice in practice. Furthermore, \cite{zhu_paralelSGD_2024} do not compare the $t$-stat method with the ASGD plug-in method. Instead, they evaluated it against the random scaling method \citep{Lee_random_scaling_2022}, which has demonstrated coverage similar to the plug-in method while offering faster computation. They also compare the $t$-stat method against the ``oracle'' Wald interval that assumes access to $J^{-1}VJ^{-1}$, while we use the data-driven sandwich estimator. Furthermore, \cite{zhu_paralelSGD_2024} only employ a single fixed hyperparameter $c=0.5$, fewer dimension sizes $d\in\{5, 20\}$, and only identity covariance; we found that when employing $d=100$ with other covariance schemes, different choices of $c$ are required to achieve correct coverage. \cite{zhu_paralelSGD_2024} test a similar range of sample sizes ($T \in \{600, \dots, 6\cdot 10^4\}$) for linear regression but larger sizes ($T \in \{600, \dots, 2\cdot 10^5\}$) for logistic regression. They use a finer grid of $T$ with a mesh size of $600$, resulting in $100$ choices of $T$ for linear regression and $333$ choices of $T$ for logistic regression---made feasible by their use of random scaling as a proxy for the computationally heavy ASGD plug-in intervals. In contrast, we only consider four sample sizes ($T \in \{10^3, 10^4, 5\cdot 10^4, 10^5\}$) for each choice of $c$.  Despite these differences, our simulations agree that the $t$-stat method achieves correct coverage for a large enough $T$, but has a wider confidence interval than the random scale method (comparable to the ASGD plug-in method).

In the remaining part of this section, we first present simulations illustrating the sensitivity of ASGD in Subsection~\ref{subs:asgd_sensitivity}, followed by simulation summaries for linear (Subsection~\ref{subs:lin_reg_ASGD}) and logistic regression (Subsection~\ref{subs:log_reg_ASGD}).

\subsection{Sensitivity of ASGD}\label{subs:asgd_sensitivity}

\begin{figure}[H]
    \centering
    \begin{subfigure}[b]{0.475\textwidth}
        \centering
        \includegraphics[width=\textwidth]{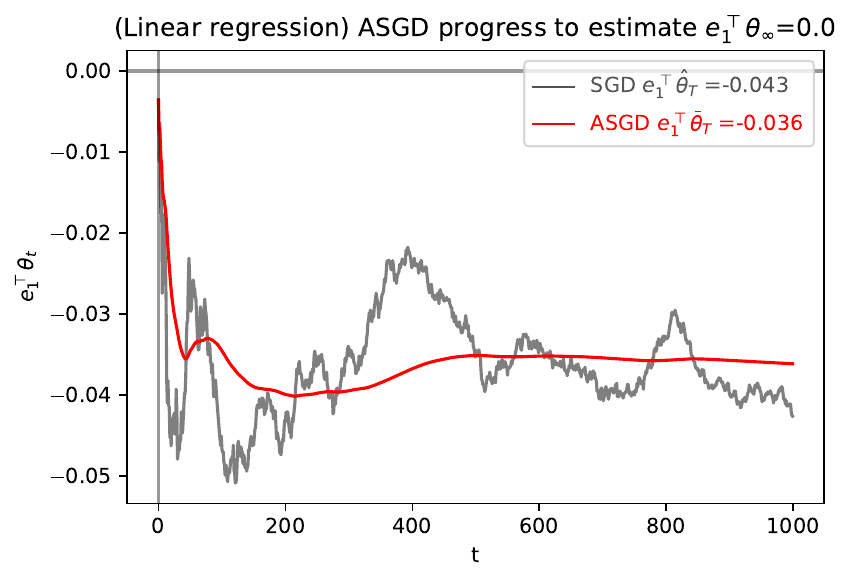}
        \caption[]%
        {{\footnotesize Single trajectory of ASGD and SGD for $e_1^{\top}\theta_{\infty}=0$}}    
        \label{fig:ASGD_n1000_smallc_theta1}
    \end{subfigure}
    \hfill
    \begin{subfigure}[b]{0.475\textwidth}  
        \centering 
        \includegraphics[width=\textwidth]{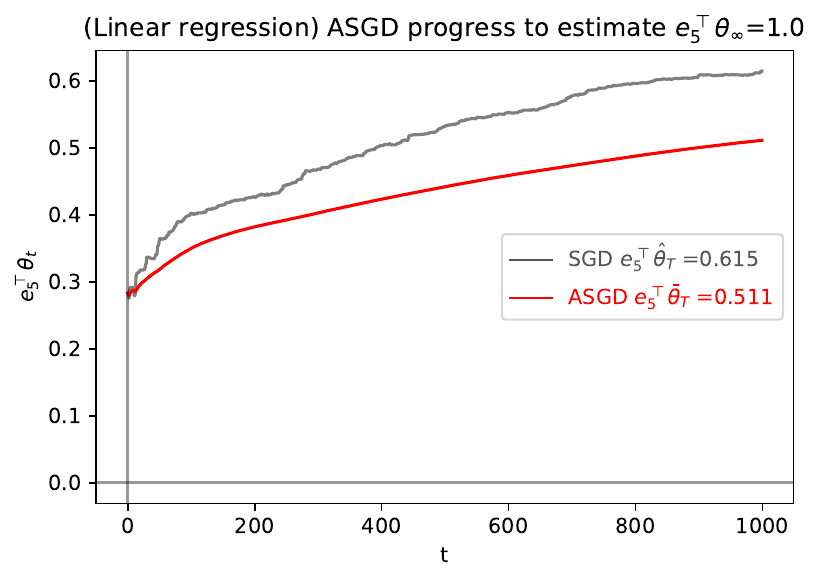}
        \caption[]%
        {{\footnotesize Single trajectory of ASGD and SGD for $e_5^{\top}\theta_{\infty}=1$}}    
        \label{fig:ASGD_n1000_smallc_theta5}
    \end{subfigure}
    \vskip\baselineskip
    \begin{subfigure}[b]{0.475\textwidth}   
        \centering 
        \includegraphics[width=\textwidth]{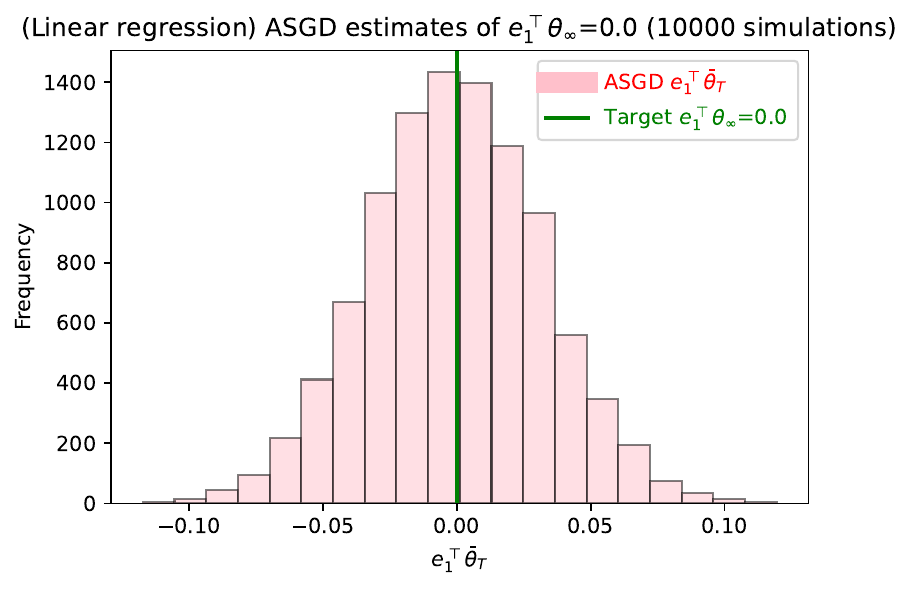}
        \caption[]%
        {{\footnotesize Histogram of ASGD estimates for $e_1^{\top}\theta_{\infty}=0$ \newline  ($10^4$ replications)}}    
        \label{fig:ASGD_n1000_smallc_theta1_hist}
    \end{subfigure}
    \hfill
    \begin{subfigure}[b]{0.475\textwidth}   
        \centering 
        \includegraphics[width=\textwidth]{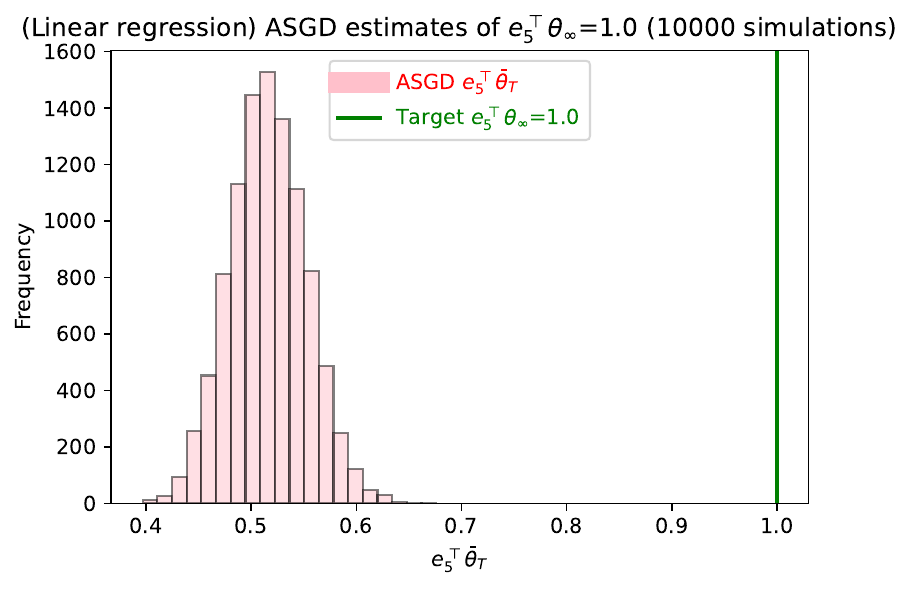}
        \caption[]%
        {{\footnotesize Histogram of ASGD estimates for $e_5^{\top}\theta_{\infty}=1$ \newline  ($10^4$ replications)}}    
        \label{fig:ASGD_n1000_smallc_theta5_hist}
    \end{subfigure}
    \caption
    {\small For the linear regression task with identity covariance, $d=5$, $T=10^3$, and a \textbf{small step size hyperparameter} $c=0.01$, both the ASGD and SGD estimators for $\theta_{\infty}$ fail to converge. The distance from the target is worse for the larger coordinate, $e_5^{\top}\theta_{\infty}=1$, compared to the first coordinate, $e_1^{\top}\theta_{\infty}=0$,  likely due to the initialization procedure, which favors smaller coordinates. In plot (c), which presents a histogram of $10,000$ repetitions of ASGD, there is no systematic bias: the mean ASGD estimates for $e_1^{\top}\theta_{\infty}=0$ is $-0.0006$. However, there is systematic bias for $e_5^{\top}\theta_{\infty}$ (as well as all larger coordinates), as seen in (d): the mean ASGD estimate for $e_5^{\top}\theta_{\infty}=1$ is $0.516$.} 
\label{fig:ASGD_n1000_smallc_nonconvergence}
\end{figure}

\begin{figure}[H]
    \centering
    \begin{subfigure}[b]{0.475\textwidth}
        \centering
        \includegraphics[width=\textwidth]{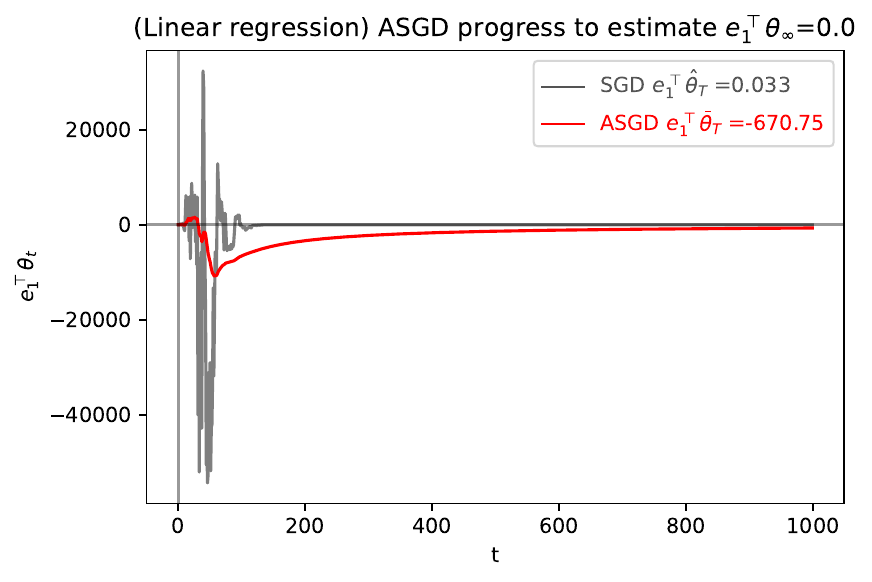}
        \caption[]%
        {{\footnotesize Single trajectory of ASGD and SGD for $e_1^{\top}\theta_{\infty}=0$}}    
        \label{fig:ASGD_n1000_largec_theta1}
    \end{subfigure}
    \hfill
    \begin{subfigure}[b]{0.475\textwidth}  
        \centering 
        \includegraphics[width=\textwidth]{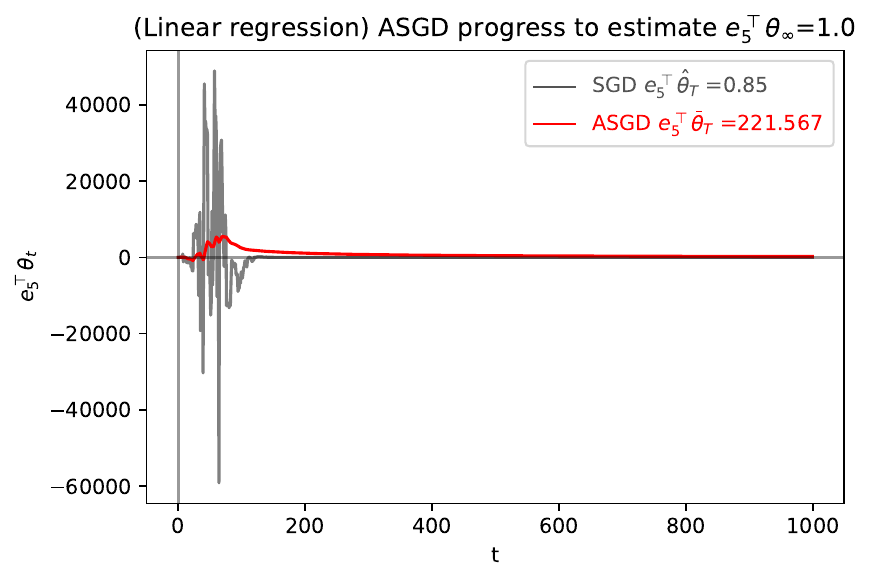}
        \caption[]%
        {{\footnotesize Single trajectory of ASGD and SGD for $e_5^{\top}\theta_{\infty}=1$}}    
        \label{fig:ASGD_n1000_largec_theta5}
    \end{subfigure}
    \vskip\baselineskip
    \begin{subfigure}[b]{0.475\textwidth}   
        \centering 
        \includegraphics[width=\textwidth]{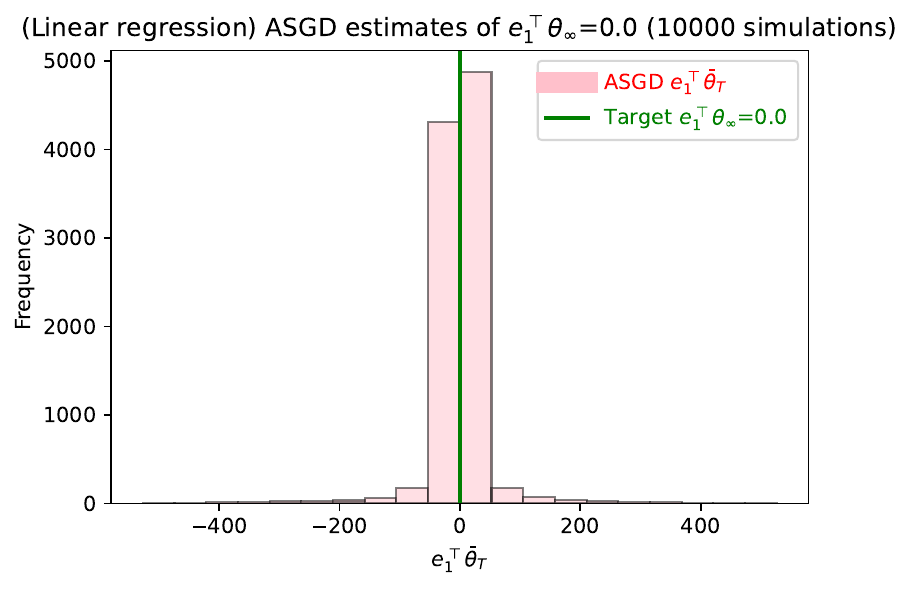}
        \caption[]%
        {{\footnotesize Histogram of ASGD estimates for $e_1^{\top}\theta_{\infty}=0$ \newline  ($10^4$ replications)}}    
        \label{fig:ASGD_n1000_largec_theta1_hist}
    \end{subfigure}
    \hfill
    \begin{subfigure}[b]{0.475\textwidth}   
        \centering 
        \includegraphics[width=\textwidth]{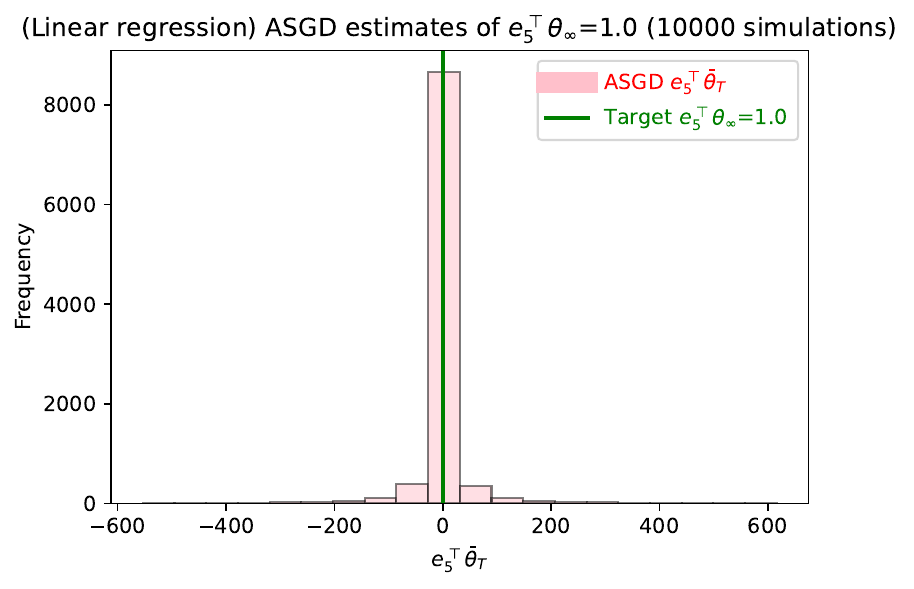}
        \caption[]%
        {{\footnotesize Histogram of ASGD estimates for $e_5^{\top}\theta_{\infty}=1$ \newline  ($10^4$ replications)}}    
        \label{fig:ASGD_n1000_largec_theta5_hist}
    \end{subfigure}
    \caption
    {\small For the linear regression task with identity covariance, $d=5$, $T=10^3$, and a \textbf{large step size hyperparameter} $c=2$, the SGD estimator tends to converge, but not the ASGD estimator, due to the large initial ``wrong SGD points'' at the start of the trajectory. The ASGD estimator is off by a large margin from the target parameter for both $e_1^{\top}\theta_{\infty}=0$ (a) and $e_5^{\top}\theta_{\infty}=1$ (b). In addition, like the case when $c$ is small (Figure \ref{fig:ASGD_n1000_smallc_nonconvergence}), there is systematic bias: in the histograms of $10^4$ replications, shown in plots (c) and (d), which both exclude the largest 1\% outliers, the mean ASGD estimates for $e_1^{\top}\theta_{\infty}=0$ and $e_5^{\top}\theta_{\infty}=1$ are respectively $4.296$ and $4.687$.} 
\label{fig:ASGD_n1000_largec_nonconvergence}
\end{figure}

\begin{figure}[H]
    \centering
    \begin{subfigure}[b]{0.475\textwidth}
        \centering
        \includegraphics[width=\textwidth]{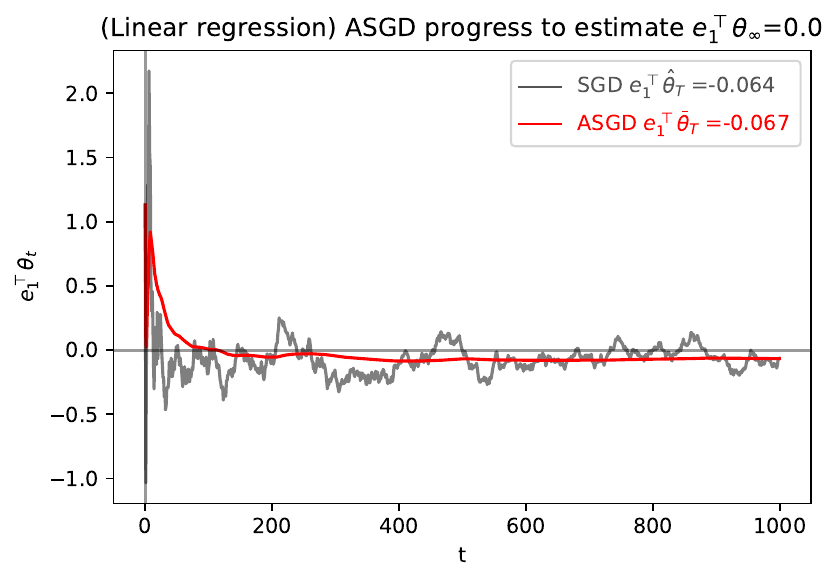}
        \caption[]%
        {{\footnotesize Single trajectory of ASGD and SGD estimates for $e_1^{\top}\theta_{\infty}=0$}}    
        \label{fig:ASGD_n1000_goodc_theta1}
    \end{subfigure}
    \hfill
    \begin{subfigure}[b]{0.475\textwidth}  
        \centering 
        \includegraphics[width=\textwidth]{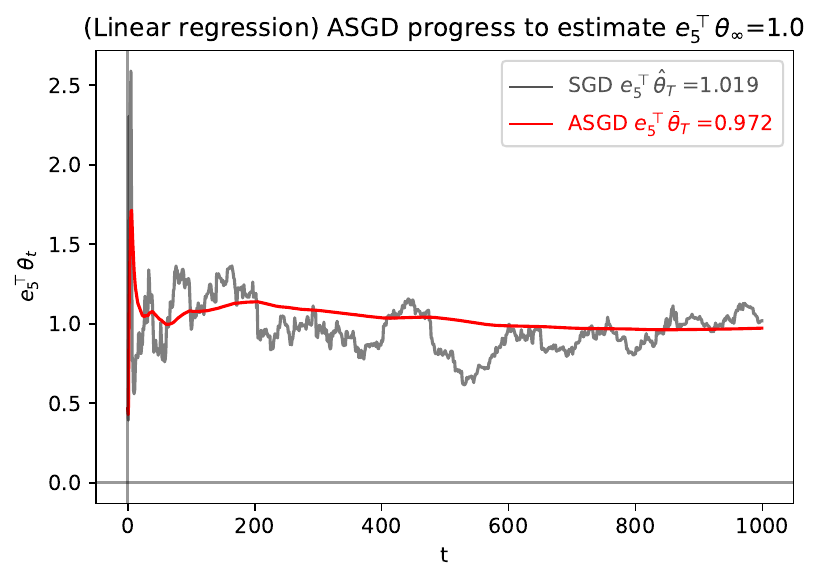}
        \caption[]%
        {{\footnotesize Single trajectory of ASGD and SGD estimates for $e_5^{\top}\theta_{\infty}=1$}}    
        \label{fig:ASGD_n1000_goodc_theta5}
    \end{subfigure}
    \vskip\baselineskip
    \begin{subfigure}[b]{0.475\textwidth}   
        \centering 
        \includegraphics[width=\textwidth]{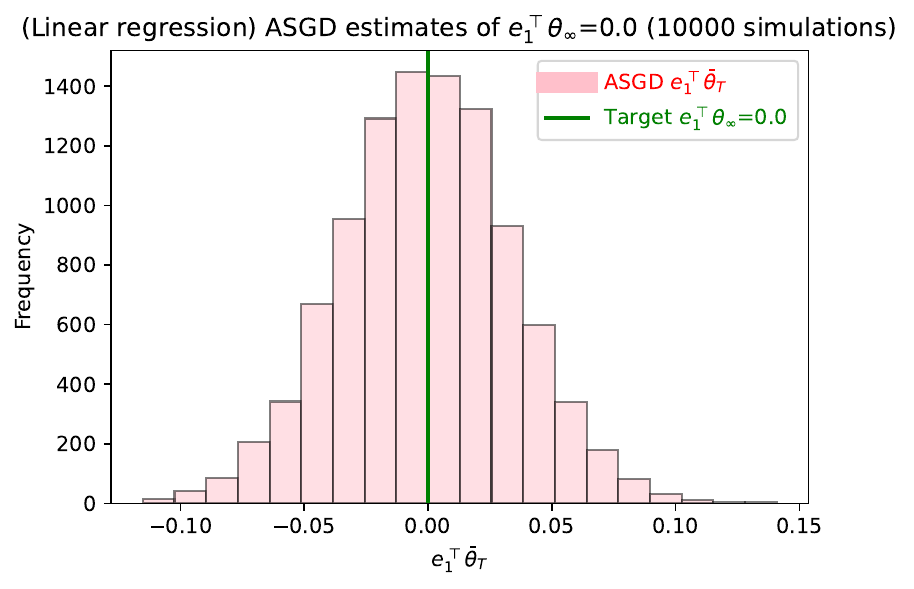}
        \caption[]%
        {{\footnotesize Histogram of ASGD estimates for $e_1^{\top}\theta_{\infty}=0$ \newline  ($10^4$ replications)}}    
        \label{fig:ASGD_n1000_goodc_theta1_hist}
    \end{subfigure}
    \hfill
    \begin{subfigure}[b]{0.475\textwidth}   
        \centering 
        \includegraphics[width=\textwidth]{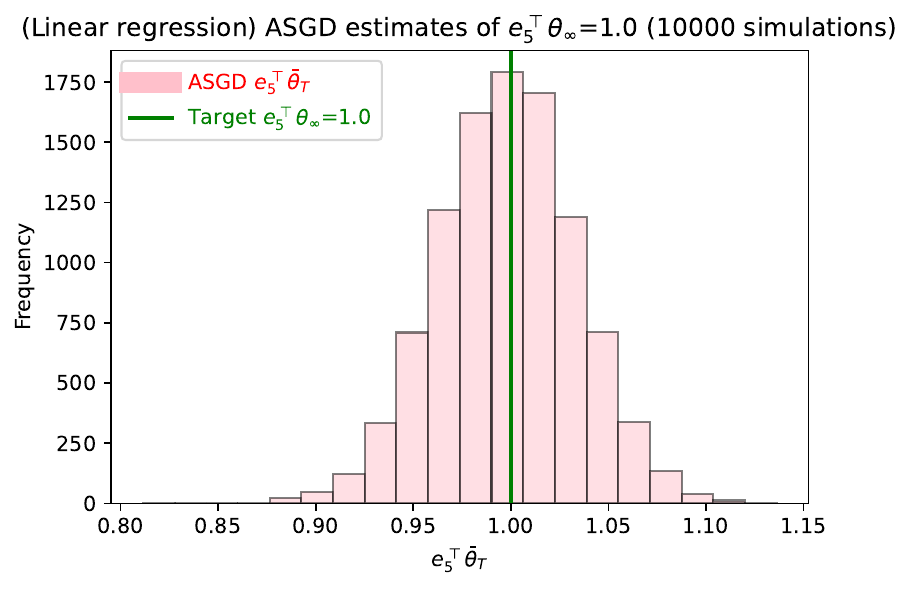}
        \caption[]%
        {{\footnotesize Histogram of ASGD estimates for $e_5^{\top}\theta_{\infty}=1$ \newline  ($10^4$ replications)}}    
        \label{fig:ASGD_n1000_goodc_theta5_hist}
    \end{subfigure}
    \caption
    {\small For the linear regression task with identity covariance, $d=5$, $T=10^3$, and a \textbf{``mid-range'' step size hyperparameter} $c=0.5$, both the SGD and ASGD estimators tend to converge. The ASGD estimator appears unbiased: in the histograms of $10^4$ replications in plots (c) and (d), which both exclude the largest 1\% outliers, the mean ASGD estimates for $e_1^{\top}\theta_{\infty}=0$ and $e_5^{\top}\theta_{\infty}=1$ are respectively $-0.0005$ and $0.998$.} 
\label{fig:ASGD_n1000_goodc_nonconvergence}
\end{figure}

As mentioned in our discussion above, the ASGD algorithm is sensitive to step size  $\eta_t = ct^{-\gamma}$ with hyperparameters $c$ and $\gamma$. While $\gamma$ is the crucial parameter for theoretical analysis ($\gamma \in (1/2, 1)$), the hyperparameter $c$ plays a more prominent role in practical performance. \cite{Chen_SGD_2021} use $\gamma = 0.501$, and we use $\gamma = 0.505$. If $c$ is too small, the ASGD algorithm generally does not converge, leading to systematic bias (see Figure \ref{fig:ASGD_n1000_smallc_nonconvergence}). If $c$ is too large, the ASGD estimate is completely off target, again showing systematic bias (see Figure \ref{fig:ASGD_n1000_largec_nonconvergence}). In both cases, the confidence interval produced using the plug-in variance estimator $\widetilde{J}_T^{-1} \widetilde{V}_T \widetilde{J}_T^{-1}$ does not cover the true parameter. Hence, there tends to be a particular ``basin of attraction'' of hyperparameter $c$ for which ASGD performs well, which is specific to the particular dataset and loss function. In our linear regression setup with $d=5$, $T=10^3$, and identity covariance, setting $c=0.5$, we found that ASGD achieved the desired $95\%$ coverage (see Figure \ref{fig:ASGD_n1000_goodc_nonconvergence}).

This systematic bias may be the result of our initialization choice, the zero vector; this initialization is closer to the truth for some coordinates and farther for some other coordinates. As can be expected, initialization closer to the truth can help the convergence but may not always be practical. \cite{Chen_SGD_2021} (in a personal communication) suggested running the SGD for a few steps with a fixed (constant) step size and using that as an initialization with decreasing step sizes. Formally, we start with $\widetilde{\theta}^{(0)}=0$, and run SGD with a fixed step size of $\eta_t = 0.001$ for $\lfloor T/3\rfloor$ steps, producing our true initial value ${{\theta}}^{(0)}={\widetilde{\theta}}^{(\lfloor T/3 \rfloor)}$, before the ASGD loop begins.

\subsection{Simulations for linear regression setting}\label{subs:lin_reg_ASGD}

In our simulations, linear regression is the easiest of the two modeling tasks. The Wald interval demonstrated correct coverage (i.e., approximately $95\%$) across all coordinates and sample sizes, dimensions, and covariance schemes, making it a useful baseline. For the other three methods, the choice of the step size hyperparameter $c$ has a huge impact on both coverage and confidence interval widths.  In particular, the ASGD plug-in confidence interval consistently demonstrates poor coverage, while both the HulC and the $t$-stat methods generally produce correct coverage for appropriately chosen $c$. Although the confidence interval of the ASGD plug-in is typically narrower than those of the HulC and the $t$-stat methods, the width ratios for $t$-stat and HulC are not excessively large when 
$c$ is appropriately chosen. Moreover, as the sample size increases, the ratios generally decrease, aligning with the intuition that larger datasets yield more precise confidence intervals.

%d larger
As the dimension $d$ increases, the coverage and width ratios are considerably more sensitive to the step size hyperparameter $c$ and the sample size $T$ compared to the low-dimensional case ($d=5$). Of the three methods, HulC is the most flexible with respect to $c$ both in terms of coverage and width. The ASGD plug-in interval typically undercovers the target while the $t$-stat method sometimes overcovers.

\begin{figure}[H]
\centering
 % \par\medskip
\includegraphics[width=1\textwidth]{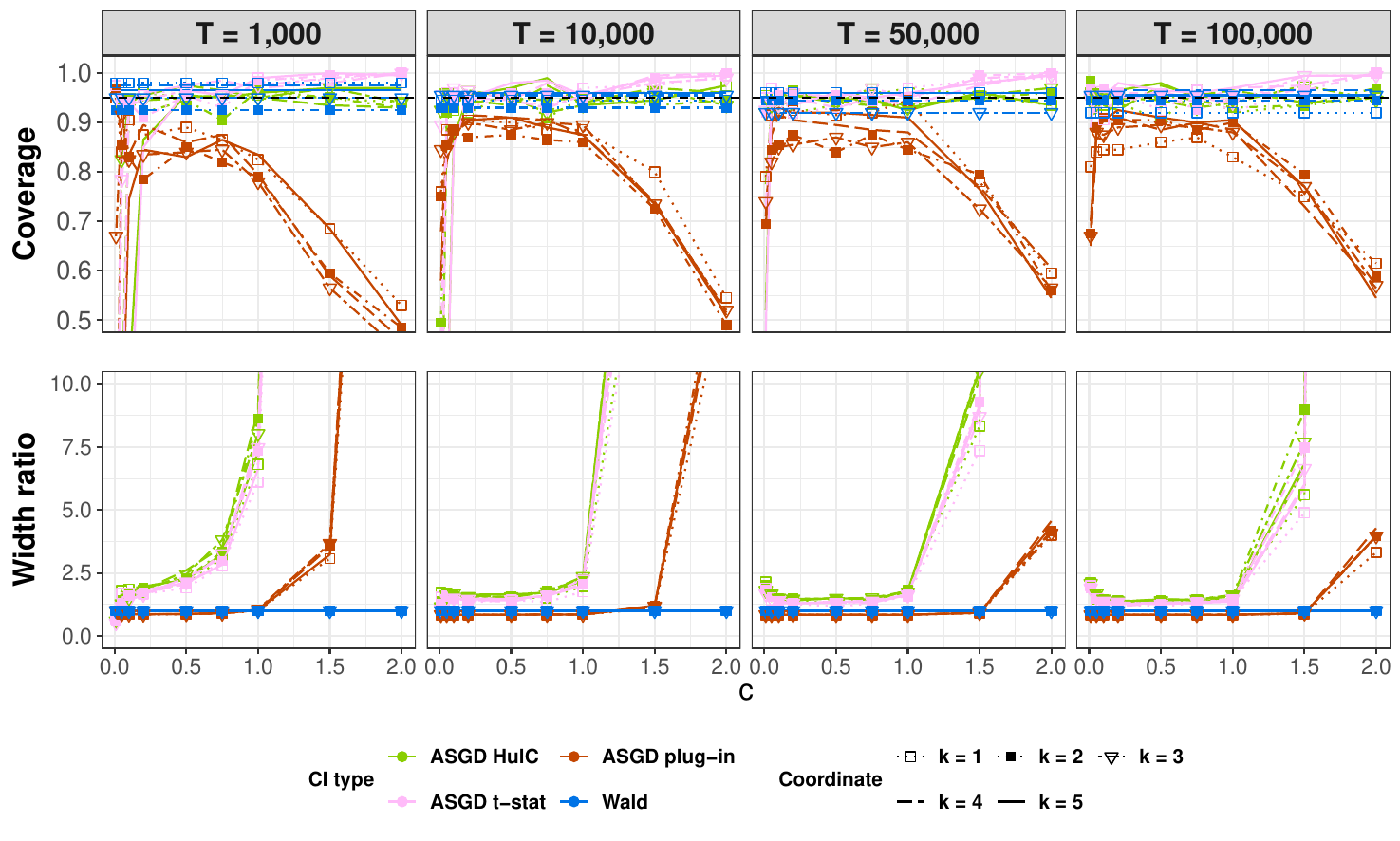}
\caption{Comparison of Wald, AGSD plug-in, HulC, and $t$-stat methods in the linear regression setting with a Toeplitz covariance structure and dimension $d=5$. The ASGD plug-in confidence interval consistently demonstrates poor coverage, while both the HulC and the $t$-stat methods generally produce correct coverage for appropriately chosen $c$. Meanwhile, the width ratios for $t$-stat and HulC are not excessively large when $c$ is appropriately chosen; as the sample size $T$ increases, the ratios decrease.}
\label{fig:linear_D5_Toeplitz_cov_wr}
\end{figure}

\subsubsection{Small dimension ($d=5$)}

We begin by examining the simplest scenario where the dimension is relatively small ($d = 5$).  Figure \ref{fig:linear_D5_Toeplitz_cov_wr} summarizes the median coverage and width ratios across $200$ replications for linear regression under Toeplitz covariance. We first focus on the top row, which displays four plots, each corresponding to a different sample size $T$, illustrating the median coverage for each coordinate $k \in \{1, \dots, 5\}$.  By varying the step size $c$ (x-axis), we compare the coverage for each confidence interval type (``CI type''), distinguished by different colors.

Note that the Wald interval has correct coverage (i.e., approximately $95\%$) across all coordinates and sample sizes. Recalling that the Wald interval is not a function of $c$, its coverage is therefore constant with respect to $c$. Furthermore, sample size $T$ does not have a significant effect on the coverage of the Wald interval: instead, coverages range from $92.0\%$ to $98.0\%$ depending on the coordinate $k$. Similarly, when the covariance matrix is the identity (Figure \ref{fig:linear_D5_I_cov_wr}) or equicorrelation (Figure \ref{fig:linear_D5_EquiCorr_cov_wr}), the Wald interval again has approximately $95\%$ coverage, making it a reliable baseline to compare the other three techniques with.

In contrast, the ASGD plug-in confidence interval consistently demonstrates poor coverage. Under Toeplitz covariance (Figure \ref{fig:linear_D5_Toeplitz_cov_wr}), its coverage is highest for $c \in [.25, 1]$, but still tends to significantly undercover the target: in this ``optimal'' range of $c$, the maximum coverage across the five dimensions is around $89\%$ for all $T\ge 10^4$. We attribute this low coverage to the tendency of the ASGD estimator to deviate from the target ${\theta}_{\infty}$, given that the ASGD plug-in confidence interval is not sufficiently wide to encompass ${\theta}_\infty$, as demonstrated in Figures \ref{fig:ASGD_n1000_smallc_nonconvergence} and \ref{fig:ASGD_n1000_largec_nonconvergence}. It could be that a finer grid of $c$ values would increase the ASGD plug-in estimator coverage, highlighting that the exact ``ideal'' $c$ is data- and model-dependent. There is one exception: when $T=10^3$ and $c$ is small, the ASGD plug-in estimator has ostensibly high coverage for coordinates $k=1$ and $k=2$. This appears to be a fluke result of our particular initialization procedure in the ASGD algorithm, as described in section \ref{subs:asgd_sensitivity}. Therefore, we do not emphasize the simulation results for small $T$ and small $c$.

% Meanwhile, the ASGD plug-in estimator has wildly unstable coverage with respect to $c$, with a maximum \textit{average} coverage across the five dimensions of between $88.8\%$ and $89.9\%$ achieved when $c \in [0.2, 0.5]$ for all $T\ge 10^4$; but even in this ``Goldilocks'' range of $c$,  the maximum ASGD plug-in coverage is $92.5\%$ (achieved for $c=0.2$, $T=10^5$, $k=5$). It could be that a finer grid of $c$ values would increase the ASGD plug-in estimator coverage, noting that the exact ``ideal'' range of $c$ is hard to find. Outside this range of $c$, the ASGD plug-in estimator's coverage is generally much lower

Unlike the ASGD plug-in estimator, both the HulC and the $t$-stat methods generally produce \textit{correct} coverage for certain values of $c$. For HulC, the empirical coverage is in the ``ballpark'' (between $93.5\%$ and $96.5\%$) for most coordinates $k$, for all $c \ge 0.2$ and all $T$. For example, at $c=0.5$,  HulC coverage is reasonably close to the nominal $95\%$ target (ranging from $90.5\%$ for $k=1$ and $T=10^5$ to $98.0\%$ for $k=5$ and $T=10^5$).  The $t$-stat interval produces the same desired ballpark coverage for most coordinates $k$ and $c \in [0.2, 0.5]$, but has over-coverage for $c>0.5$---for example, when $c=1$ and $T=10^3$, the $t$-stat coverage is at least $97\%$ for all coordinates $k$, and the coverage shoots up to nearly $100\%$ when $c=2$. Such overcoverage property of the $t$-stat method was also highlighted in~\cite{zhu_paralelSGD_2024}

To assess the width of the confidence intervals relative to the Wald interval, the second row of Figure \ref{fig:linear_D5_Toeplitz_cov_wr} presents the median width ratios across $200$ replications for linear regression under Toeplitz covariance and $d=5$. The ``width ratio'' is the width of the ``CI type'' divided by the width of the Wald interval, fixing a particular coordinate $k \in \{1, \dots, 5\}$ and sample size $T$. When ``CI type'' is ``Wald'', the ratio is $1$, which we plot for reference. 

In an ideal scenario, the three $c$-dependent ratios (ASGD plug-in, HulC, and $t$-stat) would not greatly exceed $1$---that is, the proposed methods ideally would not produce wider confidence intervals relative to the Wald. All else equal, we prefer \textit{smaller} confidence intervals so that the range of candidates for the unknown ${\theta}_{\infty}$ is tight.

For all values of $c$ and sample size $T$ under Toeplitz covariance, the ASGD plug-in interval exhibits a smaller width ratio compared to the HulC and the $t$-stat methods. However, the $t$-stat and the HulC width ratios are not \textit{egregiously} large for appropriately chosen $c$.
When $c\le .5$, all four ratios are less than $2.5$ uniformly across any sample size $T$. In addition, as $T$ increases, the ratios tend to shrink across all coordinates $k$ -- supporting the intuition that more data leads to greater precision in the estimate. When $c$ is too large, the ratios for the three $c$-dependent methods ``blow up": in Figure \ref{fig:linear_D5_Toeplitz_cov_wr}, the ratios appear as near vertical lines when $c \ge 1$ for the HulC and the $t$-stat methods and when $c>1.5$ for the ASGD plug-in method. In these cases, the width ratios surpass $10^3$. We attribute this to the systematic bias of the ASGD, as illustrated in Figure \ref{fig:ASGD_n1000_largec_nonconvergence}.
% : since both HulC and the $t$-stat confidence intervals are dependent on a relatively small set of $B^* \in \{5, 6\}$ ASGD plug-in estimators $\{\bar{\boldsymbol{\theta}}_{n,j}\}_{j=1}^{B^*}$, where each estimator fails to convergence for large $c$, then the confidence intervals produced tend to be wide. 
% %\begin{cmt}
%     Are we sure ASGD has zero median bias for all values of $c$? It seems we showed this is not true empirically.
% %\end{cmt} 

While the HulC intervals typically exhibit the highest width ratios compared to the ASGD plug-in and the $t$-stat intervals, the difference is minimal. For width ratios less than $2.5$, the maximum difference in width ratio is $0.31$ between HulC ($2.45$) and the $t$-stat ($2.14$) methods. On the other hand, the ASGD plug-in interval has width ratios near $1$ (``ideal'') for most values of $c$ and all $T$ (Figure \ref{fig:linear_D5_Toeplitz_cov_wr}), but its corresponding coverage is unsatisfactory (less than $90\%$). For both the HulC and the $t$-stat intervals, the optimal balance between a low width ratio ($\le 2.5$) and accurate coverage seems to be achieved when $c\approx 0.5$ across all sample sizes in this particular linear data simulation under the Toeplitz covariance scheme, recalling that $d=5$.

The results are similar for the other two covariance schemes, identity and equicorrelation (see Figures~\ref{fig:linear_D5_I_cov_wr} and \ref{fig:linear_D5_EquiCorr_cov_wr} in Section~\ref{app:OLS_plots}). 
% The HulC intervals achieves the correct coverage rate for all $c \geq 0.5$, the $t$-stat intervals maintains correct coverage for $c \le 1.0$ but overcovers for larger $c$, and the ASGD plug-in method consistently undercovers the target across all $c$.  Width ratios are small ($\le 2.5$) across all CI types and $T$ when $c\le 0.5$. In general, the ``basin of attraction for $c$'' for which coverage is achieved is $c \in [.2, .5]$ for the HulC and the $t$-stat methods. The HulC intervals exhibits a width ratio nearly identical to that of the $t$-stat method across all values of $c$ and $T$. Among the covariance structures, the identity matrix consistently yields the lowest overall width ratios across all $c$ and $T$ compared to the Toeplitz and equicorrelation cases. This downward shift in width ratio curves for the identity covariance may be attributed to the reduced number of nuisance parameters involved in estimating $\boldsymbol{\theta}_{\infty}$.

\subsubsection{Larger dimensions ($d= 20, 100$)}

In higher-dimensional linear regression scenarios, our simulations reveal that the coverage and width ratio outcomes are considerably more sensitive to the step size hyperparameter $c$ and the sample size $T$ compared to the low-dimensional case ($d=5$). For $d=20$, only small values of $c$ ($c < 0.375$) and larger sample sizes produce reasonable width ratios---less than $2.5$---for all covariance schemes (Figures \ref{fig:linear_D20_I_cov_wr}, \ref{fig:linear_D20_EquiCorr_cov_wr}, and \ref{fig:linear_D20_Toeplitz_cov_wr}). Coverage is likewise sensitive to $c$ and $T$, with ``ballpark'' coverage achieved when $c \in [0.125, 0.25]$ for the HulC and the $t$-stat methods, while the ASGD plug-in estimator systematically undercovers the parameter across all coordinates $k$.

In very high dimensions ($d=100$), these hyperparameters become even more restrictive: achieving width ratios below $2.5$ requires a large sample size ($T \geq 5 \cdot 10^4$), and the range of permissible values of $c$ narrows further to $c \in [0.005, 0.1]$. In this range, both the HulC and the $t$-stat methods achieve ballpark coverage. As before, the HulC width ratios are generally comparable to the $t$-statistic ratios across all covariance structures. In other words, the HulC intervals achieve coverages comparable to the $t$-stat intervals with only a minimal increase in width ratios, provided that $c$ is well-chosen and $T$ is sufficiently large. However, when $c$ is too large, the width ratios across all methods become excessively high. In such cases, only HulC maintains correct coverage, while the ASGD plug-in estimator undercovers, and the $t$-stat intervals tend to overcover the true parameter. For instance, when $c = 0.25$ and $T = 5 \cdot 10^4$, all width ratios exceed $16$, with ASGD plug-in coverage ranging from $60.0\%$ to $74.0\%$, $t$-statistic coverage ranging from $97.5\%$ to $100\%$, and HulC coverage ranging from $93.0\%$ to $96.5\%$. Furthermore, if $c$ becomes excessively large ($c > 0.3$), all methods fail, likely due to divergence in the ASGD algorithm.

\subsection{Simulations for logistic regression setting}\label{subs:log_reg_ASGD}

The empirical findings for the logistic regression setting are similar to those in the linear regression setting. However, the Wald interval frequently proves to be unsuitable as a baseline for comparison when $T$ is small and $d$ is large: in all simulations under equicorrelation covariance (and in half of the simulations under Toeplitz covariance) with $T=10^3$ and $d=100$, the data matrix is ill-conditioned, rendering the Wald interval uncomputable in these cases.

Similarly to the linear regression setting, the ASGD plug-in confidence interval consistently undercovers the target in our logistic regression setting. When dimension size $d$ is small/moderate ($d\in\{5,20\}$), the HulC and the $t$-stat intervals have correct coverage if $c$ is appropriately chosen --- but this task becomes much harder for larger dimensions ($d=100$). Furthermore, if $d$ is small/moderate, both the HulC and the $t$-stat intervals have width ratios less than $2$ for all sample sizes and covariance types. When $d=100$, sometimes only one value of $c$ across the grid of tested values achieves ballpark coverage and small width ratios across most coordinates of ${\theta}_{\infty}$, and this ``best'' value of $c$ fluctuates greatly depending on the sample size $T$ and covariance type. As expected, width ratios generally decrease as the sample size $T$ increases. There is no discernible pattern in the width ratios as the covariance type changes. 

HulC proves to be the most adaptable method with respect to $c$, both in terms of coverage and interval widths. Notably, in high-dimensional settings ($d=100$), there are cases where the $t$-stat fails to achieve coverage, but HulC succeeds. Conversely, there are no instances where the $t$-stat achieves coverage while the HulC does not. For example, when $d=100$, $T=5\cdot10^4$, under Toeplitz covariance (see Figure~\ref{fig:logistic_D100_Toeplitz_cov_wr}),  and $c=0.75$, the median HulC coverage is 94.5\% across the coordinates (and a median width ratio of $2.6$), whereas the median $t$-stat coverage is 89.5\% (and a median width ratio of $1.8$).

% \begin{figure}[H]
% \centering
%  \caption{ Logistic regression, Covariance = Toeplitz, d = 5}\par\medskip
% \includegraphics[width=1\textwidth]{figures/logistic_D5_Toeplitz_cov_wr}\label{fig:logistic_D5_Toeplitz_cov_wr}
% \end{figure}

\begin{figure}[H]
\centering
 % \par\medskip
\includegraphics[width=1\textwidth]{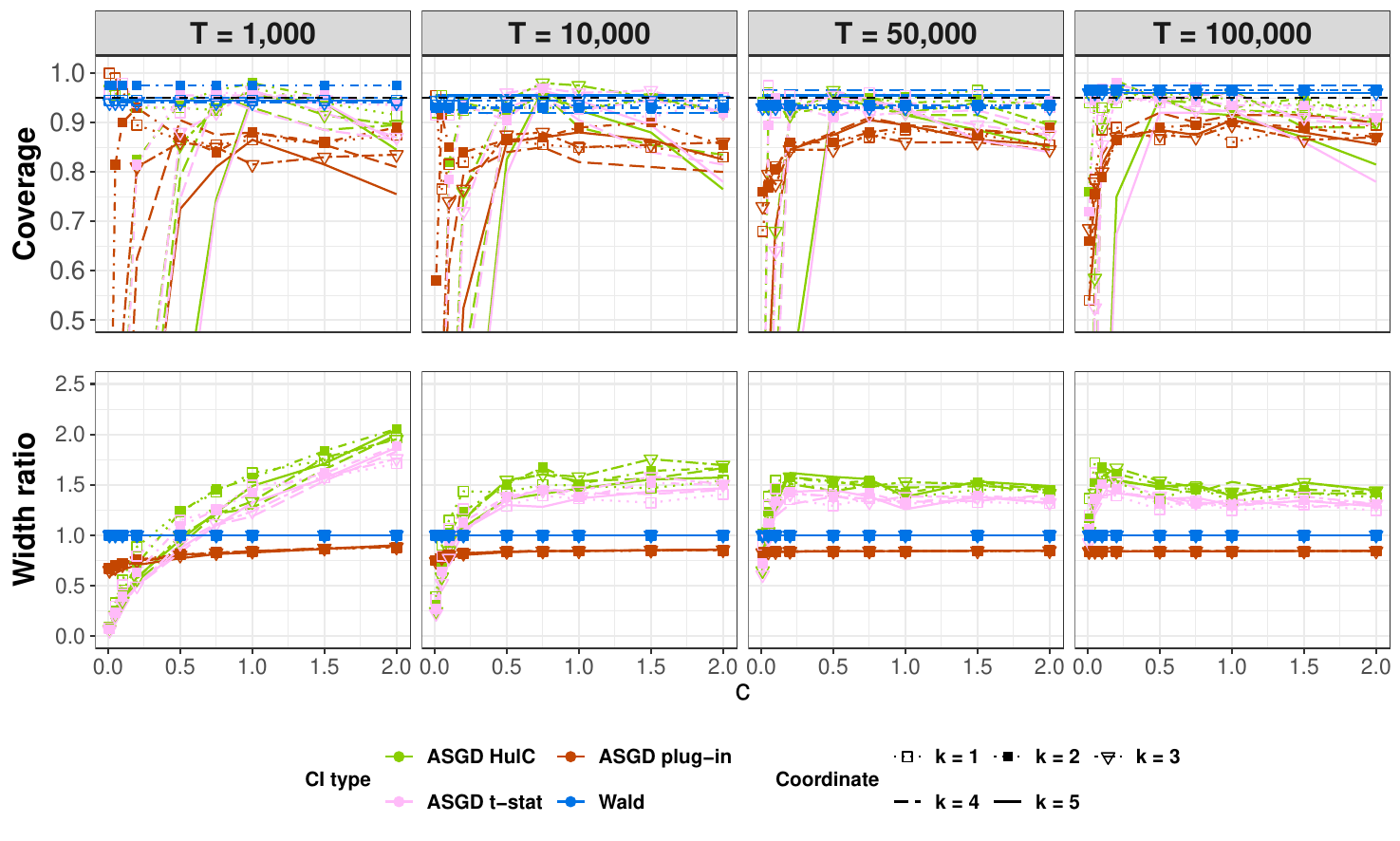}
\caption{Comparison of Wald, AGSD plug-in, HulC, and $t$-stat methods in the logistic regression setting with a Toeplitz covariance structure and dimension $d=5$. Like the linear regression case (Figure \ref{fig:linear_D5_Toeplitz_cov_wr}), the ASGD plug-in confidence interval consistently demonstrates poor coverage, while the HulC and $t$-stat methods produce correct coverage for appropriately chosen $c$. The width ratios for $t$-stat and HulC are not excessively large when $c$ is appropriately chosen; as the sample size $T$ increases, the ratios decrease.}
\label{fig:logistic_D5_Toeplitz_cov_wr}
\end{figure}

\subsubsection{Small dimension ($d=5$)}

When the dimension size is small ($d=5$),  Figure \ref{fig:logistic_D5_Toeplitz_cov_wr} summarizes the median coverage and width ratios across $200$ replications for logistic regression under Toeplitz covariance. Focusing on the top row, we can see that the Wald interval has correct coverage (approximately $95\%$) across all coordinates $k$ and sample sizes. Similarly, when the covariance matrix is identity (Figure \ref{fig:logistic_D5_I_cov_wr}) or equicorrelation (Figure \ref{fig:logistic_D5_EquiCorr_cov_wr}), the Wald interval again has approximately $95\%$ coverage. This observation again justifies our use of the Wald interval as a baseline upon which to compare the other $c$-dependent methods. 

We find that for logistic regression, both the HulC and the $t$-stat intervals achieve \textit{correct} coverage for $c$ in the basin of attraction, although the basin of attraction is narrower than in the linear regression case. Under Toeplitz covariance, Figure \ref{fig:logistic_D5_Toeplitz_cov_wr} reveals that for both methods, empirical coverages are in the ``ballpark'' (between $93.5\%$ and $96.5\%$) for most coordinates $k$ when $c=1$ and $T=10^3$; for larger sample sizes, ballpark coverage is achieved when $c \in [0.5, 0.75]$. Outside this range, both the HulC and the $t$-stat intervals tend to undercover the target, with coverage levels similar to those of the ASGD plug-in estimator. In contrast to linear regression (Figure~\ref{fig:linear_D5_Toeplitz_cov_wr}), our simulations reveal that for the logistic regression setting, the HulC and the $t$-stat coverages fluctuate more noticeably across the coordinates $k$ outside the basin of attraction, as illustrated by the vertical dispersion of green and pink lines in Figure~\ref{fig:logistic_D5_Toeplitz_cov_wr}.

The ASGD plug-in confidence interval persistently undercovers the target in the logistic regression setting. For the Toeplitz covariance in Figure~\ref{fig:logistic_D5_Toeplitz_cov_wr}, we observe that when the sample size is small ($T=10^3$), the maximum coverage for the ASGD plug-in confidence interval is $90.5\%$. For the largest sample size $T=10^5$, the maximum coverage for the ASGD plug-in confidence interval is $92\%$. Unlike the findings in the linear regression setting (Figure~\ref{fig:linear_D5_Toeplitz_cov_wr}), we observe that the coverage of the ASGD plug-in confidence interval is above $80\%$ for all $c \ge .75$.

Focusing on the bottom row of Figure \ref{fig:logistic_D5_Toeplitz_cov_wr}, we observe that the width ratios are reasonably stable over $c$ in the case of logistic regression for dimension $d=5$ with Toeplitz covariance. Here, almost all width ratios are less than $2$ for all $c\in [0.0005, 2]$ and all sample sizes $T$. The same phenomenon is observed for the identity (Figure \ref{fig:logistic_D5_I_cov_wr}) and equicorrelation (Figure \ref{fig:logistic_D5_EquiCorr_cov_wr}) covariance types. As in the case of linear regression, ratios tend to shrink as $T$ increases. The HulC width ratio is the highest across all methods, but only slightly higher than the $t$-stat ratio.

\subsubsection{Larger dimensions ($d=20,100$)}

When the dimension is high ($d=20, 100$), our simulations reveal that the coverage and width ratios are extremely sensitive to the step size hyperparameter $c$ and the sample size $T$, compared to the low-dimensional case ($d=5$). Although this was also observed in the linear regression setting, logistic regression exhibits even greater variability across the coordinates for all $c$-dependent methods in higher dimensions, especially with respect to $T$.

When $d=20$ (Figures \ref{fig:logistic_D20_I_cov_wr}, \ref{fig:logistic_D20_EquiCorr_cov_wr}, and \ref{fig:logistic_D20_Toeplitz_cov_wr}), the Wald interval consistently achieves correct coverage across all sample sizes and coordinates, reaffirming its suitability as a baseline method. As previously observed, the ASGD plug-in method continues to undercover the target for the majority of coordinates. For the HulC and the $t$-stat methods, the coverage is optimized within a narrower range of $c$ compared to the low-dimensional setting. Similar results hold for the equicorrelation (Figure \ref{fig:logistic_D20_EquiCorr_cov_wr}) and the Toeplitz (Figure \ref{fig:logistic_D20_Toeplitz_cov_wr}) covariance schemes. Furthermore, compared with linear regression with $d=20$ (Figures \ref{fig:linear_D20_I_cov_wr}, \ref{fig:linear_D20_EquiCorr_cov_wr}, and \ref{fig:linear_D20_Toeplitz_cov_wr}), the ``valid'' range of $c$ (in terms of coverage) is significantly narrower for logistic regression across all covariance schemes. This suggests that, in practice, selecting an appropriate $c$ is more challenging for logistic regression in higher dimensions than for linear regression, at least within the scope of our simulations.

When $d=100$ for logistic regression (Figures \ref{fig:logistic_D100_I_cov_wr}, \ref{fig:logistic_D100_EquiCorr_cov_wr}, and \ref{fig:logistic_D100_Toeplitz_cov_wr}), correct coverage is harder to achieve. The Wald interval cannot be computed for $T=10^3$ under equicorrelation covariance (Figure \ref{fig:logistic_D100_EquiCorr_cov_wr}) due to the ill-conditioning of the data matrix. In this case, Wald intervals cannot serve as a baseline. For the Toeplitz covariance (Figure \ref{fig:logistic_D100_Toeplitz_cov_wr}), in about half ($98$ of $200$) of the experiments, the same conditioning issues occur when $T=10^3$. Even when the Wald interval is computable, coverage rates are poor for $T=10^3$. Since the $c$-dependent methods do not involve matrix inversion, they avoid this problem. 

The coverage of the ASGD plug-in method is generally the lowest in all experiments when $d=100$ for logistic regression. To visually assess the agreement of coordinate-wise coverage rates for any given $c$, we can examine the vertical spread of the red lines: in Figures \ref{fig:logistic_D100_I_cov_wr}, \ref{fig:logistic_D100_EquiCorr_cov_wr}, and \ref{fig:logistic_D100_Toeplitz_cov_wr}, the red lines show substantial vertical dispersion, even for the best-performing $c$. For instance, with the largest sample size $T=10^5$ and Toeplitz covariance (Figure \ref{fig:logistic_D100_Toeplitz_cov_wr}), the ASGD plug-in coverages range from $49.5\%$ ($k=95$) to $86\%$ ($k=2$) when $c=0.75$. In comparison, for $d=20$ and $T=10^5$ (Figure \ref{fig:logistic_D20_Toeplitz_cov_wr}), the best-performing step size $c=0.75$ resulted in coordinate-wise coverages ranging from $74.7\%$ ($k=20$) to $84.5\%$ ($k=2$). Similarly, for linear regression with $d=100$, Toeplitz covariance, and $T=10^5$ (Figure \ref{fig:linear_D100_Toeplitz_cov_wr}), coordinate-wise coverage rates ranged from $81.0\%$ ($k=98$) to $91.5\%$ ($k=91$) for the best-performing step size $c=0.1$. These results suggest that in high dimensions, the ASGD plug-in method for logistic regression can be particularly unreliable.
%\begin{cmt}
%    check reason why ASGD is performing badly for logistic regression. Is it due to a systematic bias in theta estimates?
%\end{cmt}

Similarly, the HulC and the $t$-stat methods fail to reliably achieve accurate coverage for high-dimensional ($d=100$) logistic regression, even with the best-performing $c$. Specifically, HulC coordinate-wise coverage rates remain within a reasonable range for identity and Toeplitz covariances but fall short for equicorrelation covariance. The $t$-stat method, meanwhile, does not provide adequate coverage for $d=100$ under any covariance scheme. Despite these shortcomings, both methods exhibit much tighter coordinate-wise coverage ranges compared to the ASGD plug-in method. For instance, focusing on Toeplitz covariance (Figure \ref{fig:logistic_D100_Toeplitz_cov_wr}) with $T=10^5$ and the best-performing $c=.75$, HulC coverage rates range from to $87.0\%$ ($k=84$) to $98.5\%$ ($k=25$), while $t$-stat coverage rates vary between $78.0\%$ ($k=100$) and $94.0\%$ ($k=25$).  These findings suggest that HulC could provide a more reliable approach for inference in high-dimensional logistic regression.

%\textbf{Width ratios}: 
Similar to linear regression, width ratios for logistic regression are generally larger when $d=20$ compared to $d=5$, but they are still reasonable. Except in the case of $T=10^3$, width ratios are all less than $2$ for appropriately chosen $c$ under any covariance scheme. For example, when covariance is equicorrelation (Figure \ref{fig:logistic_D20_EquiCorr_cov_wr}), width ratios are maximally $1.94$ when $c=1$ for all $T\ge 10^4$. As before, the HulC width ratio is the highest among all methods, albeit only marginally exceeding the $t$-stat ratio. Interestingly, in the case of logistic regression, the width ratios remain controlled and do not exceed $10$ for any choice of $c\in (0,2]$, unlike in linear regression. The width ratio behavior appears relatively consistent across all three covariance schemes, with no noticeable differences.

In high dimensions ($d=100$), achieving width ratios below $2.5$ requires $c \in (0, 0.75]$ for most sample sizes $T$ and covariance schemes.  But even in this range, as discussed above, the HulC and the $t$-stat methods do not reliably achieve ballpark coverage. For example, focusing on Toeplitz covariance (Figure \ref{fig:logistic_D100_Toeplitz_cov_wr}) with $T=10^5$ and the best-performing $c=.75$, where HulC achieves its highest coverage rates (ranging from $87.0\%$ to $98.5\%$), the corresponding HulC width ratios span from $1.91$ to $2.97$, with a median width ratio of $2.50$ across all $k$ coordinates. In this case, $t$-stat width ratios range from $1.33$ to $2.11$, though its coverage rates are lower than those of HulC. This indicates that while HulC typically has higher width ratios compared to $t$-stat, its coverage rates are more reliably in the ballpark in high-dimensional logistic regression.

\section{Additional SGD Methods}\label{sec:add-methods}

We run additional simulations testing four common SGD algorithms:

\begin{enumerate}
    \item Implicit-SGD~\citep{toulis2017asymptotic}:
    \[
    \theta^{(t)} = \theta^{(t-1)} - \eta_t\nabla\ell(Z_t; \theta^{(t)}) \quad \mbox{and} \quad \bar{\theta}_T = \begin{cases}
        \theta^{(T)}, &\mbox{last-iterate or}\\
        T^{-1}\sum_{t=1}^T \theta^{(t)}, &\mbox{average-iterate.}
    \end{cases}
    \]
    Here $\theta^{(t)}$ is implicitly defined.

    \item ROOT-SGD \citep{Li-Mou-Wainwright-Jordon-2020-ROOT-SGD}

    The Recursive One-Over-T SGD (ROOT-SGD) has a two-line update rule:

    \[
    \begin{aligned}
    v^{(t)} &= \nabla\ell(Z_t; \theta^{(t-1)}) + \frac{t-1}{t} (v^{(t-1)} - \nabla\ell(Z_t; \theta^{(t-2)})) \\
    \theta^{(t)} &= \theta^{(t-1)} - \eta_t v^{(t)}
    \end{aligned}
    \]

    \item Truncated-SGD~\citep{langford_truncated_2009, noisyt_zhou_2021}:

    Update $\theta^{(t)}$ using the sparse (``truncated'') gradient $\tilde{g}^{(t)}$ defined in Algorithms~\ref{alg:GT} and~\ref{algo:soft_cut} (see Section~\ref{sec:algo}) controlled by hyperparameter $\varepsilon>0$:

    \[
    \theta^{(t)} = \theta^{(t-1)} - \eta_t \tilde{g}^{(t)}
    \]
     
    \item Noisy-truncated-SGD~\citep{noisyt_zhou_2021}:
    \newline

    Update $\theta^{(t)}$ using sparse (``truncated'') gradient $\tilde{g}^{(t)}$ with additional Gaussian noise defined in Algorithm~\ref{algo:ntrunc} (see Section~\ref{sec:algo}) controlled by hyperparameters $\varepsilon>0$, $\sigma>0$, $\beta \in [0, 0.5]$:

    \[
    \theta^{(t)} = \theta^{(t-1)} - \eta_t \tilde{g}^{(t)} + \eta_t^{\frac{1}{2}+\beta} b_t, \quad \text{where } b_t \sim(N, \sigma^2 I) \text{ and } \beta \in [0, 0.5].
    \]

\end{enumerate}

All simulations use the same grid of settings described in \ref{sec:simulation}, except for $c$ (see Table~\ref{tab:c-extra} in Supplement~\ref{sec:annex_plots} for the grid of $c$ values we tested). For truncated- and noisy-truncated-SGD, we use hyperparameters $\varepsilon=0.8$, $\sigma=1$, and $\beta=.25$. Note that we use the same initialization procedure for $\theta^{(0)}$ as described in section~\ref{subs:asgd_sensitivity}. In general, the noisy-truncated-SGD algorithm produced large width ratios for the chosen range of $c$. It appears that we needed either smaller $c$ values or adjustments to hyperparameters $\varepsilon=0.8$, $\sigma=1$, and $\beta=0.25$. We present the noisy-truncated SGD plots exclusively in the online tool.\footnote{Please refer to the interactive tool available at \url{https://public.tableau.com/app/profile/selina.carter6629/viz/OnlineinferencesimulationsOLSandlogisticregression/Coverageandwidthratio_paper}.} 

\subsection{Simulation results}

\begin{figure}[H]
\centering
 % \par\medskip
\includegraphics[width=1\textwidth]{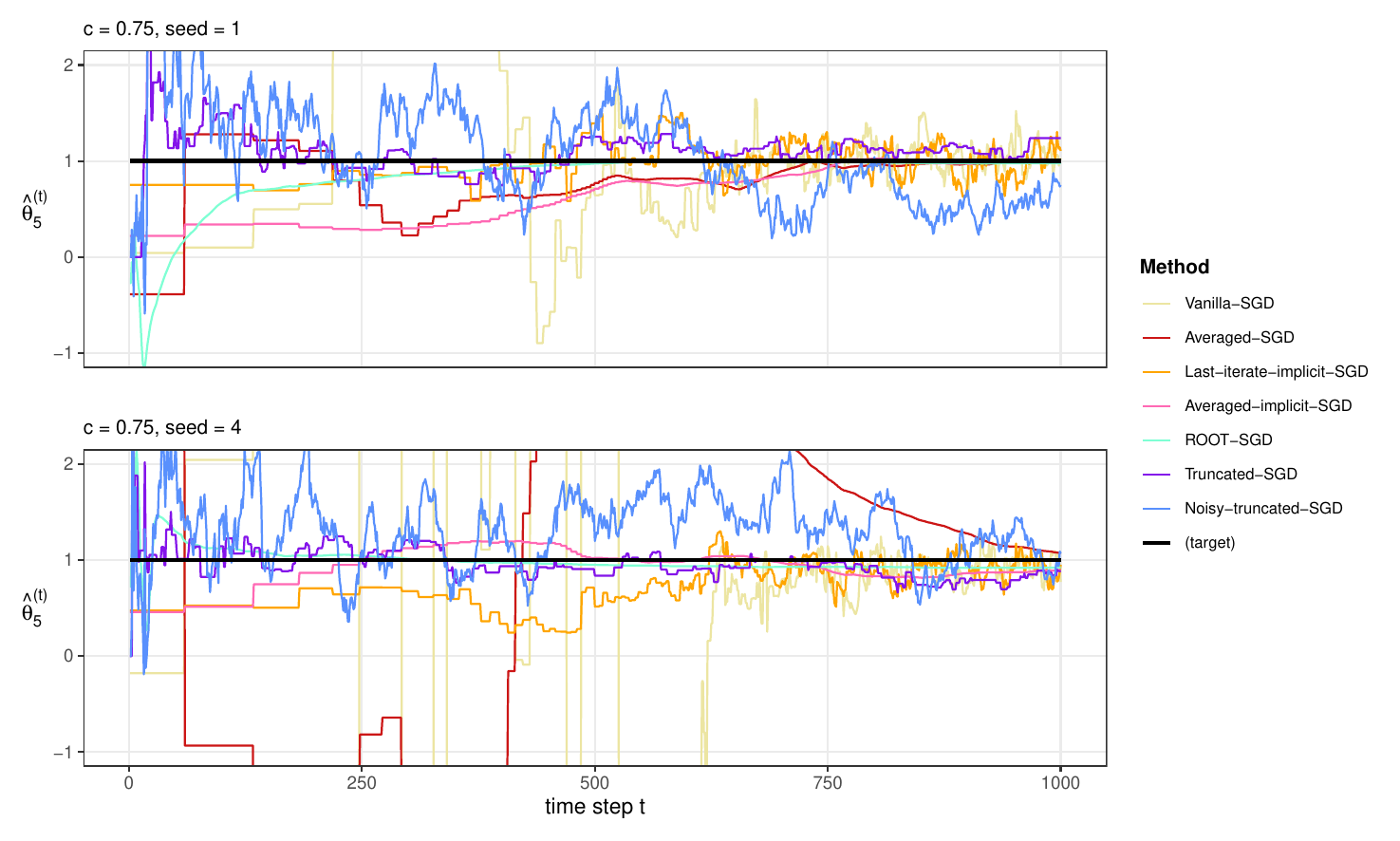}
\caption{Using step size hyperparameter $c=0.75$, we compare all SGD methods in the linear regression setting with a Toeplitz covariance structure, $T=10^3$, and dimension $d=5$, viewing coordinate $k=5$. The top panel shows the trajectories setting the randomization seed to $1$ (resulting in a common initialization $\hat{\theta}^{(0)}$) while the bottom panel has seed $4$ (and a different common initialization $\hat{\theta}^{(0)}$). In both cases, all algorithms achieve estimates that are relatively close to the target value $\theta_5 = 1.0$. The maximum distance is $0.25$ for noisy-truncated-SGD for seed=1 ($\hat{\theta}_5^{(T)}=0.75$).}
\label{fig:other_methods_examples_D5}
\end{figure}

Our findings at a high level are as follows:

\begin{itemize}
    \item \textbf{For larger dimensions, there is no single value of step size hyperparameter $c$ that works uniformly for all algorithms}: In the small dimension case ($d=5$) with OLS regression,  Figure~\ref{fig:other_methods_examples_D5} compares trajectories for a ``good'' choice of $c=0.75$; all methods generally approach the target. But when the dimension size is larger, most algorithms are more sensitive to $c$. Moreover, there is no single value of $c$ that works uniformly for all algorithms (see Figure~\ref{fig:other_methods_examples_D20} in the Appendix for examples).

    \item \textbf{The HulC and the $t$-stat methods generally produce correct coverage for appropriately chosen $c$.} HulC width ratios are only slightly larger than the $t$-stat width ratios across all settings -- around $10\%$ wider on average across all settings for which ballpark coverage is achieved.

    \item \textbf{As the sample size increases, width ratios can decrease or increase depending on the algorithm}. Width ratios decrease with sample size for ASGD and ROOT-SGD; but they increase for last-iterate-implicit-SGD, truncated-SGD, and noisy-truncated-SGD, and they are relatively stable across $T$ for average-iterate-implicit-SGD. The reasons are unclear, but in the case of last-iterate-implicit-SGD, it may be the case that drastically larger step sizes are needed as the sample size increases, unlike ASGD and ROOT-SGD.

    \item \textbf{As was the case for ASGD, linear regression is the easiest of the two modeling tasks, but with caveats.} Both in terms of achieving correct coverage and achieving small width ratios (i.e., less than $2.5$), linear regression tends to accept a wider range of $c$ than logistic regression. However, for ROOT-SGD and truncated-SGD, logistic regression has smaller width ratios.

    \item \textbf{For linear regression, the most reliable algorithm in terms of hyperparameter $c$ is average-iterate-implicit-SGD.} In this case, across all values of $c \in [0.0005, 2]$, both the HulC and $t$-stat methods produce correct coverage and width ratios are less than $1.8$. 

    \item \textbf{For linear regression, truncated- and noisy-truncated-SGD have the smallest ranges of $c$ for which the HulC and $t$-stat methods achieve reasonable width ratios.} This region is tighter as the dimension increases. Coverage is generally correct, however.

    \item \textbf{For logistic regression, achieving correct coverage is typically more challenging; but, sometimes width ratios are smaller than for linear regression.} For logistic regression, most algorithms have a relatively tighter range of $c$ for which HulC and $t$-stat target coverage is achieved, compared to linear regression. This range shrinks as dimension $d$ increases. While last-iterate-implicit-SGD is the most reliable in terms of coverage for a wide range of $c$, average-iterate-implicit-SGD has smaller width ratios for a tighter interval of $c$. Using ROOT-SGD and truncated-SGD, however, logistic regression produces smaller width ratios than linear regression.

\end{itemize}

\subsubsection{Comparison to ASGD results}

Compared to the HulC and $t$-stat intervals computed with ASGD (Section~\ref{sec:simulation}), the intervals produced by the other SGD methods show broadly similar coverage and width patterns. Since the ASGD plug-in confidence interval \citep{Chen_SGD_2021} is used only with the ASGD algorithm, here we focus solely on the HulC and $t$-stat coverage and width ratios.

First, as before, linear regression is the easiest of the two modeling tasks both in terms of achieving correct coverage and achieving small width ratios (less than around $2.5$). Logistic regression is more challenging; all algorithms have a relatively tighter range of $c$ for which HulC and $t$-stat target coverage is achieved. As was the case for ASGD, this range of $c$ shrinks as dimension $d$ increases. The same is true for noisy-truncated-SGD, which has additional hyperparameters $\varepsilon$, $\sigma$, and $\beta$.

Second, like ASGD, the choice of the step size hyperparameter $c$ has a large impact on both coverage and confidence interval widths -- but flexibility in the choice of $c$ varies by algorithm. Average-iterate-implicit-SGD is the most flexible with respect to $c$, although for small sample size ($T=10^3$) it undercovers the target if the dimension is large (see Figures \ref{fig:linear_D20_Toeplitz_cov_wr_AISGD_initTRUE} and \ref{fig:linear_D100_Toeplitz_cov_wr_AISGD_initTRUE}). Last-iterate-implicit-SGD works best for larger values of $c$, and coverage is nearly always close the target $95\%$; however, given any fixed $c$, the width ratio increases as the sample size increases, a counter-intuitive result (for $d=5$, see Figure~\ref{fig:linear_D5_Toeplitz_cov_wr_ISGD_initTRUE} and \ref{fig:logistic_D5_Toeplitz_cov_wr_ISGD_initTRUE}). This is likely to due needing a larger step size than the grid of $c$ values we tested ($c \in [0.0005, 2]$). ROOT-SGD is the only method besides ASGD that has a pronounced ``U''-shaped width ratio plot for linear regression, indicating a tight basin of attraction for optimal $c$ (for $d=5$, see Figure \ref{fig:linear_D5_Toeplitz_cov_wr_rootSGD_initTRUE}); for logistic regression, the shape is flatter (for $d=5$, see Figure \ref{fig:logistic_D5_Toeplitz_cov_wr_rootSGD_initTRUE}). Like ASGD, this basin of $c$ in which width ratios are low and coverage meets the target shrinks as the dimension size $d$ increases (see Figures \ref{fig:linear_D20_Toeplitz_cov_wr_rootSGD_initTRUE} and \ref{fig:linear_D100_Toeplitz_cov_wr_rootSGD_initTRUE} for linear regression and Figure \ref{fig:logistic_D20_Toeplitz_cov_wr_rootSGD_initTRUE} for logistic regression). Truncated-SGD, meanwhile, is also highly sensitive to $c$, but its width ratio plot rises dramatically as $c$ increases (see Figures \ref{fig:linear_D5_Toeplitz_cov_wr_truncatedSGD_initTRUE} and \ref{fig:logistic_D5_Toeplitz_cov_wr_truncatedSGD_initTRUE} for $d=5$ examples); noisy-truncated-SGD produces the same pattern. 

Third, like the ASGD simulations, the HulC and the $t$-stat methods generally produce correct coverage for appropriately chosen $c$. Moreover, as previously, HulC width ratios are only slightly larger than the $t$-stat width ratios across all settings. When coverage is ``ballpark'' (between $92.5\%$ and $97.5\%$), across all settings and coordinates $k$, the HulC width ratio exceeds that of the $t$-stat method on average by $10.1\%$ (with a median of $10.0\%$). This rate is approximately the same for both linear and logistic regression. Looking at the ``worst case'' differences between HulC and $t$-stat width ratios when coverage is ``ballpark,'' for linear regression, the HulC width ratio exceeds that of the $t$-stat method \textit{maximally} by $28.0\%$ (occurring when $T=5\cdot10^4$, $d=100$, $c=.5$, Equicorrelation covariance, and coordinate $k=6$ using noisy-truncated-SGD, resulting in a HulC width ratio of $146.0$ and $t$-stat width ratio of $114.0$). For logistic regression,  the HulC width ratio exceeds that of the $t$-stat method \textit{maximally} by $23.3\%$ (occurring when $T=10^3$, $d=100$, $c=1.0$, Toeplitz covariance, and coordinate $k=84$ using noisy-truncated-SGD, resulting in a HulC width ratio of $1.42$ and $t$-stat width ratio of $1.15$).

\subsubsection{Linear regression: additional simulations}

\begin{figure}[H]
\centering
 % \par\medskip
\includegraphics[width=1\textwidth]{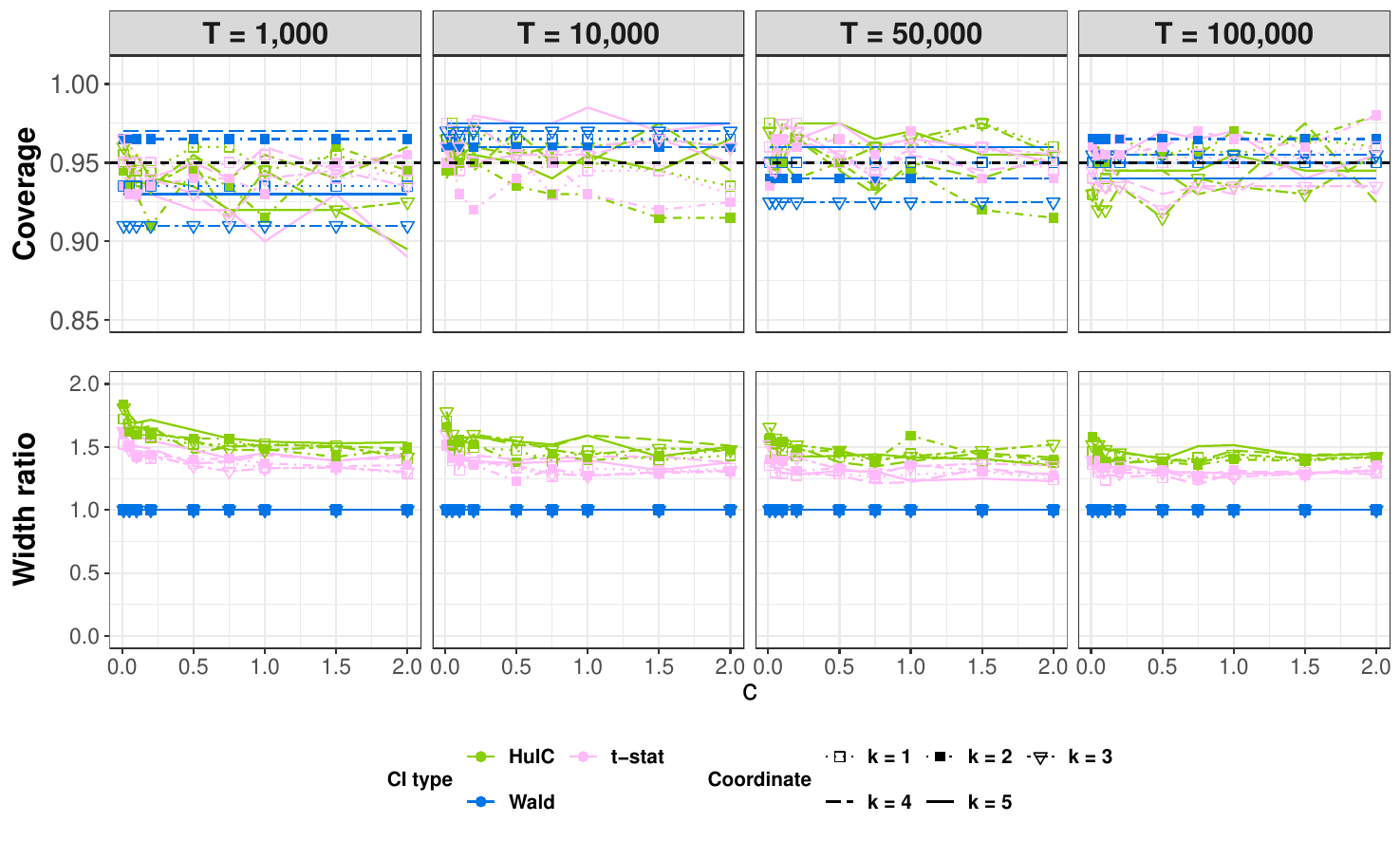}
\caption{Using average-iterate-implicit-SGD, we compare Wald, HulC, and $t$-stat methods in the linear regression setting with a Toeplitz covariance structure and dimension $d=5$. Both the HulC and the $t$-stat methods generally produce correct coverage for any value of $c$. Meanwhile, the width ratios for $t$-stat and HulC are small and do not shrink considerably as the sample size $T$ increases.}
\label{fig:linear_D5_Toeplitz_cov_wr_AISGD_initTRUE}
\end{figure}

\begin{figure}[H]
\centering
 % \par\medskip
\includegraphics[width=1\textwidth]{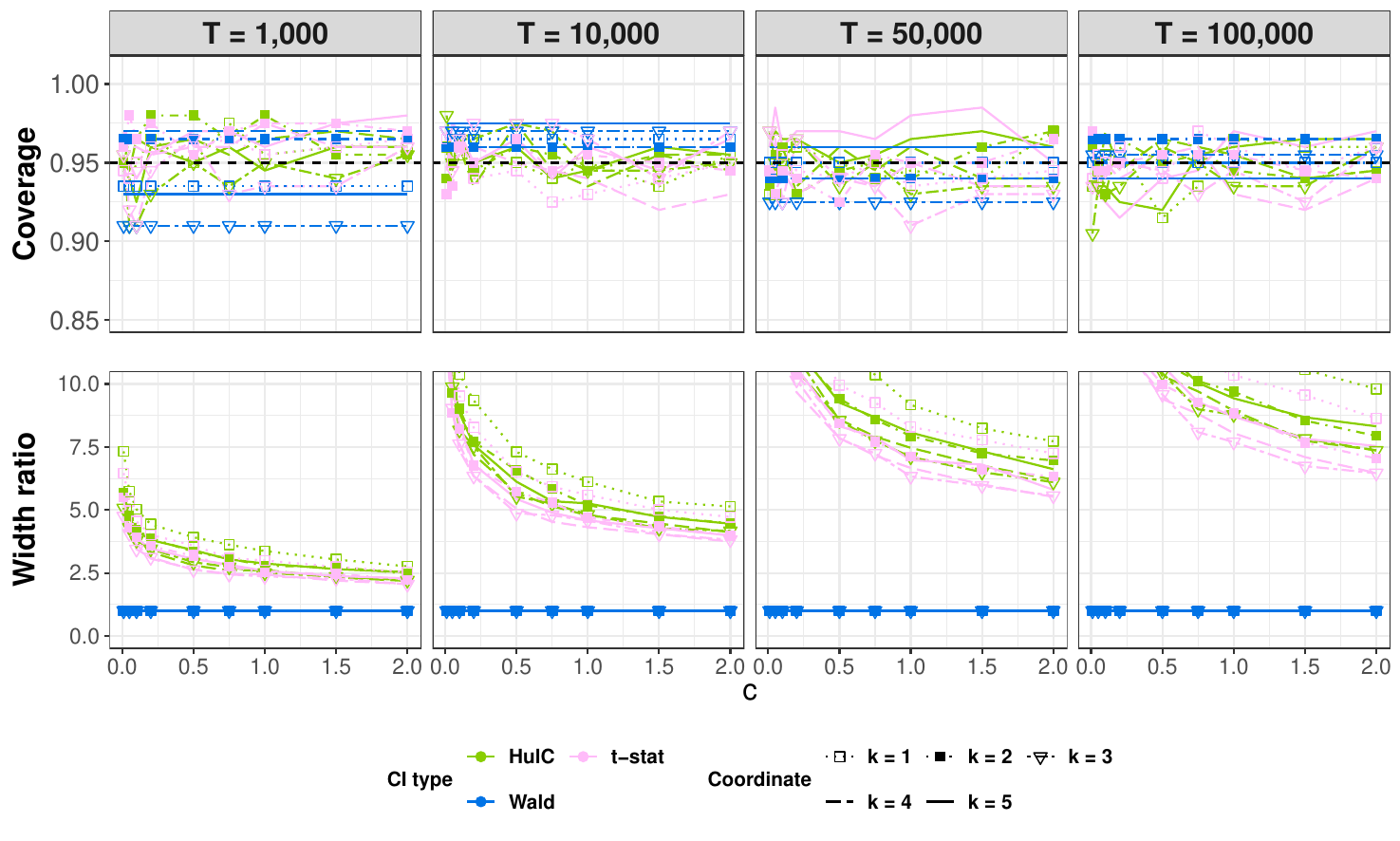}
\caption{Using last-iterate-implicit-SGD, we compare Wald, HulC, and $t$-stat methods in the linear regression setting with a Toeplitz covariance structure and dimension $d=5$. Both the HulC and the $t$-stat methods generally produce correct coverage for appropriately chosen $c$. Meanwhile, the width ratios for $t$-stat and HulC are not excessively large when $c$ is large enough. As the sample size $T$ increases, the ratios increase, requiring (possibly) larger choices of $c$.}
\label{fig:linear_D5_Toeplitz_cov_wr_ISGD_initTRUE}
\end{figure}

\begin{figure}[H]
\centering
 % \par\medskip
\includegraphics[width=1\textwidth]{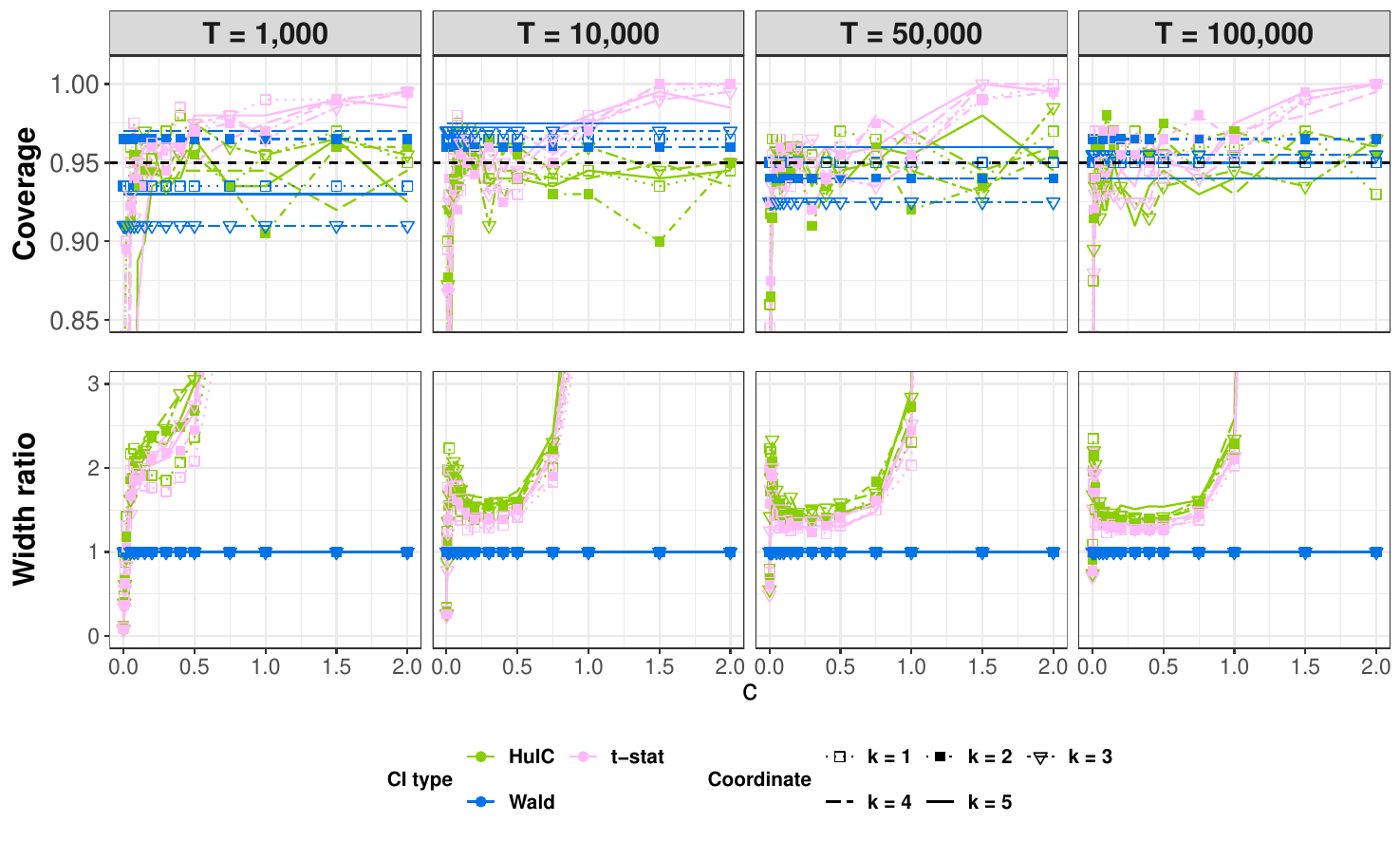}
\caption{Using ROOT-SGD, we compare Wald, HulC, and $t$-stat methods in the linear regression setting with a Toeplitz covariance structure and dimension $d=5$. Both the HulC and the $t$-stat methods generally produce correct coverage for any $c>0.1$.   Meanwhile, the width ratios for $t$-stat and HulC are not excessively large when $c$ is within the appropriate  ``basin of attraction'' centered at around $0.25$; as the sample size $T$ increases, the range of suitable values of $c$ expands.}
\label{fig:linear_D5_Toeplitz_cov_wr_rootSGD_initTRUE}
\end{figure}

\begin{figure}[H]
\centering
 % \par\medskip
\includegraphics[width=1\textwidth]{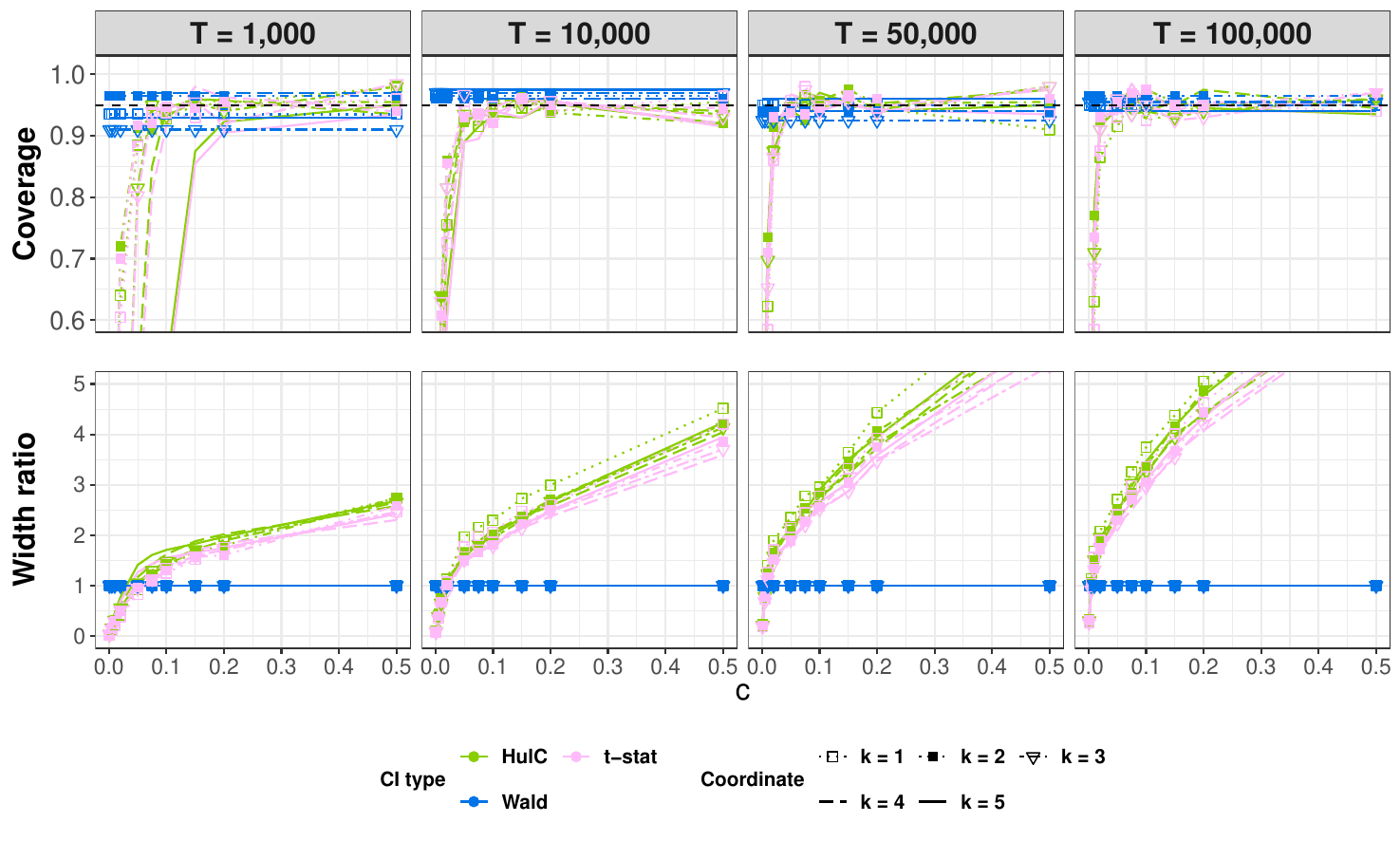}
\caption{Using truncated-SGD, we compare Wald, HulC, and $t$-stat methods in the linear regression setting with a Toeplitz covariance structure and dimension $d=5$. Both the HulC and the $t$-stat methods generally produce correct coverage for $c>0.1$. However, the width ratios for $t$-stat and HulC rise sharply as $c$ grows larger than $0.2$, which is exacerbated as $T$ increases.}
\label{fig:linear_D5_Toeplitz_cov_wr_truncatedSGD_initTRUE}
\end{figure}

As previously for ASGD, we find that linear regression is the easiest of the two modeling tasks. While the best step size hyperparameter $c$ is usually dependent on the particular settings, the shape of the width ratio plots can vary substantially. For example, focusing on $d=5$ and Toeplitz covariance, average-iterate-implicit-SGD produces a flat width ratio curve with respect to $c$ (see Figure~\ref{fig:linear_D5_Toeplitz_cov_wr_AISGD_initTRUE}), indicating that the choice of $c$ only affects the coverage but not the width ratios. Last-iterate-implicit-SGD's width ratio plot is ``L''-shaped (see Figure~\ref{fig:linear_D5_Toeplitz_cov_wr_ISGD_initTRUE}), ROOT-SGD's plot is ``U''-shaped (see Figure~\ref{fig:linear_D5_Toeplitz_cov_wr_rootSGD_initTRUE}), and truncated-SGD has a roughly ``logarithmic''-shaped plot (see Figure~\ref{fig:linear_D5_Toeplitz_cov_wr_truncatedSGD_initTRUE}). Except for average-iterate-implicit-SGD, these shapes are made ``skinnier'' (more pronounced) as the dimension $d$ increases (see Sections~\ref{app:OLS_plots_isgd}--\ref{app:OLS_plots_truncsgd} in the Appendix).

Unlike ASGD, which has an ``inverted U''-shaped coverage plot indicating $c$ can't be too large or too small, the additional methods show that for $c$ larger than some minimum value, both HulC and the $t$-stat methods achieve the target coverage  (up to $c=2$ in our grid of test points). That is, $c$ is less restrictive in terms of coverage than in the ASGD case.

As the sample size increases, the width ratios only decrease for certain algorithms. Width ratios decrease with sample size for ASGD and ROOT-SGD, while they increase for last-iterate-implicit-SGD, truncated-SGD, and noisy-truncated-SGD, and ratios are relatively stable across $T$ for average-iterate-implicit-SGD. The reasons are unclear, but in the case of last-iterate-implicit-SGD, it may be that drastically larger step sizes are needed as the sample size increases.

As the sample size $T$ increases, the width ratios decrease for some algorithms and increase for others. The ratios decrease only for ASGD and ROOT-SGD. The ratios increase with $T$ for last-iterate-implicit-SGD, truncated-SGD, and noisy-truncated-SGD, and ratios are stable across $T$ for average-iterate-implicit-SGD. The reasons are unclear, but in the case of last-iterate-implicit-SGD, it may be the case that drastically larger step sizes are needed as the sample size increases.

\subsubsection{Logistic regression: additional simulations}

As was the case for ASGD, the simulations for logistic regression show that finding the correct hyperparameter $c$ is sometimes more challenging compared to linear regression in terms of achieving the target coverage. Furthermore, the Wald interval frequently proves to be unsuitable as a baseline for comparison when $T$ is small and $d$ is large: in all simulations under equicorrelation covariance (and in roughly half of the simulations under Toeplitz covariance) with $T=10^3$ and $d=100$, the data matrix is ill-conditioned, rendering the Wald interval incomputable in these cases (in particular, width ratios for Toeplitz covariance are incomputable).

\begin{figure}[H]
\centering
 % \par\medskip
\includegraphics[width=1\textwidth]{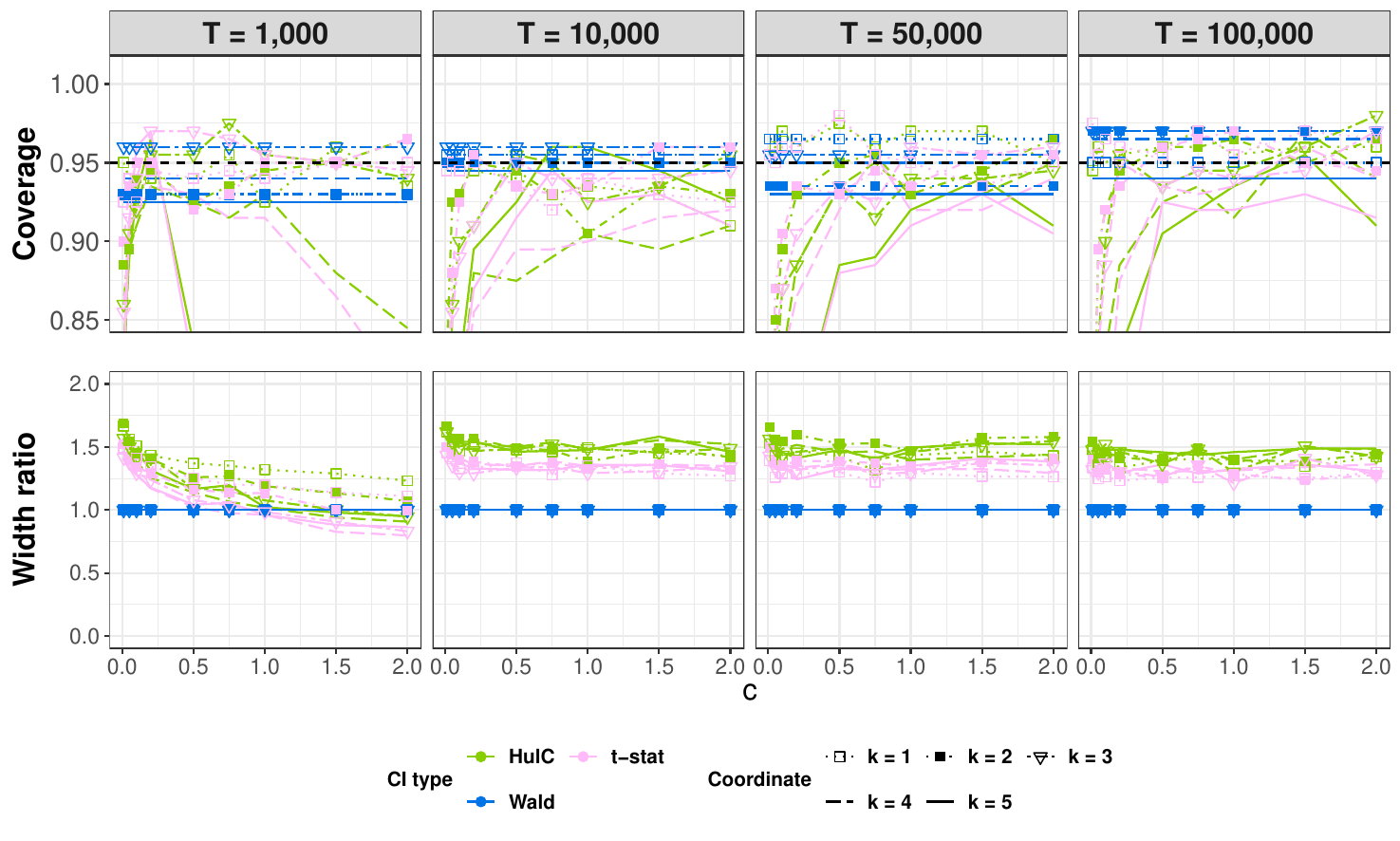}
\caption{Using averaged-iterate-implicit-SGD, we compare Wald, HulC, and $t$-stat methods in the logistic regression setting with a Toeplitz covariance structure and dimension $d=5$. Both the HulC and the $t$-stat methods generally produce correct coverage for any value of $c$. Meanwhile, the width ratios for $t$-stat and HulC are small but do not shrink considerably as the sample size $T$ increases.}
\label{fig:logistic_D5_Toeplitz_cov_wr_AISGD_initTRUE}
\end{figure}

\begin{figure}[H]
\centering
 % \par\medskip
\includegraphics[width=1\textwidth]{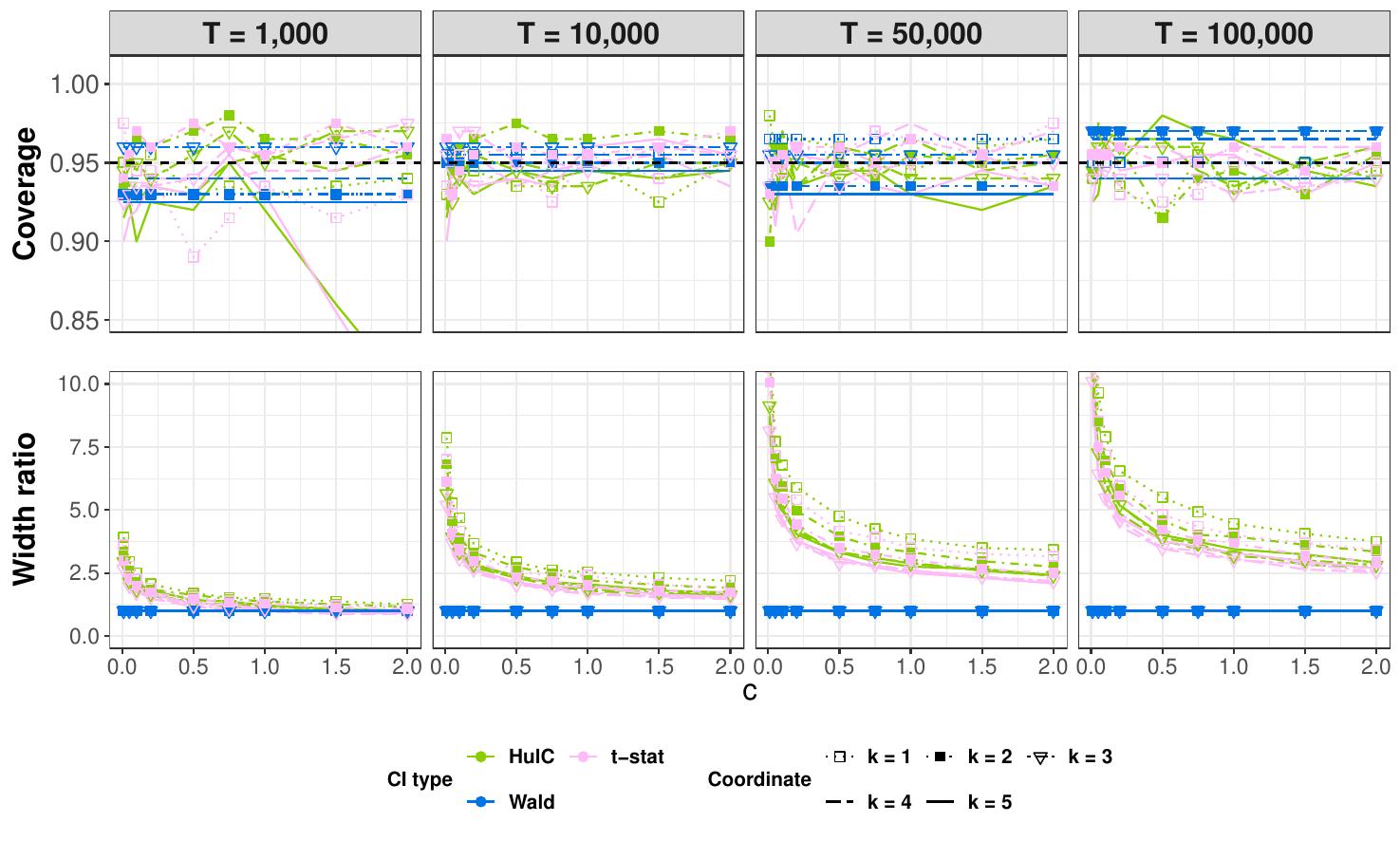}
\caption{Using last-iterate-implicit-SGD, we compare Wald, HulC, and $t$-stat methods in the linear regression setting with a Toeplitz covariance structure and dimension $d=5$. Both the HulC and the $t$-stat methods generally produce correct coverage for appropriately chosen $c$. Meanwhile, the width ratios for $t$-stat and HulC are not excessively large when $c$ is large enough; as the sample size $T$ increases, the ratios increase, requiring (possibly) larger choices of $c$.}
\label{fig:logistic_D5_Toeplitz_cov_wr_ISGD_initTRUE}
\end{figure}

For average-iterate-implicit-SGD, logistic regression has a reduced range of $c$ for which correct coverage is achieved compared to linear regression, but the width ratios remain flat in both cases (see Figures \ref{fig:logistic_D5_Toeplitz_cov_wr_AISGD_initTRUE} and \ref{fig:logistic_D20_Toeplitz_cov_wr_AISGD_initTRUE}). For last-iterate-implicit-SGD, logistic regression tends to achieve correct coverage, but the width ratios are higher compared to linear regression and remain ``L''-shaped (see Figures \ref{fig:logistic_D5_Toeplitz_cov_wr_ISGD_initTRUE} and \ref{fig:logistic_D20_Toeplitz_cov_wr_ISGD_initTRUE}).

\begin{figure}[H]
\centering
 % \par\medskip
\includegraphics[width=1\textwidth]{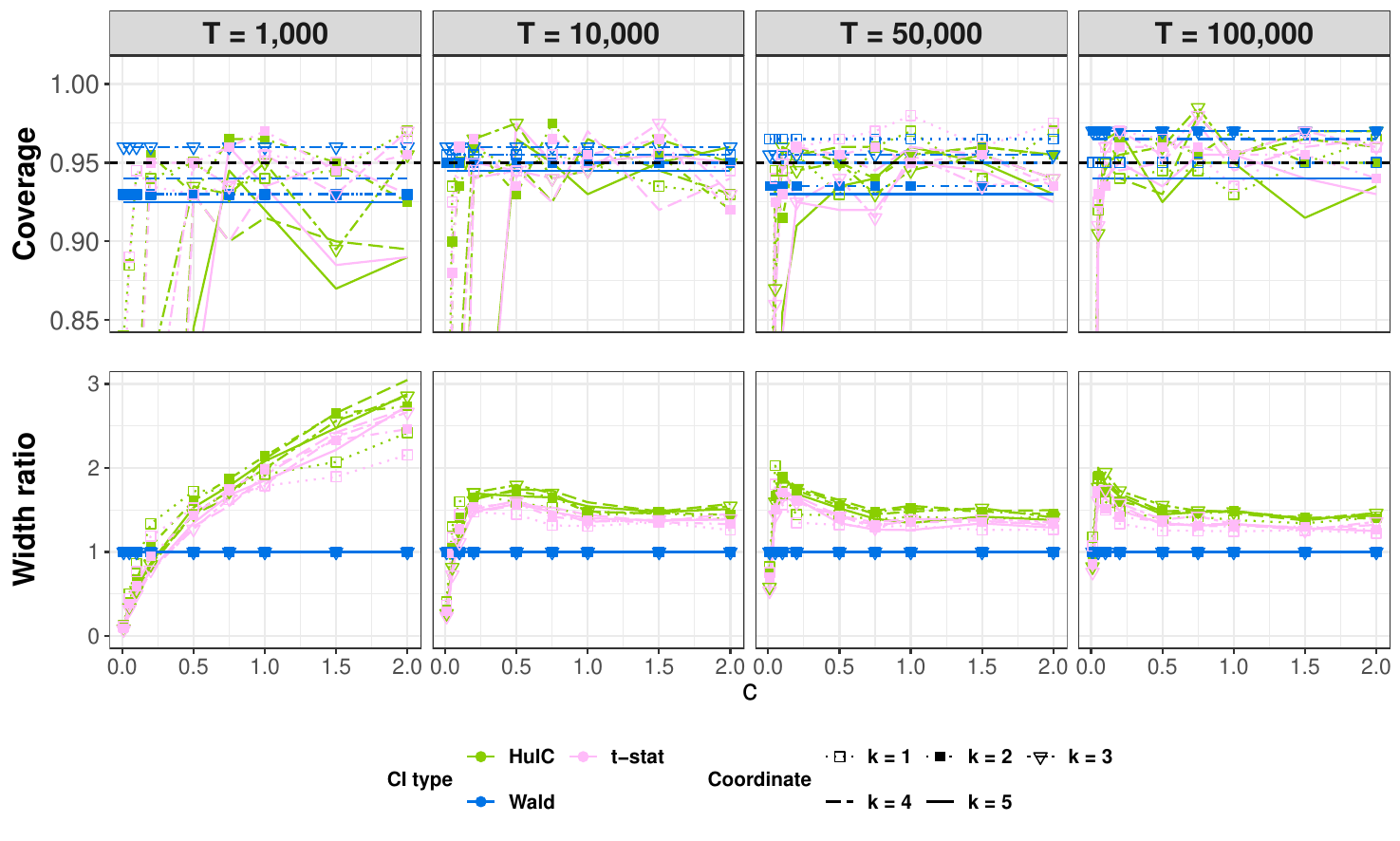}
\caption{Using ROOT-SGD, we compare Wald, HulC, and $t$-stat methods in the logistic regression setting with a Toeplitz covariance structure and dimension $d=5$. Both the HulC and the $t$-stat methods generally produce correct coverage for appropriately chosen $c$. Meanwhile, the width ratios for $t$-stat and HulC are not excessively large when $c$ is appropriately chosen; as the sample size $T$ increases, the ratios decrease.}
\label{fig:logistic_D5_Toeplitz_cov_wr_rootSGD_initTRUE}
\end{figure}

For ROOT-SGD, logistic regression has a comparable coverage behavior compared to linear regression -- however, width ratios are in general much smaller (see Figures \ref{fig:logistic_D5_Toeplitz_cov_wr_rootSGD_initTRUE} and \ref{fig:logistic_D20_Toeplitz_cov_wr_rootSGD_initTRUE}). For truncated-SGD, although logistic regression requires a tighter range of $c$ than linear regression to achieve correct coverage, the width ratios are likewise substantially smaller. Noisy-truncated-SGD exhibits pattern similar to that of truncated-SGD, but logistic regression width ratios are not necessarily smaller than in linear regression.

\begin{figure}[H]
\centering
 % \par\medskip
\includegraphics[width=1\textwidth]{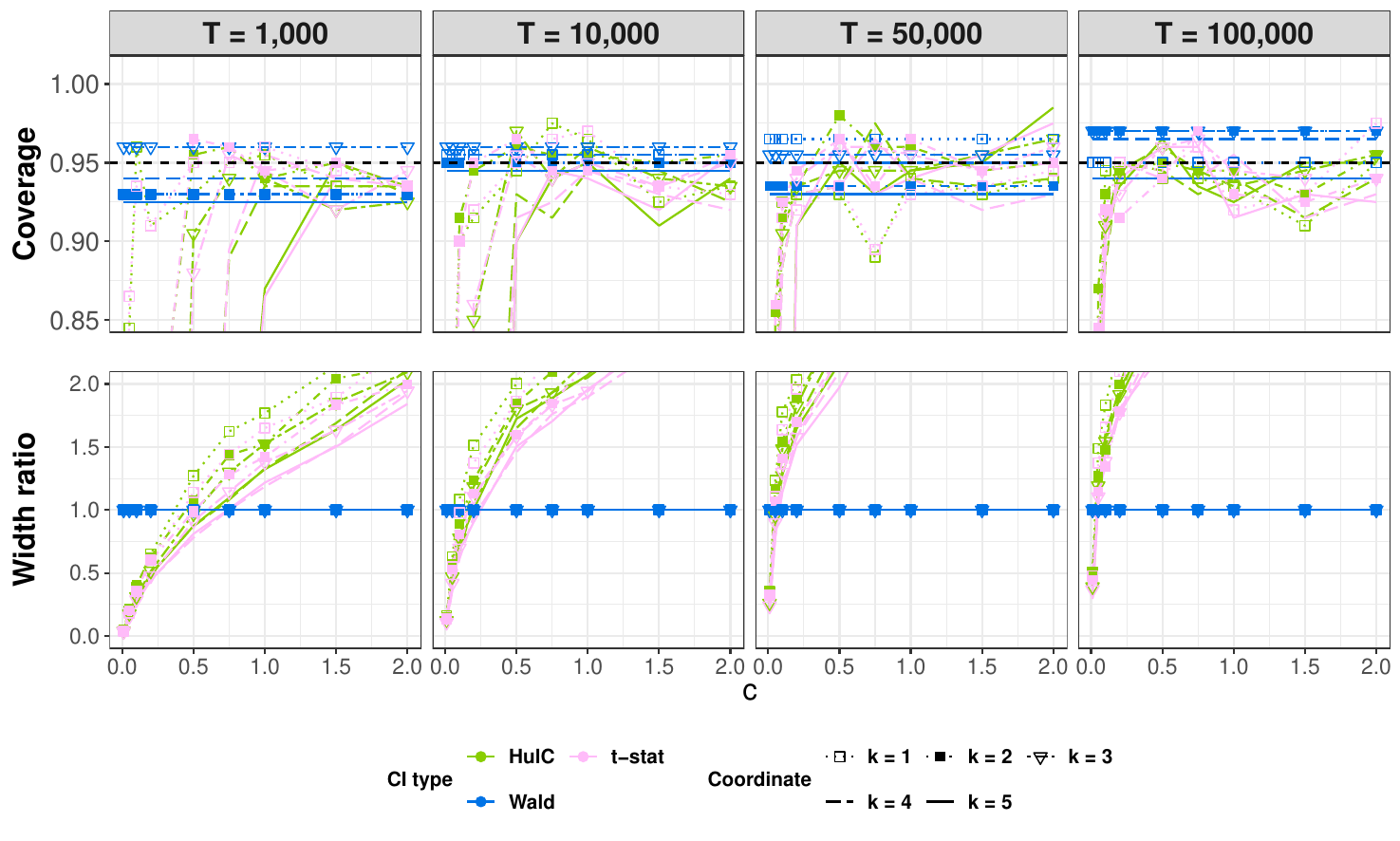}
\caption{Using truncated-SGD, we compare Wald, HulC, and $t$-stat methods in the logistic regression setting with a Toeplitz covariance structure and dimension $d=5$. Both the HulC and the $t$-stat methods generally produce correct coverage for large enough $c$. However, the width ratios for $t$-stat and HulC rise sharply as $T$ increases.}
\label{fig:logistic_D5_Toeplitz_cov_wr_truncatedSGD_initTRUE}
\end{figure}

\section{Concluding Remarks}\label{sec:conc-remarks}
In this paper, we present a reliable, computationally efficient, variance-free inference method for online algorithms under mild regularity conditions. The proposed method, HulC, achieves reasonably good coverage in comparison to existing inference methods with a minimal increase in width. Our simulations show that with a non-stochastic step size scheme, the averaged stochastic gradient descent method (ASGD) can perform poorly as an estimation scheme. This subpar performance directly impacts the performance of any inference method using SGD. In particular, ASGD is highly sensitive to the choice of step size hyperparameters. Similar conclusions continue to hold other online algorithms we have tested in our experiments including implicit-SGD, ROOT-SGD, and other robustified SGD methods (including noisy-truncated-SGD). Out of all the methods tested, the average-iterate-implicit-SGD method of~\citep{toulis2017asymptotic} has some robustness to step size hyperparameters.

With stochastic step sizes, one might need more assumptions on the underlying optimization problem (i.e., loss function) to achieve asymptotic normality and construct theoretically valid confidence intervals. SGD iterations are derived using a local quadratic expansion of the loss function, and some authors have proposed a local non-quadratic expansion to alleviate the poor performance issues. The exploration of such variants of SGD is beyond the scope of the current manuscript and would be an interesting research topic to consider in the future. 
\bibliography{proposal-references}
\bibliographystyle{apalike}
\newpage
\setcounter{section}{0}
\setcounter{equation}{0}
\setcounter{figure}{0}
\renewcommand{\thesection}{S.\arabic{section}}
\renewcommand{\theequation}{E.\arabic{equation}}
\renewcommand{\thefigure}{A.\arabic{figure}}
\renewcommand{\theHsection}{S.\arabic{section}}
\renewcommand{\theHequation}{E.\arabic{equation}}
\renewcommand{\theHfigure}{A.\arabic{figure}}
% \tableofcontents
% \titlelabel{\thetitle: }
% \cftsetindents{section}{1em}{2.5em}
% \cftsetindents{subsection}{1.5em}{3em}
% \setcounter{page}{1}
  \begin{center}
  \Large {\bf Supplement to ``Statistical Inference for Online Algorithms''}
  \end{center}
       
\begin{abstract}
This supplement contains the proofs of all the main results in the paper and some additional simulation results.
\end{abstract}
\section{Proof of Theorem~\ref{thm:SGD-expansion}}\label{appsec:proof-of-SGD-expansion}
By the mean value theorem and assumptions~\ref{assump:zero-gradient},~\ref{assump:holder-cont} imply
\begin{align*}
\nabla M(\theta^{(t-1)}) &= \nabla M(\theta_{\infty}) + \nabla^2 M(\theta_{\infty})(\theta^{(t-1)} - \theta_{\infty}) + R_{t-1}\\
&= \nabla^2 M(\theta_{\infty})(\theta^{(t-1)} - \theta_{\infty}) + R_{t-1},
\end{align*}
with the remainder $R_{t-1}$ satisfying $\|J^{-1/2}R_{t-1}\|_2 \le L\|\theta^{(t-1)} - \theta_{\infty}\|_J^{1 + \mu}$. This implies that $$\xi_t = \nabla \ell(\theta^{(t-1)}, X_t) - J(\theta^{(t-1)} - \theta_{\infty}) - R_{t-1}.$$ Therefore, the SGD iteration becomes
\begin{equation*}
\theta^{(t)} = \theta^{(t-1)} - \eta_t[\xi_t - J(\theta^{(t-1)} - \theta_{\infty}) - R_{t-1}],
\end{equation*}
which yields
\begin{equation*}
\frac{1}{T}\sum_{t = 1}^T \eta_t^{-1}(\theta^{(t)} - \theta^{(t-1)}) = -\frac{1}{T}\sum_{t=1}^T \xi_t + J\left(\frac{1}{T}\sum_{t = 1}^T \theta^{(t-1)} - \theta_{\infty}\right) + \frac{1}{T}\sum_{t=1}^T R_{t-1}.
\end{equation*}
Rearranging, we conclude
\begin{align*}
\left( \frac{1}{T}\sum_{t=1}^T \theta^{(t)} - \theta_{\infty} \right)
&= \frac{1}{T}\sum_{t=1}^T(\theta^{(t)} - \theta^{(t-1)} + \theta^{(t-1)} -\theta_{\infty})\\
&= \frac{1}{T}\sum_{t=1}^T(\theta^{(t-1)} -\theta_{\infty}) + \frac{\theta^{(T)} - \theta^{(0)}}{T}\\
&= \frac{1}{T}\sum_{t=1}^T J^{-1}\xi_t - \frac{1}{T}\sum_{t=1}^T J^{-1}\eta_t^{-1}(\theta^{(t)} - \theta^{(t-1)}) - \frac{1}{T}\sum_{t=1}^T J^{-1}R_{t-1} + \frac{\theta^{(T)} - \theta^{(0)}}{T}.
\end{align*}
Additionally, by simple rearrangement, we get
\begin{align*}
    \frac{1}{T}\sum_{t=1}^T J^{-1}\eta_t^{-1}(\theta^{(t)} - \theta^{(t-1)}) &= \frac{1}{T}\sum_{t=1}^{T-1}  J^{-1}(\eta_t^{-1} - \eta_{t+1}^{-1})(\theta^{(t)} - \theta_{\infty})\\ 
    &\quad+ \frac{J^{-1}\eta_T^{-1}(\theta^{(T)} - \theta_{\infty})}{T} - \frac{J^{-1}\eta_1^{-1}(\theta^{(0)} - \theta_{\infty})}{T}. 
\end{align*}
Hence,
\begin{equation}\label{eq:SGD-influence}
\begin{split}
\sqrt{T}\left\|\bar{\theta}_T - \theta_{\infty} - \frac{1}{T}\sum_{t=1}^T J^{-1}\xi_t\right\|_{J} &\le \sqrt{T}\left\|\frac{1}{T}\sum_{t=1}^{T-1} J^{-1/2}(\eta_t^{-1} - \eta_{t+1}^{-1})(\theta^{(t)} - \theta_{\infty})\right\|_2\\
&\quad+ \sqrt{T}\left\|\frac{J^{-1/2}\eta_T^{-1}(\theta^{(T)} - \theta_{\infty})}{T}\right\|_2 + \sqrt{T}\left\|\frac{J^{-1/2}\eta_1^{-1}(\theta^{(0)} - \theta_{\infty})}{T}\right\|_2\\
&\quad+ \sqrt{T}\left\|\frac{1}{T}\sum_{t=1}^{T} J^{-1/2}R_{t-1}\right\|_2 + \sqrt{T}\left\|\frac{\theta^{(T)} - \theta^{(0)}}{T}\right\|_J.
\end{split}
\end{equation}
This completes the proof.
\section{Proof of Theorem~\ref{thm:rate-bound-SGD}}\label{appsec:rate-bound-SGD}
    Recall the SGD iteration 
    \[
    \theta^{(t)} = \theta^{(t-1)} - \eta_t\nabla \ell(Z_t; \theta^{(t-1)}).
    \]
    Set $\mathcal{F}_s = \{\theta^{(0)}, Z_1, \ldots, Z_s\}$ and $\Delta_t = M(\theta^{(t)}) - M(\theta_{\infty})$. We need the following lemma.
    \begin{lemma}[Implication of~\eqref{eq:ES-c} and~\eqref{eq:smooth-c}]\label{lem:intermediate-rate-bound}
    For any $t \ge 1$, we have
    \[
    \mathbb{E}[\|\nabla\ell(Z_t; \theta^{(t-1)})\|_{J^{-1}}^2|\mathcal{F}_{t-1}] \le 2(A + B\kappa)\Delta_{t-1} + C.
    \]
\end{lemma}
\begin{proof}
    From~\eqref{eq:ES-c}, we get
    \[
    \mathbb{E}[\|\nabla\ell(Z_t; \theta^{(t-1)})\|_{J^{-1}}^2|\mathcal{F}_{t-1}] \le 2A\Delta_{t-1} + B\|\nabla M(\theta^{(t-1)})\|_{J^{-1}}^2 + C.
    \]
    From~\eqref{eq:smooth-c} and the proof of Lemma 1 of~\cite{karandikar2024convergence}, we get
    \[
    \|\nabla M(\theta^{(t-1)})\|_{J^{-1}}^2 \le 2\kappa(M(\theta^{(t-1)}) - M(\theta_{\infty})) = 2\kappa\Delta_{t-1}.
    \]
    Combining these inequalities, we get the result.
\end{proof}
    The SGD iteration implies that
    \begin{align*}
    M(\theta^{(t)}) ~=~ &M(\theta^{(t-1)}) - \eta_t \nabla \ell(Z_t; \theta^{(t-1)}) \\
    ~\overset{\eqref{eq:smooth-c}}{\le} &M(\theta^{(t-1)}) - \nabla M(\theta^{(t-1)})^\top \eta_t \nabla \ell(Z_t; \theta^{(t-1)})  + \frac{\kappa}{2} \|\eta_t\nabla \ell(Z_t; \theta^{(t-1)}) \|_{J}^2\\
    ~\le~ &M(\theta^{(t-1)}) - \nabla M(\theta^{(t-1)})^\top \eta_t \nabla \ell(Z_t; \theta^{(t-1)})  + \frac{\kappa}{2} \|\tilde{\eta}_t\|_{\mathrm{op}}^2 \|\nabla \ell(Z_t; \theta^{(t-1)}) \|_{J^{-1}}^2.
    \end{align*}
    Taking conditional expectation given $\mathcal{F}_{t-1}$, we get
    \begin{align*}
\mathbb{E}[M(\theta^{(t)}) | \mathcal{F}_{t-1}] &\le M(\theta^{(t-1)}) - \nabla M(\theta^{(t-1)})^\top \eta_t \nabla M(\theta^{(t-1)})  + \frac{\kappa}{2} \|\tilde{\eta}_t\|_{\mathrm{op}}^2 \mathbb{E}[\|\nabla \ell(Z_t; \theta^{(t-1)}) \|_{J^{-1}}^2 | \mathcal{F}_{t-1}]
\end{align*}
From Lemma~\ref{lem:intermediate-rate-bound}, we get
\begin{align*}
 \mathbb{E}[\Delta_t|\mathcal{F}_{t-1}] &~{\le}~ \Delta_{t-1}  + \frac{\kappa}{2}  \|\tilde{\eta}_t\|_{\mathrm{op}}^2 (2(A + B\kappa)\Delta_{t-1} + C) - \|\nabla M(\theta^{(t-1)})\|_{\eta_t}^2\\
 &~= \Delta_{t-1}\left(1 + (A + B\kappa)\kappa\|\tilde{\eta}_t\|_{\mathrm{op}}^2\right) + C\kappa\|\tilde{\eta}_t\|_{\mathrm{op}}^2/2 - \|\nabla M(\theta^{(t-1)})\|_{\eta_t}^2.
% \\
%&\overset{\eqref{eq:PL-c}}{\le} \Delta_{t-1}\left(1 + (A + B\kappa)\kappa\|\tilde{\eta}_t\|_{\mathrm{op}}^2 - 2\mu_{\mathrm{PL}}\|\tilde{\eta}_t\|_{\mathrm{op}}\right) + C\kappa\|\tilde{\eta}_t\|_{\mathrm{op}}^2/2.
% \\
% &~= \left(1 - 2\mu_{\mathrm{PL}}\|\tilde{\eta}_t\|_{\mathrm{op}} + \kappa(A + B\mu_{\mathrm{PL}})\|\tilde{\eta}_t\|_{\mathrm{op}}^2 \right) \Delta_{t-1} + {C \kappa \|\tilde{\eta}_t\|_{\mathrm{op}}^2}/{2}.
\end{align*}
Hence, setting
\begin{equation}\label{eq:application-of-Robbins-Siegmund}
\beta_{t-1} := (A + B\kappa)\kappa\|\tilde{\eta}_t\|_{\mathrm{op}}^2,\quad X_{t-1} = \|\nabla M(\theta^{(t-1)})\|_{\eta_t}^2,\quad\mbox{and}\quad Z_{t-1} = C\kappa\|\tilde{\eta}_t\|_{\mathrm{op}}^2/2,\quad t\ge1,
\end{equation}
we get the recursion
\begin{equation}\label{eq:basic-recursion-without-PL}
\mathbb{E}[\Delta_t|\mathcal{F}_{t-1}] \le (1 + \beta_{t-1})\Delta_{t-1} + Z_{t-1} - X_{t-1}\quad\mbox{for all}\quad t \ge 1.
\end{equation}
Recall that $\eta_t$ is $\mathcal{F}_{t-1}$-measurable by the assumption and it is clear that $\beta_{t-1} \ge 0, X_{t-1} \ge 0, Z_{t-1} \ge 0$.
This implies that $\{\Delta_t\}$ is a non-negative almost supermartingale satisfying the assumptions of the Robbins-Siegmund theorem~\citep{Robbins1971}. This implies that $\{\Delta_t\}$ converges almost surely (as $t\to\infty$), and $\sum_{t=0}^{l} X_t$ converges almost surely (as $l\to\infty$). To find a quantitative rate of convergence, we follow the proof of Theorem 4.3 of~\cite{neri2024quantitative}. To this end, for $\xi\in\{0, 1\}$, set
\[
R_t(\xi) := \Delta_t\left(\prod_{s=1}^t (1 + \beta_{s-1})\right)^{-1} - \sum_{s=1}^t (Z_{s-1} - \xi X_{s-1})\left(\prod_{m=1}^s (1 + \beta_{m-1})\right)^{-1},\quad t\ge1. 
\]
From~\eqref{eq:basic-recursion-without-PL}, it follows that $\{R_t(\xi)\}_{t\ge0}$ is a supermartingale with respect to the natural filtration of the sequence $\theta^{(0)}, Z_1, Z_2, \ldots$.  Although $R_t(\xi)$ is not a non-negative supermartingale, it is almost non-negative because for $\xi\ge0$,
\[
\sum_{s=1}^t (Z_{s-1} - \xi X_{s-1})\left(\prod_{m=1}^s (1 + \beta_{m-1})\right)^{-1} \le \sum_{s=1}^{t} Z_{s-1}\left(\prod_{m=1}^s (1 + \beta_{m-1})\right)^{-1} \le \sum_{s=1}^t Z_{s-1} = \frac{C\kappa}{2}\sum_{s=1}^t \|\tilde{\eta}_s\|_{\mathrm{op}}^2,
\]
which is finite almost surely as $t\to\infty$. Hence, following the proof of Theorem 4.3 of~\cite{neri2024quantitative} (see, in particular, the first display on page 18), we get
\begin{align*}
\mathbb{P}\left(\sup_{t\ge1}|R_t(1)| \ge \frac{4(\mathbb{E}[\Delta_0] + \zeta)}{\lambda}\right) \le \frac{\lambda}{4} + \mathbb{P}\left(\frac{C\kappa}{2}\sum_{s=1}^{\infty} \|\tilde{\eta}_s\|_{\mathrm{op}}^2 \ge \zeta\right),\\
\mathbb{P}\left(\sup_{t\ge1} |R_t(0)| \ge \frac{4(\mathbb{E}[\Delta_0] + \zeta)}{\lambda}\right) \le \frac{\lambda}{4} + \mathbb{P}\left(\frac{C\kappa}{2}\sum_{s=1}^{\infty} \|\tilde{\eta}_s\|_{\mathrm{op}}^2 \ge \zeta\right).
\end{align*}
Together, we get
\[
\mathbb{P}\left(\sup_{t\ge1}\, |R_t(1) - R_t(0)| \ge \frac{8(\mathbb{E}[\Delta_0] + \zeta)}{\lambda}\right) \le \frac{\lambda}{2} + 2\mathbb{P}\left(\frac{C\kappa}{2}\sum_{s=1}^{\infty} \|\tilde{\eta}_s\|_{\mathrm{op}}^2 \ge \zeta\right).
\]
Also, observe that
\[
R_t(1) - R_t(0) = \sum_{s=1}^t X_{s-1}\left(\prod_{m=1}^s (1 + \beta_{m-1})\right)^{-1} \ge 0.
\]
Moreover, 
\[
\left(\prod_{m=1}^s (1 + \beta_{m-1})\right)^{-1} \ge \exp\left(-\sum_{m=1}^t \beta_{m-1}\right) \ge \exp\left(-\frac{2A + 2B\kappa}{C}\times\frac{C\kappa}{2}\sum_{s=1}^t \|\tilde{\eta}_s\|_{\mathrm{op}}^2\right).
\]
Combining this inequality with the tail probability bound for $R_t(1) - R_t(0)$, we conclude
%\[
%\mathbb{P}\left(\sup_{t\ge1}\sum_{s=1}^t X_{s-1}\left(\prod_{m=1}^s (1 + \beta_{m-1})\right)^{-1} \ge  \frac{8(\mathbb{E}[\Delta_0] + \zeta)}{\lambda}\right) \le \frac{\lambda}{2} + 2\mathbb{P}\left(\frac{C\kappa}{2}\sum_{s=1}^{\infty} \|\tilde{\eta}_s\|_{\mathrm{op}}^2 \ge \zeta\right).
%\]
%Additionally, because $\beta_{m-1} \ge 0$ for all $m\ge1$, we have
%\[
%\left(\prod_{m=1}^t (1 + \beta_{m-1})\right)^{-1}\sum_{s=1}^t X_{s-1} \le \sum_{s=1}^t X_{s-1}\left(\prod_{m=1}^s (1 + \beta_{m-1})\right)^{-1} \le \sum_{s=1}^t X_{s-1},
%\]
%and hence, using the inequality $1 + x \le e^x$ for all $x\ge0$,
%\begin{align*}
%\sum_{s=1}^t X_{s-1} &\le \exp\left(\sum_{m=1}^t \beta_{m-1}\right)\sum_{s=1}^t X_{s-1}\left(\prod_{m=1}^s (1 + \beta_{m-1})\right)^{-1}\\
%&\le \exp\left(\frac{2A + 2B\kappa}{C}\times\frac{C\kappa}{2}\sum_{s=1}^t \|\tilde{\eta}_s\|_{\mathrm{op}}^2\right)\sum_{s=1}^t X_{s-1}\left(\prod_{m=1}^s (1 + \beta_{m-1})\right)^{-1}.
%\end{align*}
%we conclude
\[
\mathbb{P}\left(\sum_{s=1}^{\infty} \|\nabla M(\theta^{(s-1)})\|_{\eta_s}^2 \ge \frac{8(\mathbb{E}[\Delta_0] + \zeta)}{\lambda}\exp\left(\frac{(2A + 2B\kappa)\zeta}{C}\right)\right) \le \lambda + 3\mathbb{P}\left(\frac{C\kappa}{2}\sum_{s=1}^{\infty} \|\tilde{\eta}_s\|_{\mathrm{op}}^2 \ge \zeta\right).
\]

% This implies that
% \[
% \mathbb{E}[R_t] \le \mathbb{E}[R_1] \le \mathbb{E}[M(\theta^{(1)}) - M(\theta_{\infty})] < \infty.
% \]
% Now note that
% \[
% \Delta_t = \left(\prod_{s=1}^t (1 + \beta_{s-1})\right)R_t + \frac{C\kappa}{2}\sum_{s=1}^t \|\tilde{\eta}_s\|_{\mathrm{op}}^2\left(\prod_{m=s+1}^t (1 + \beta_{m-1})\right)^{-1}.
% \]
% Because $B\ge2$ and that $\kappa \ge \mu_{\mathrm{PL}}$, we get
% \[
% 1 + (A + B\kappa)\kappa\|\tilde{\eta}_t\|_{\mathrm{op}}^2 - 2\mu_{\mathrm{PL}}\|\tilde{\eta}_t\|_{\mathrm{op}} \ge 1 + 2\mu_{\mathrm{PL}}^2\|\tilde{\eta}_t\|_{\mathrm{op}}^2 - 2\mu_{\mathrm{PL}}\|\tilde{\eta}_t\|_{\mathrm{op}} \ge 1/2,
% \]
% which, in particular, implies that the coefficient of $\Delta_{t-1}$ in our recursive inequality is non-negative. The sequence $\{R_t\}$ can be verified to be a supermartingale by substitution.
\section{Proof of Corollary~\ref{cor:under-PL-convergence}}
Inequality~\eqref{eq:norm-of-gradients} follows readily from inequality~\eqref{eq:tail-bound-sum-of-gradients} using the fact that
\begin{align*}
\|\nabla M(\theta^{(t-1)})\|_{\eta_t}^2 &= \nabla M(\theta^{(t-1)})^{\top}J^{-1/2}(J^{1/2}\eta_tJ^{1/2})J^{-1/2}\nabla M(\theta^{(t-1)})\\ 
&= \nabla M(\theta^{(t-1)})^{\top}J^{-1/2}\tilde{\eta}_tJ^{-1/2}\nabla M(\theta^{(t-1)})\\ 
&\ge \frac{\|\nabla M(\theta^{(t-1)})\|_{J^{-1}}^2}{\|\tilde{\eta}_t^{-1}\|_{\mathrm{op}}}.
\end{align*}
Inequality~\eqref{eq:norm-of-iterates} follows from~\eqref{eq:norm-of-gradients} under~\ref{eq:PL} following the proof of (PL)$\Rightarrow$(EB) in Theorem 3.1 of~\cite{liao2024error}. In particular, it follows from the proof in~\cite{liao2024error} that~\ref{eq:PL} implies
\[
\|\nabla M(\theta)\|_{J^{-1}}^2 \ge \mu_{\mathrm{PL}}\|\theta - \theta_{\infty}\|_J^2.
\]
This yields~\eqref{eq:norm-of-iterates} from~\eqref{eq:norm-of-gradients}.

\newpage

\section{Algorithms}\label{sec:algo}

The following two algorithms have been adapted to the context of online updates from \cite{noisyt_zhou_2021}.

Algorithm~\ref{alg:GT} calculates the cut threshold $\kappa_{\varepsilon, t}$ with inputs $g:=\nabla\ell(Z_t; \theta^{(t)})$ and hyperparameter $\varepsilon>0$. Algorithm~\ref{algo:soft_cut} produces the sparse gradient $\tilde{g}^{(t)}$ and updates $\theta^{(t)}$ accordingly. Algorithm~\ref{algo:ntrunc} also produces the sparse gradient $\tilde{g}^{(t)}$ and updates $\theta^{(t)}$ with additional Gaussian noise.

\begin{algorithm}[h]
    \caption{Gradient Truncation (GT)}
    \begin{algorithmic}[1] \label{alg:GT}
    \STATE  \textbf{Input}: Gradient $g := \left[g_1,,...,g_d \right] \in \mathbb{R}^{d}$, cut rate $\varepsilon^2$
    \STATE Sort the squares of gradient coordinates $g_1^2,\dots, g_d^2$ by descending order: $g_{(1)}^2 \geq, ..., \geq g_{(p)}^2$.
    \STATE $g_{CumSum} = 0$
    \FOR{$i =1,...,d$}
    \IF{$g_{CumSum} \geq (1-\varepsilon^2)\|\bold{g}\|^2$ }
    \STATE \textbf{Return} $| g_{(i)}|$ and \textbf{Halt}.
    \ENDIF
    \STATE $g_{CumSum} = g_{CumSum} + g_i^2$
    \ENDFOR
    %\STATE \textbf{Return}: 
    \end{algorithmic}
    \end{algorithm}

    \begin{algorithm}[h] 
\caption{Truncated-SGD}
	\begin{algorithmic}[1] \label{algo:soft_cut}
		\STATE \textbf{Input}: Online data $\{Z_t\}_{t=1}^T$, loss function $\ell(\cdot)$, initial point $\theta^{(0)}$
		\STATE \textbf{Set}:  Learning rate $\eta_t$, cut rate $\varepsilon^2$.
		\FOR{$t = 0,\dots,T$}
		\STATE Calculate gradient $g^{(t)} := \nabla \ell(Z_t;\theta^{(t)})$.
		\STATE  Call $\textbf{GT}(g^{(t)}, \varepsilon^2)$ to calculate the cut threshold $\kappa_{\varepsilon, t}$.
		\FOR{$i = 1,\dots,d$}
		    \STATE If $|g^{(t)}_i|< \kappa_{\varepsilon, t}$, then $\tilde{g}^{(t)}_i = 0 $, else $\tilde g^{(t)}_i = g^{(t)}_i$.
		\ENDFOR
		\STATE Update parameter using sparse gradient $ \tilde g_t$: $\theta^{(t+1)}=\theta^{(t)}-\eta_t \tilde{g}^{(t)}$.
		\ENDFOR 
	\end{algorithmic}
\end{algorithm}

 \begin{algorithm}[h] 
\caption{Noisy-truncated-SGD}

	\begin{algorithmic}[1] \label{algo:ntrunc}
		\STATE \textbf{Input}: Online data $\{Z_t\}_{t=1}^T$, loss function $\ell(\cdot)$, initial point $\theta^{(0)}$
		\STATE \textbf{Set}: Noise parameter $\sigma>0$, learning rate $\eta_t$, cut rate $\varepsilon^2$, $\beta \in [0, 0.5]$.
		\FOR{$t = 0,\dots,T$}
		\STATE Calculate gradient $g^{(t)} := \nabla \ell(Z_t;\theta^{(t)})$.
		\STATE  Call $\textbf{GT}(g^{(t)}, \varepsilon^2)$ to calculate the cut threshold $\kappa_{\varepsilon, t}$.
		\FOR{$i = 1,\dots,d$}
		    \STATE If $|g^{(t)}_i|< \kappa_{\varepsilon, t}$, then $\tilde{g}^{(t)}_i = 0 $, else $\tilde g^{(t)}_i = g^{(t)}_i$.
		\ENDFOR
		\STATE Update parameter using sparse gradient $ \tilde g_t$: $\theta^{(t+1)}=\theta^{(t)}-\eta_t \tilde{g}^{(t)} + \eta_t^{\frac{1}{2} + \beta}b_t$, where $b_t \sim N(0, \sigma^2 I$
		\ENDFOR 
	\end{algorithmic}
\end{algorithm}

\newpage

%%%%%%%%%%%%PLOTS%%%%%%%%%%%%%%%
\section{Plots from additional simulations}\label{sec:annex_plots}
In this section, we provide plots for identity and equicorrelation covariance types. The findings are already summarized in the main article. Note that all plots are available in the online interactive tool.\footnote{\url{https://public.tableau.com/app/profile/selina.carter6629/viz/OnlineinferencesimulationsOLSandlogisticregression/Coverageandwidthratio_paper}}

We describe the simulation settings in Section \ref{sec:simulation} of the main paper, which are based on experiments by \cite{Chen_SGD_2021}. Note that we choose the following values of step size hyperparameter $c$ according to Table \ref{tab:c-ASGD}. We did this to increase the granularity of the grid of $c$ points in places we noticed in the graphs that coverages were maximized across the different methods and width ratios were lowest.

\begin{table}[H]
\centering
\begin{tabular}{|r|l|}
\hline
\multicolumn{1}{|l|}{\textbf{Linear regression}}   & \multicolumn{1}{c|}{Grid of $c$} \\ \hline
$d=5$                                              &  $c \in \{.01, .05, .1, .2, .5, .75, 1, 1.5, 2\}$     \\ \hline
$d=20$                                             &  $c \in \{.0025, .0125, .025, .05, .125, .1875, .25, .375, .5, .75, 1, 1.5, 2\}$    \\ \hline
$d=100$                                            &  $c \in \{.0005, .0025, .005, .01, .025, .0375, .05, .075, .1, .15, .2, .25, .3, .4, .5, .75, 1, 1.5, 2 \}$  \\ \hline
\multicolumn{1}{|l|}{\textbf{Logistic regression}} &                                  \\ \hline
$d=5$                                              &   (same as linear regression)       \\ \hline
$d=20$                                             &   (same as linear regression)      \\ \hline
$d=100$                                            &    $c \in \{.15, .2, .25, .3, .4, .5, .75, 1, 1.5, 2\}$        \\ \hline
\end{tabular}
\vspace{.2cm}
\caption{Grid of values for step size hyperparameter $c$ for the ASGD simulations}
\label{tab:c-ASGD}
\end{table}

For all other simulations, we use use the same grid of settings described in Section \ref{sec:simulation}. For truncated- and noisy-truncated-SGD, we use hyperparameters $\varepsilon=0.8$, $\sigma=1$, and $\beta=.25$. Note that we use the same initialization procedure for $\theta^{(0)}$ as described in section~\ref{subs:asgd_sensitivity}.  Note also that we choose the following values of step size hyperparameter $c$ according to Table \ref{tab:c-extra}. We did this to increase the granularity of the grid of $c$ points in places we noticed in the graphs that coverages were maximized across the different algorithms and width ratios were lowest.

\begin{table}[H]
\centering
\begin{tabular}{|r|l|}
\hline
\multicolumn{1}{|l|}{}   & \multicolumn{1}{c|}{Grid of $c$} \\ \hline
 \textbf{Linear regression}         &  $c \in \{.001, .005, .01, .02, .05, .075, .1, .15, .2, .3, .4, .5, .75, 1, 1.5, 2 \}$     \\ \hline
\multicolumn{1}{|l|}{\textbf{Logistic regression}} &   $c \in \{.001, .005, .01, .02, .05, .075, .1, .15, .2, .5, .75, 1, 1.5, 2 \}$                               \\ \hline
\end{tabular}
\vspace{.2cm}
\caption{Grid of values for step size hyperparameter $c$ for all SGD simulations excluding ASGD}
\label{tab:c-extra}
\end{table}

\begin{figure}[H]
\centering
 % \par\medskip
\includegraphics[width=1\textwidth]{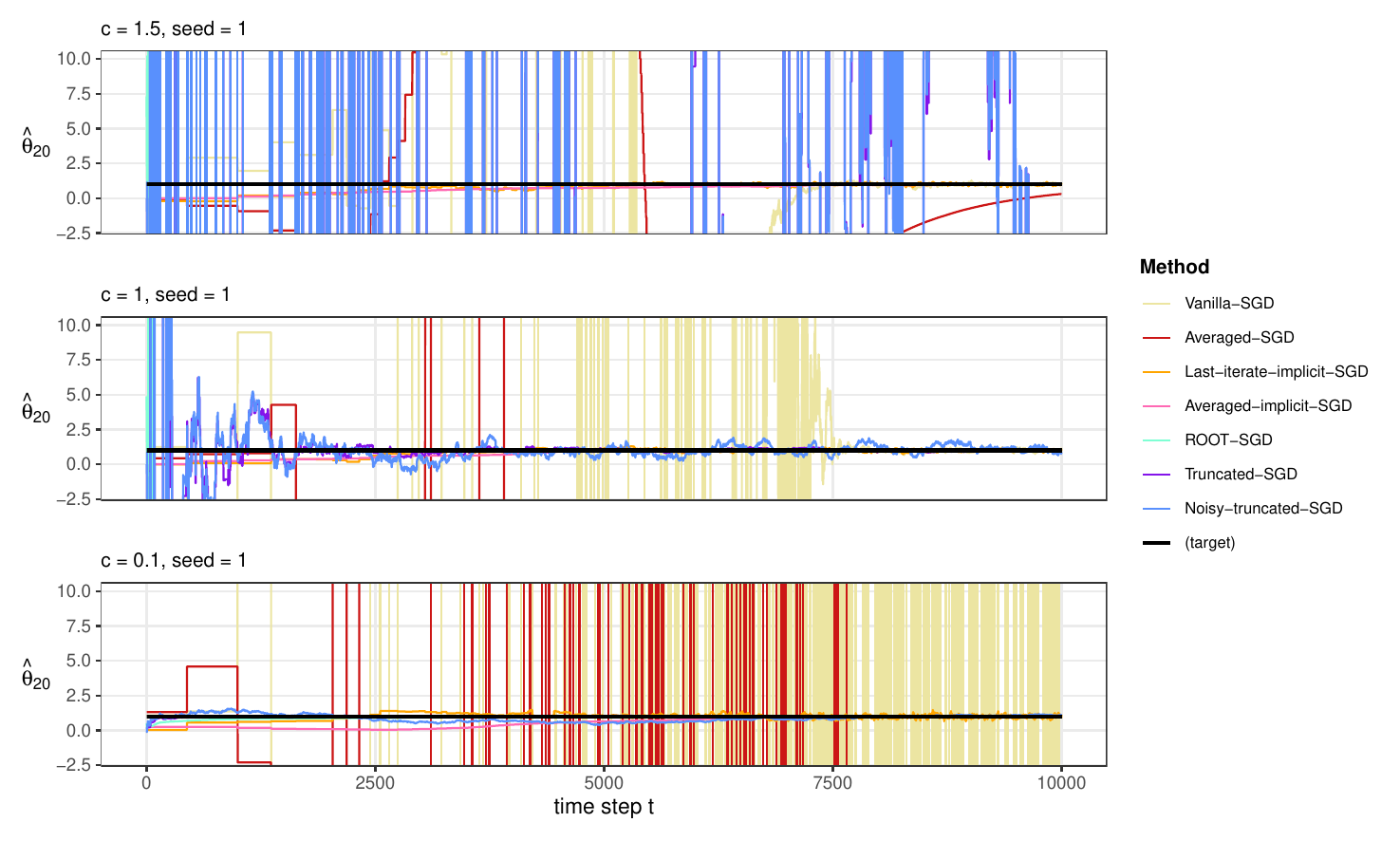}
\caption{Using varying step size hyperparameters $c \in \{0.25, 1.0, 1.5\}$, we compare all SGD methods in the linear regression setting with a Toeplitz covariance structure, $T=10^4$, and dimension $d=20$, viewing coordinate $k=20$. All have a common seed of $1$ (and the same initialization $\hat{\theta}^{(0)}$). In the top panel, $c=1.5$ is too large for certain algorithms (particularly truncated and noisy-truncated-SGD, which respectively estimate $\hat{\theta}_{20}^{(T)}\approx -45$ and $\approx -55$); all other algorithms are within $0.41$ of the target $\theta_{20}=1.0$. The middle panel illustrates that a ``modest value'' of $c=1.0$ doesn't work for ROOT-SGD and ASGD, which have final estimates of  $\hat{\theta}_{20}^{(T)}\approx 1542$ and $\approx 98$, respectively. In the bottom panel, $c=0.25$ is too small for vanilla SGD and ASGD, which each estimate $\hat{\theta}_{20}^{(T)}>10^7$; all the other algorithms are within $0.08$ of the target value. }
\label{fig:other_methods_examples_D20}
\end{figure}

\subsection{ASGD}

\subsubsection{Linear regression}\label{app:OLS_plots}
\vspace{-20 pt}

\begin{figure}[H]
\centering
\par\medskip
\includegraphics[width=1\textwidth]{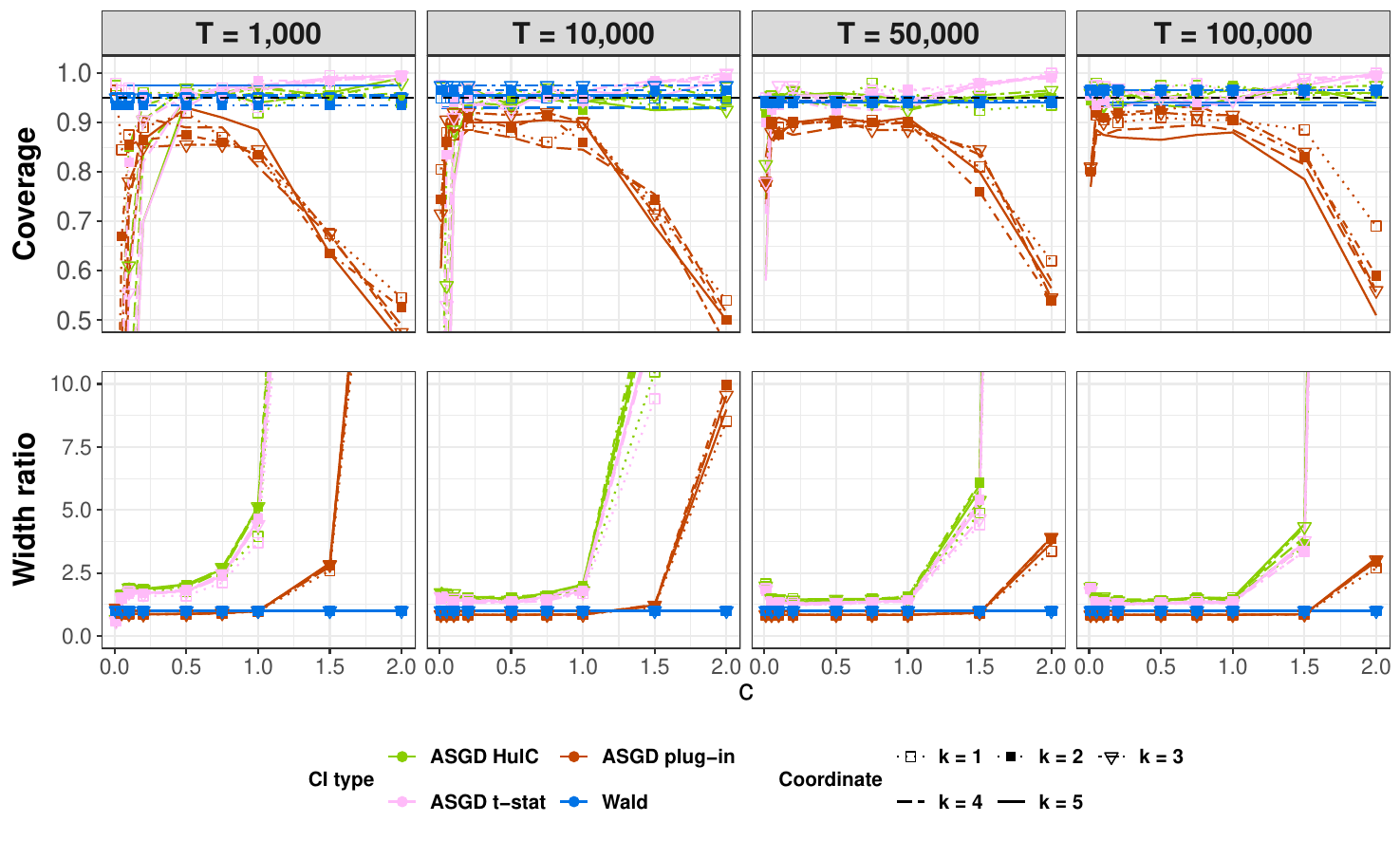}
\caption{ Linear regression, Covariance = I, d = 5}
\label{fig:linear_D5_I_cov_wr}
\end{figure}
\vspace{-25 pt}

\begin{figure}[H]
\centering
\par\medskip
\includegraphics[width=1\textwidth]{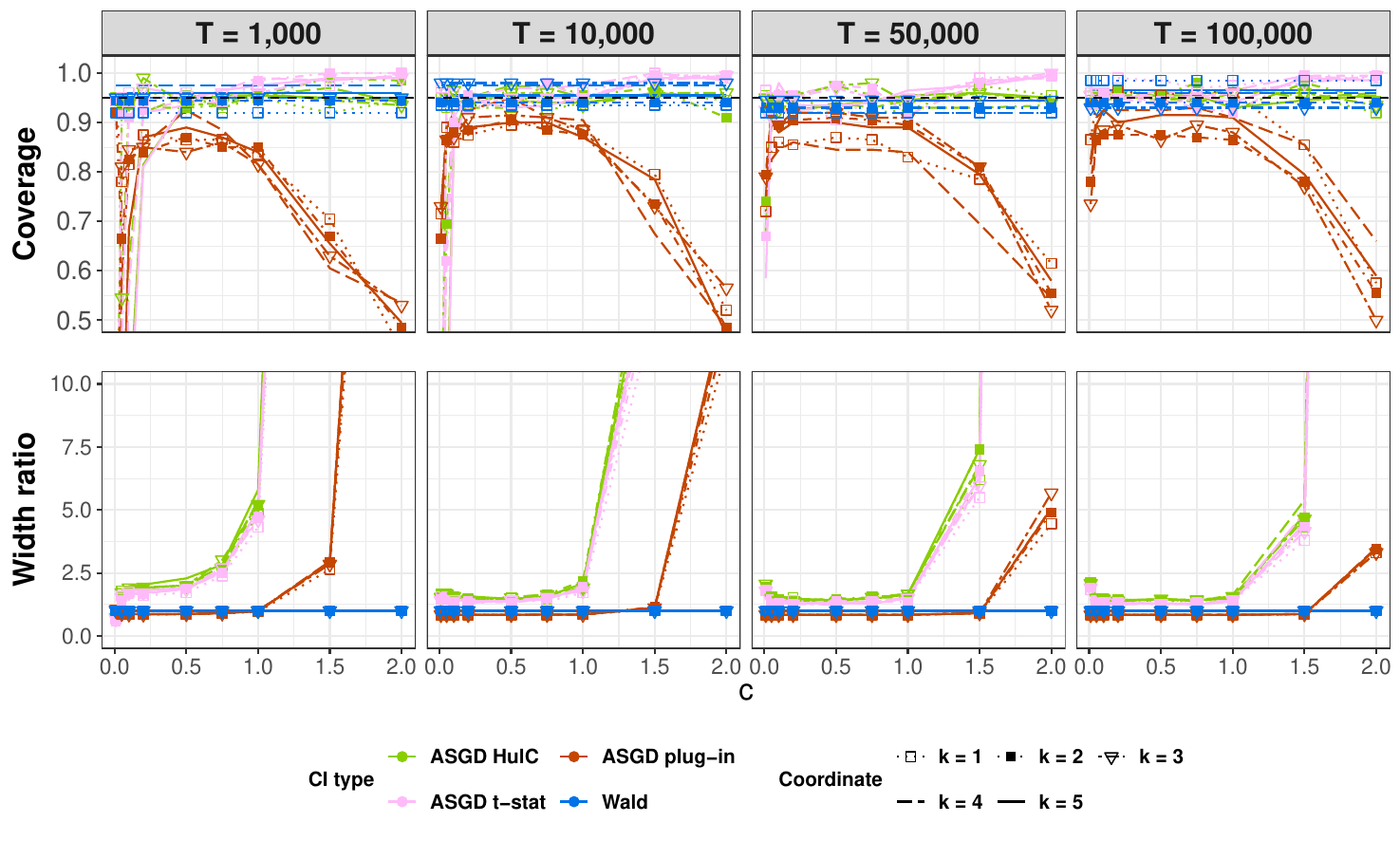}
\caption{ Linear regression, Covariance = Equicorrelation, d = 5}
\label{fig:linear_D5_EquiCorr_cov_wr}
\end{figure}

For linear regression using ASGD, $d=5$, and Toeplitz covariance, see Figure~\ref{fig:linear_D5_Toeplitz_cov_wr}.

%%%%%%%%%%%%%%%%%%%%%%%%%%%%%%%%
\newpage 

\begin{figure}[H]
\centering
\par\medskip
\includegraphics[width=1\textwidth]{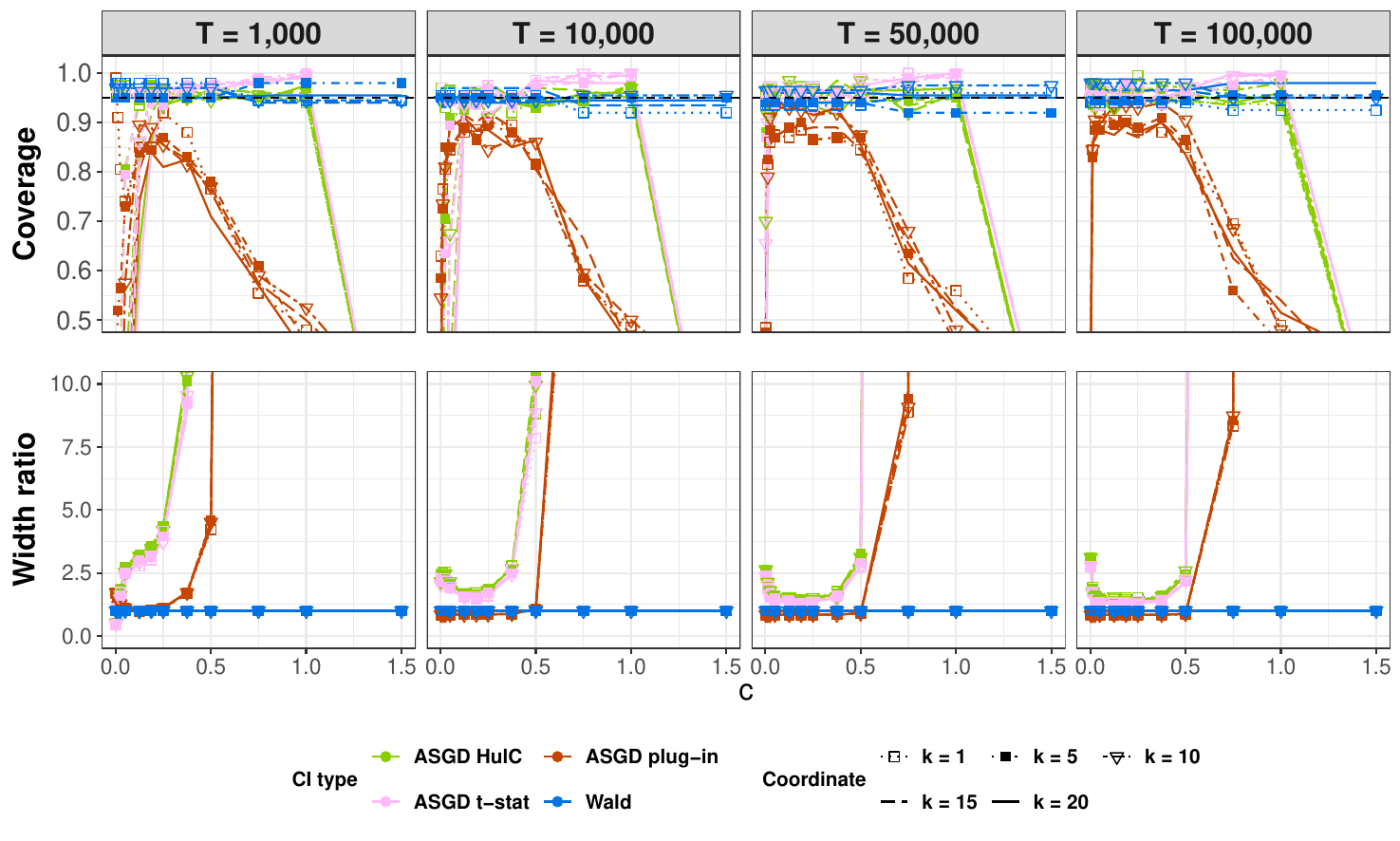}
\caption{ Linear regression, Covariance = I, d = 20}
\label{fig:linear_D20_I_cov_wr}
\end{figure}
\vspace{-25 pt}

\begin{figure}[H]
\centering
\par\medskip
\includegraphics[width=1\textwidth]{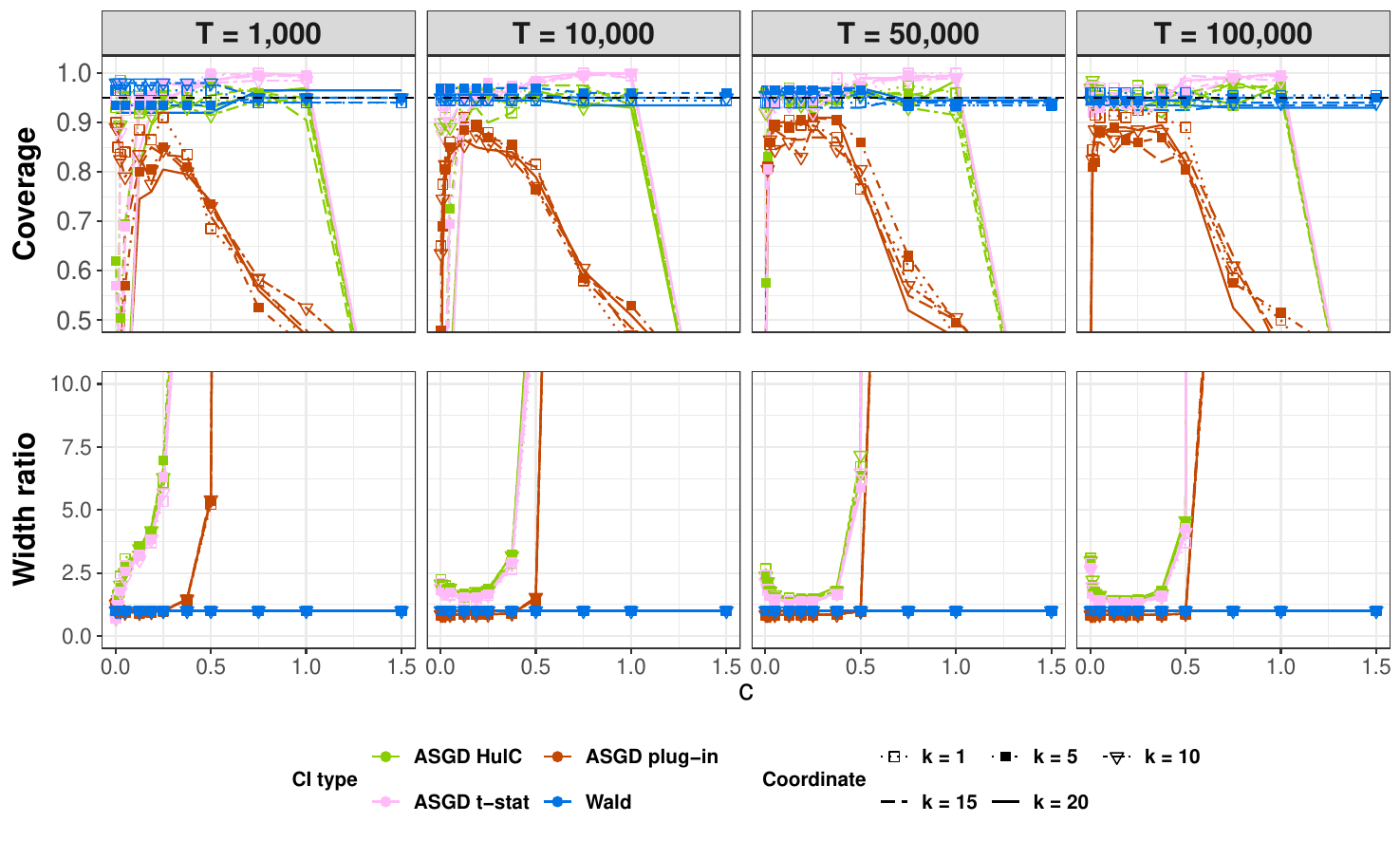}
\caption{ Linear regression, Covariance = Equicorrelation, d = 20}
\label{fig:linear_D20_EquiCorr_cov_wr}
\end{figure}

\begin{figure}[H]
\centering
\par\medskip
\includegraphics[width=1\textwidth]{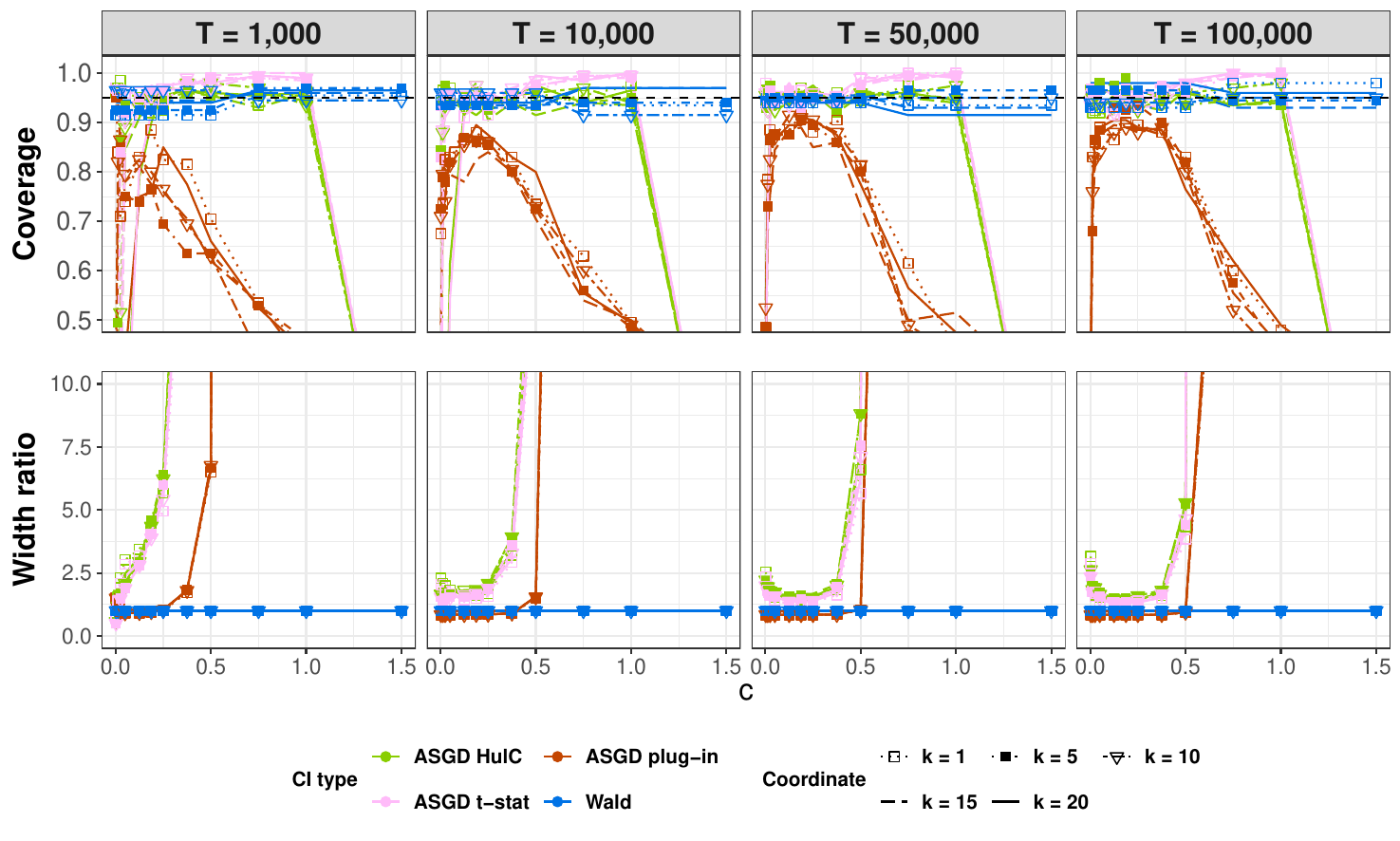}
\caption{ Linear regression, Covariance = Toeplitz, d = 20}
\label{fig:linear_D20_Toeplitz_cov_wr}
\end{figure}

%%%%%%%%%%%%%%%%%%%%%%
%%%%%%%%%%%%%%%%%%%%%%%%%%%%%%%%
\newpage 

\begin{figure}[H]
\centering
\par\medskip
\includegraphics[width=1\textwidth]{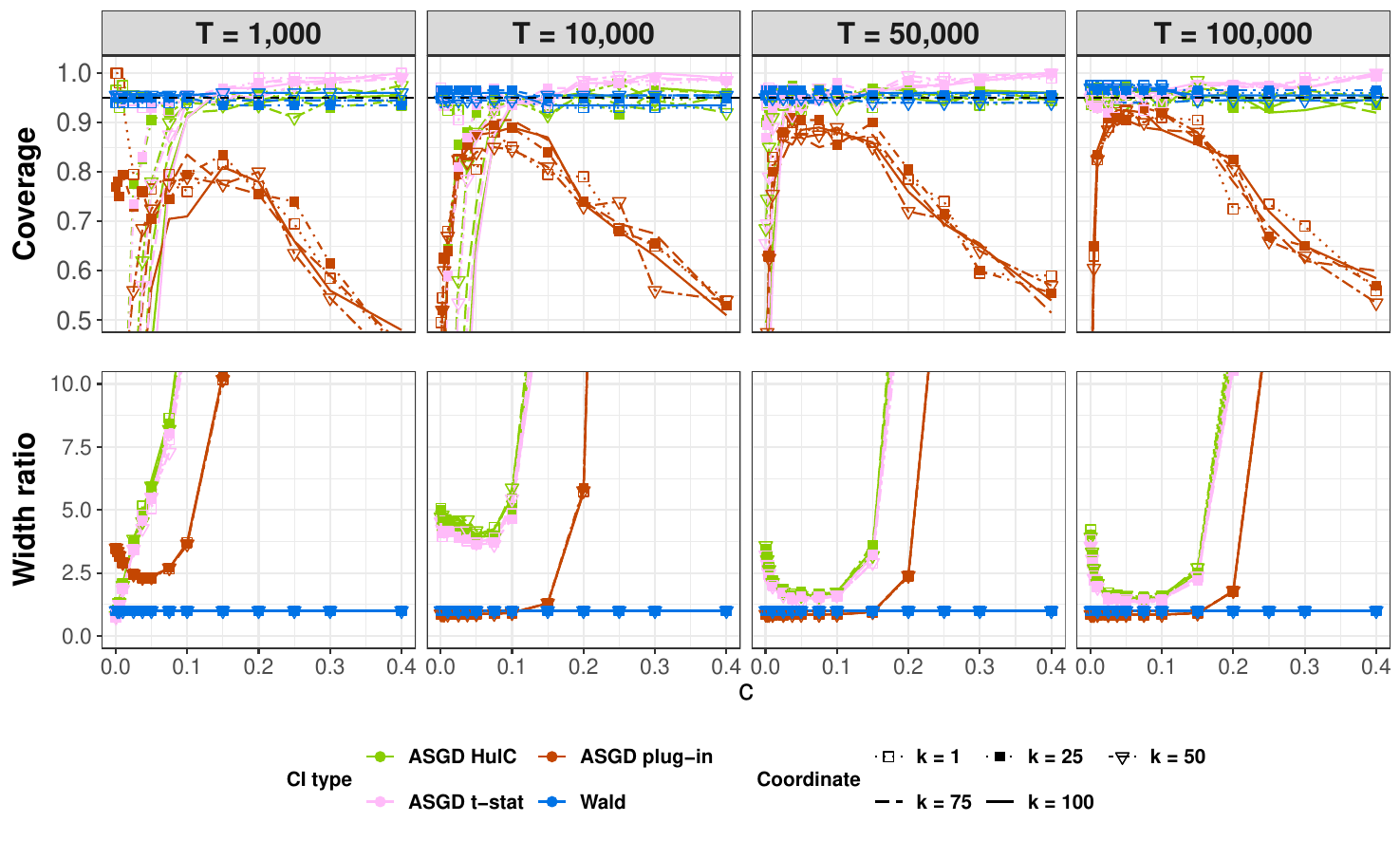}
\caption{ Linear regression, Covariance = I, d = 100}
\label{fig:linear_D100_I_cov_wr}
\end{figure}
\vspace{-25 pt}

\begin{figure}[H]
\centering
\par\medskip
\includegraphics[width=1\textwidth]{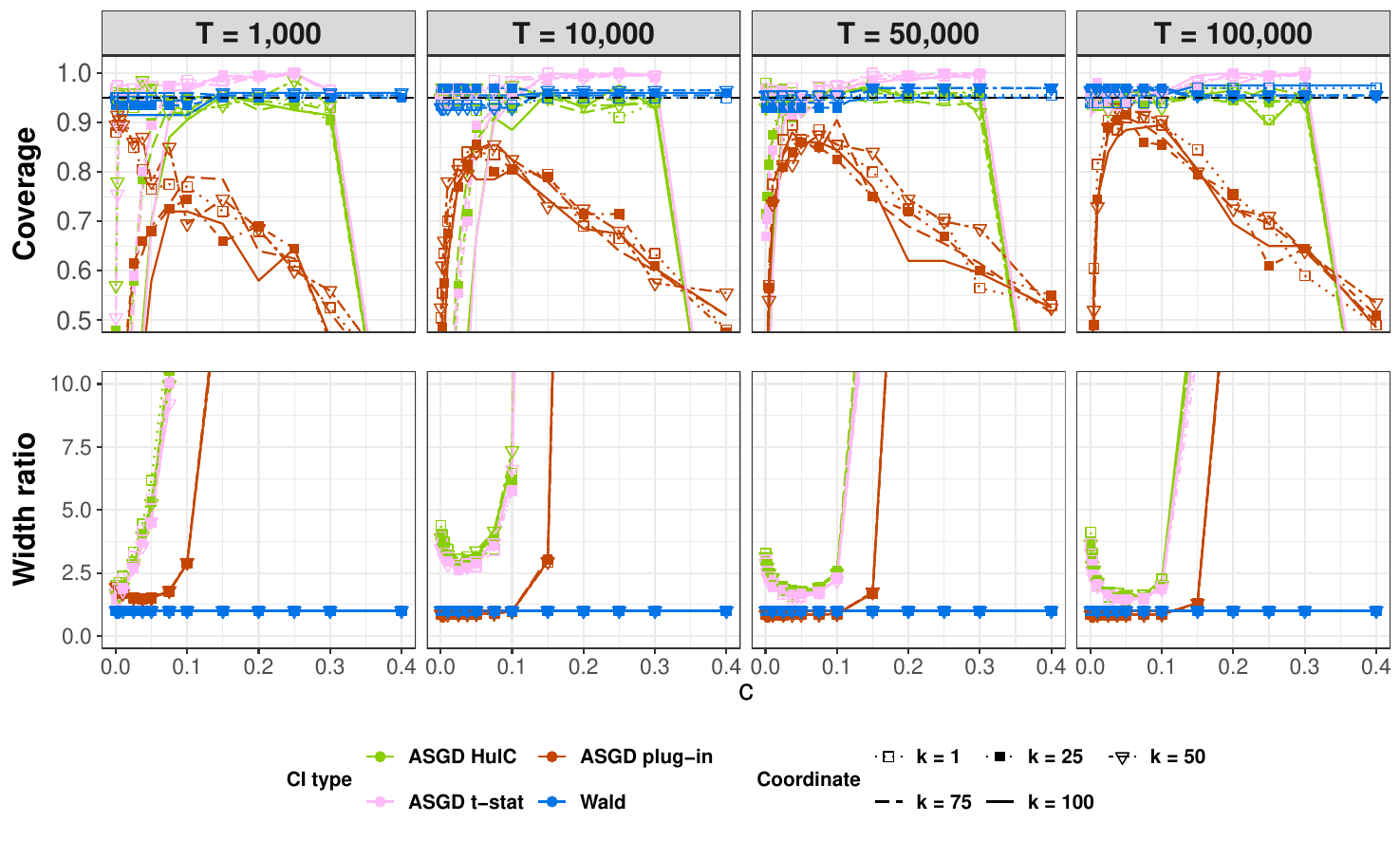}
\caption{ Linear regression, Covariance = Equicorrelation, d = 100}
\label{fig:linear_D100_EquiCorr_cov_wr}
\end{figure}

\begin{figure}[H]
\centering
\par\medskip
\includegraphics[width=1\textwidth]{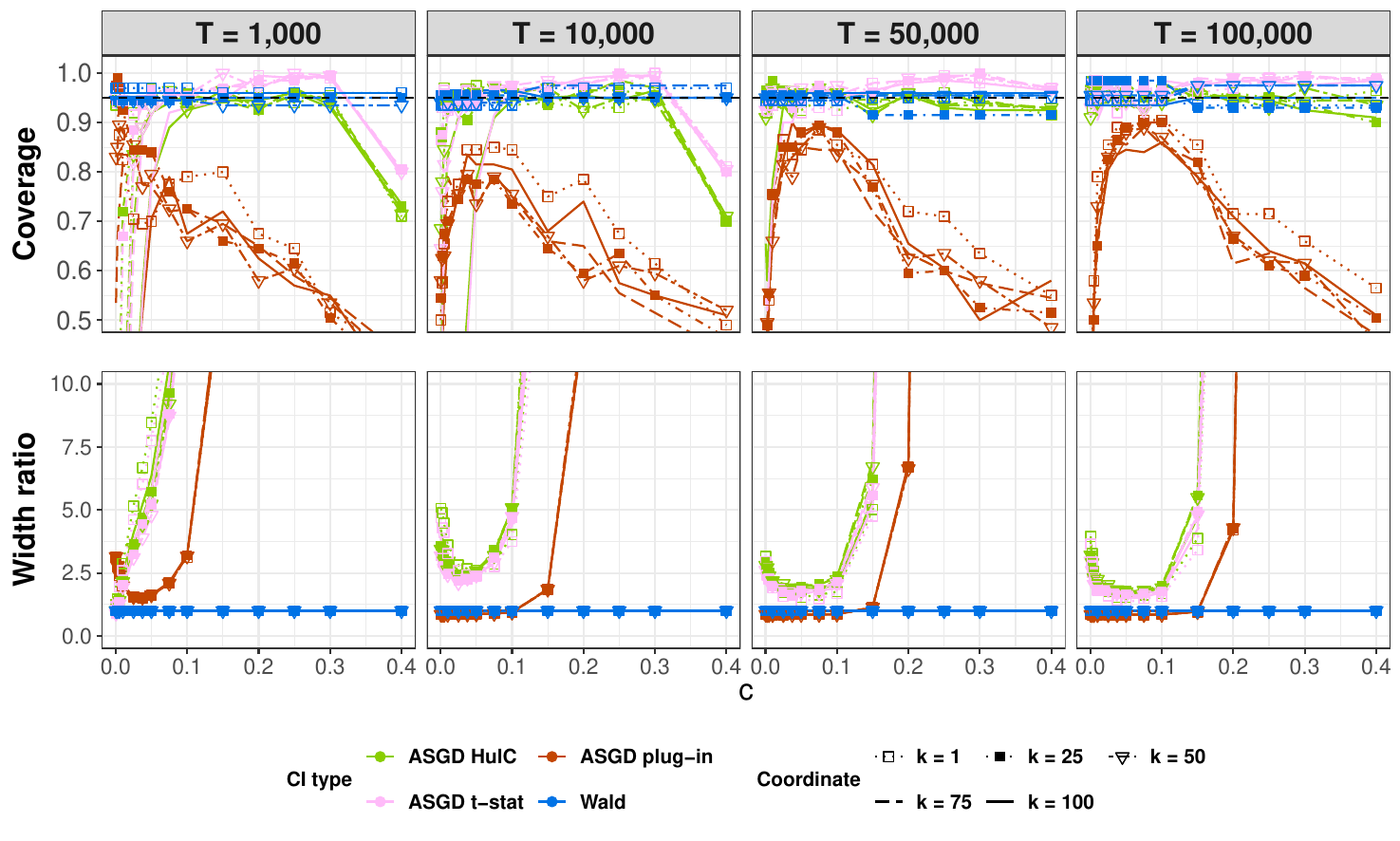}
\caption{ Linear regression, Covariance = Toeplitz, d = 100}
\label{fig:linear_D100_Toeplitz_cov_wr}
\end{figure}

%%%%%%%%%%%%%%%%%%%%%%%%

%%%%%%%%%%%%%%%%%%%%%%%%
\newpage
\subsubsection{Logistic regression}\label{app:log_plots}
\vspace{-20 pt}
\begin{figure}[H]
\centering
\par\medskip
\includegraphics[width=1\textwidth]{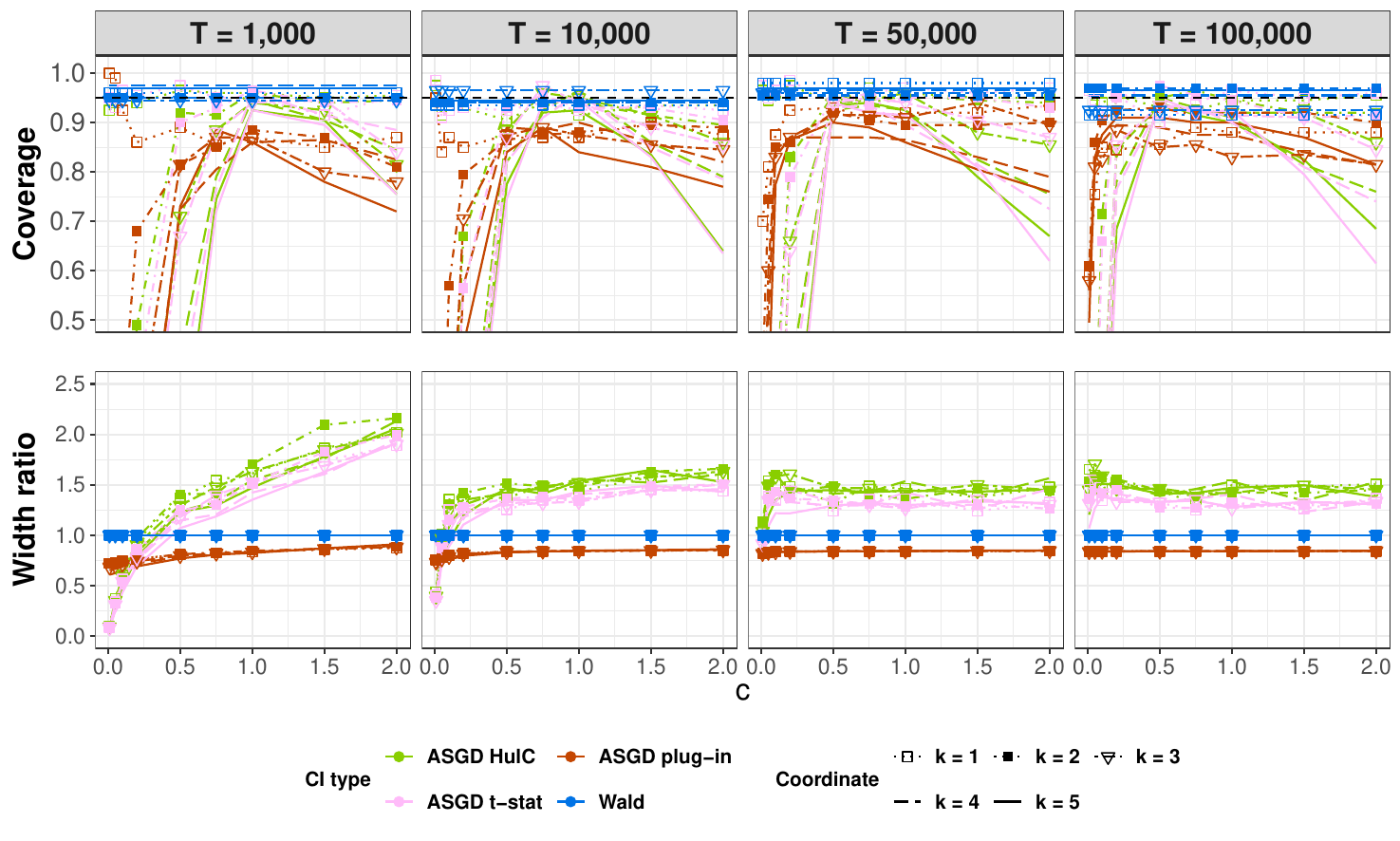}
\caption{ Logistic regression, Covariance = I, d = 5}
\label{fig:logistic_D5_I_cov_wr}
\end{figure}
\vspace{-25 pt}

\begin{figure}[H]
\centering
\par\medskip
\includegraphics[width=1\textwidth]{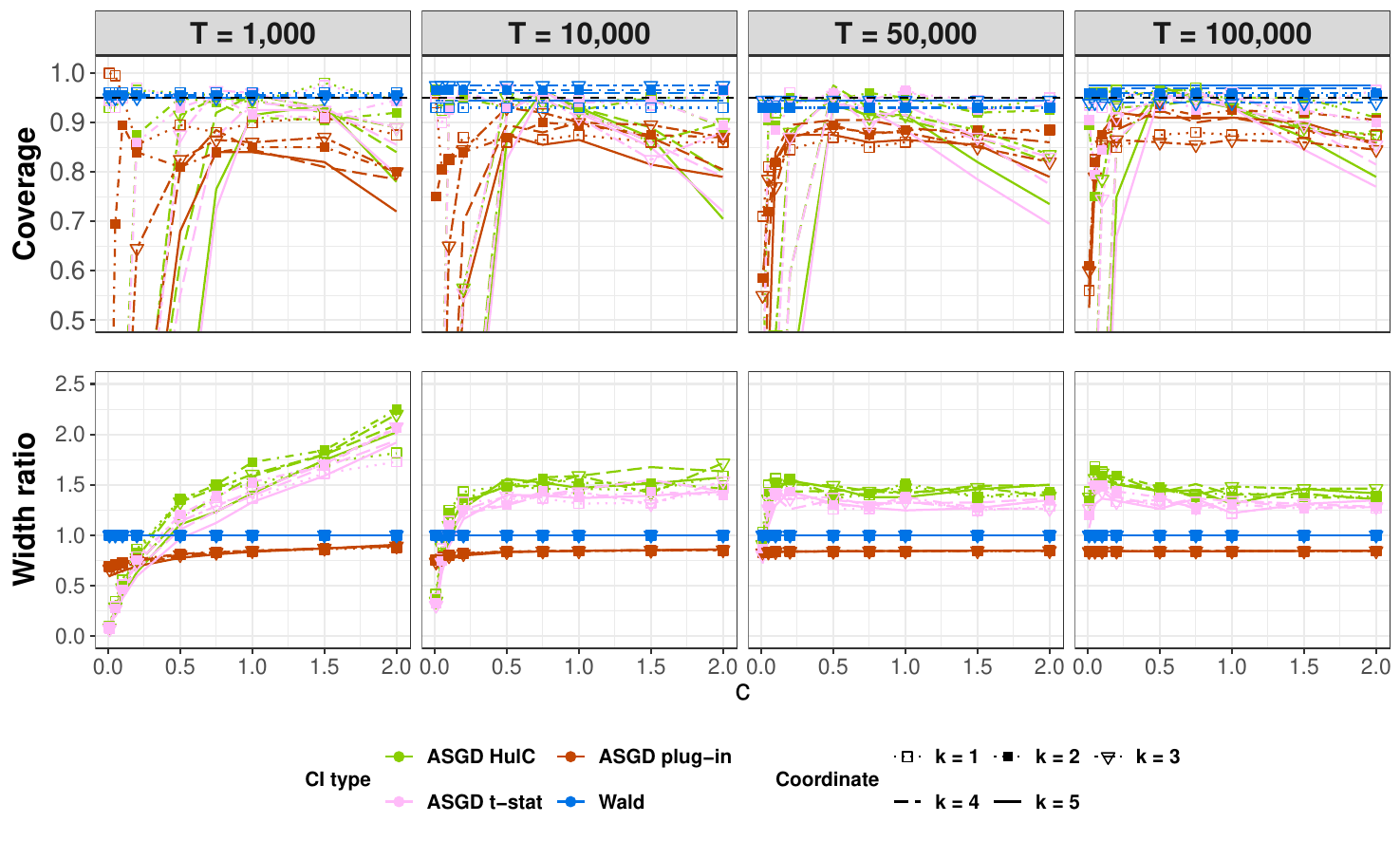} 
\caption{ Logistic regression, Covariance = Equicorrelation, d = 5}
\label{fig:logistic_D5_EquiCorr_cov_wr}
\end{figure}

For logistic regression using ASGD, $d=5$, and Toeplitz covariance, see Figure~\ref{fig:logistic_D5_Toeplitz_cov_wr}.

%%%%%%%%%%%%%%%%%%%%%%%
\newpage
\begin{figure}[H]
\centering
\par\medskip
\includegraphics[width=1\textwidth]{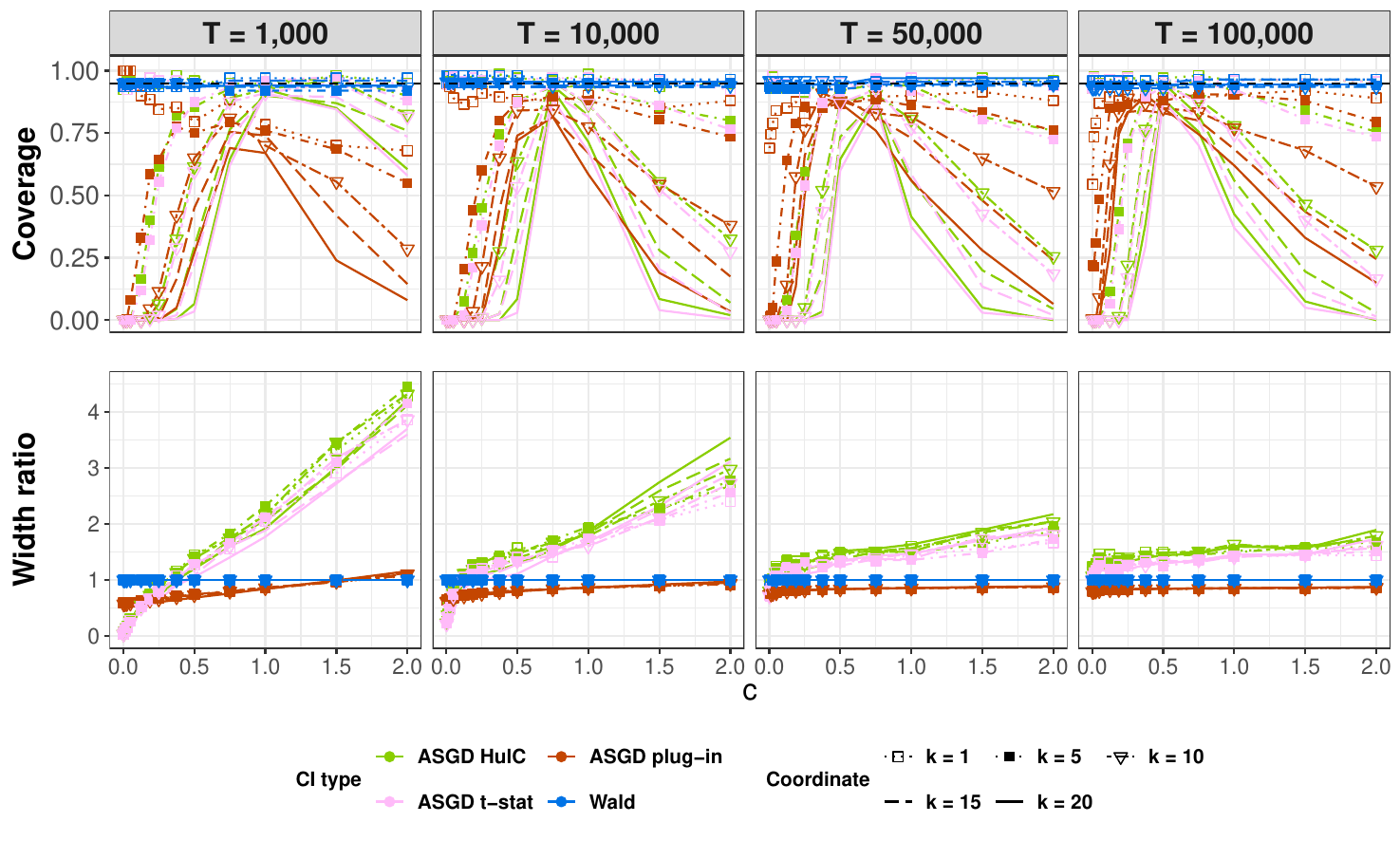}
\caption{ Logistic regression, Covariance = I, d = 20}
\label{fig:logistic_D20_I_cov_wr}
\end{figure}
\vspace{-25 pt}

\begin{figure}[H]
\centering
\par\medskip
\includegraphics[width=1\textwidth]{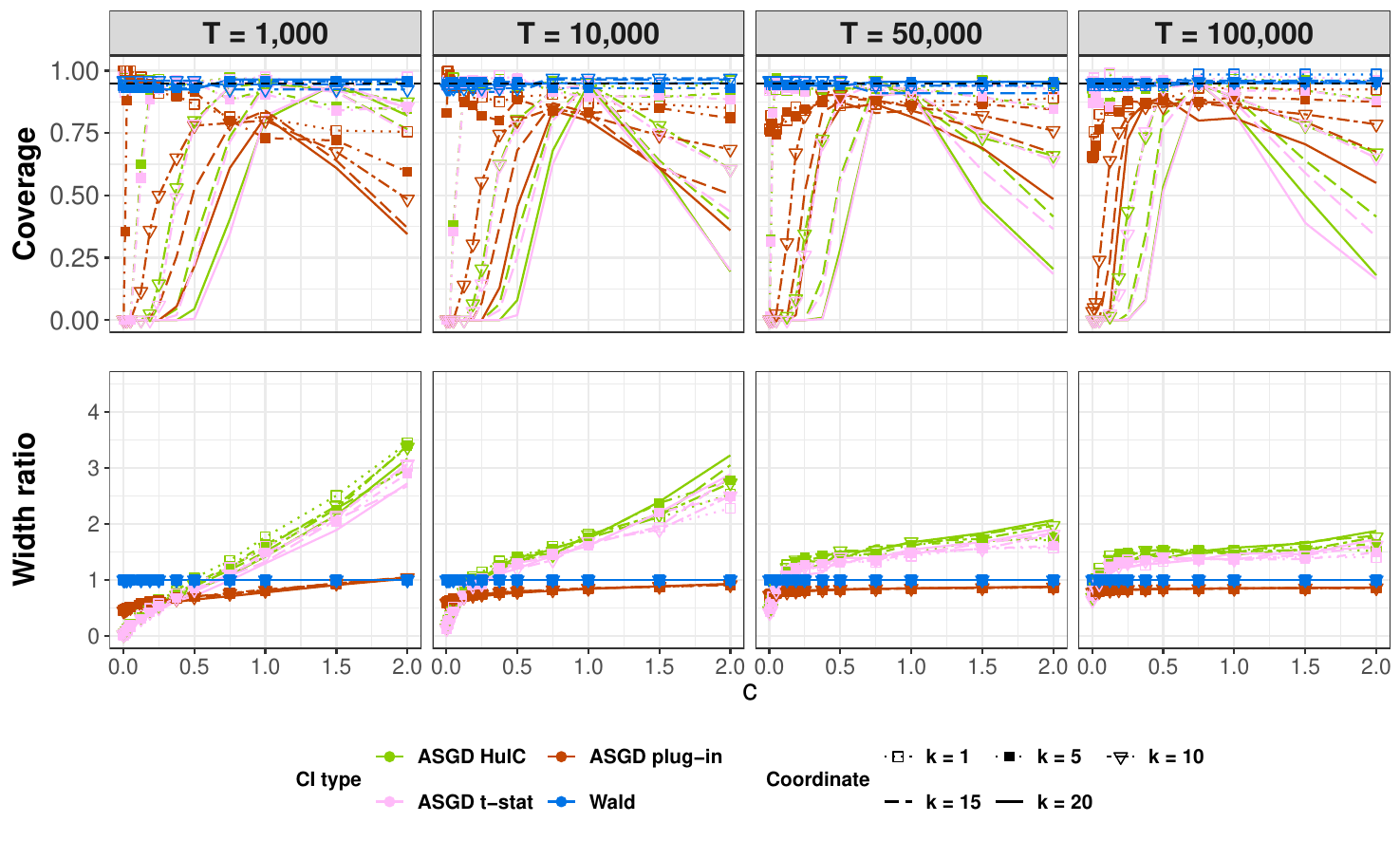}
\caption{ Logistic regression, Covariance = Equicorrelation, d = 20}
\label{fig:logistic_D20_EquiCorr_cov_wr}
\end{figure}

\begin{figure}[H]
\centering
\par\medskip
\includegraphics[width=1\textwidth]{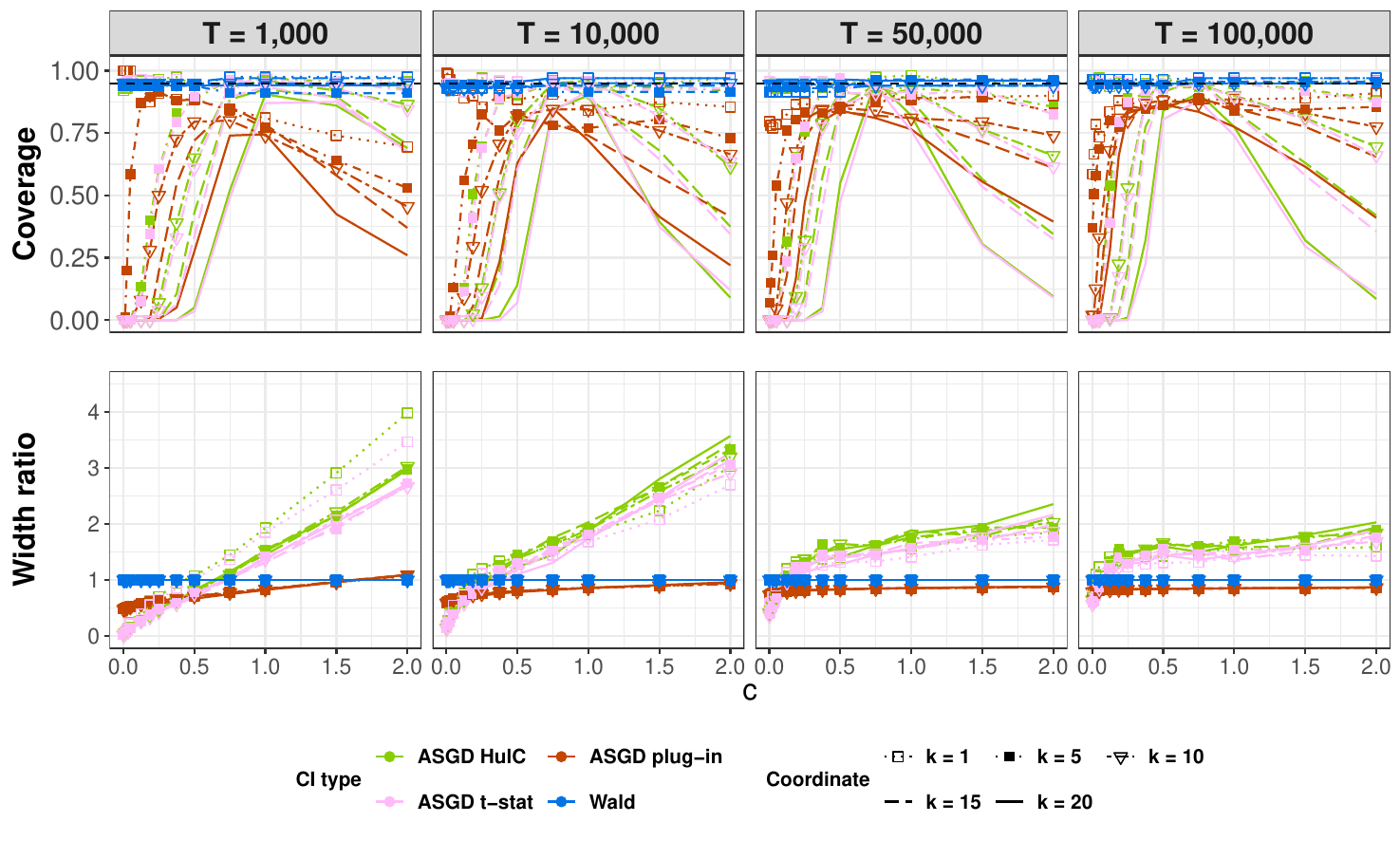}
\caption{ Logistic regression, Covariance = Toeplitz, d = 20}
\label{fig:logistic_D20_Toeplitz_cov_wr}
\end{figure}

%%%%%%%%%%%%%%%%
\newpage
\begin{figure}[H]
\centering
\par\medskip
\includegraphics[width=1\textwidth]{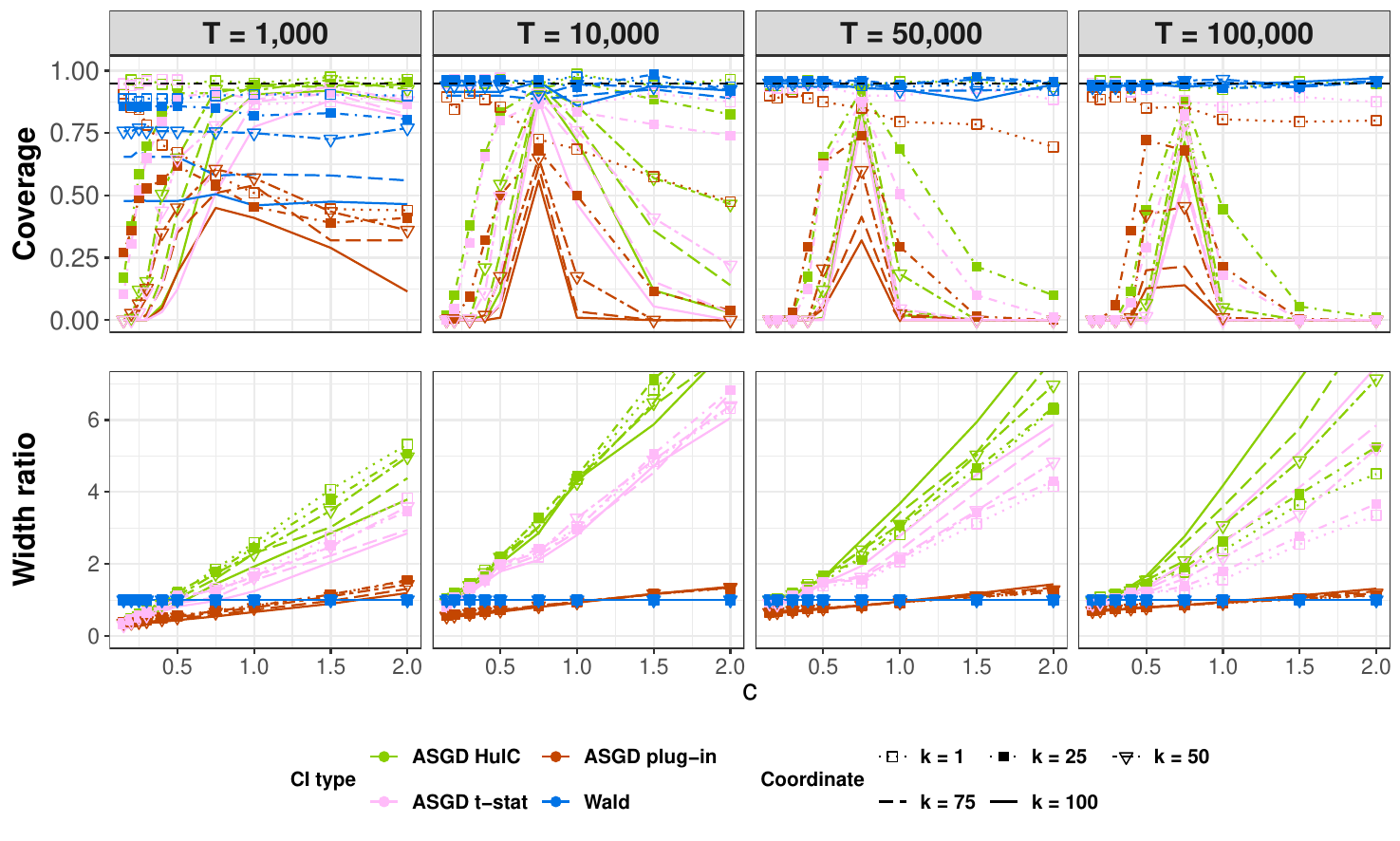}
\caption{ Logistic regression, Covariance = I, d = 100}
\label{fig:logistic_D100_I_cov_wr}
\end{figure}
\vspace{-25 pt}

\begin{figure}[H]
\centering
\par\medskip
\includegraphics[width=1\textwidth]{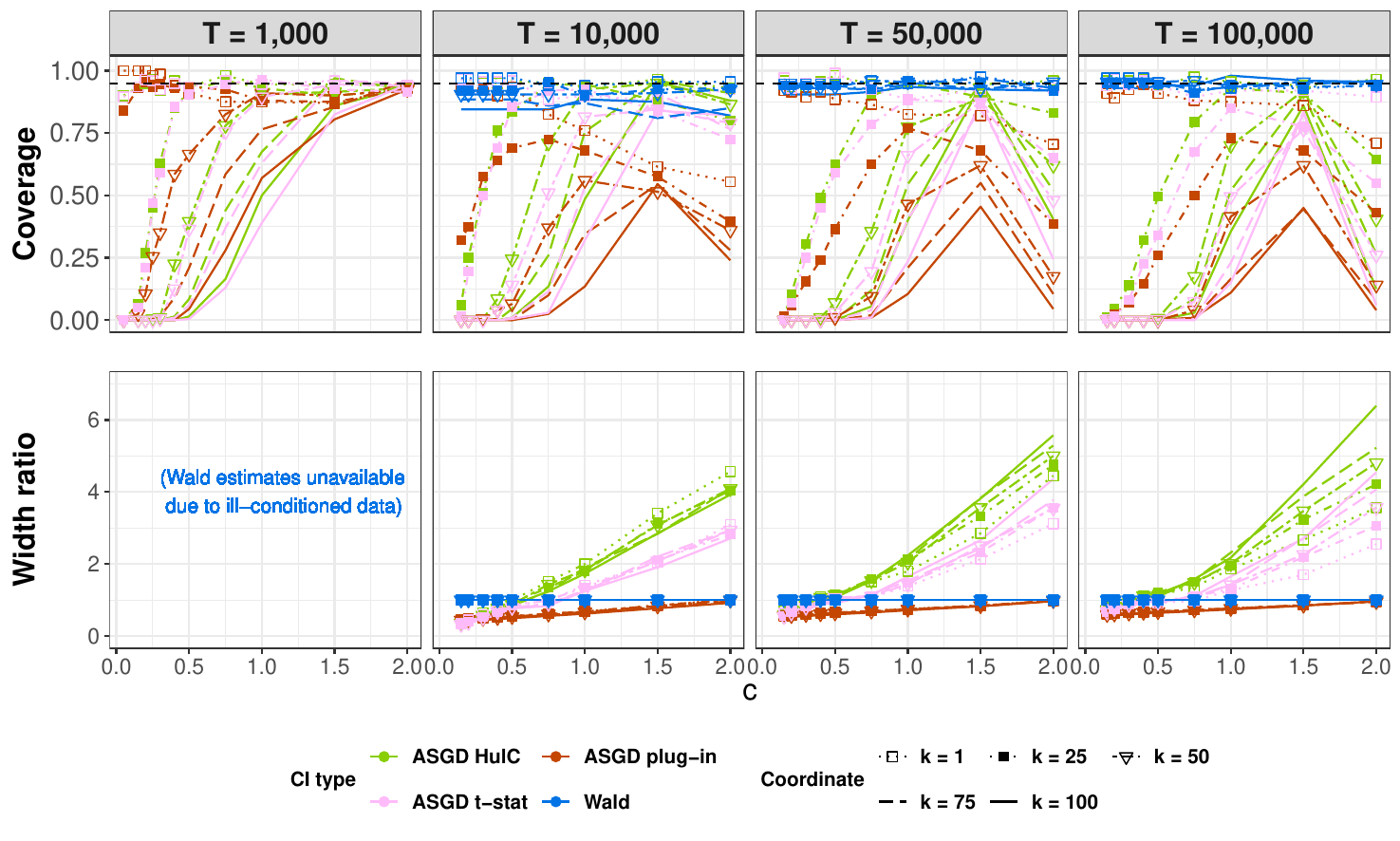}
\caption{ Logistic regression, Covariance = Equicorrelation, d = 100}
\label{fig:logistic_D100_EquiCorr_cov_wr}
\caption*{\footnotesize Note: The Wald estimates are unavailable as a baseline for $n=10^3$ given that the matrix $X^\top X$ was ill-conditioned for all $200$ experiments.}
\end{figure}

\begin{figure}[H]
\centering
\par\medskip
\includegraphics[width=1\textwidth]{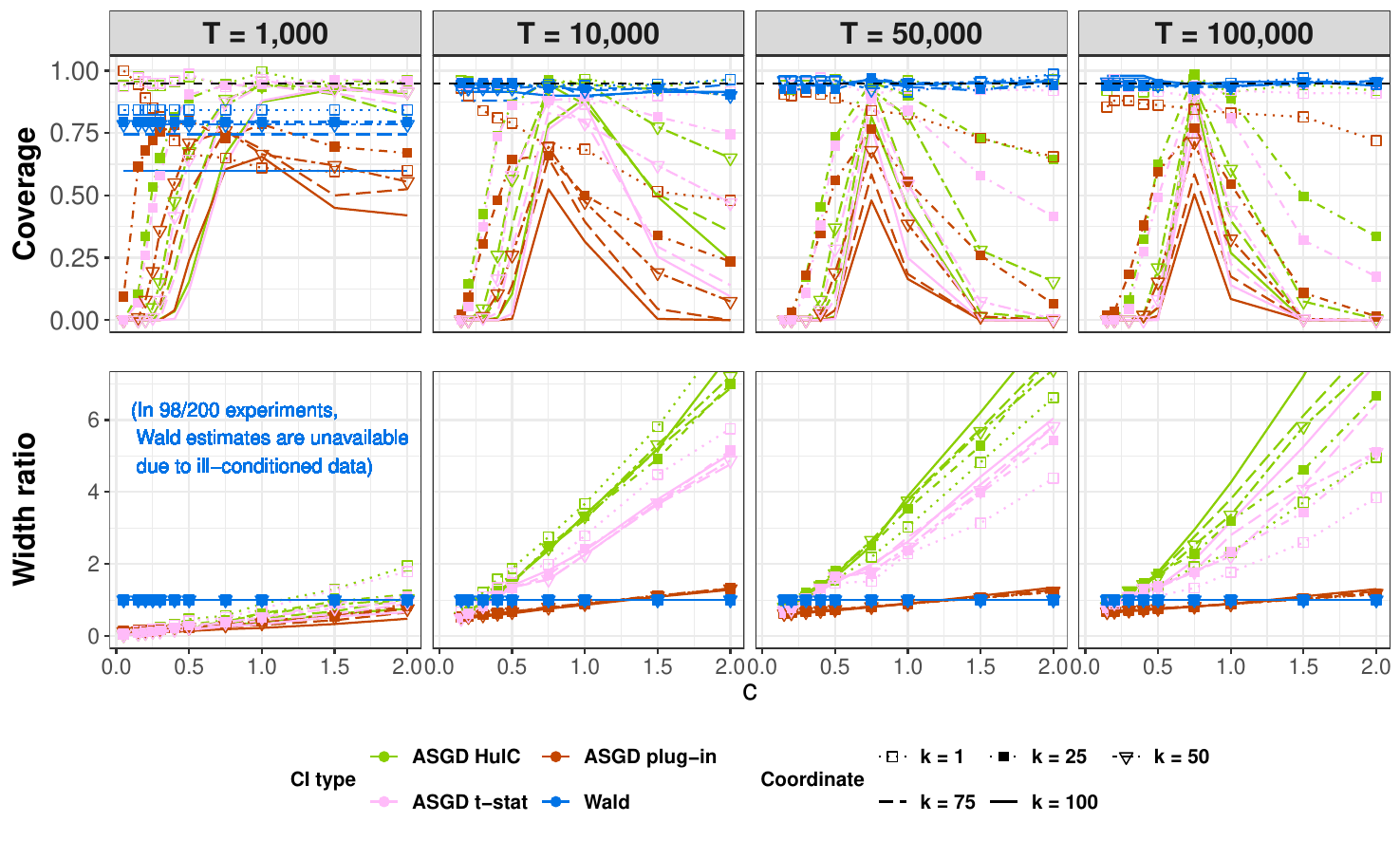}
\caption{ Logistic regression, Covariance = Toeplitz, d = 100}
\label{fig:logistic_D100_Toeplitz_cov_wr}
\caption*{\footnotesize Note: In $98$ out of $200$ experiments, the matrix $X^\top X$ was ill-conditioned, hence the Wald estimates could not be produced. The bottom left plot (width ratios for $n=10^3$) represents median width ratios across the $102$ experiments for which the Wald estimates were available. The top left plot (coverage for $n=10^3$) shows median coverage rates across all $200$ experiments for $c$-dependent methods (HulC, $t$-stat, and ASGD plug-in) and across $102$ experiments for the Wald estimator.}
\end{figure}

\subsection{Last-Iterate-Implicit-SGD}

\subsubsection{Linear regression}\label{app:OLS_plots_isgd}
\vspace{-20 pt}

\begin{figure}[H]
\centering
 % \par\medskip
\includegraphics[width=1\textwidth]{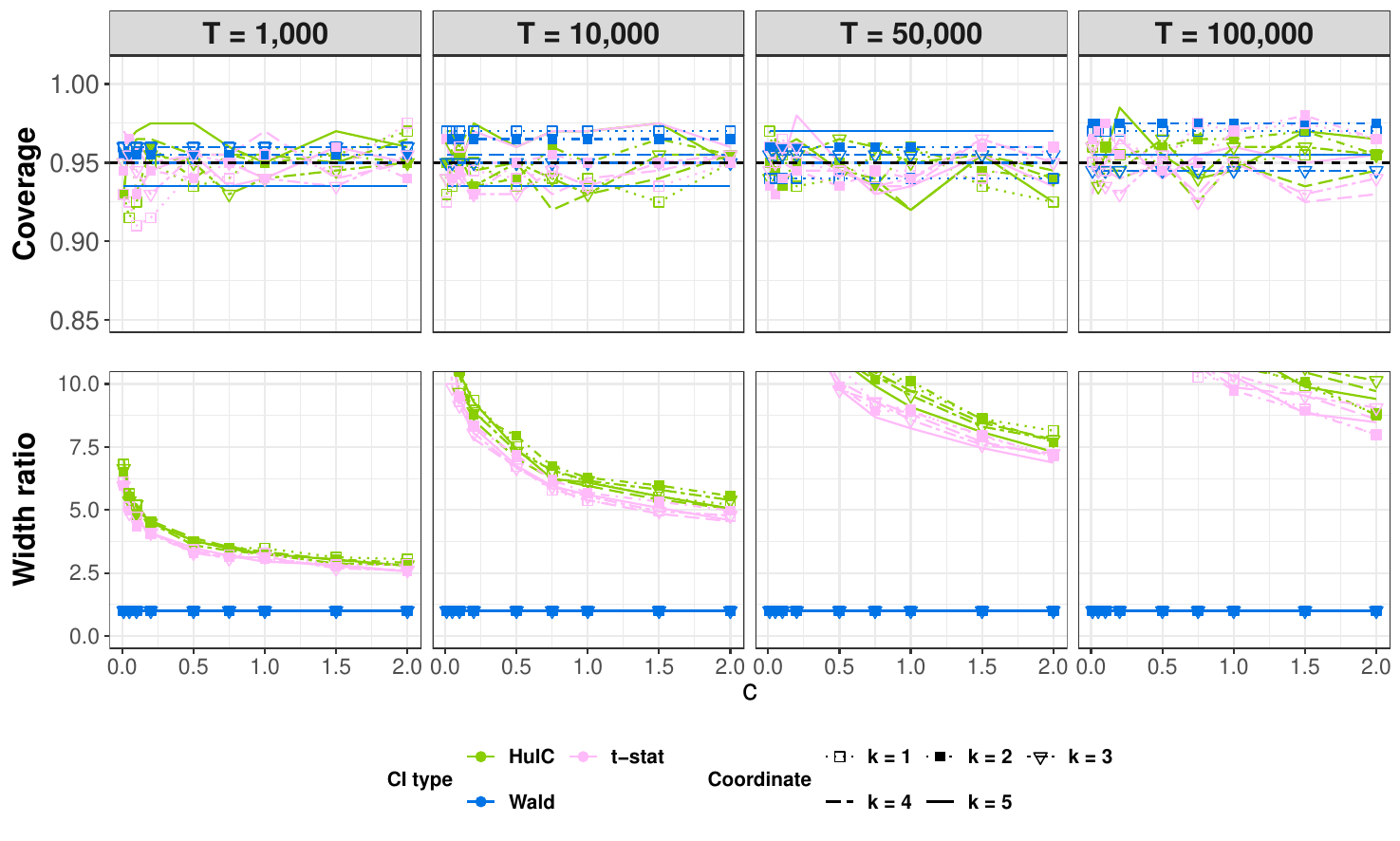}
\caption{Linear regression (last-iterate-implicit-SGD), Covariance = I, d = 5}
\label{fig:linear_D5_I_cov_wr_ISGD_initTRUE}
\end{figure}

\begin{figure}[H]
\centering
 % \par\medskip
\includegraphics[width=1\textwidth]{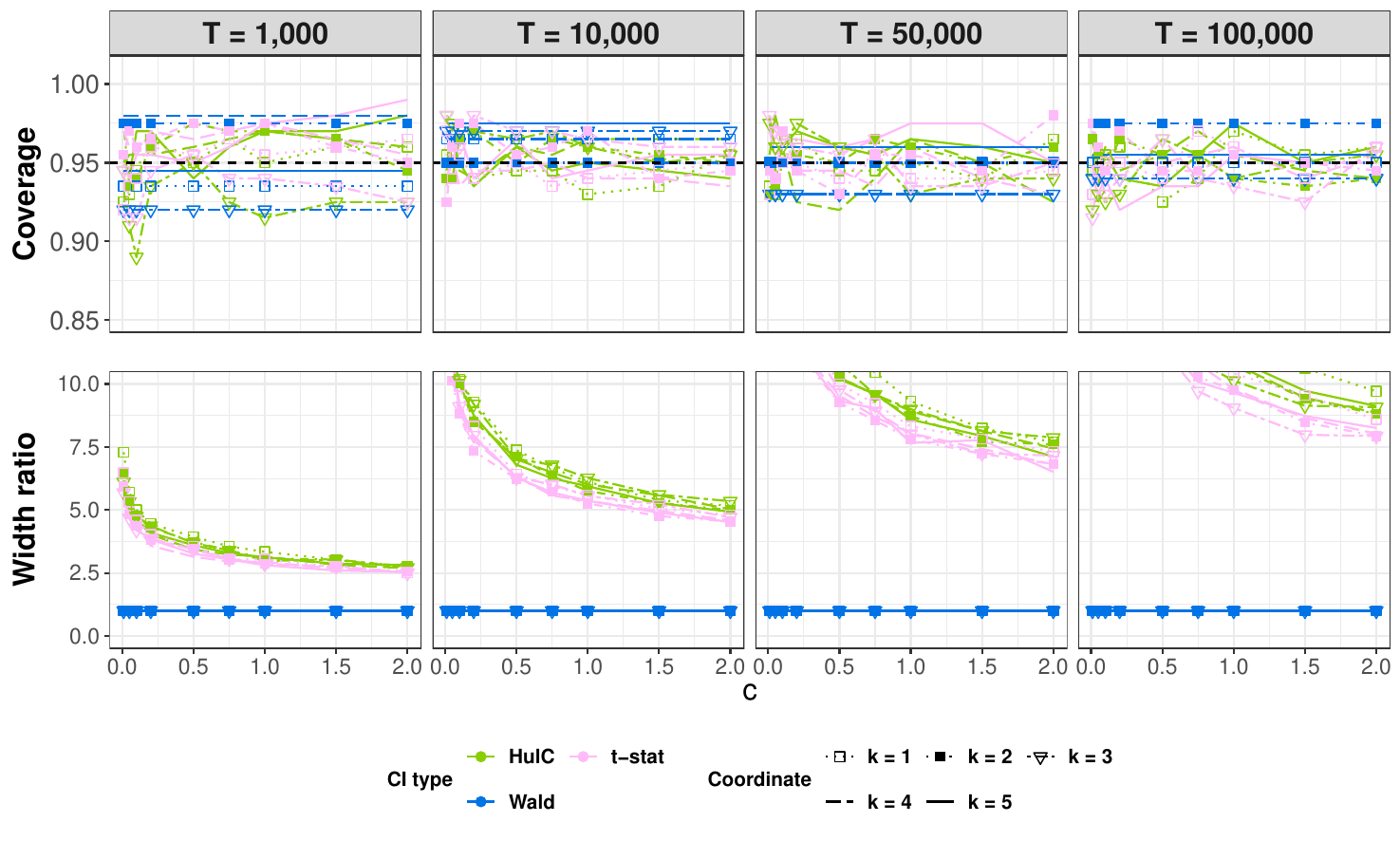}
\caption{Linear regression (last-iterate-implicit-SGD), Covariance = Equicorrelation, d = 5}
\label{fig:linear_D5_EquiCorr_cov_wr_ISGD_initTRUE}
\end{figure}

For linear regression using last-iterate-implicit-SGD, $d=5$, and Toeplitz covariance, see Figure~\ref{fig:linear_D5_Toeplitz_cov_wr_ISGD_initTRUE}.

\begin{figure}[H]
\centering
 % \par\medskip
\includegraphics[width=1\textwidth]{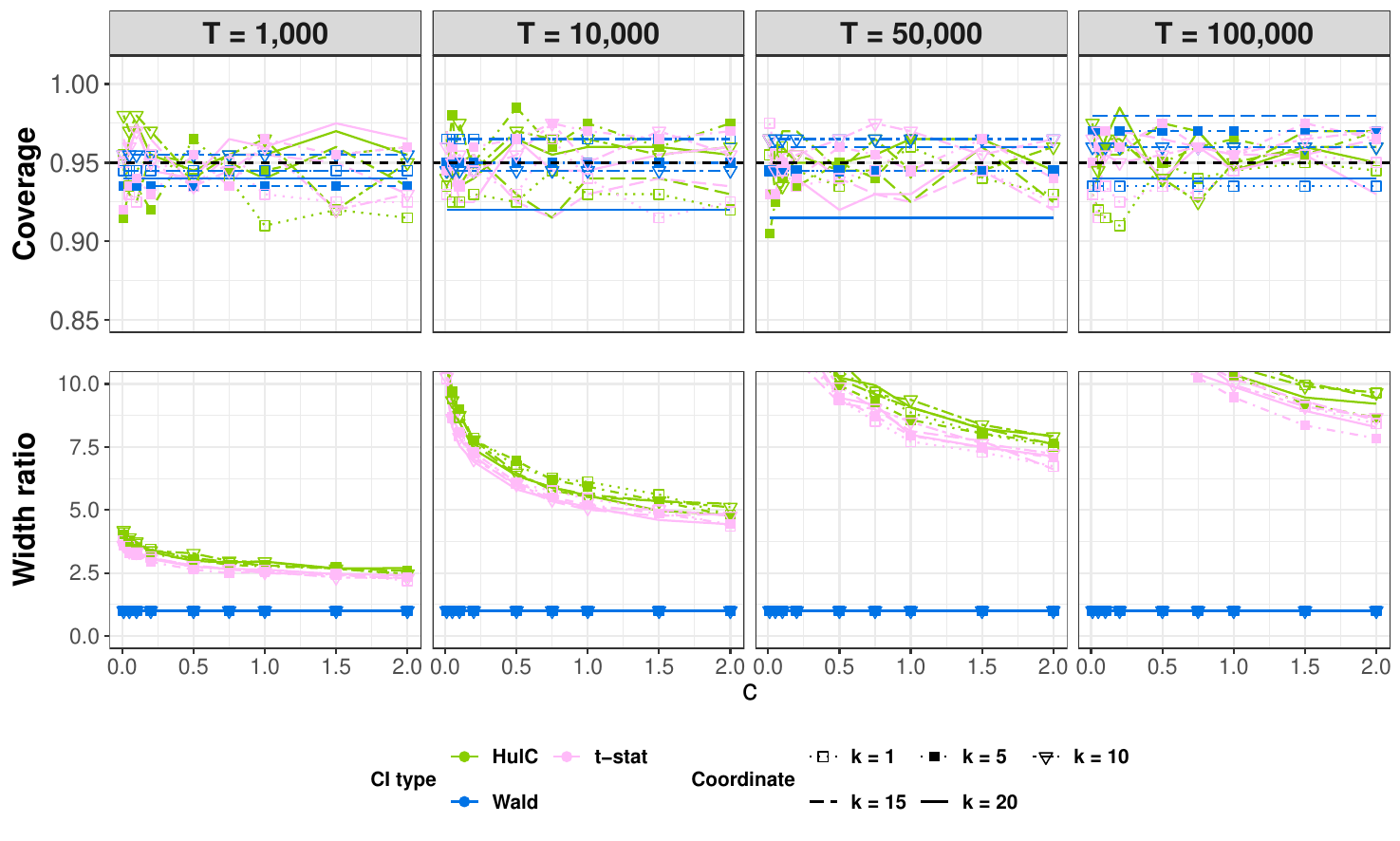}
\caption{Linear regression (last-iterate-implicit-SGD), Covariance = I, d = 20}
\label{fig:linear_D20_I_cov_wr_ISGD_initTRUE}
\end{figure}

\begin{figure}[H]
\centering
 % \par\medskip
\includegraphics[width=1\textwidth]{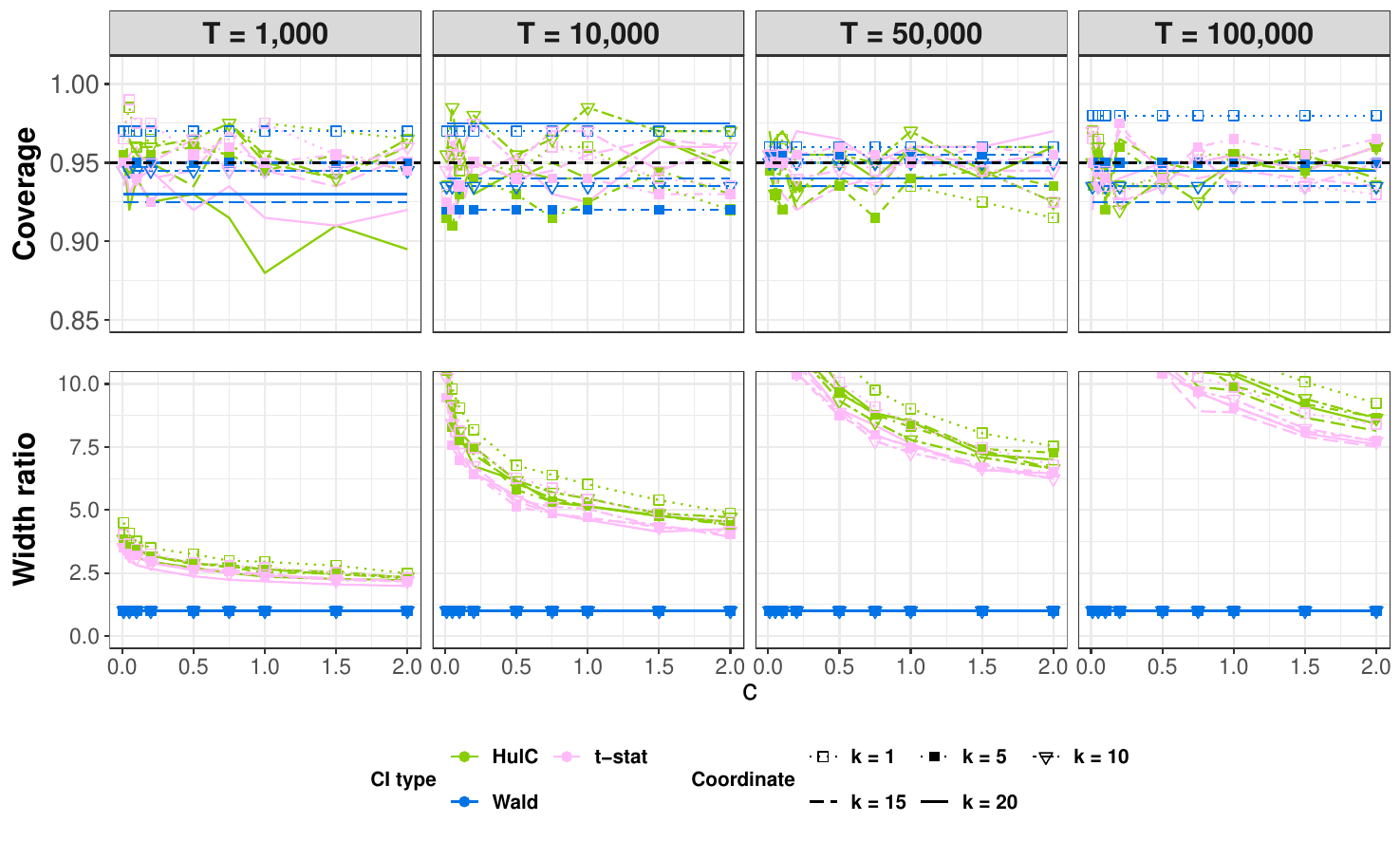}
\caption{Linear regression (last-iterate-implicit-SGD), Covariance = Equicorrelation, d = 20}
\label{fig:linear_D20_EquiCorr_cov_wr_ISGD_initTRUE}
\end{figure}

\begin{figure}[H]
\centering
 % \par\medskip
\includegraphics[width=1\textwidth]{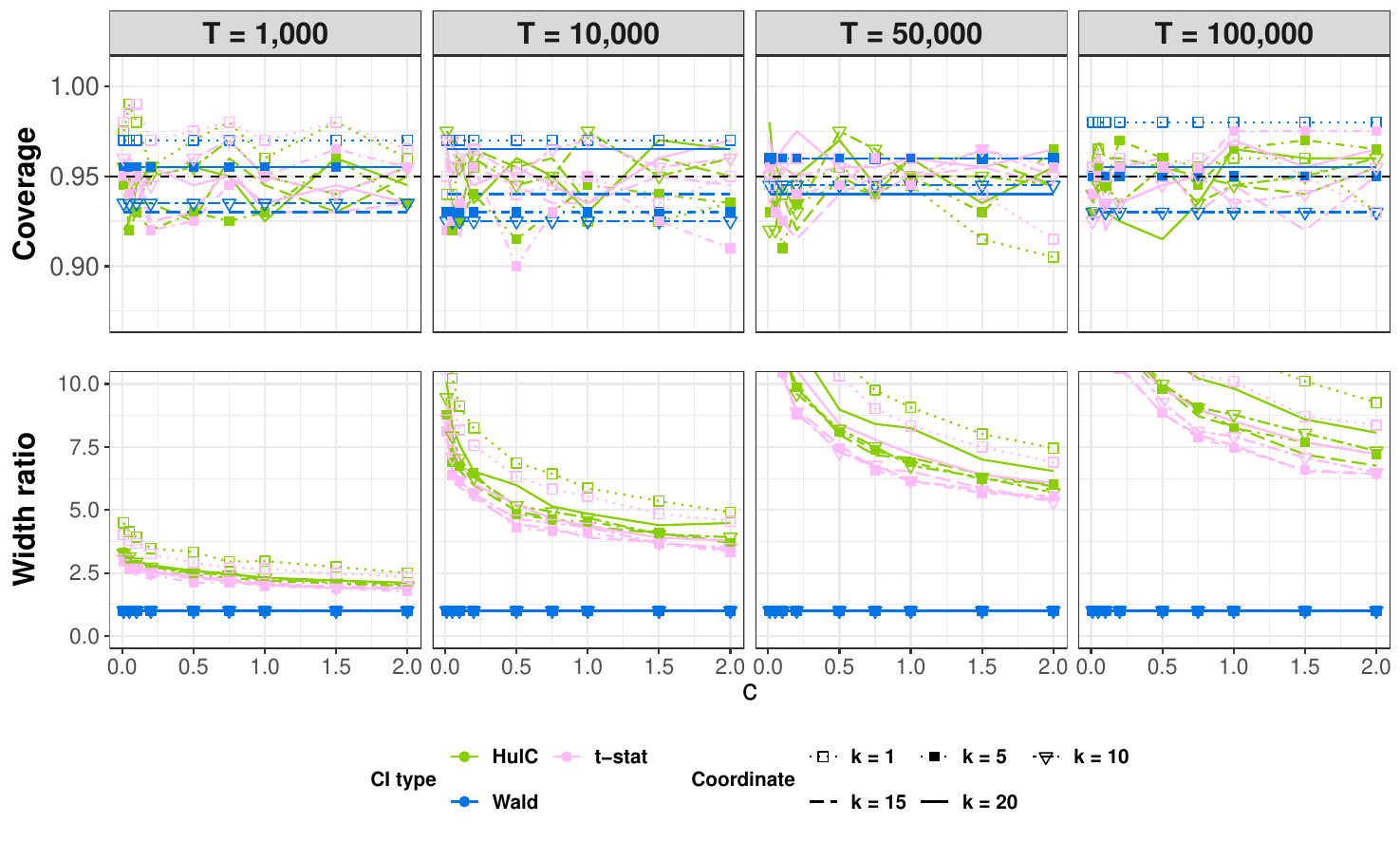}
\caption{Linear regression (last-iterate-implicit-SGD), Covariance = Toeplitz, d = 20}
\label{fig:linear_D20_Toeplitz_cov_wr_ISGD_initTRUE}
\end{figure}

\begin{figure}[H]
\centering
 % \par\medskip
\includegraphics[width=1\textwidth]{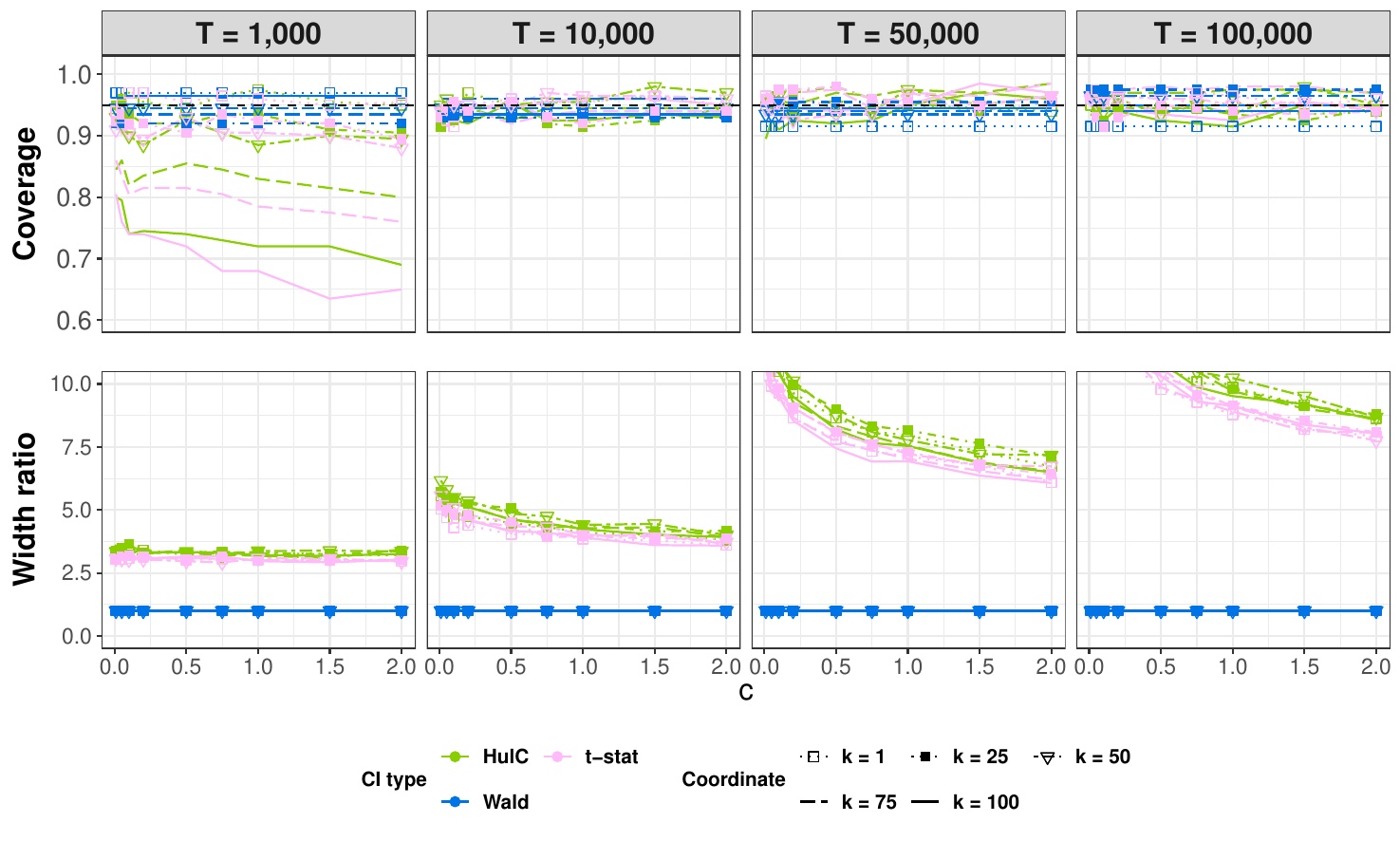}
\caption{Linear regression (last-iterate-implicit-SGD), Covariance = I, d = 100}
\label{fig:linear_D100_I_cov_wr_ISGD_initTRUE}
\end{figure}

\begin{figure}[H]
\centering
 % \par\medskip
\includegraphics[width=1\textwidth]{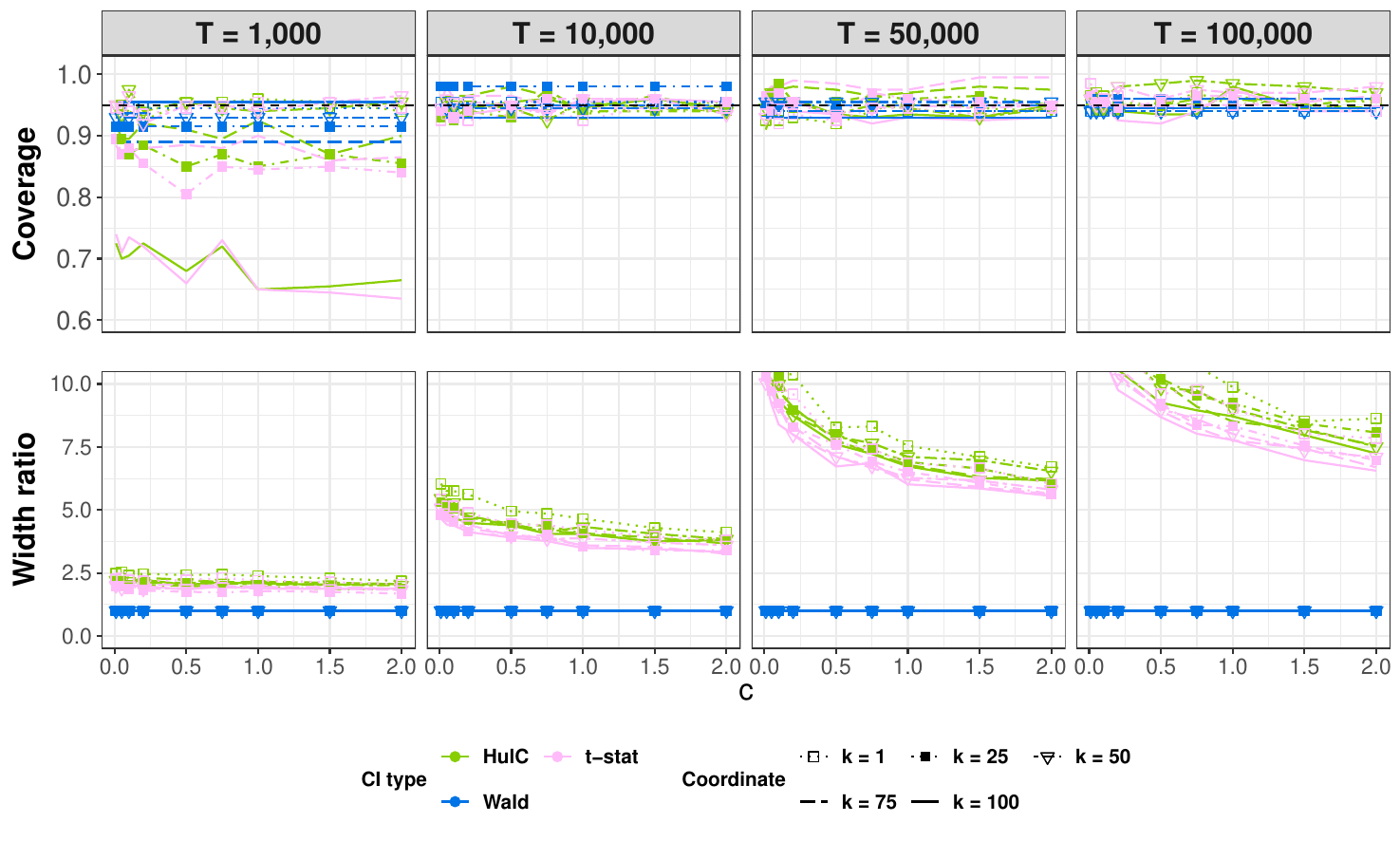}
\caption{Linear regression (last-iterate-implicit-SGD), Covariance = Equicorrelation, d = 100}
\label{fig:linear_D100_EquiCorr_cov_wr_ISGD_initTRUE}
\end{figure}

\begin{figure}[H]
\centering
 % \par\medskip
\includegraphics[width=1\textwidth]{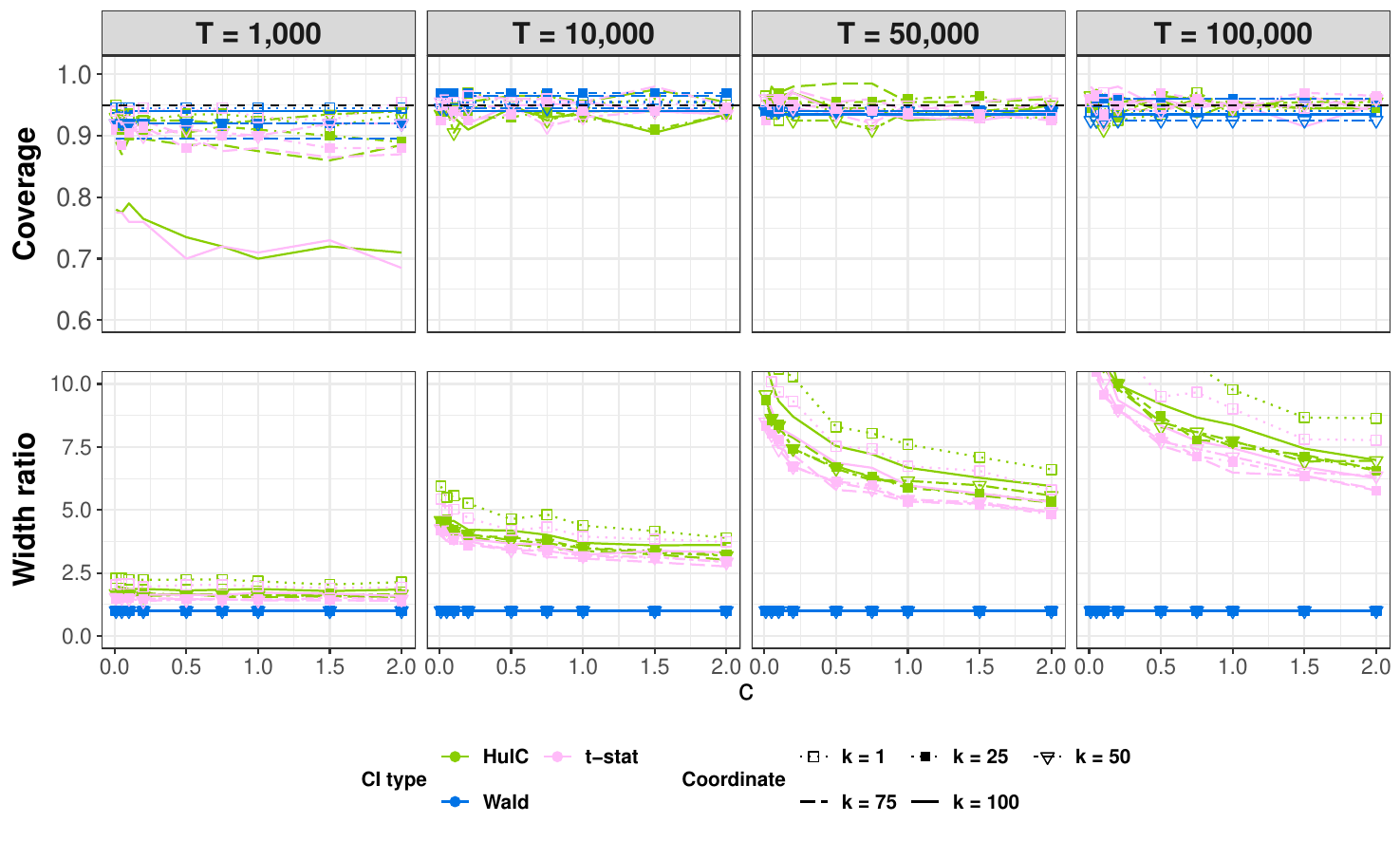}
\caption{Linear regression (last-iterate-implicit-SGD), Covariance = Toeplitz, d = 100}
\label{fig:linear_D100_Toeplitz_cov_wr_ISGD_initTRUE}
\end{figure}

\subsubsection{Logistic regression}\label{app:Logistic_plots_ISGD}
\vspace{-20 pt}

\begin{figure}[H]
\centering
 % \par\medskip
\includegraphics[width=1\textwidth]{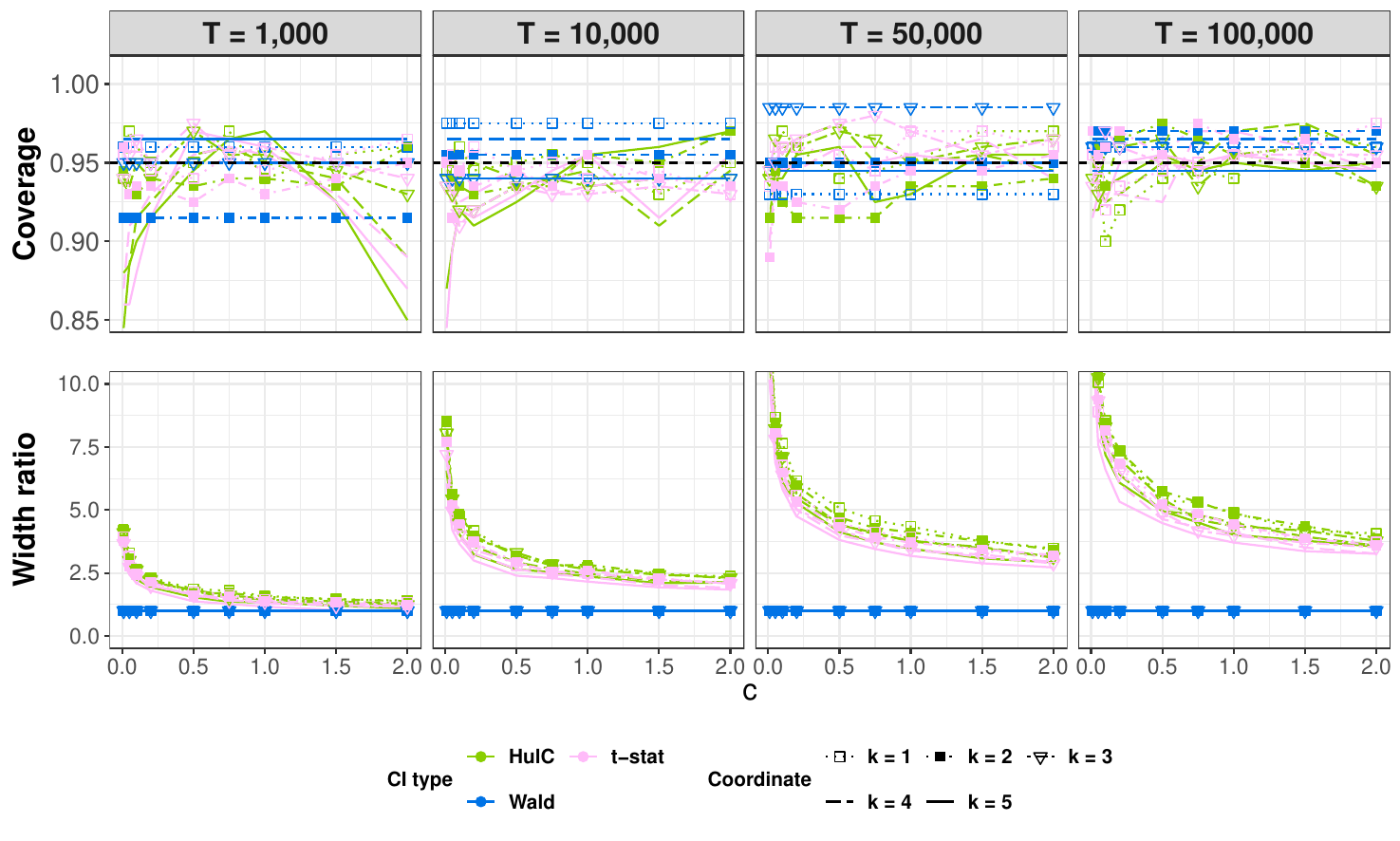}
\caption{Logistic regression (last-iterate-implicit-SGD), Covariance = I, d = 5}
\label{fig:logistic_D5_I_cov_wr_ISGD_initTRUE}
\end{figure}

\begin{figure}[H]
\centering
 % \par\medskip
\includegraphics[width=1\textwidth]{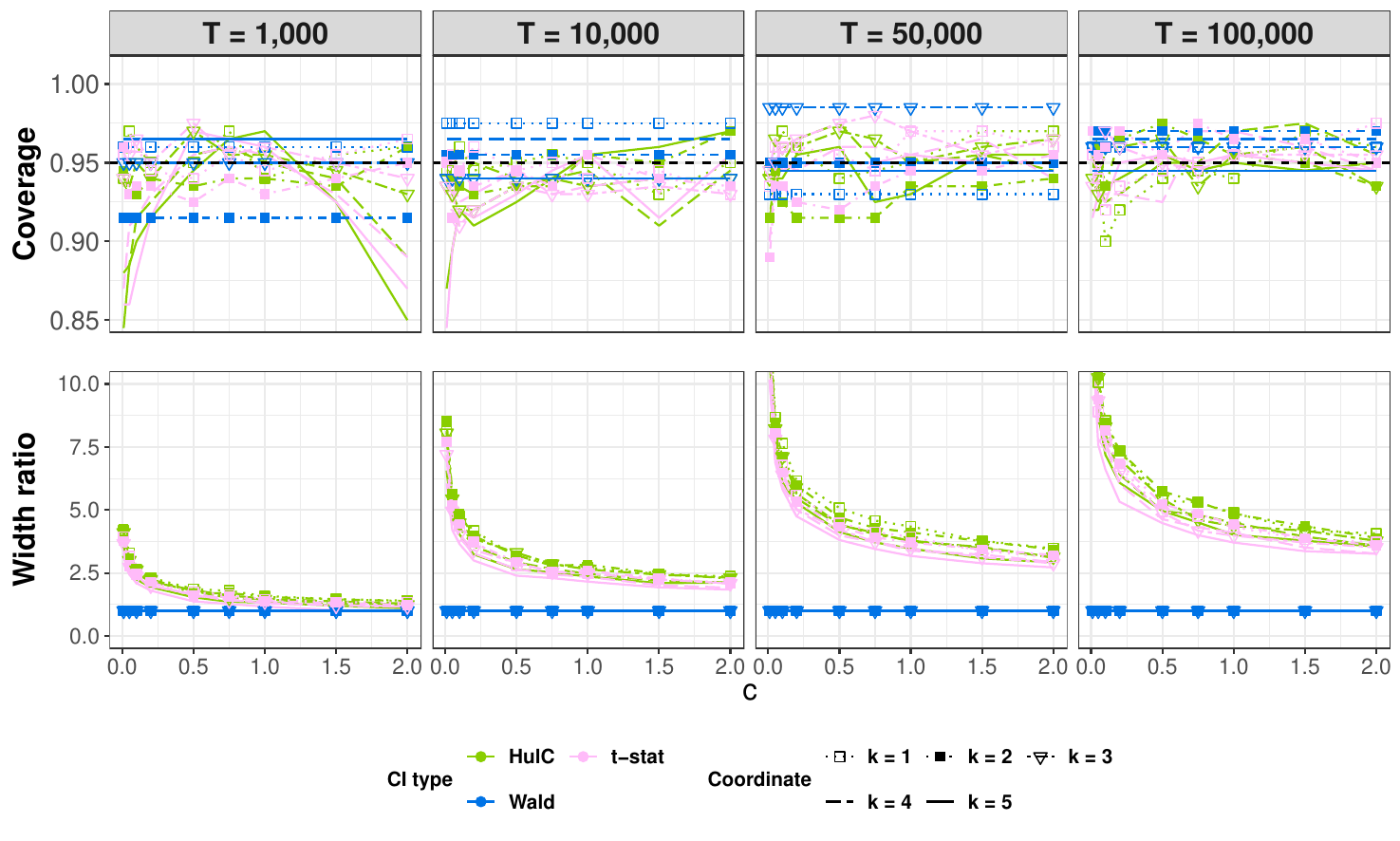}
\caption{Logistic regression (last-iterate-implicit-SGD), Covariance = Equicorrelation, d = 5}
\label{fig:logistic_D5_Equicorr_cov_wr_ISGD_initTRUE}
\end{figure}

For logistic regression using last-iterate-implicit-SGD, $d=5$, and Toeplitz covariance, see Figure~\ref{fig:logistic_D5_Toeplitz_cov_wr_ISGD_initTRUE}.

\begin{figure}[H]
\centering
 % \par\medskip
\includegraphics[width=1\textwidth]{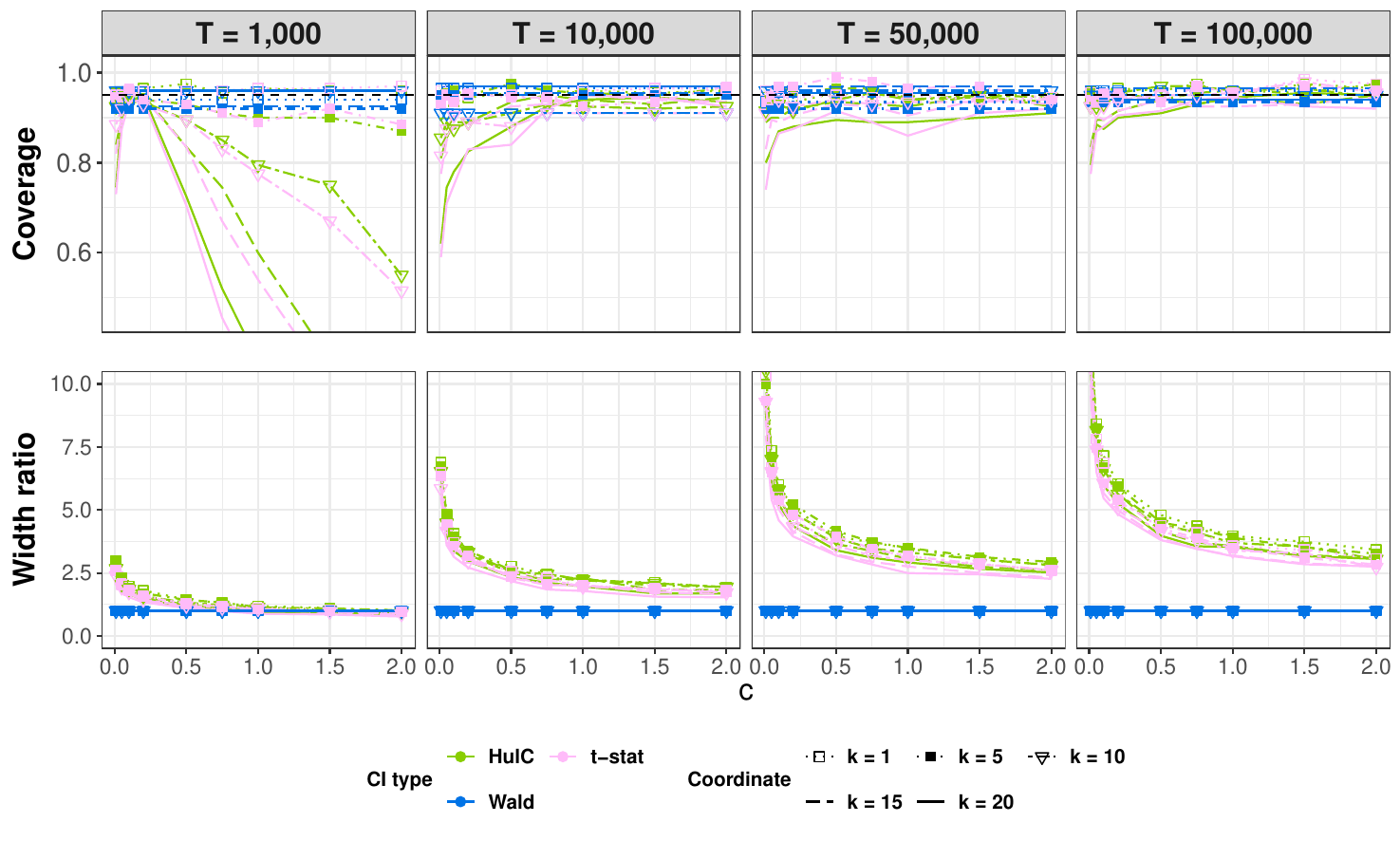}
\caption{Logistic regression (last-iterate-implicit-SGD), Covariance = I, d = 20}
\label{fig:logistic_D20_I_cov_wr_ISGD_initTRUE}
\end{figure}

\begin{figure}[H]
\centering
 % \par\medskip
\includegraphics[width=1\textwidth]{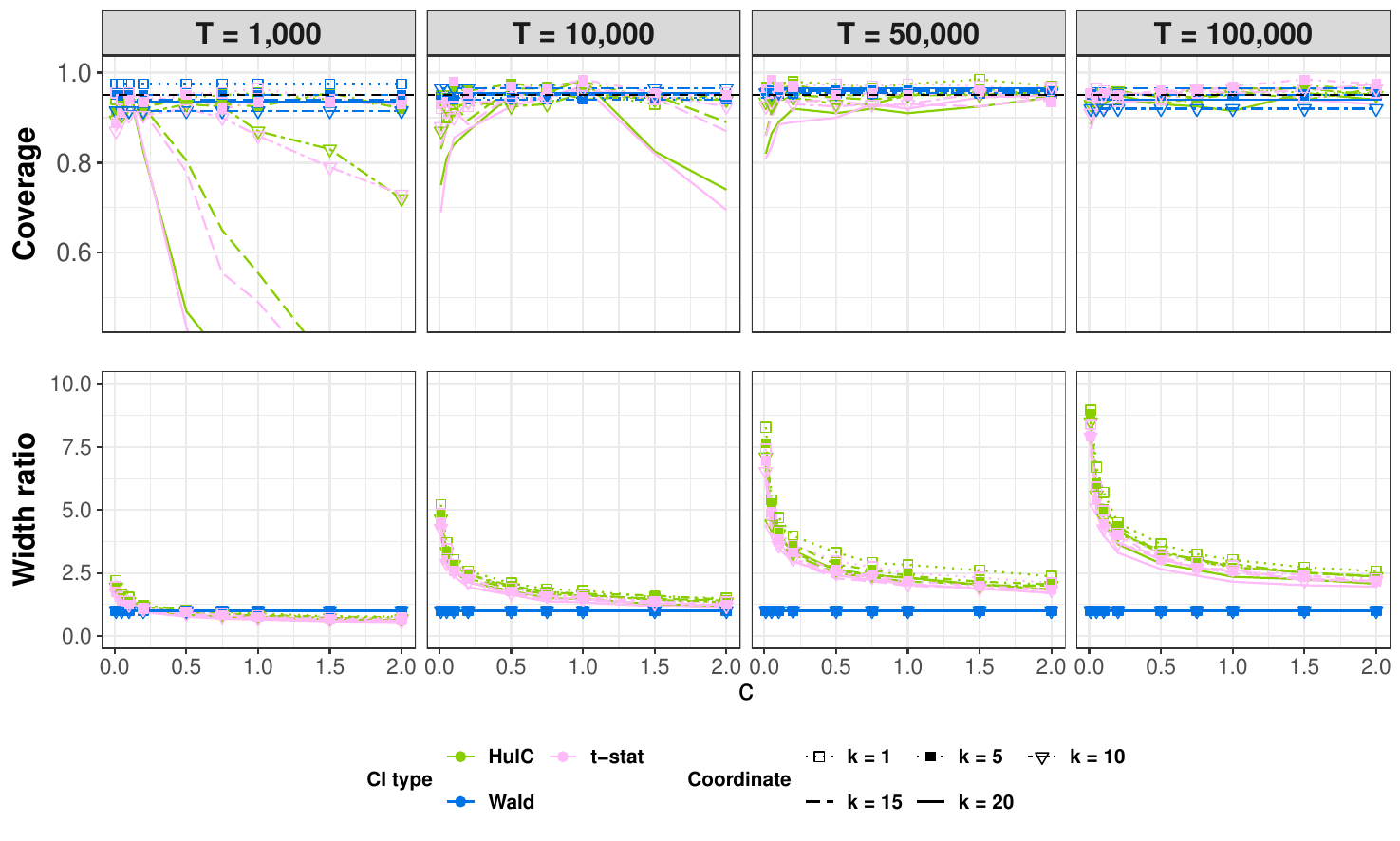}
\caption{Logistic regression (last-iterate-implicit-SGD), Covariance = Equicorrelation, d = 20}
\label{fig:logistic_D20_EquiCorr_cov_wr_ISGD_initTRUE}
\end{figure}

\begin{figure}[H]
\centering
 % \par\medskip
\includegraphics[width=1\textwidth]{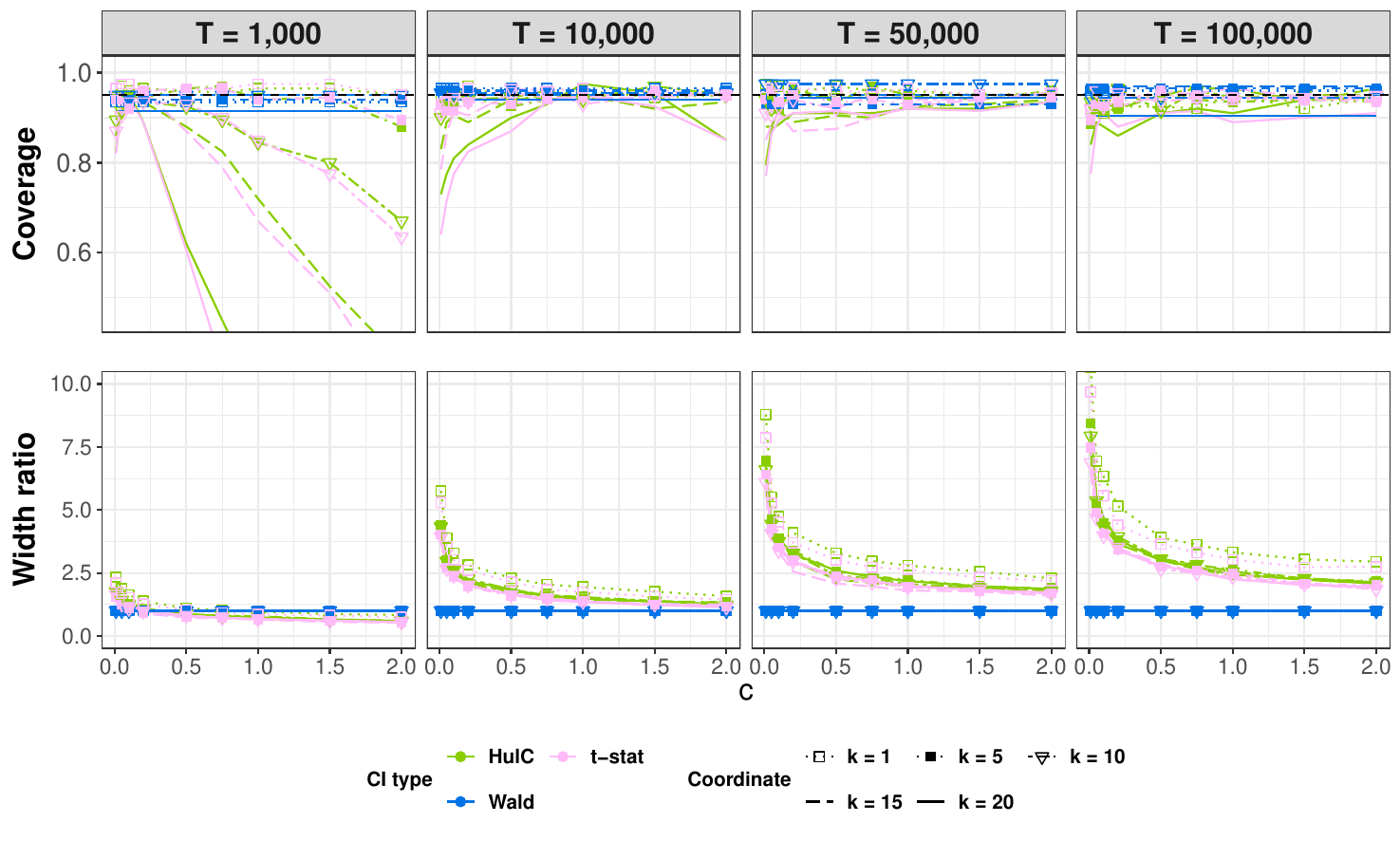}
\caption{Logistic regression (last-iterate-implicit-SGD), Covariance = Toeplitz, d = 20}
\label{fig:logistic_D20_Toeplitz_cov_wr_ISGD_initTRUE}
\end{figure}

Note: Plots for logistic regression using last-iterate-implicit-SGD, $d=100$, did not always converge and/or Wald estimates were not computable. Please see the online tool.\footnote{\url{https://public.tableau.com/app/profile/selina.carter6629/viz/OnlineinferencesimulationsOLSandlogisticregression/Coverageandwidthratio_paper}}

%TODO: update these plots 
% \begin{figure}[H]
% \centering
%  % \par\medskip
% \includegraphics[width=1\textwidth]{figures/extra_sims/logistic_D100_I_cov_wr_ISGD_initTRUE}
% \caption{Logistic regression (last-iterate-implicit-SGD), Covariance = I, d = 100}
% \label{fig:logistic_D100_I_cov_wr_ISGD_initTRUE}
% \end{figure}

% \begin{figure}[H]
% \centering
%  % \par\medskip
% \includegraphics[width=1\textwidth]{figures/extra_sims/logistic_D100_EquiCorr_cov_wr_ISGD_initTRUE}
% \caption{Logistic regression (last-iterate-implicit-SGD), Covariance = Equicorrelation, d = 100}
% \label{fig:logistic_D100_EquiCorr_cov_wr_ISGD_initTRUE}
% \end{figure}

% \begin{figure}[H]
% \centering
%  % \par\medskip
% \includegraphics[width=1\textwidth]{figures/extra_sims/logistic_D100_Toeplitz_cov_wr_ISGD_initTRUE}
% \caption{Logistic regression (last-iterate-implicit-SGD), Covariance = Toeplitz, d = 100}
% \label{fig:logistic_D100_Toeplitz_cov_wr_ISGD_initTRUE}
% \end{figure}

\subsection{Average-Iterate-Implicit-SGD}

\subsubsection{Linear regression}\label{app:OLS_plots_aisgd}
\vspace{-20 pt}

\begin{figure}[H]
\centering
 % \par\medskip
\includegraphics[width=1\textwidth]{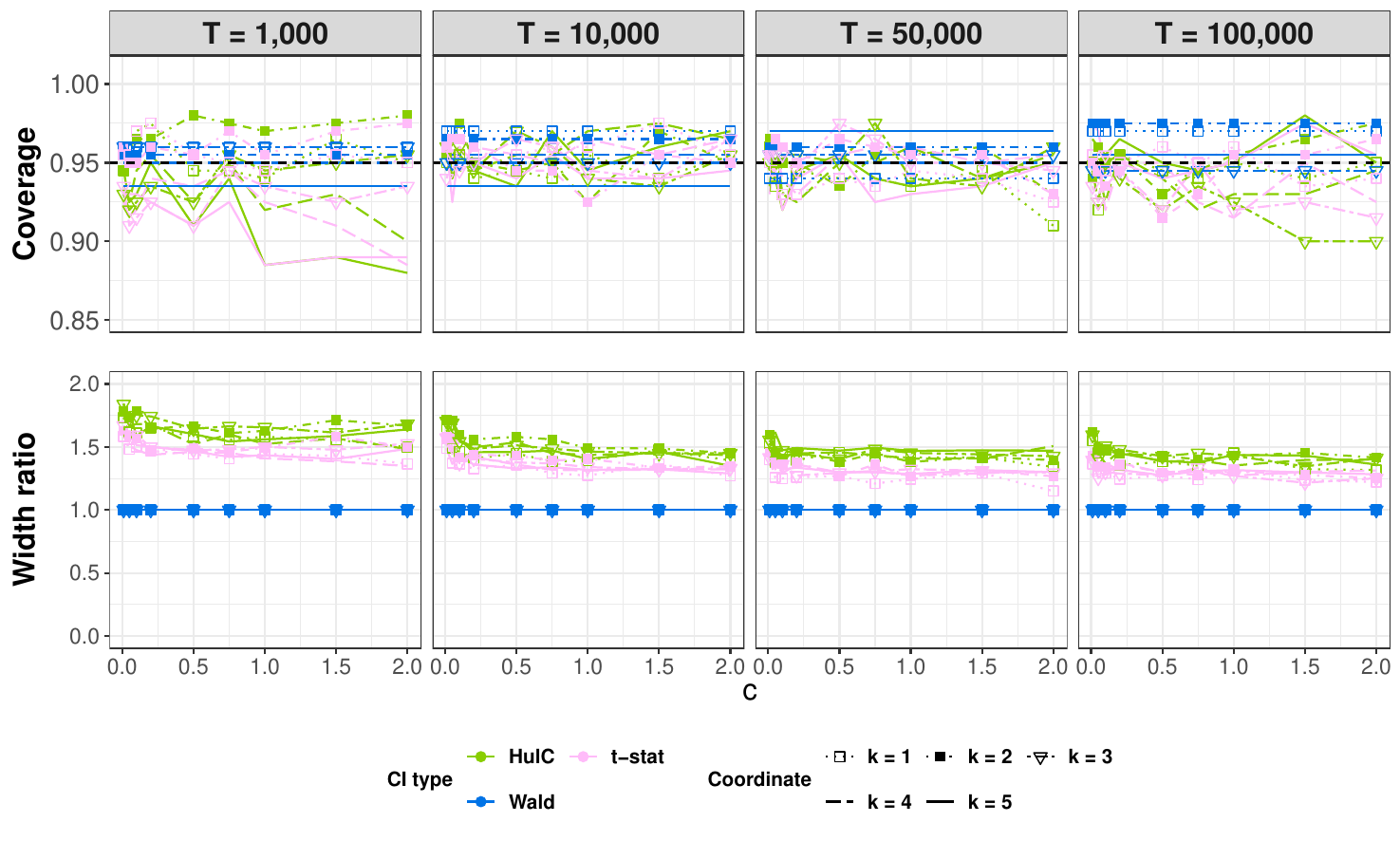}
\caption{Linear regression  (average-iterate-implicit-SGD), Covariance = I, d = 5}
\label{fig:linear_D5_I_cov_wr_AISGD_initTRUE}
\end{figure}

\begin{figure}[H]
\centering
 % \par\medskip
\includegraphics[width=1\textwidth]{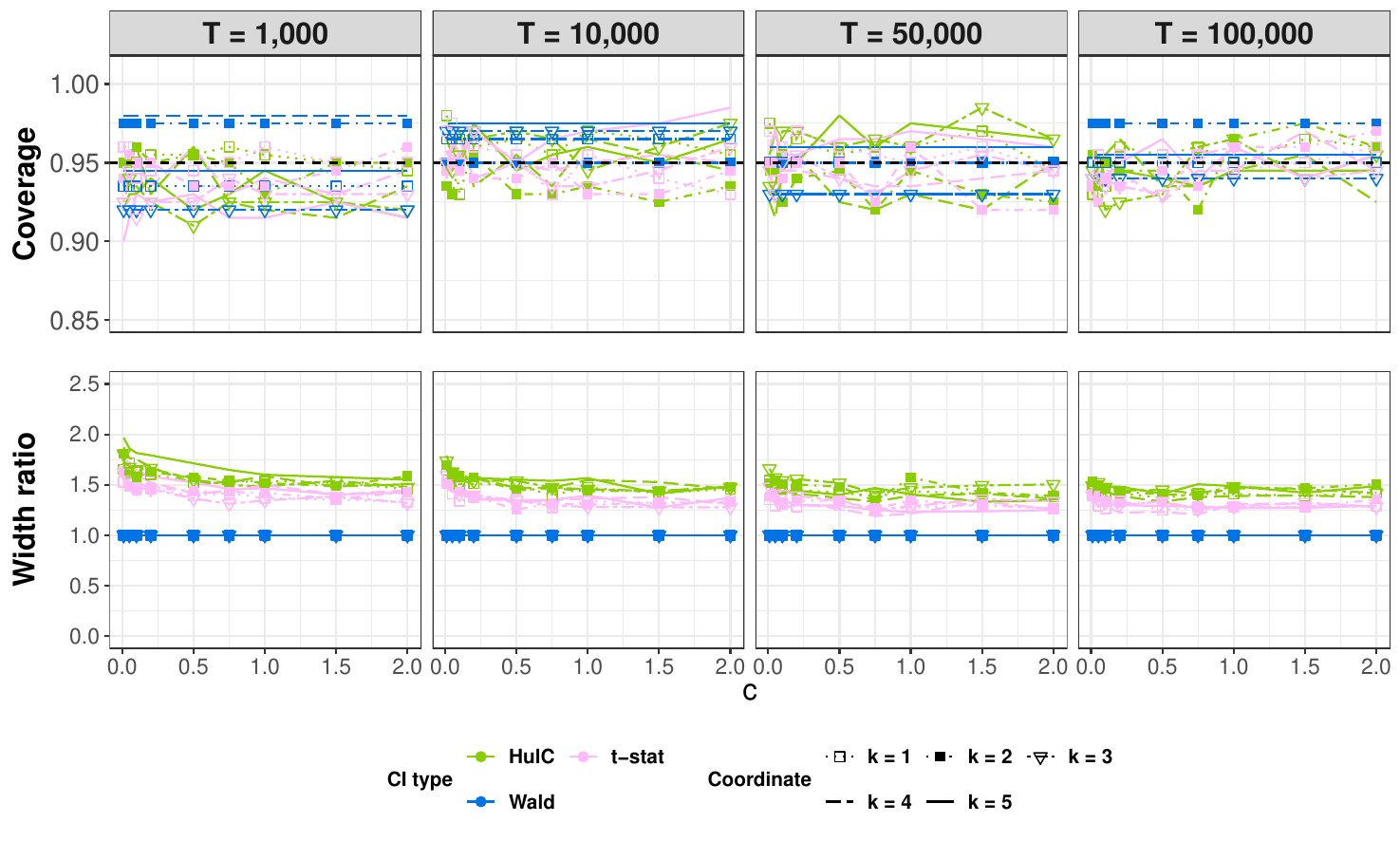}
\caption{Linear regression  (average-iterate-implicit-SGD), Covariance = Equicorrelation, d = 5}
\label{fig:linear_D5_Equicorr_cov_wr_AISGD_initTRUE}
\end{figure}

For linear regression using average-iterate-implicit-SGD, $d=5$, and Toeplitz covariance, see Figure~\ref{fig:linear_D5_Toeplitz_cov_wr_AISGD_initTRUE}.

\begin{figure}[H]
\centering
 % \par\medskip
\includegraphics[width=1\textwidth]{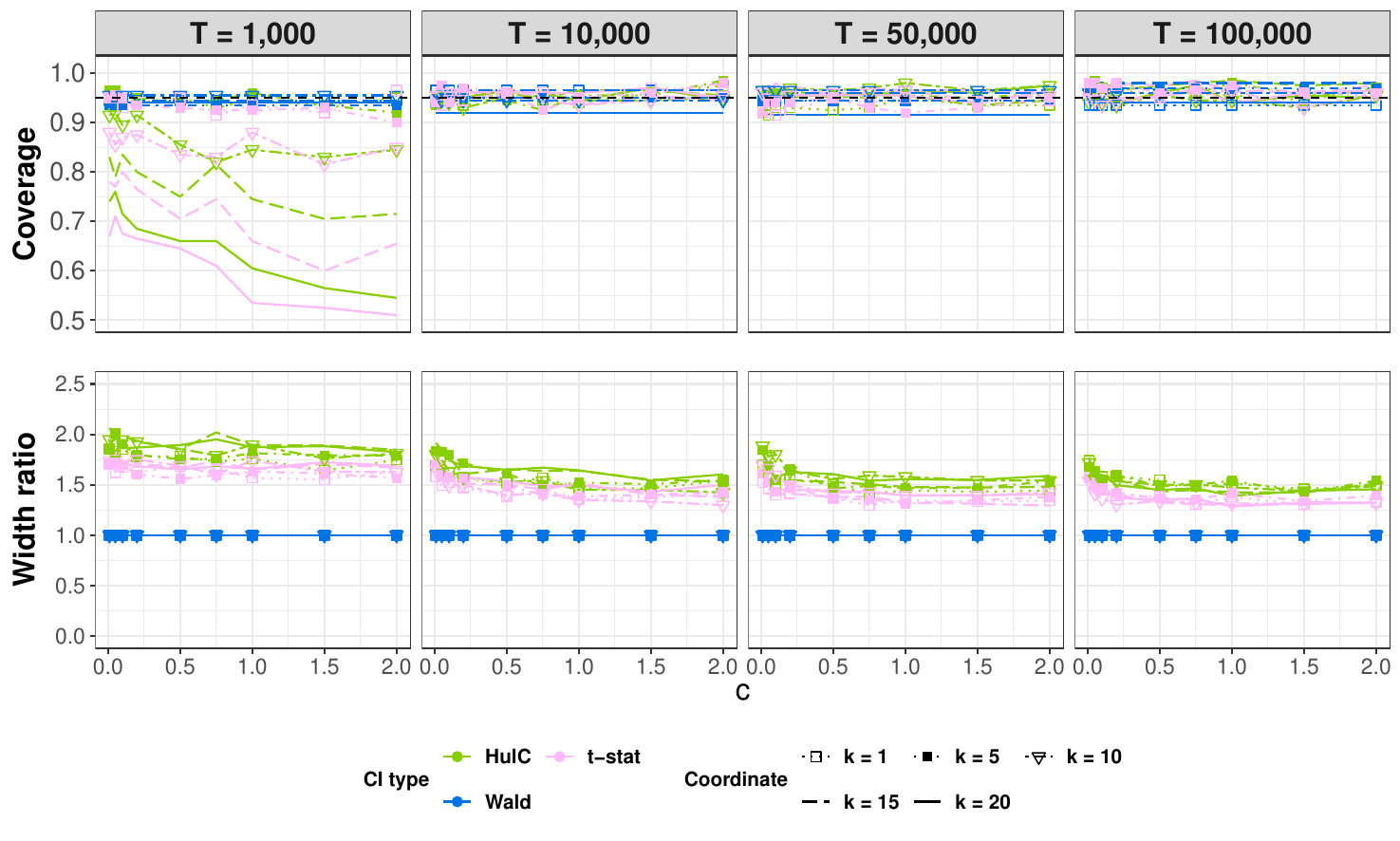}
\caption{Linear regression  (average-iterate-implicit-SGD), Covariance = I, d = 20}
\label{fig:linear_D20_I_cov_wr_AISGD_initTRUE}
\end{figure}

\begin{figure}[H]
\centering
 % \par\medskip
\includegraphics[width=1\textwidth]{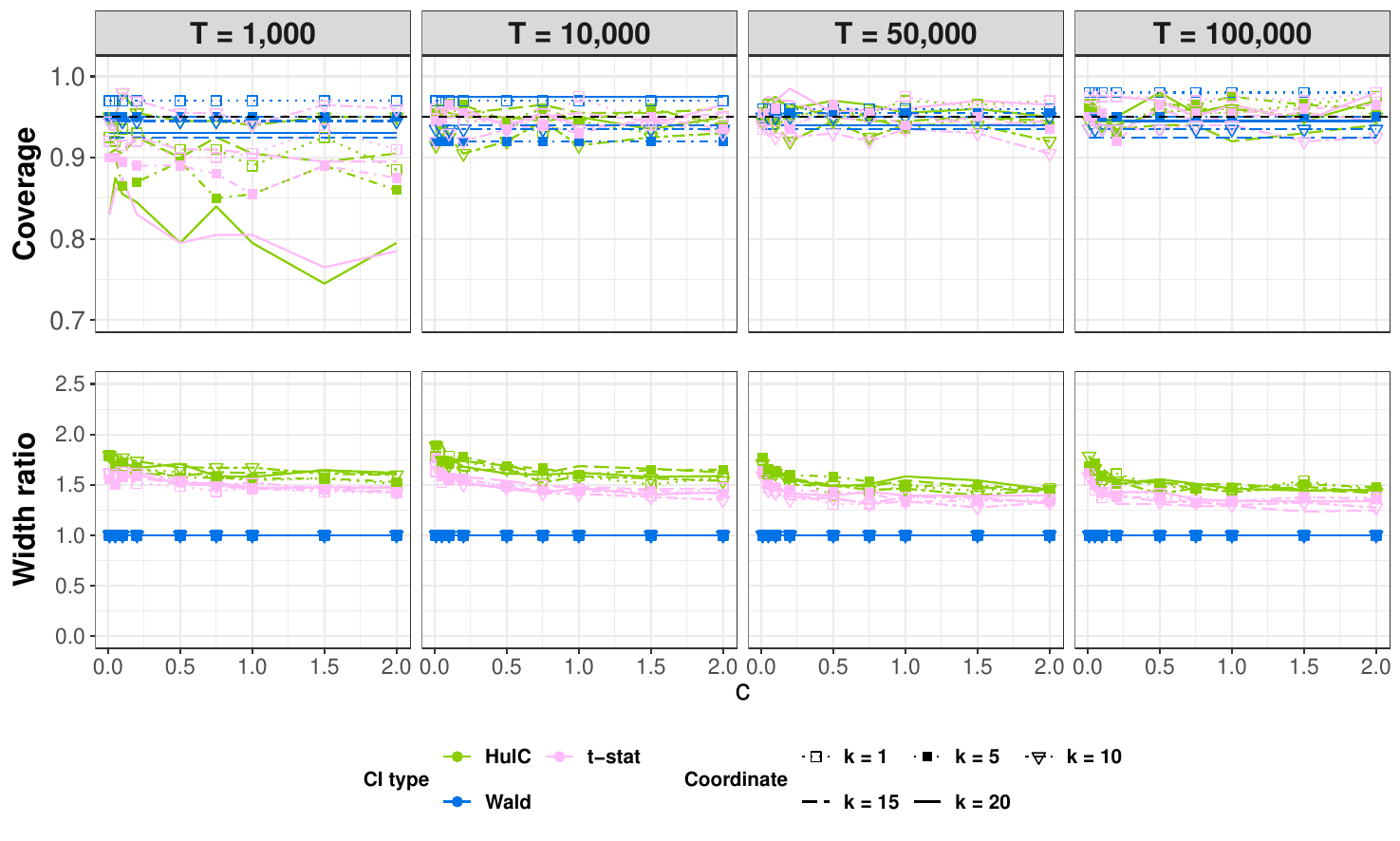}
\caption{Linear regression  (average-iterate-implicit-SGD), Covariance = Equicorrelation, d = 20}
\label{fig:linear_D20_Equicorr_cov_wr_AISGD_initTRUE}
\end{figure}

\begin{figure}[H]
\centering
 % \par\medskip
\includegraphics[width=1\textwidth]{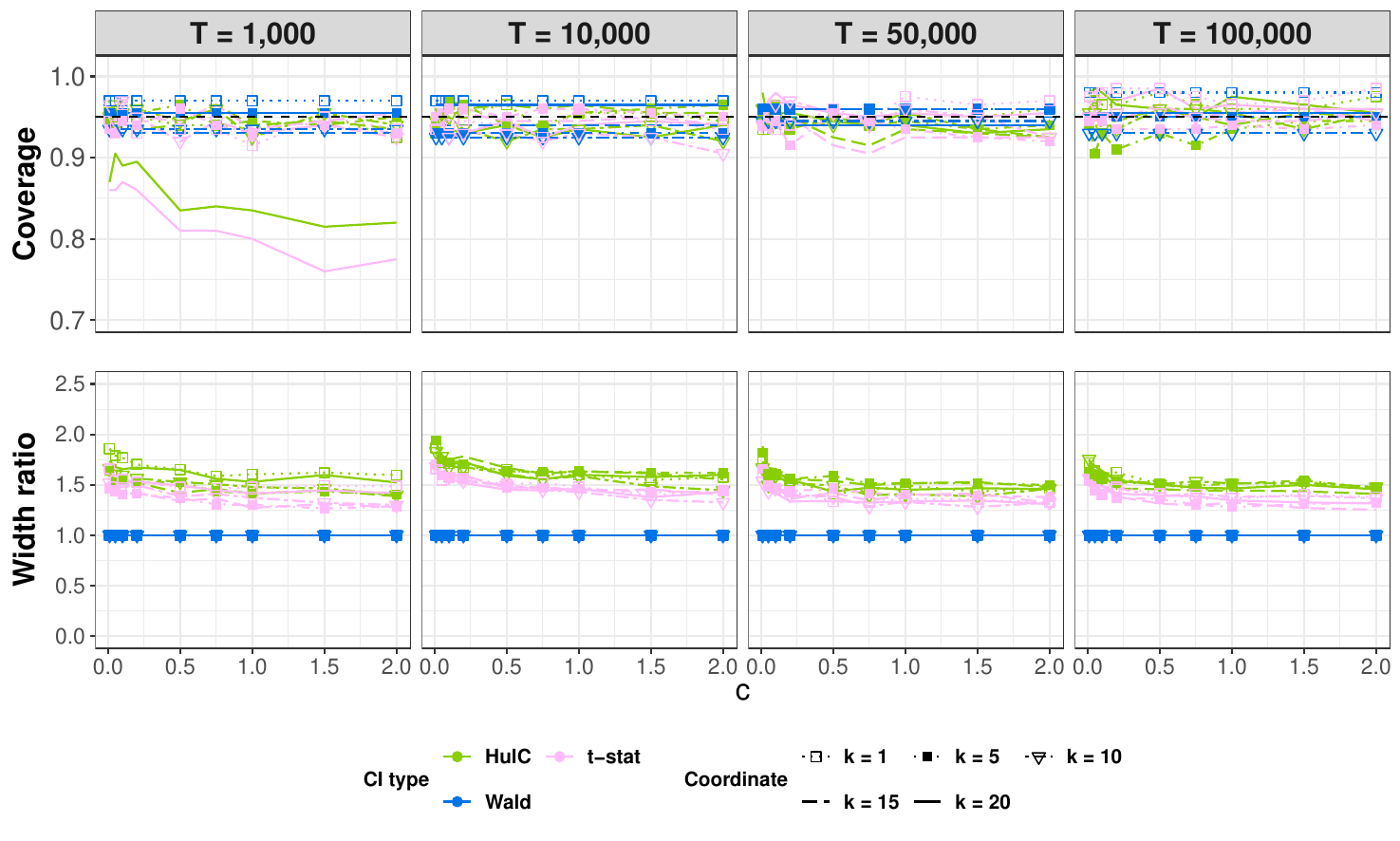}
\caption{Linear regression  (average-iterate-implicit-SGD), Covariance = Toeplitz, d = 20}
\label{fig:linear_D20_Toeplitz_cov_wr_AISGD_initTRUE}
\end{figure}

\begin{figure}[H]
\centering
 % \par\medskip
\includegraphics[width=1\textwidth]{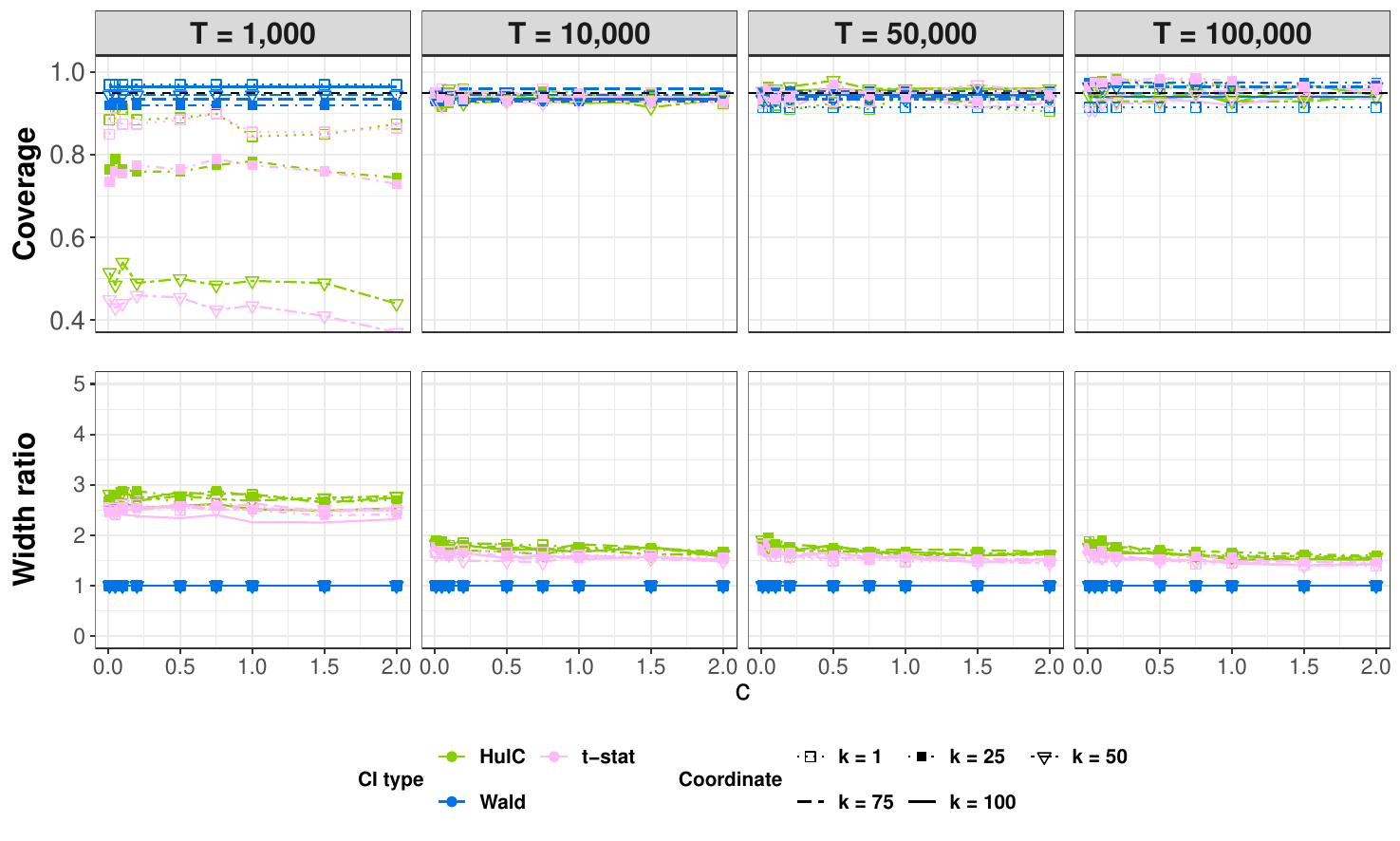}
\caption{Linear regression  (average-iterate-implicit-SGD), Covariance = I, d = 100}
\label{fig:linear_D100_I_cov_wr_AISGD_initTRUE}
\end{figure}

\begin{figure}[H]
\centering
 % \par\medskip
\includegraphics[width=1\textwidth]{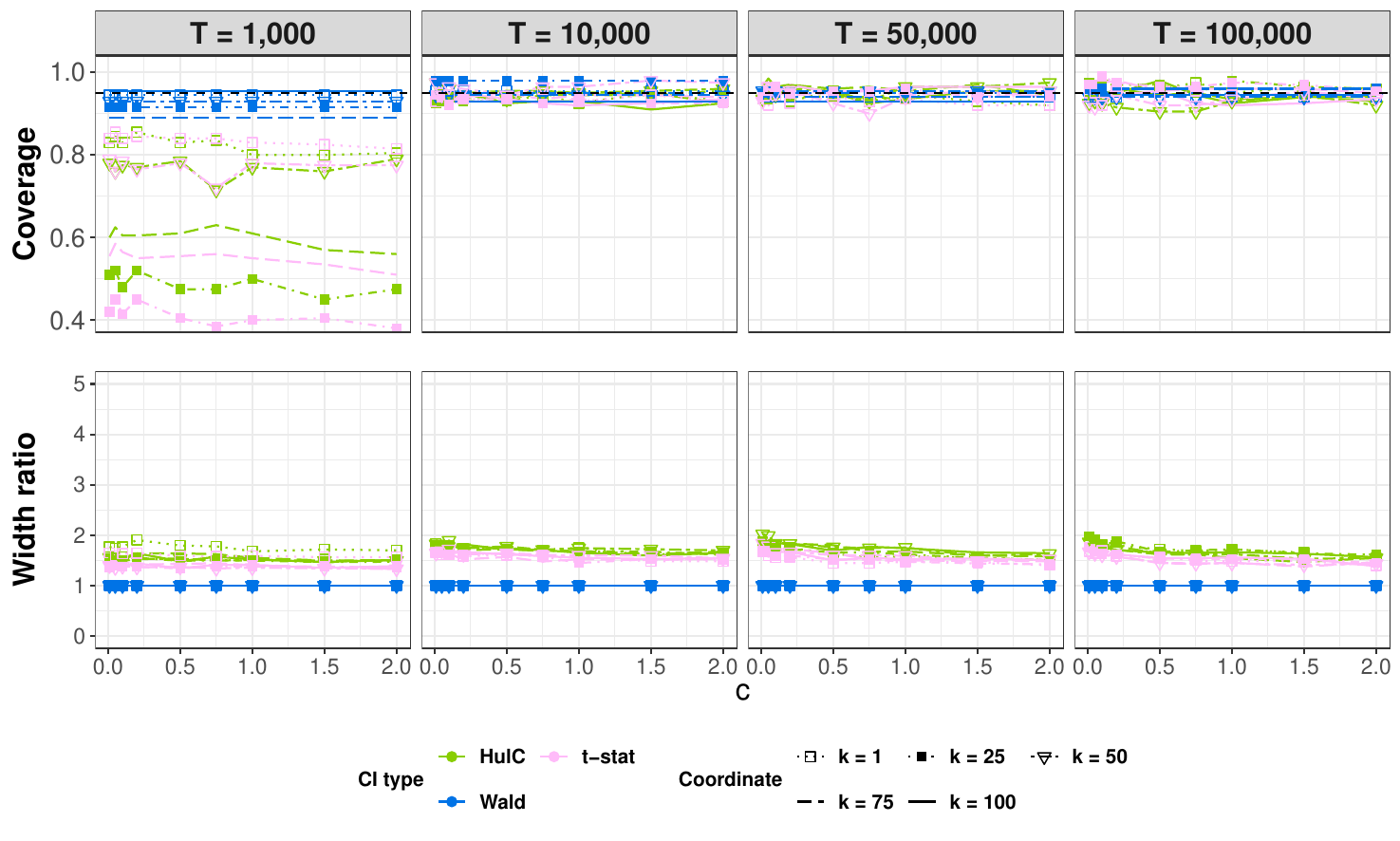}
\caption{Linear regression  (average-iterate-implicit-SGD), Covariance = Equicorrelation, d = 100}
\label{fig:linear_D100_EquiCorr_cov_wr_AISGD_initTRUE}
\end{figure}

\begin{figure}[H]
\centering
 % \par\medskip
\includegraphics[width=1\textwidth]{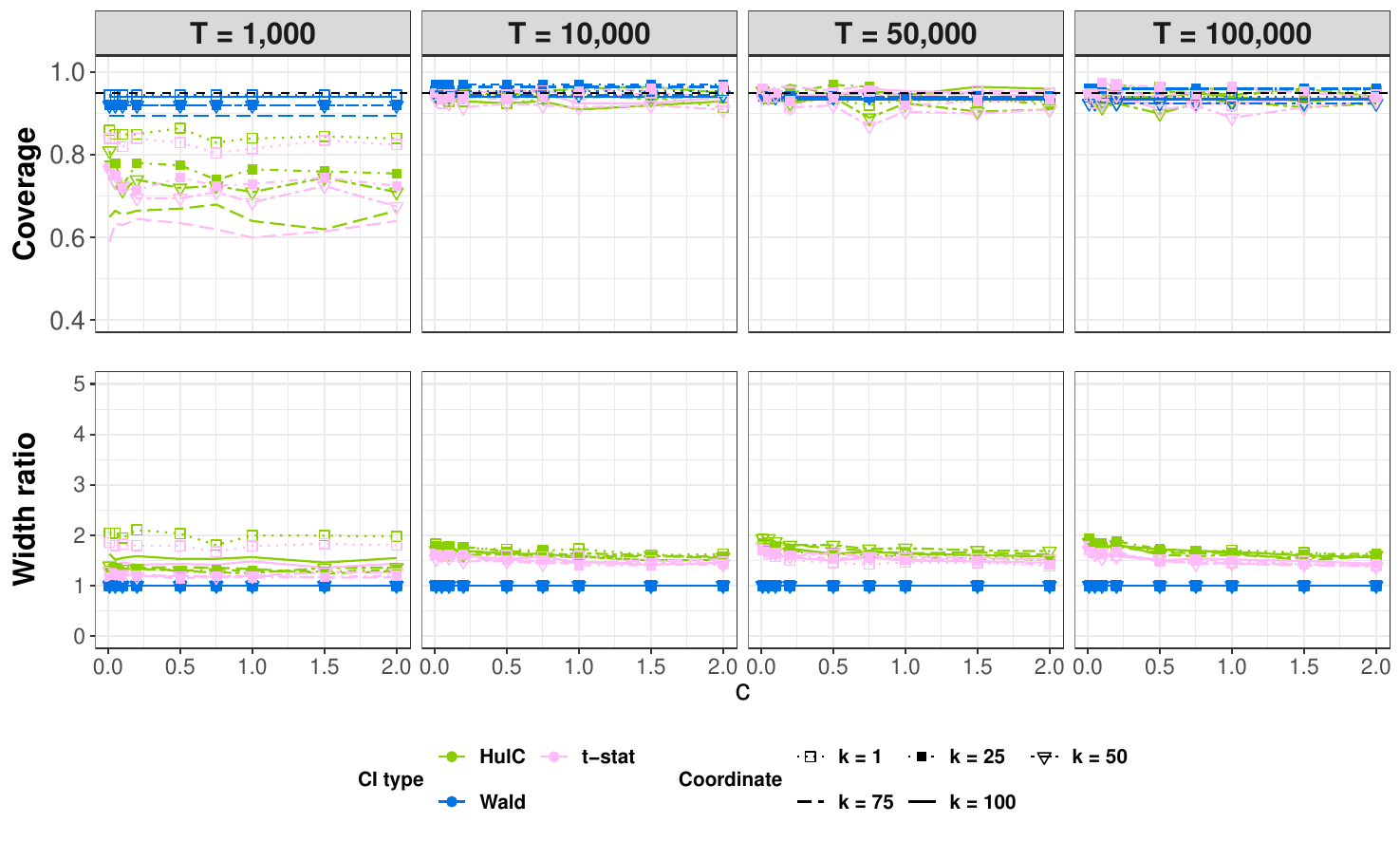}
\caption{Linear regression  (average-iterate-implicit-SGD), Covariance = Toeplitz, d = 100}
\label{fig:linear_D100_Toeplitz_cov_wr_AISGD_initTRUE}
\end{figure}

\subsubsection{Logistic regression}\label{app:Logistic_plots_AISGD}
\vspace{-20 pt}

\begin{figure}[H]
\centering
 % \par\medskip
\includegraphics[width=1\textwidth]{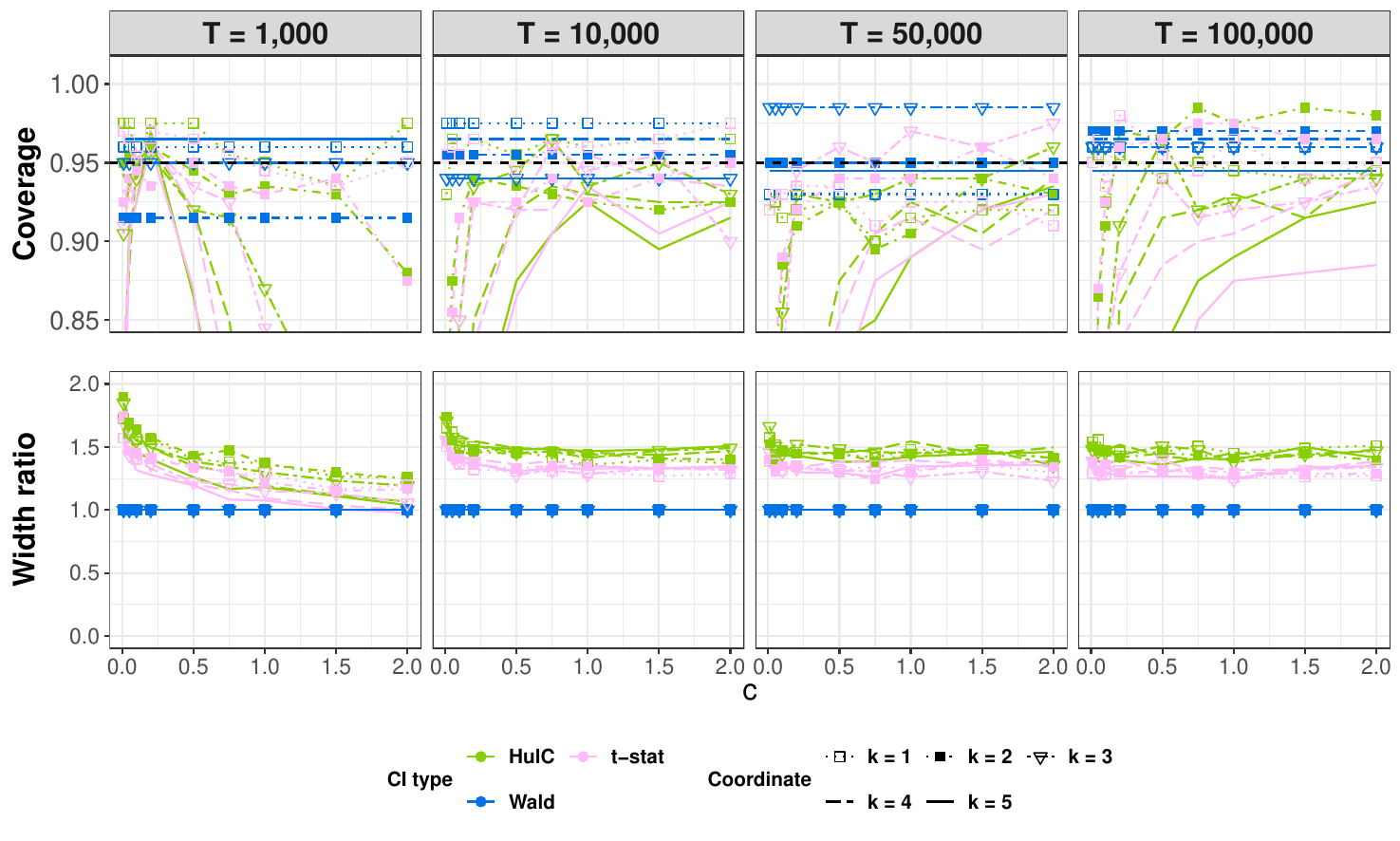}
\caption{Linear regression  (average-iterate-implicit-SGD), Covariance = I, d = 5}
\label{fig:logistic_D5_I_cov_wr_AISGD_initTRUE}
\end{figure}

\begin{figure}[H]
\centering
 % \par\medskip
\includegraphics[width=1\textwidth]{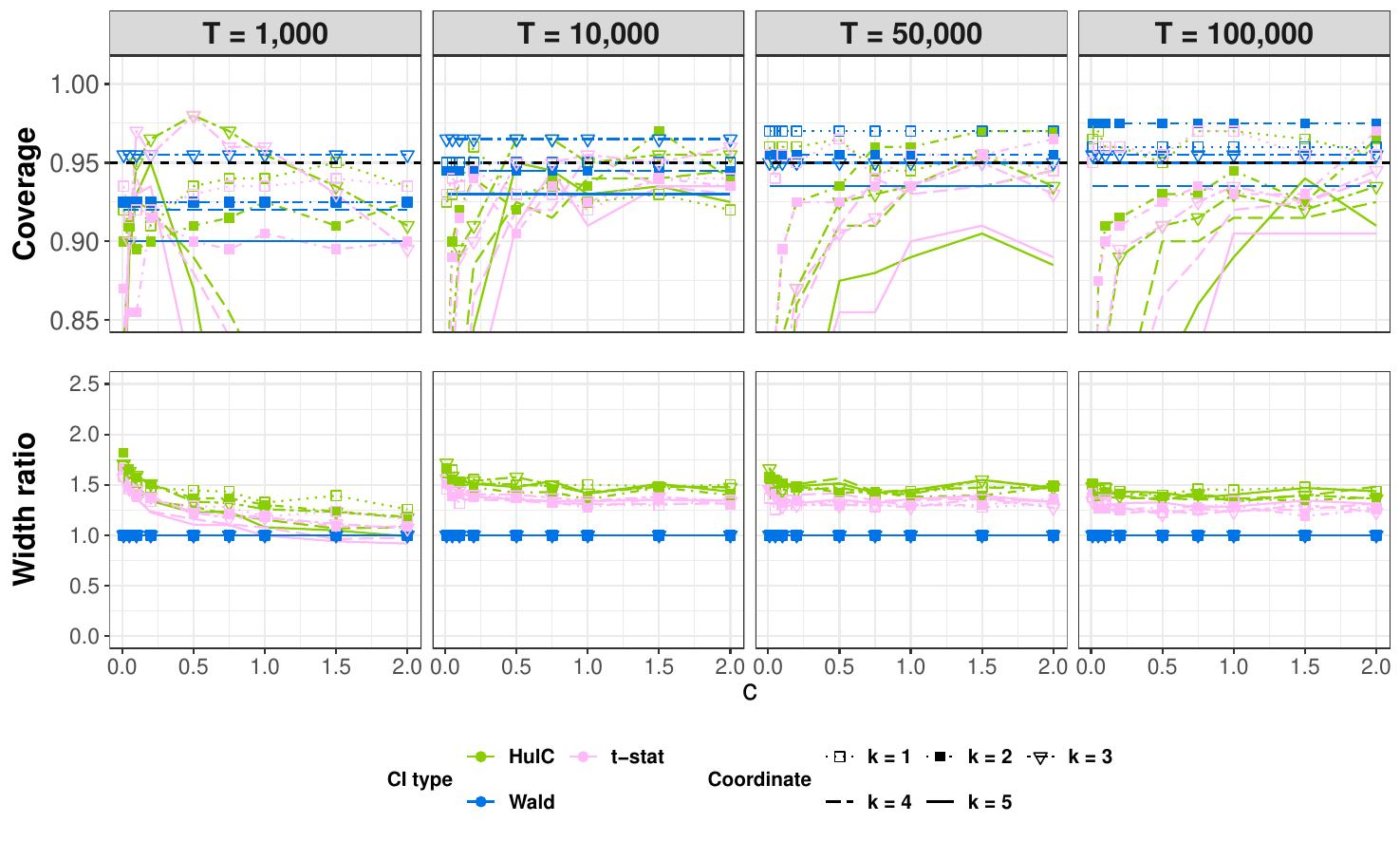}
\caption{Logistic regression  (average-iterate-implicit-SGD), Covariance = Equicorrelation, d = 5}
\label{fig:logistic_D5_Equicorr_cov_wr_AISGD_initTRUE}
\end{figure}

For logistic regression using average-iterate-implicit-SGD, $d=5$, and Toeplitz covariance, see Figure~\ref{fig:logistic_D5_Toeplitz_cov_wr_AISGD_initTRUE}.

\begin{figure}[H]
\centering
 % \par\medskip
\includegraphics[width=1\textwidth]{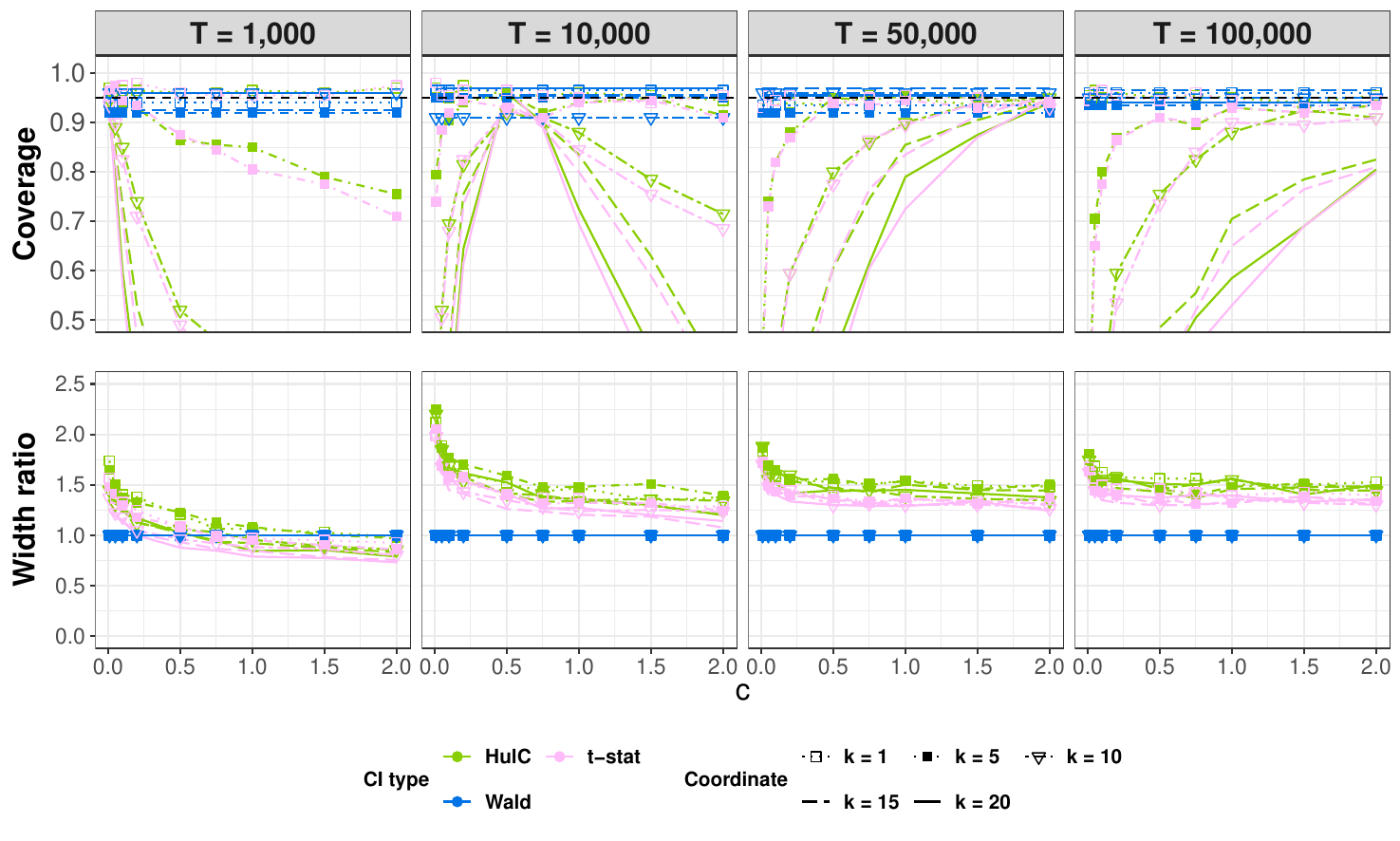}
\caption{Logistic regression  (average-iterate-implicit-SGD), Covariance = I, d = 20}
\label{fig:logistic_D20_I_cov_wr_AISGD_initTRUE}
\end{figure}

\begin{figure}[H]
\centering
 % \par\medskip
\includegraphics[width=1\textwidth]{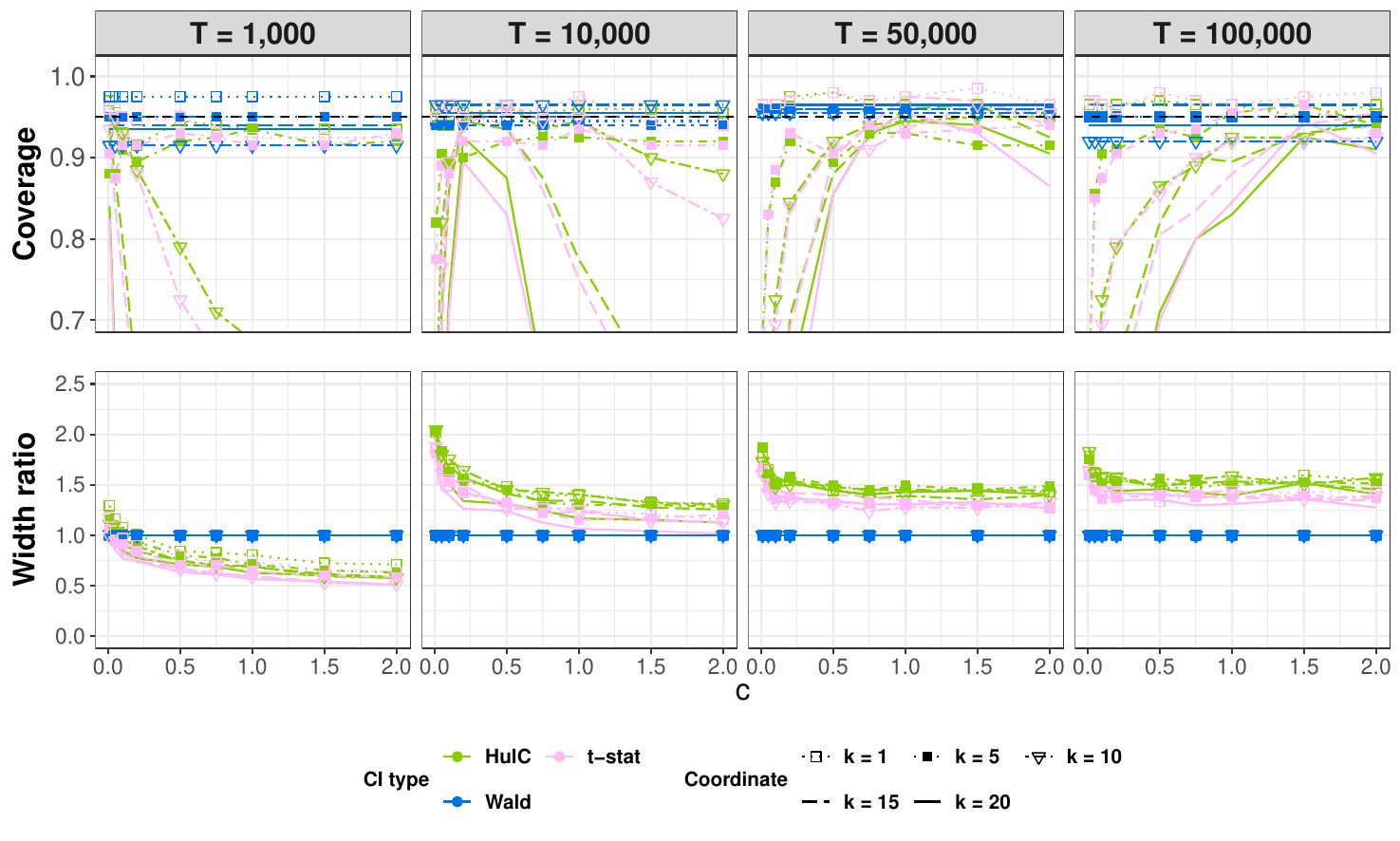}
\caption{Logistic regression  (average-iterate-implicit-SGD), Covariance = Equicorrelation, d = 20}
\label{fig:logistic_D20_EquiCorr_cov_wr_AISGD_initTRUE}
\end{figure}

\begin{figure}[H]
\centering
 % \par\medskip
\includegraphics[width=1\textwidth]{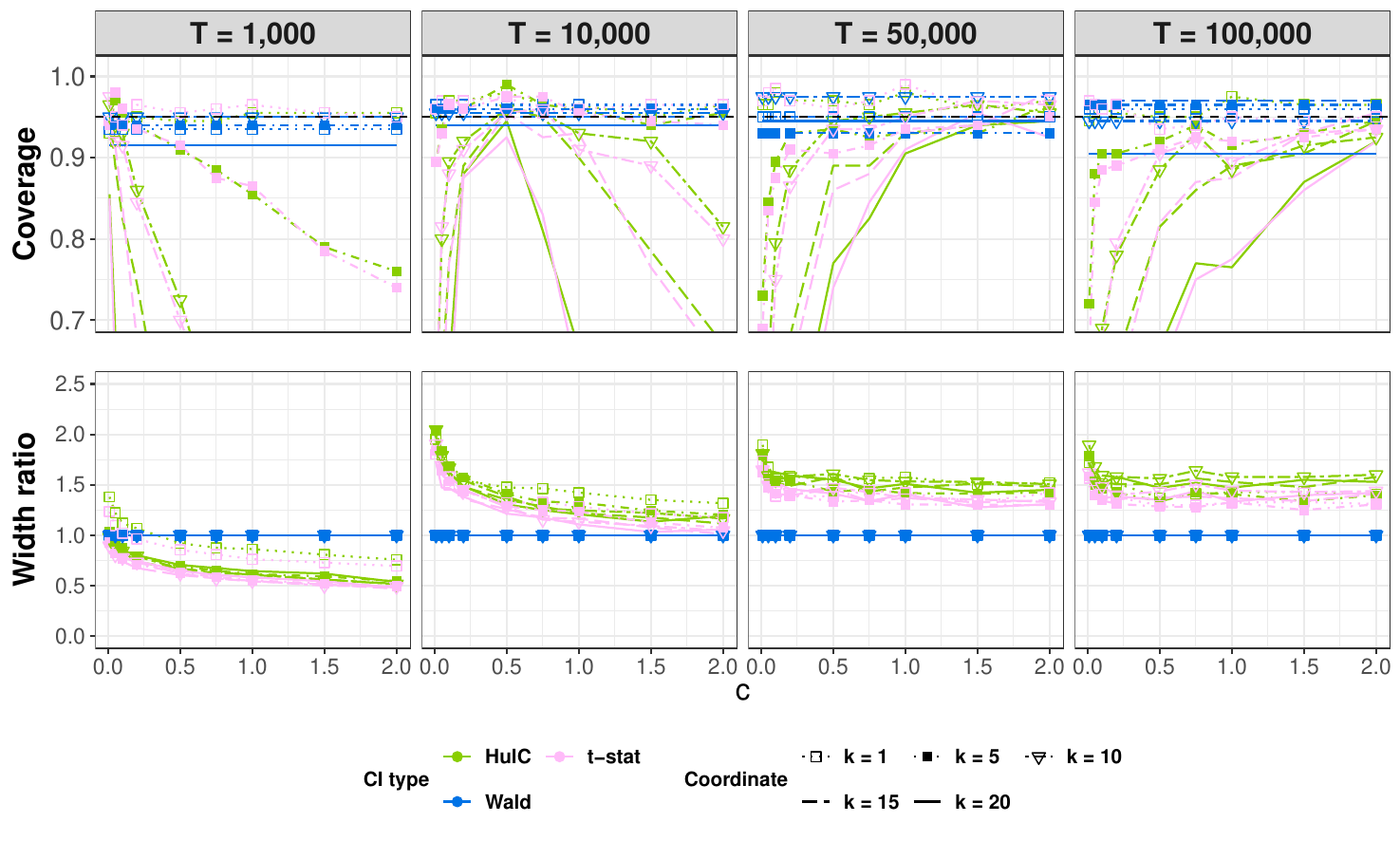}
\caption{Logistic regression  (average-iterate-implicit-SGD), Covariance = Toeplitz, d = 20}
\label{fig:logistic_D20_Toeplitz_cov_wr_AISGD_initTRUE}
\end{figure}

Note: Plots for logistic regression using average-iterate-implicit-SGD, $d=100$, did not always converge and/or Wald estimates were not computable. Please see the online tool.\footnote{\url{https://public.tableau.com/app/profile/selina.carter6629/viz/OnlineinferencesimulationsOLSandlogisticregression/Coverageandwidthratio_paper}}

%TODO: update these figures
% \begin{figure}[H]
% \centering
%  % \par\medskip
% \includegraphics[width=1\textwidth]{figures/extra_sims/logistic_D100_I_cov_wr_AISGD_initTRUE}
% \caption{Logistic regression  (average-iterate-implicit-SGD), Covariance = I, d = 100}
% \label{fig:logistic_D100_I_cov_wr_AISGD_initTRUE}
% \end{figure}

% \begin{figure}[H]
% \centering
%  % \par\medskip
% \includegraphics[width=1\textwidth]{figures/extra_sims/logistic_D100_EquiCorr_cov_wr_AISGD_initTRUE}
% \caption{Logistic regression  (average-iterate-implicit-SGD), Covariance = Equicorrelation, d = 100}
% \label{fig:logistic_D100_EquiCorr_cov_wr_AISGD_initTRUE}
% \end{figure}

% \begin{figure}[H]
% \centering
%  % \par\medskip
% \includegraphics[width=1\textwidth]{figures/extra_sims/logistic_D100_Toeplitz_cov_wr_AISGD_initTRUE}
% \caption{Logistic regression  (average-iterate-implicit-SGD), Covariance = Toeplitz, d = 100}
% \label{fig:logistic_D100_Toeplitz_cov_wr_AISGD_initTRUE}
% \end{figure}

\subsection{ROOT-SGD}

\subsubsection{Linear regression}\label{app:OLS_plots_rootsgd}
\vspace{-20 pt}

\begin{figure}[H]
\centering
 % \par\medskip
\includegraphics[width=1\textwidth]{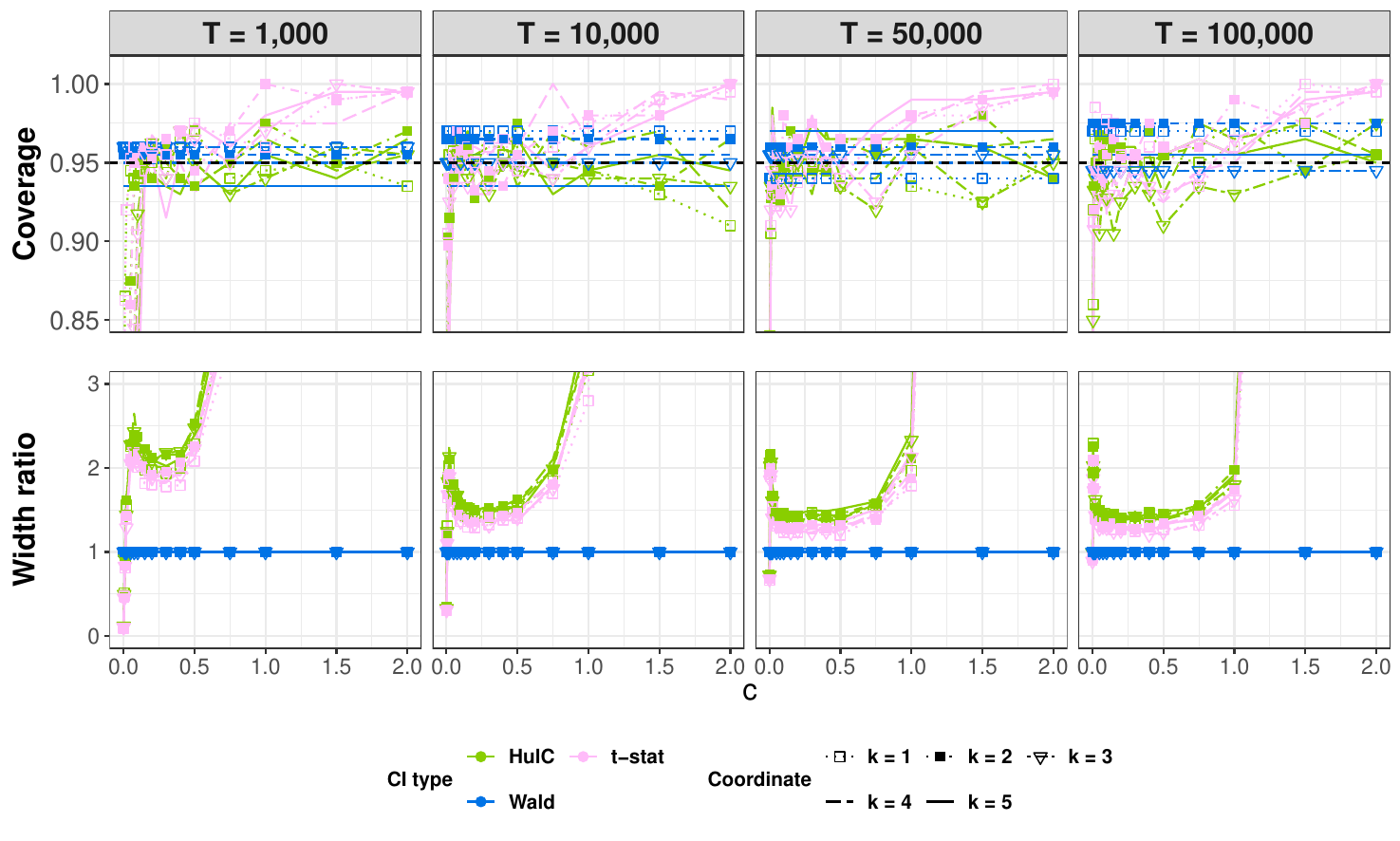}
\caption{Linear regression (ROOT-SGD), Covariance = I, d = 5}
\label{fig:linear_D5_I_cov_wr_rootSGD_initTRUE}
\end{figure}

\begin{figure}[H]
\centering
 % \par\medskip
\includegraphics[width=1\textwidth]{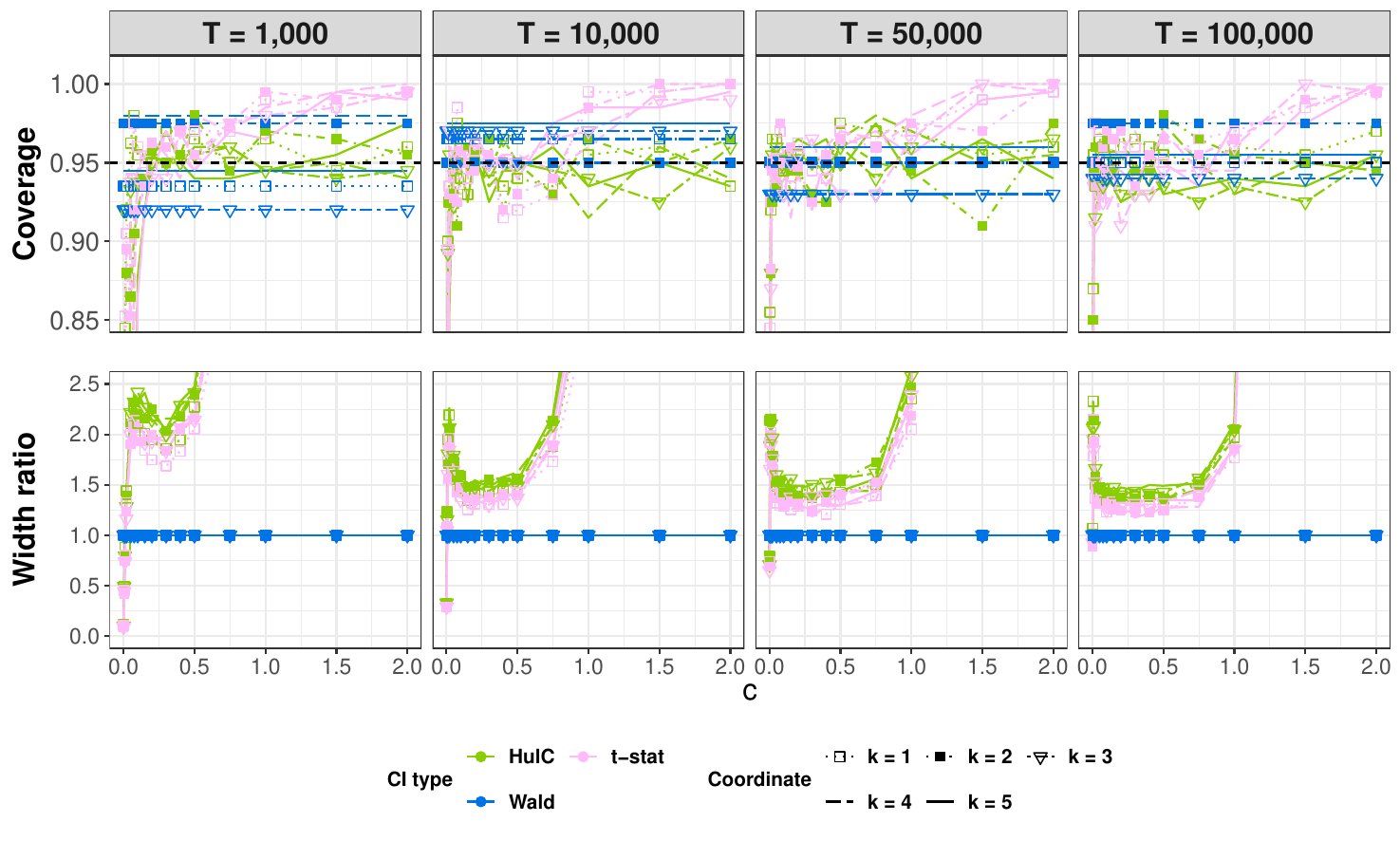}
\caption{Linear regression (ROOT-SGD), Covariance = Equicorrelation, d = 5}
\label{fig:linear_D5_Equicorr_cov_wr_rootSGD_initTRUE}
\end{figure}

For linear regression using ROOT-SGD, $d=5$, and Toeplitz covariance, see Figure~\ref{fig:linear_D5_Toeplitz_cov_wr_rootSGD_initTRUE}.

\begin{figure}[H]
\centering
 % \par\medskip
\includegraphics[width=1\textwidth]{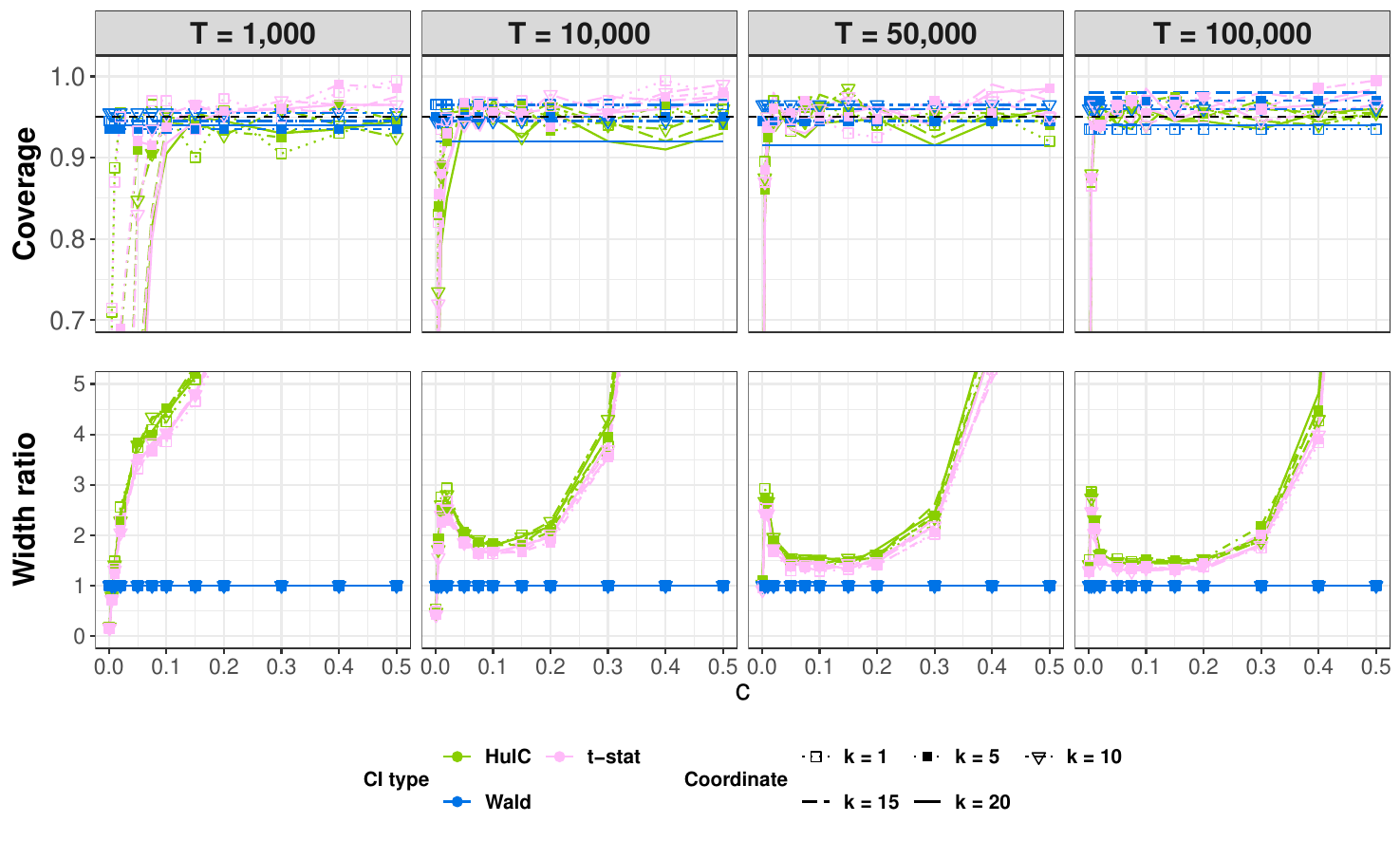}
\caption{Linear regression (ROOT-SGD), Covariance = I, d = 20}
\label{fig:linear_D20_I_cov_wr_rootSGD_initTRUE}
\end{figure}

\begin{figure}[H]
\centering
 % \par\medskip
\includegraphics[width=1\textwidth]{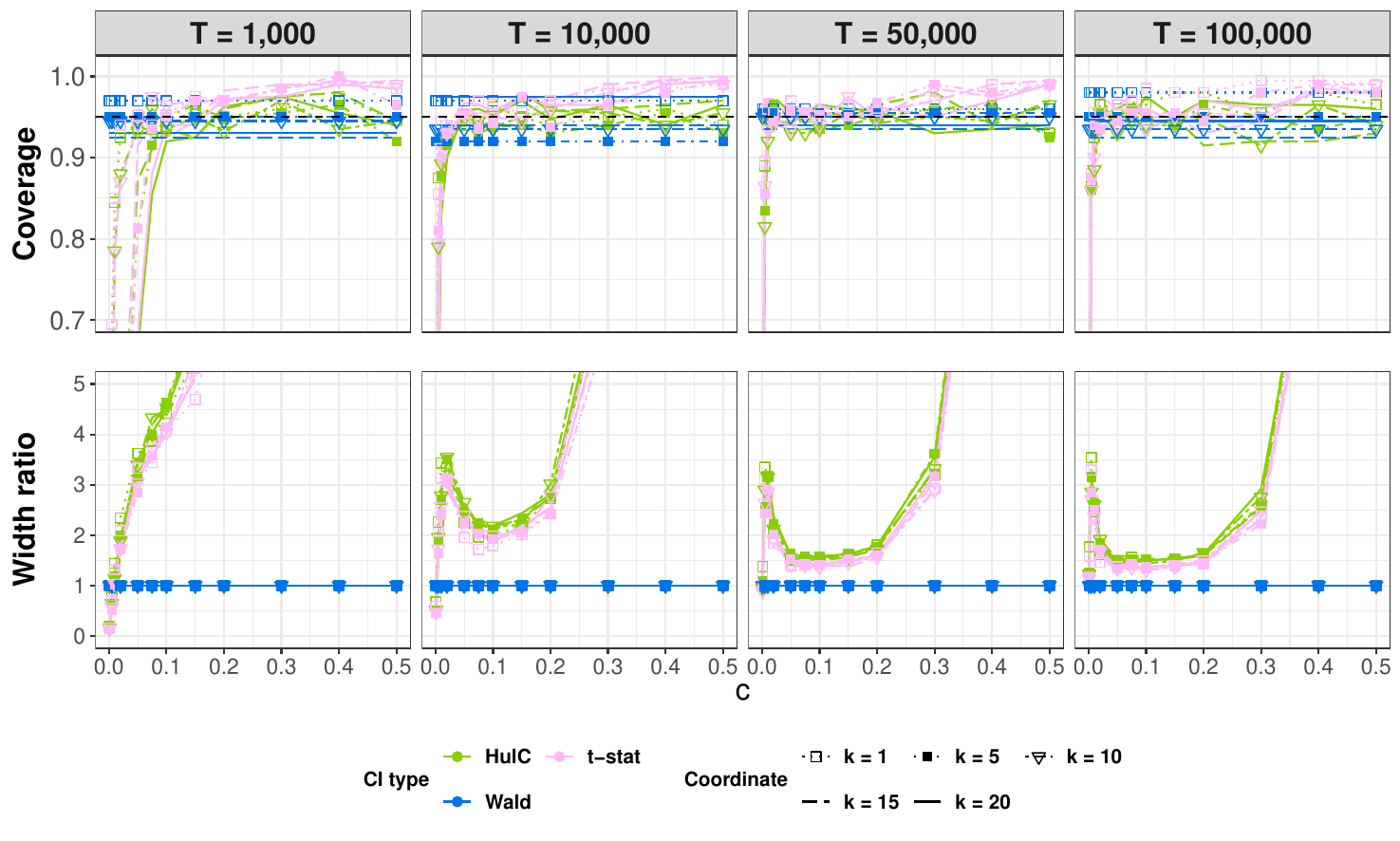}
\caption{Linear regression (ROOT-SGD), Covariance = Equicorrelation, d = 20}
\label{fig:linear_D20_EquiCorr_cov_wr_rootSGD_initTRUE}
\end{figure}

\begin{figure}[H]
\centering
 % \par\medskip
\includegraphics[width=1\textwidth]{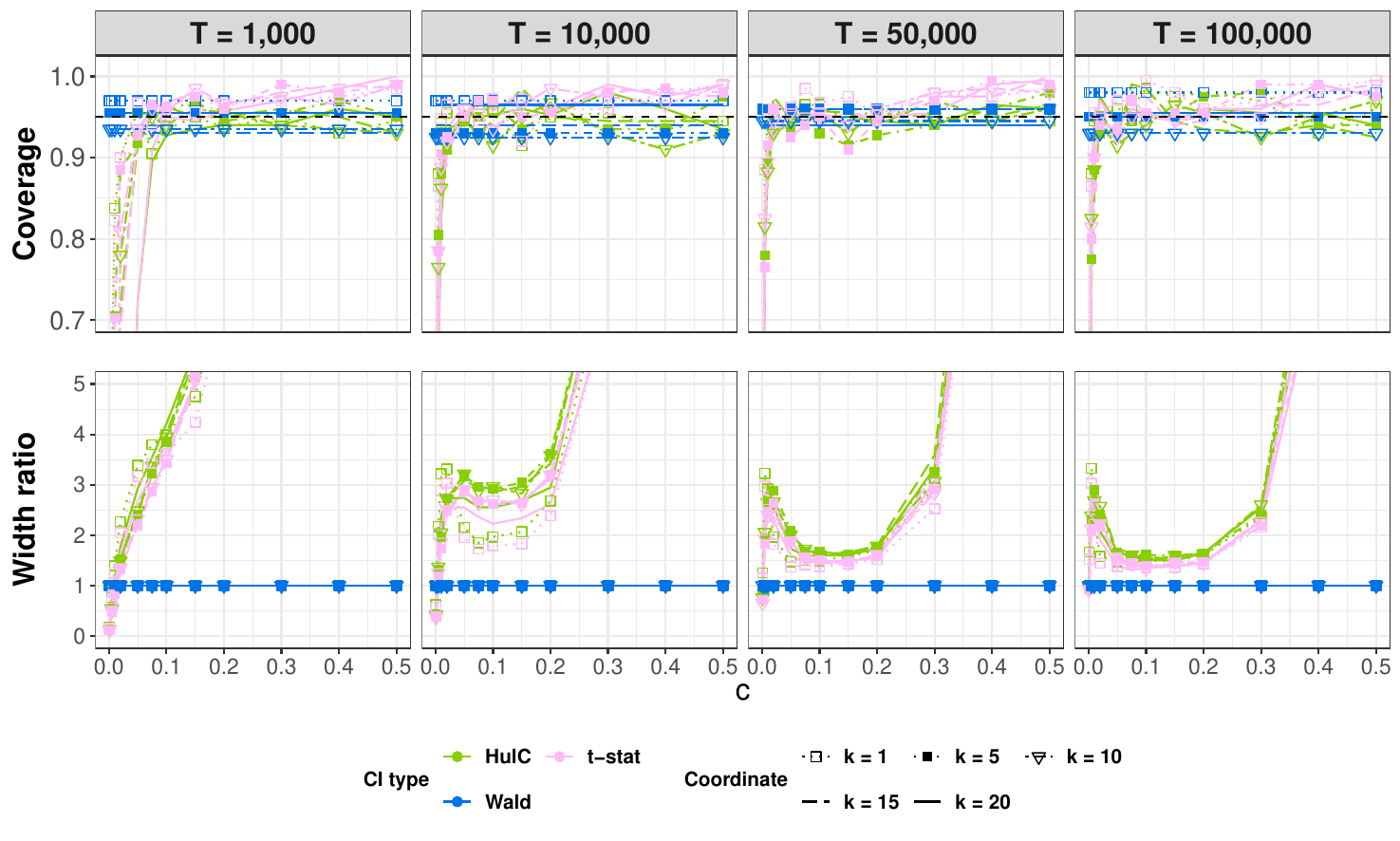}
\caption{Linear regression (ROOT-SGD), Covariance = Toeplitz, d = 20}
\label{fig:linear_D20_Toeplitz_cov_wr_rootSGD_initTRUE}
\end{figure}

\begin{figure}[H]
\centering
 % \par\medskip
\includegraphics[width=1\textwidth]{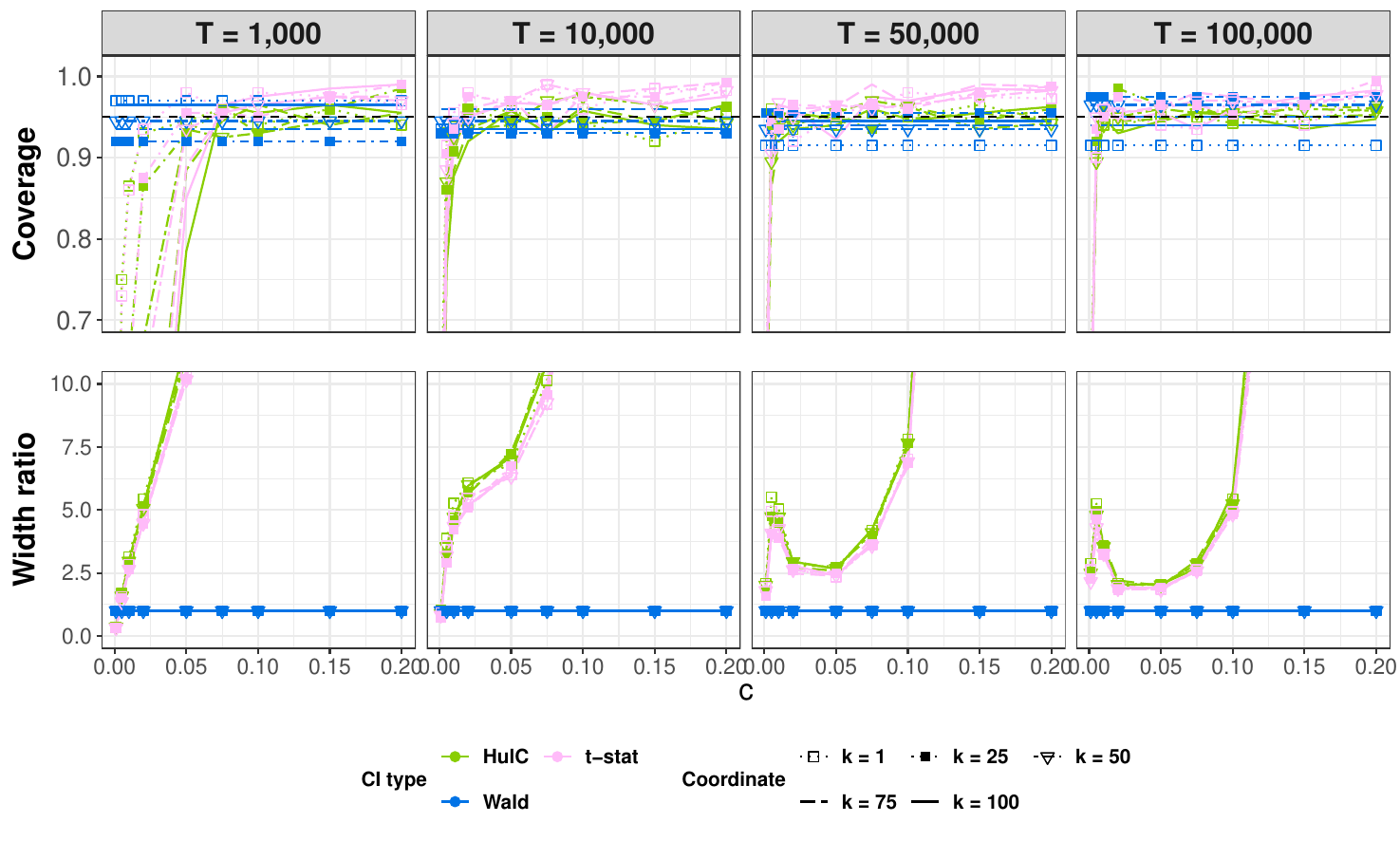}
\caption{Linear regression (ROOT-SGD), Covariance = I, d = 100}
\label{fig:linear_D100_I_cov_wr_rootSGD_initTRUE}
\end{figure}

\begin{figure}[H]
\centering
 % \par\medskip
\includegraphics[width=1\textwidth]{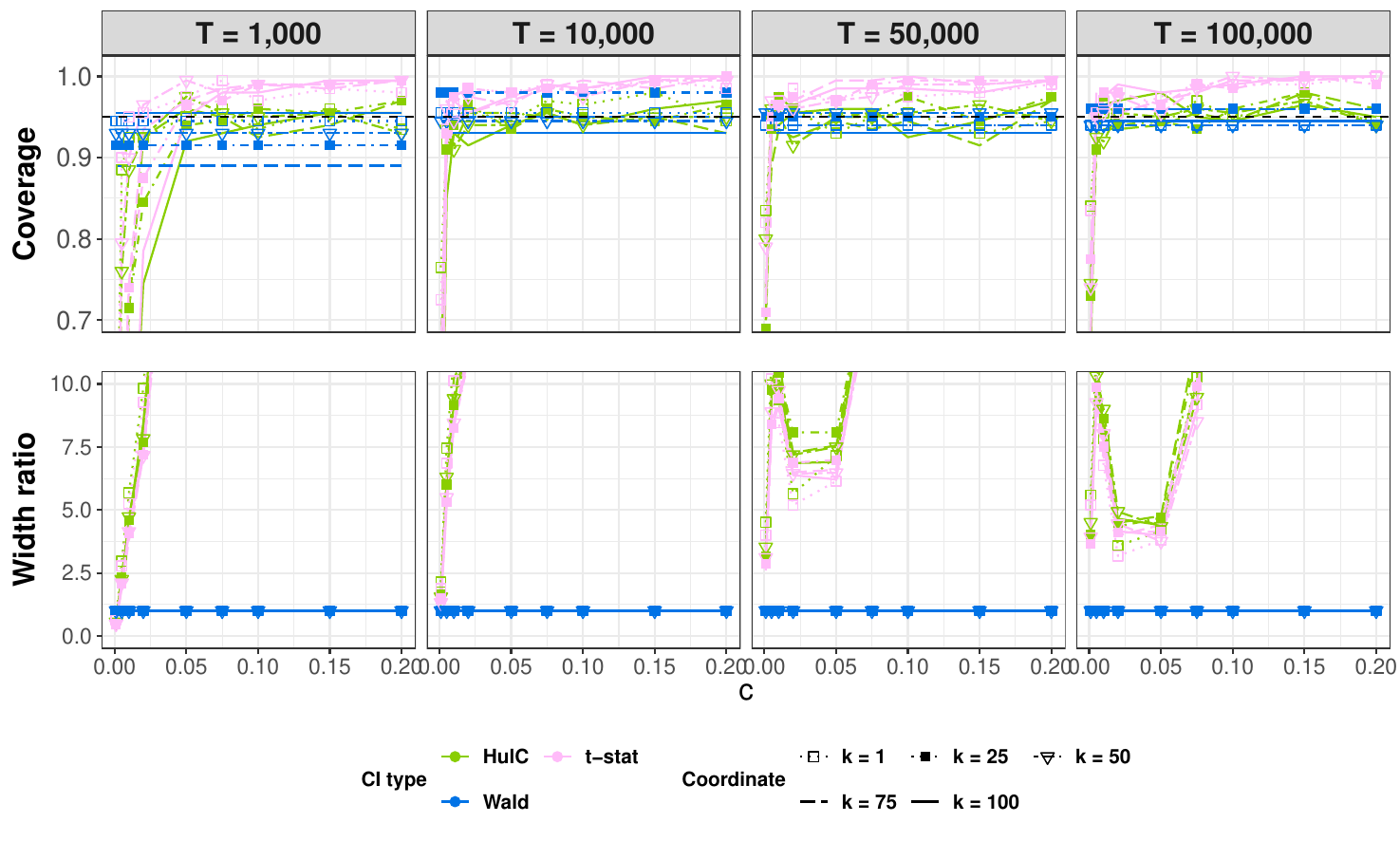}
\caption{Linear regression (ROOT-SGD), Covariance = Equicorrelation, d = 100}
\label{fig:linear_D100_EquiCorr_cov_wr_rootSGD_initTRUE}
\end{figure}

\begin{figure}[H]
\centering
 % \par\medskip
\includegraphics[width=1\textwidth]{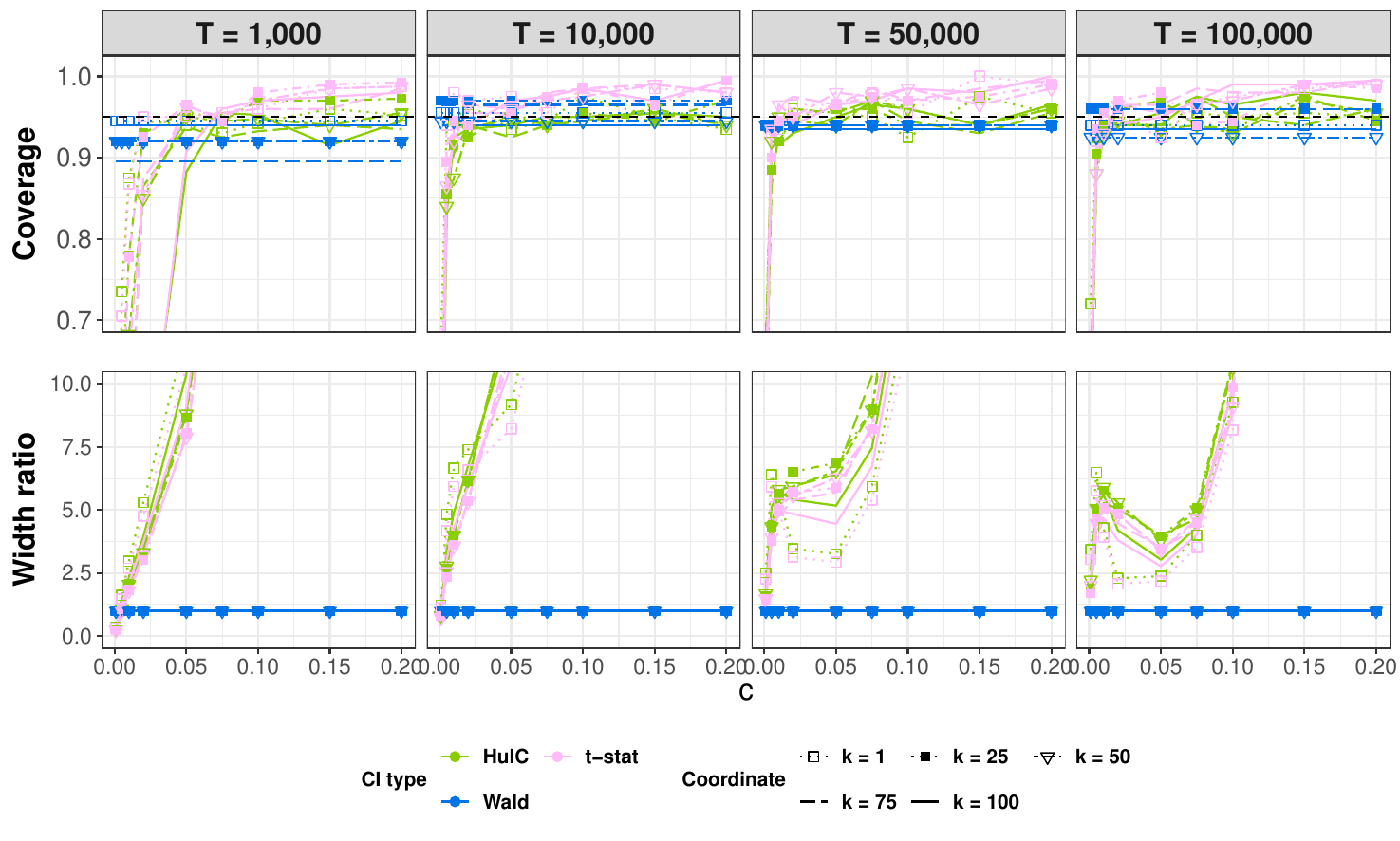}
\caption{Linear regression (ROOT-SGD), Covariance = Toeplitz, d = 100}
\label{fig:linear_D100_Toeplitz_cov_wr_rootSGD_initTRUE}
\end{figure}

\subsubsection{Logistic regression}\label{app:Logistic_plots_rootSGD}
\vspace{-20 pt}

\begin{figure}[H]
\centering
 % \par\medskip
\includegraphics[width=1\textwidth]{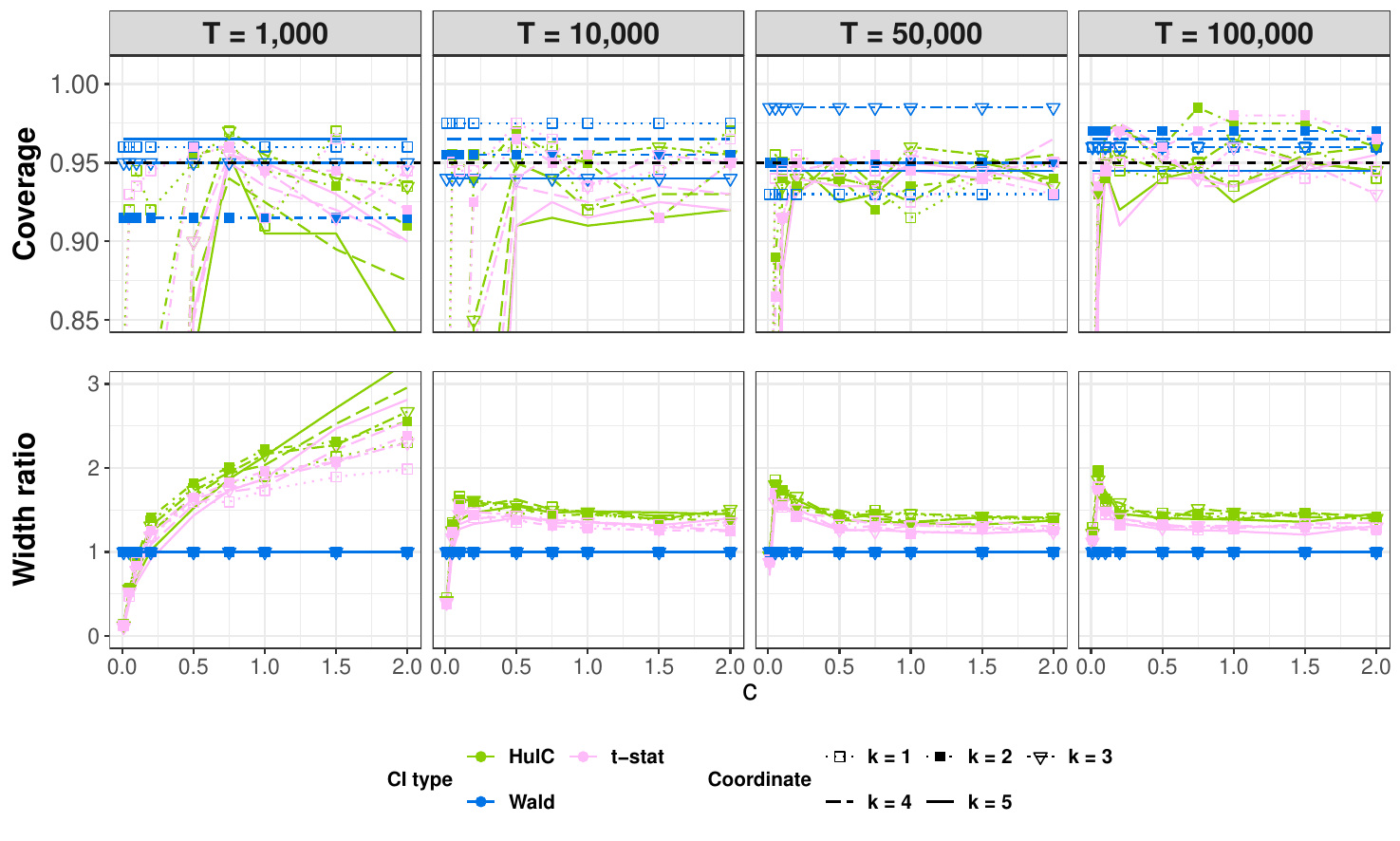}
\caption{Logistic regression (ROOT-SGD), Covariance = I, d = 5}
\label{fig:logistic_D5_I_cov_wr_rootSGD_initTRUE}
\end{figure}

\begin{figure}[H]
\centering
 % \par\medskip
\includegraphics[width=1\textwidth]{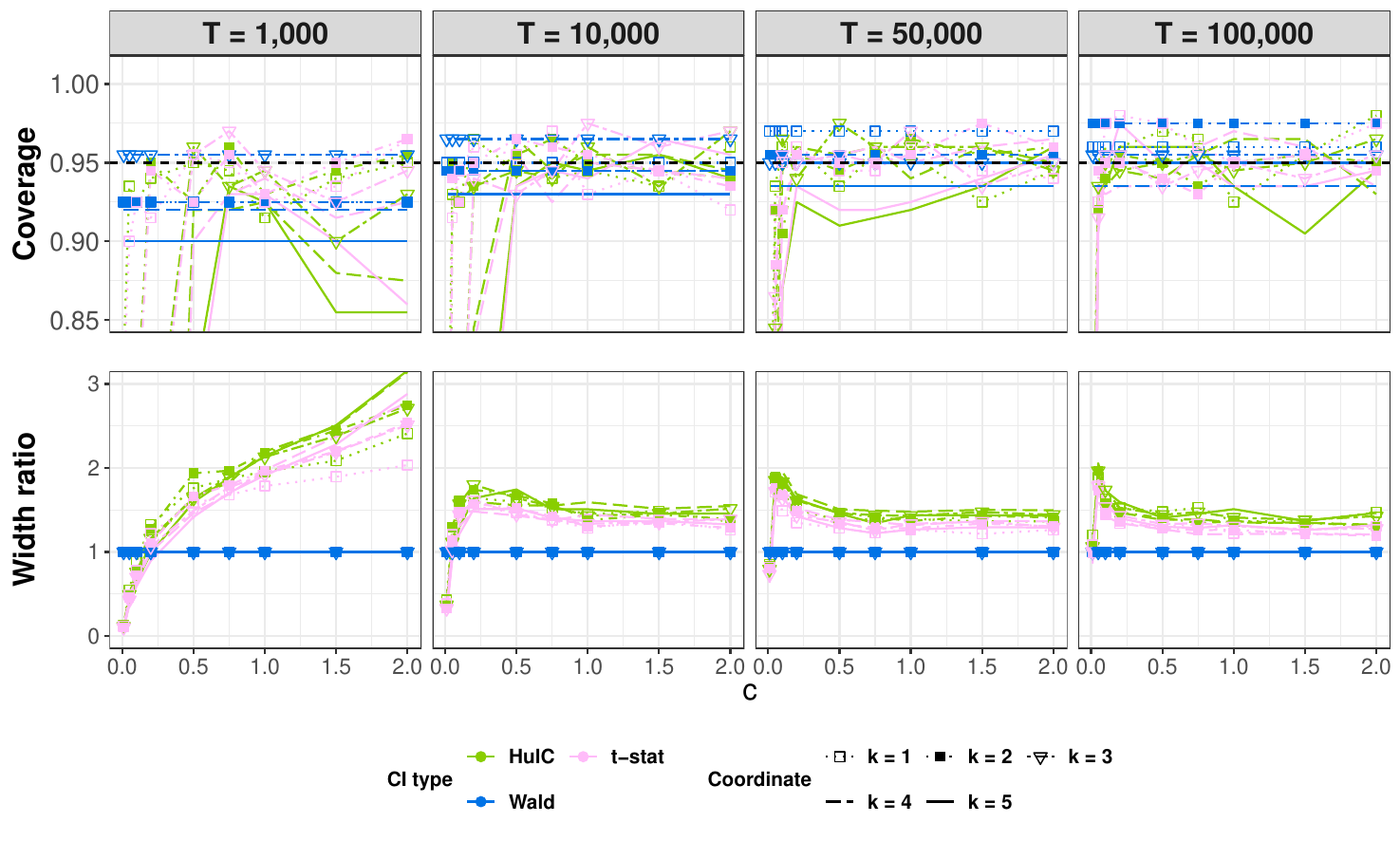}
\caption{Logistic regression (ROOT-SGD), Covariance = Equicorrelation, d = 5}
\label{fig:logistic_D5_EquiCorr_cov_wr_rootSGD_initTRUE}
\end{figure}

For logistic regression using ROOT-SGD, $d=5$, and Toeplitz covariance, see Figure~\ref{fig:logistic_D5_Toeplitz_cov_wr_rootSGD_initTRUE}.

\begin{figure}[H]
\centering
 % \par\medskip
\includegraphics[width=1\textwidth]{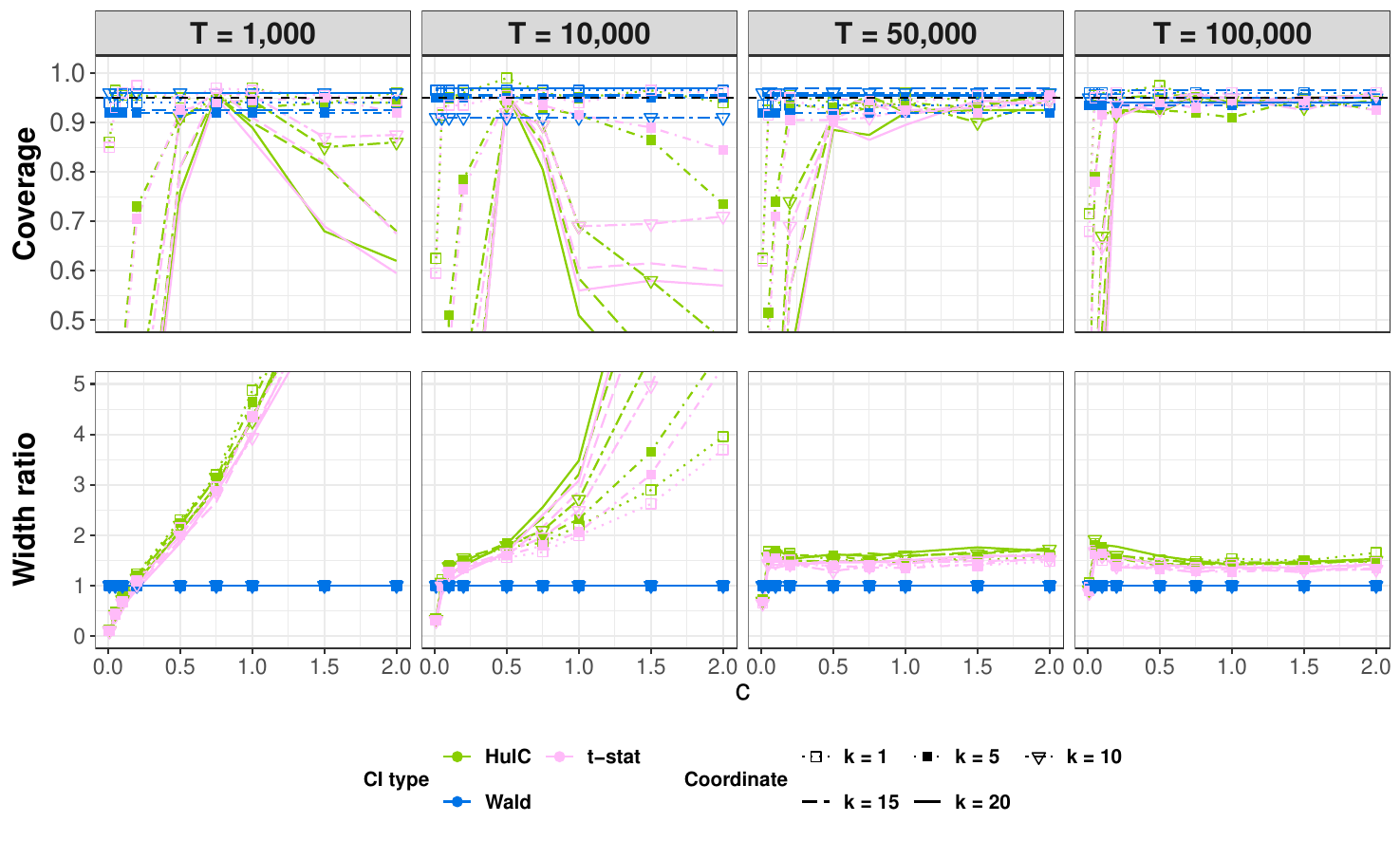}
\caption{Logistic regression (ROOT-SGD), Covariance = I, d = 20}
\label{fig:logistic_D20_I_cov_wr_rootSGD_initTRUE}
\end{figure}

\begin{figure}[H]
\centering
 % \par\medskip
\includegraphics[width=1\textwidth]{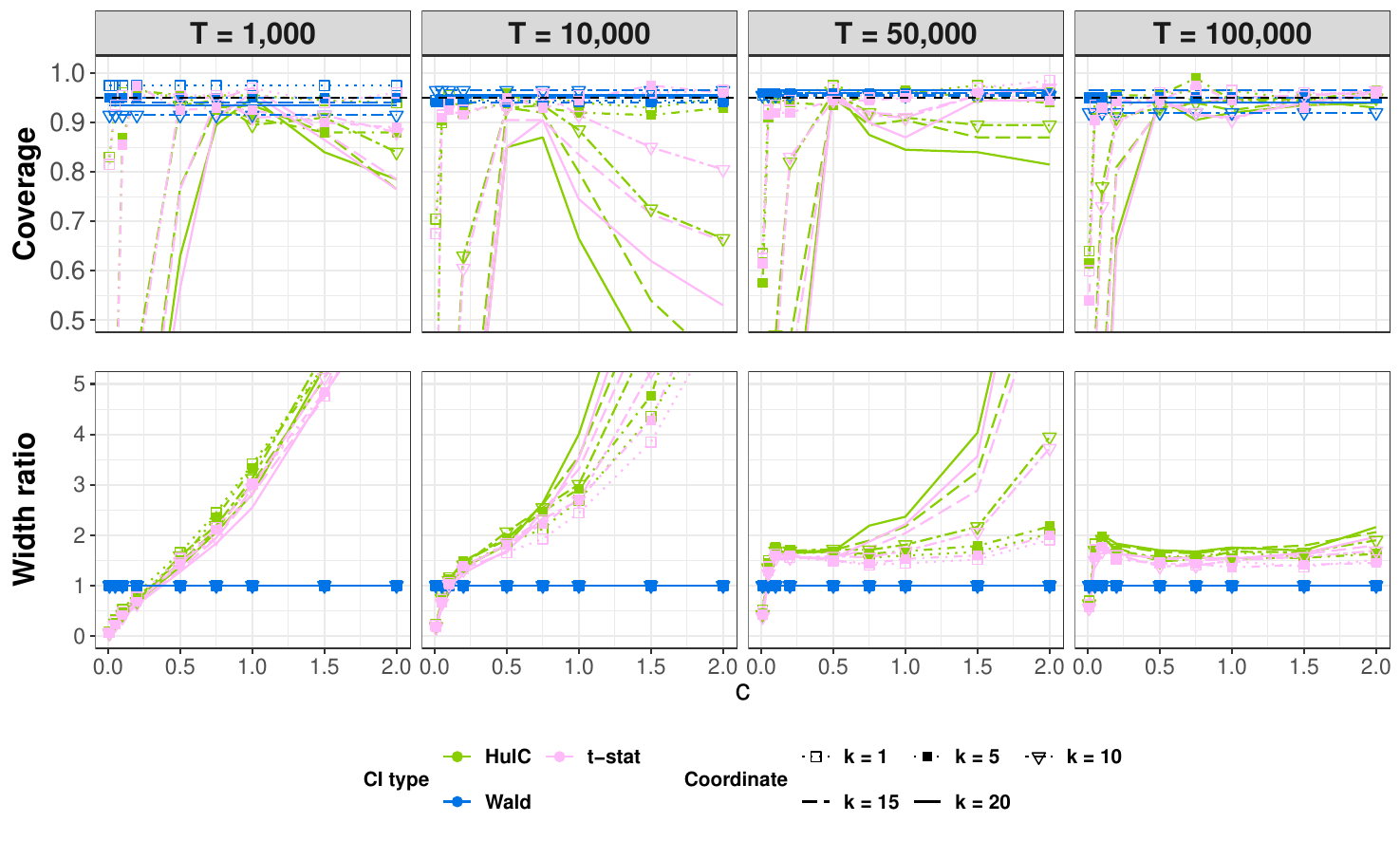}
\caption{Logistic regression (ROOT-SGD), Covariance = Equicorrelation, d = 20}
\label{fig:logistic_D20_EquiCorr_cov_wr_rootSGD_initTRUE}
\end{figure}

\begin{figure}[H]
\centering
 % \par\medskip
\includegraphics[width=1\textwidth]{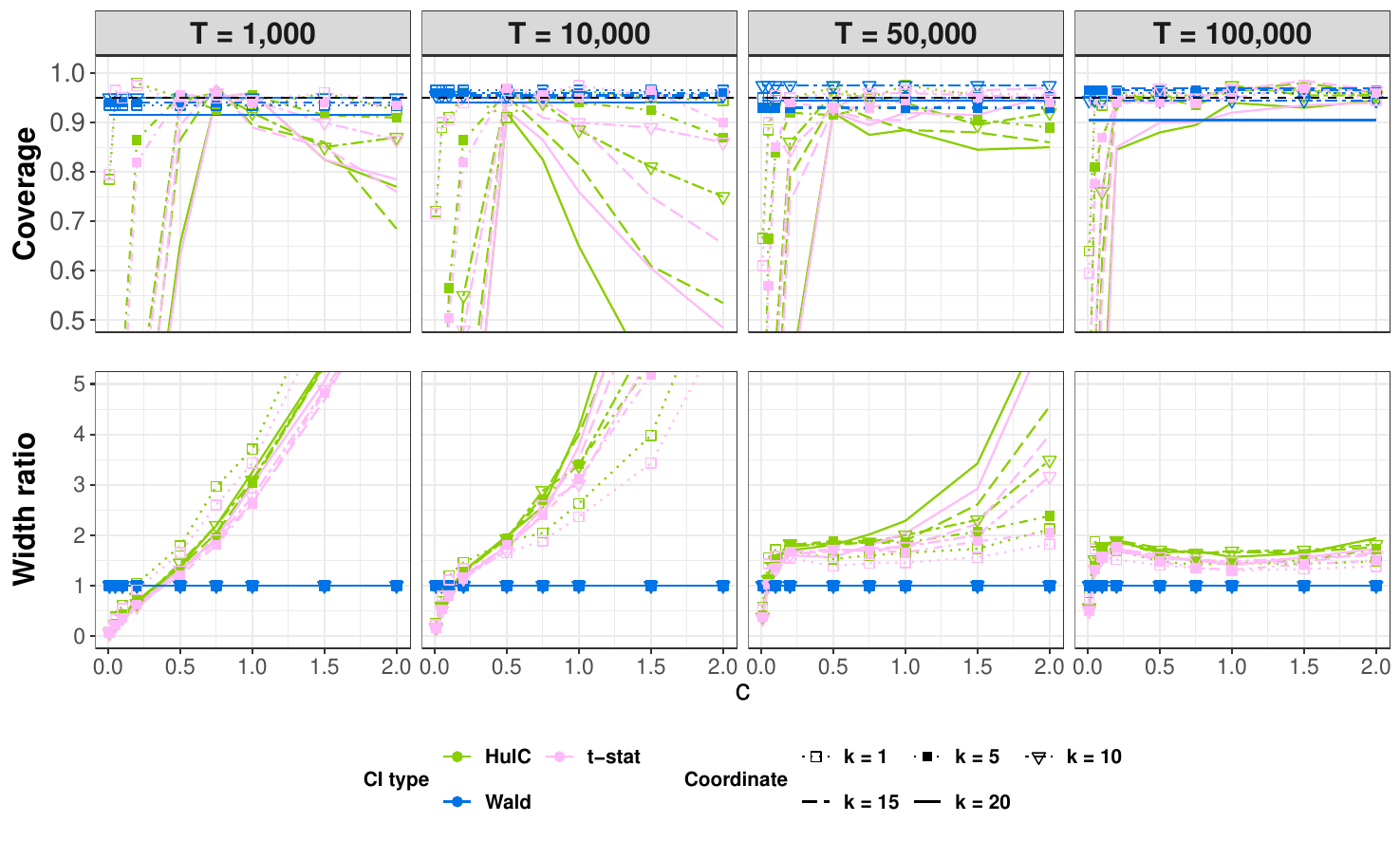}
\caption{Logistic regression (ROOT-SGD), Covariance = Toeplitz, d = 20}
\label{fig:logistic_D20_Toeplitz_cov_wr_rootSGD_initTRUE}
\end{figure}

\begin{figure}[H]
\centering
 % \par\medskip
\includegraphics[width=1\textwidth]{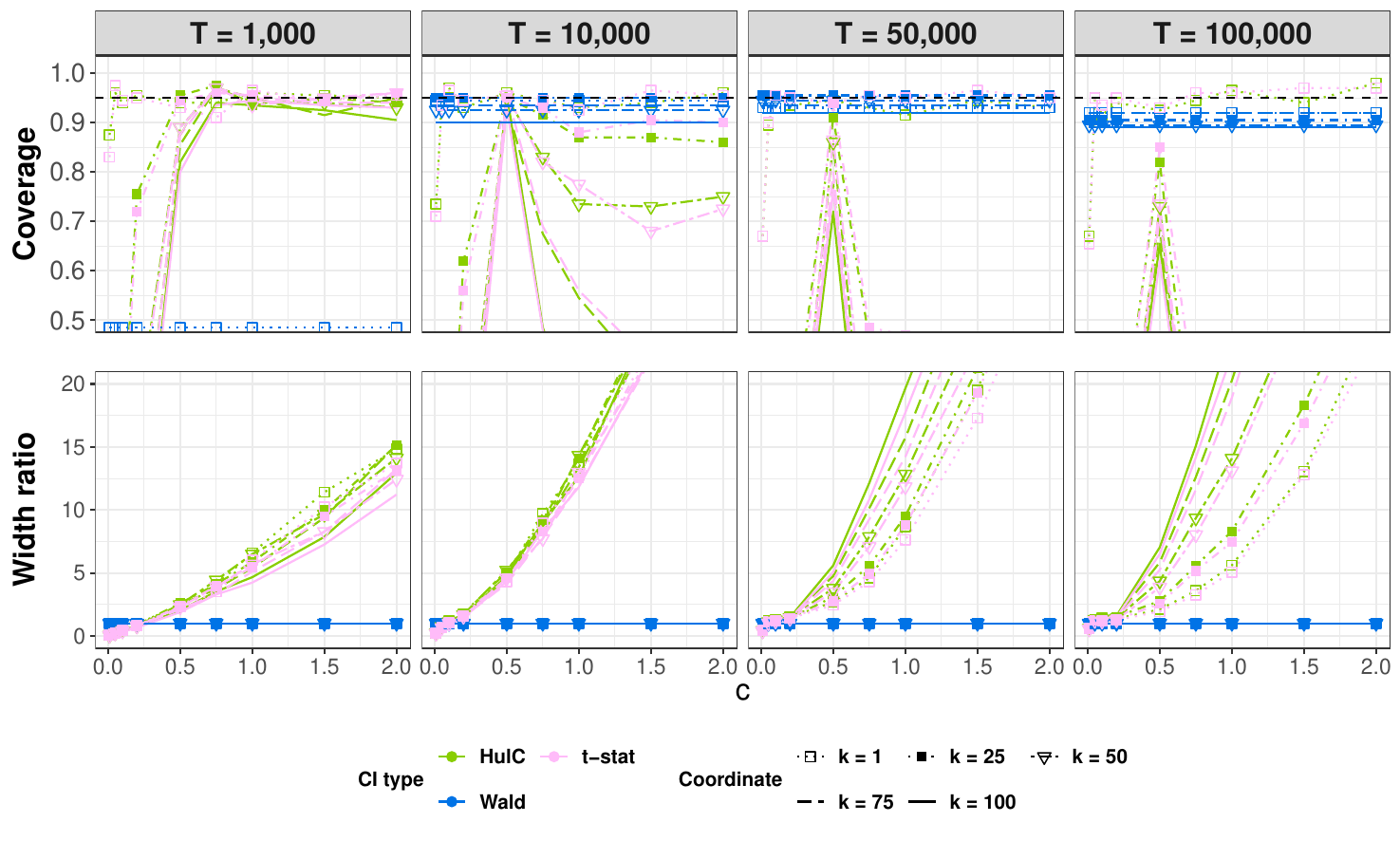}
\caption{Logistic regression (ROOT-SGD), Covariance = I, d = 100}
\label{fig:logistic_D100_I_cov_wr_rootSGD_initTRUE}
\end{figure}

Note: Wald estimates were not able to be computed for equicorrelation and Toeplitz covariances in the case of logistic regression using ROOT-SGD, $d=100$. To see raw widths, see the online tool.\footnote{\url{https://public.tableau.com/app/profile/selina.carter6629/viz/OnlineinferencesimulationsOLSandlogisticregression/Coverageandwidthratio_paper}}

% \begin{figure}[H]
% \centering
%  % \par\medskip
% \includegraphics[width=1\textwidth]{figures/extra_sims/logistic_D100_EquiCorr_cov_wr_rootSGD_initTRUE}
% \caption{Logistic regression (ROOT-SGD), Covariance = Equicorrelation, d = 100}
% \label{fig:logistic_D100_EquiCorr_cov_wr_rootSGD_initTRUE}
% \end{figure}

% \begin{figure}[H]
% \centering
%  % \par\medskip
% \includegraphics[width=1\textwidth]{figures/extra_sims/logistic_D100_Toeplitz_cov_wr_rootSGD_initTRUE}
% \caption{Logistic regression (ROOT-SGD), Covariance = Toeplitz, d = 100}
% \label{fig:logistic_D100_Toeplitz_cov_wr_rootSGD_initTRUE}
% \end{figure}

\subsection{Truncated-SGD}

\subsubsection{Linear regression}\label{app:OLS_plots_truncsgd}
\vspace{-20 pt}

\begin{figure}[H]
\centering
 % \par\medskip
\includegraphics[width=1\textwidth]{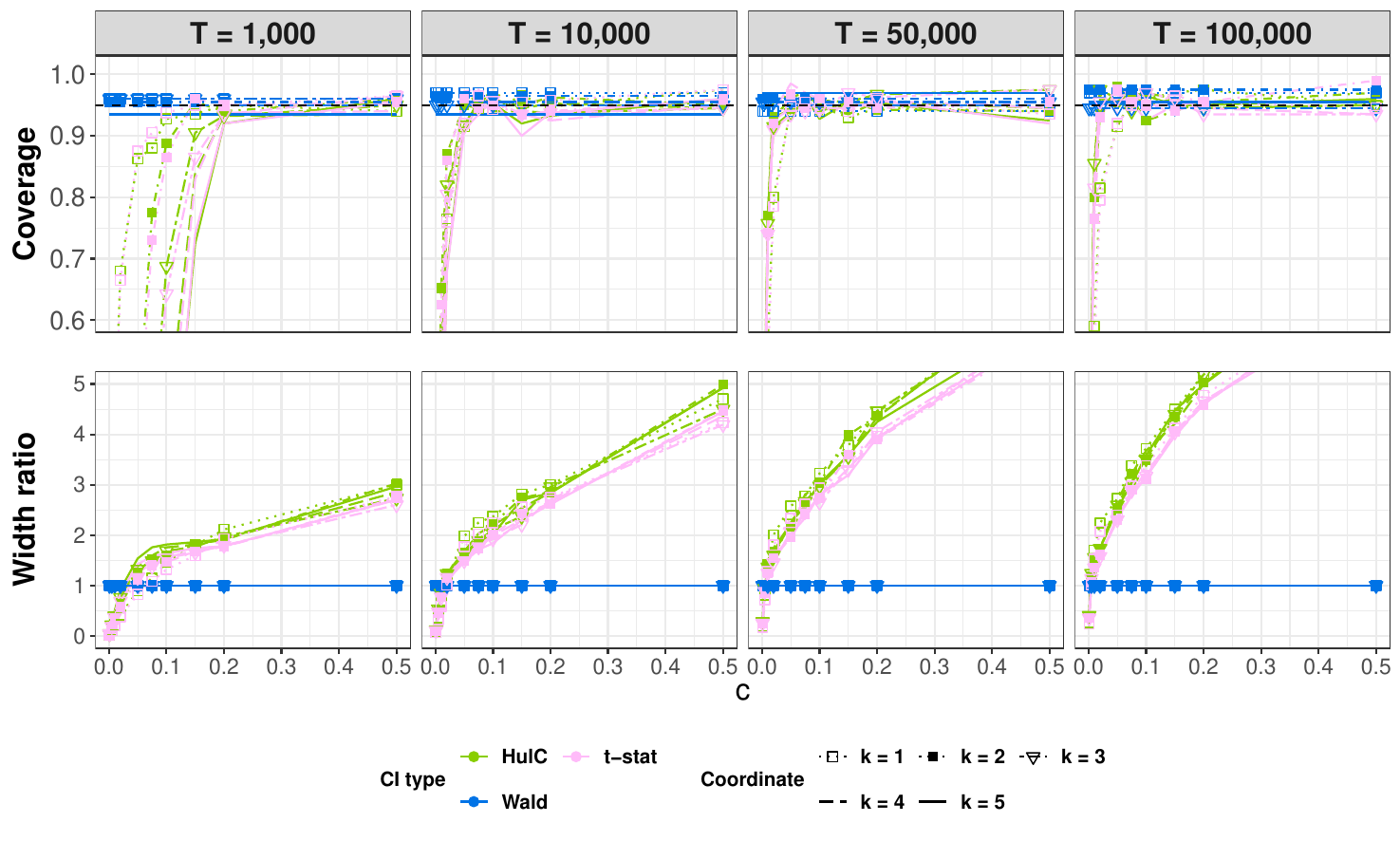}
\caption{Linear regression (Truncated-SGD), Covariance = I, d = 5}
\label{fig:linear_D5_I_cov_wr_truncatedSGD_initTRUE}
\end{figure}

\begin{figure}[H]
\centering
 % \par\medskip
\includegraphics[width=1\textwidth]{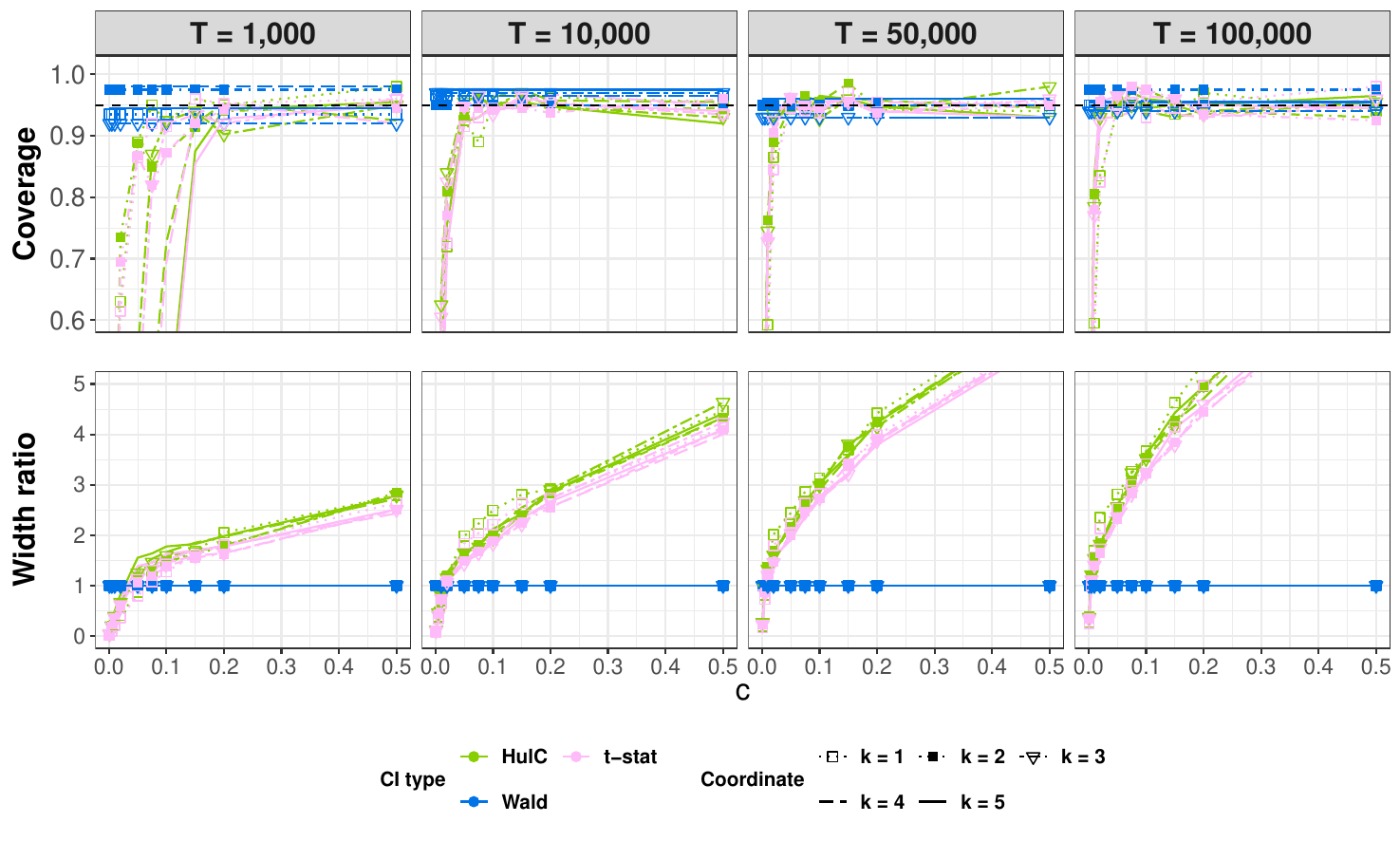}
\caption{Linear regression (Truncated-SGD), Covariance = Equicorrelation, d = 5}
\label{fig:linear_D5_EquiCorr_cov_wr_truncatedSGD_initTRUE}
\end{figure}

For linear regression using truncated-SGD, $d=5$, and Toeplitz covariance, see Figure~\ref{fig:linear_D5_Toeplitz_cov_wr_truncatedSGD_initTRUE}.

\begin{figure}[H]
\centering
 % \par\medskip
\includegraphics[width=1\textwidth]{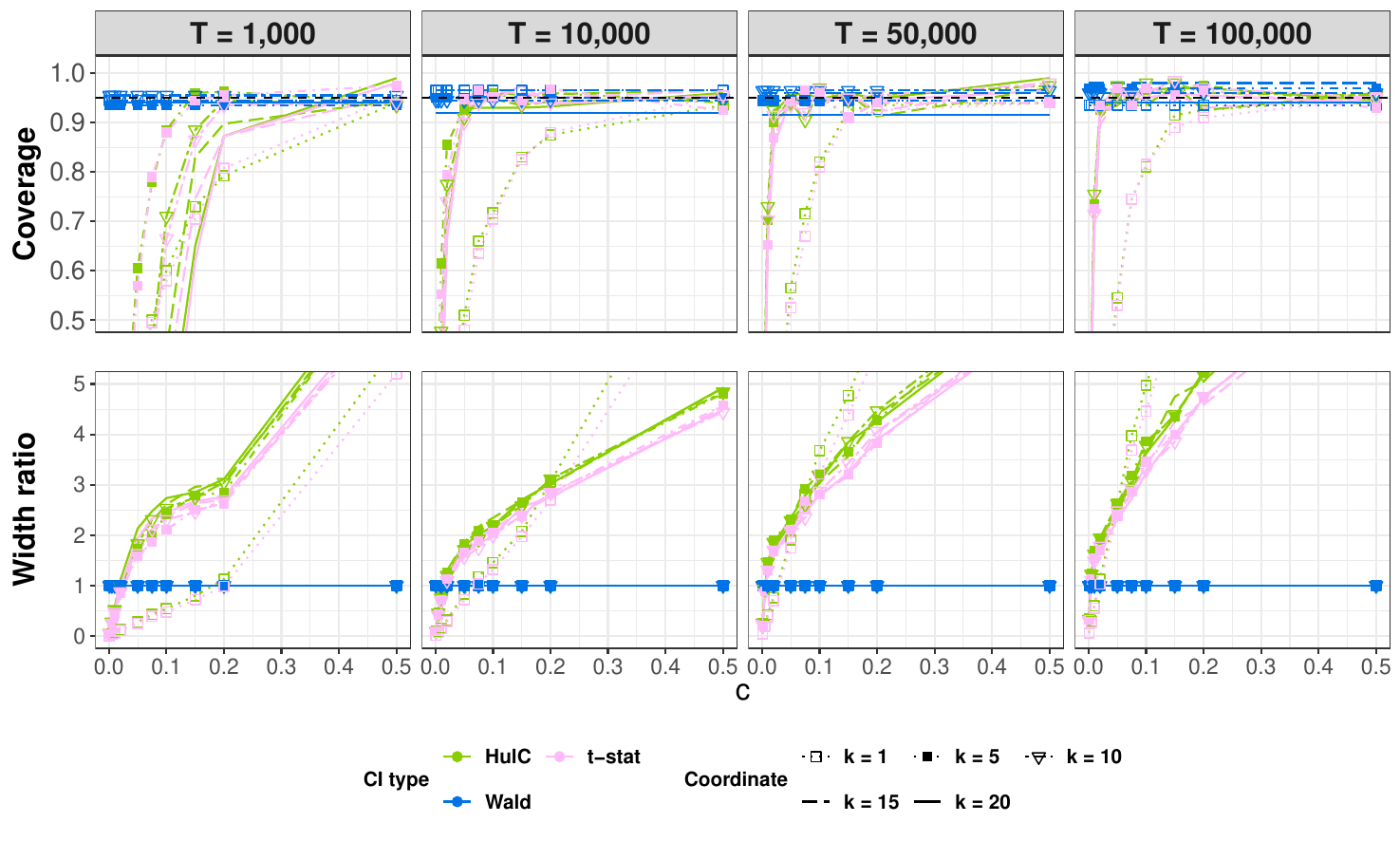}
\caption{Linear regression (Truncated-SGD), Covariance = I, d = 20}
\label{fig:linear_D20_I_cov_wr_truncatedSGD_initTRUE}
\end{figure}

\begin{figure}[H]
\centering
 % \par\medskip
\includegraphics[width=1\textwidth]{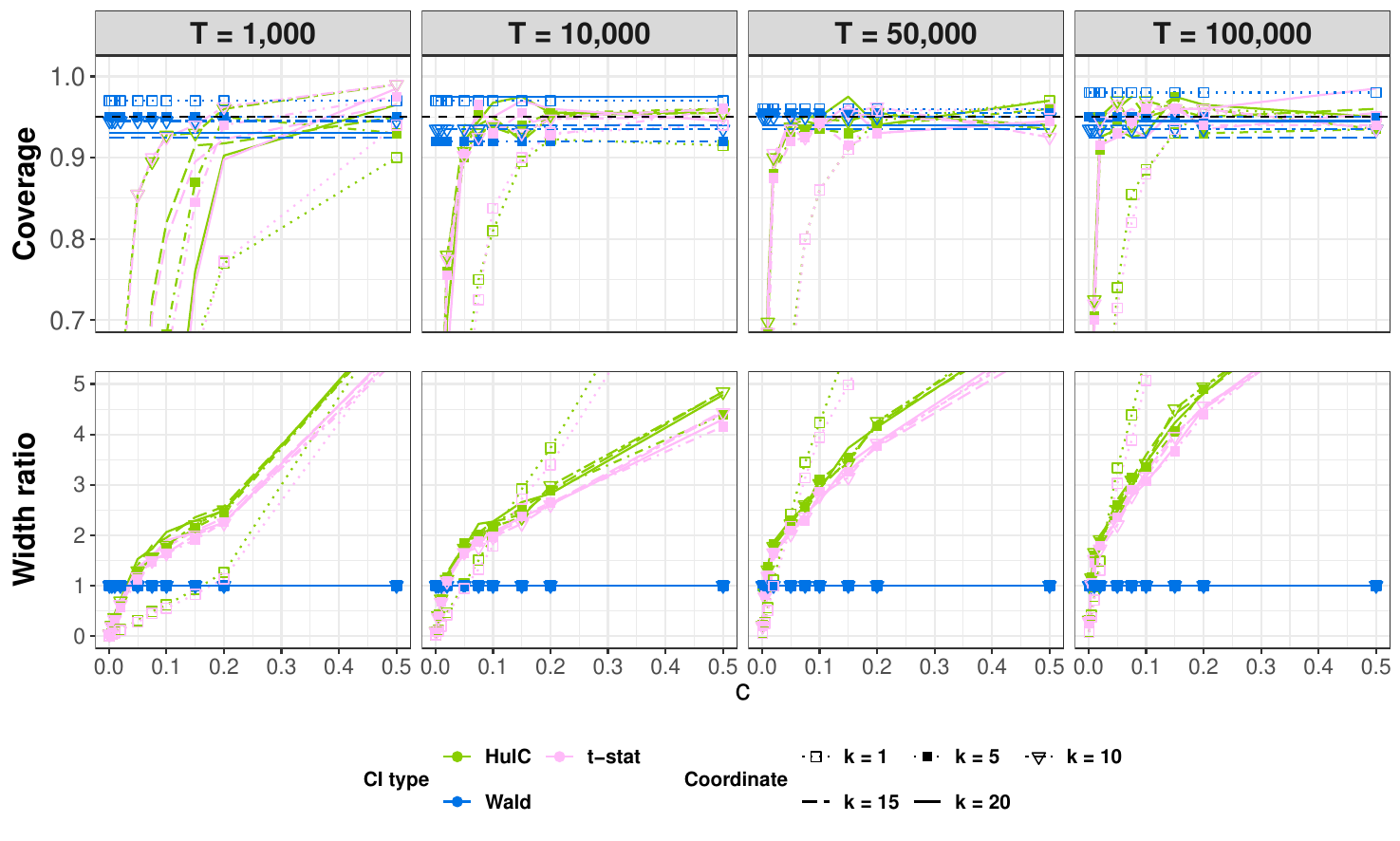}
\caption{Linear regression (Truncated-SGD), Covariance = Equicorrelation, d = 20}
\label{fig:linear_D20_EquiCorr_cov_wr_truncatedSGD_initTRUE}
\end{figure}

\begin{figure}[H]
\centering
 % \par\medskip
\includegraphics[width=1\textwidth]{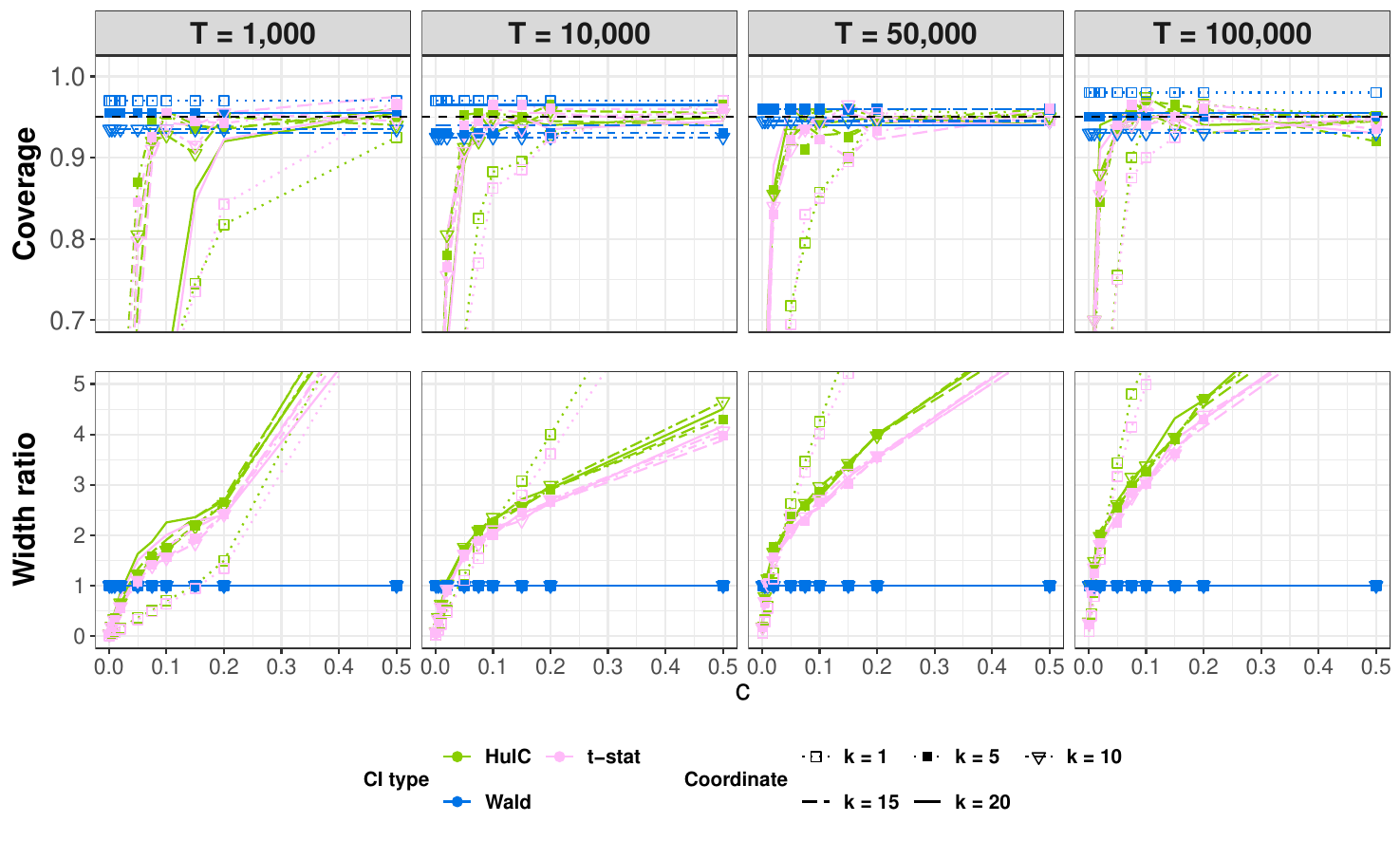}
\caption{Linear regression (Truncated-SGD), Covariance = Toeplitz, d = 20}
\label{fig:linear_D20_Toeplitz_cov_wr_truncatedSGD_initTRUE}
\end{figure}

\begin{figure}[H]
\centering
 % \par\medskip
\includegraphics[width=1\textwidth]{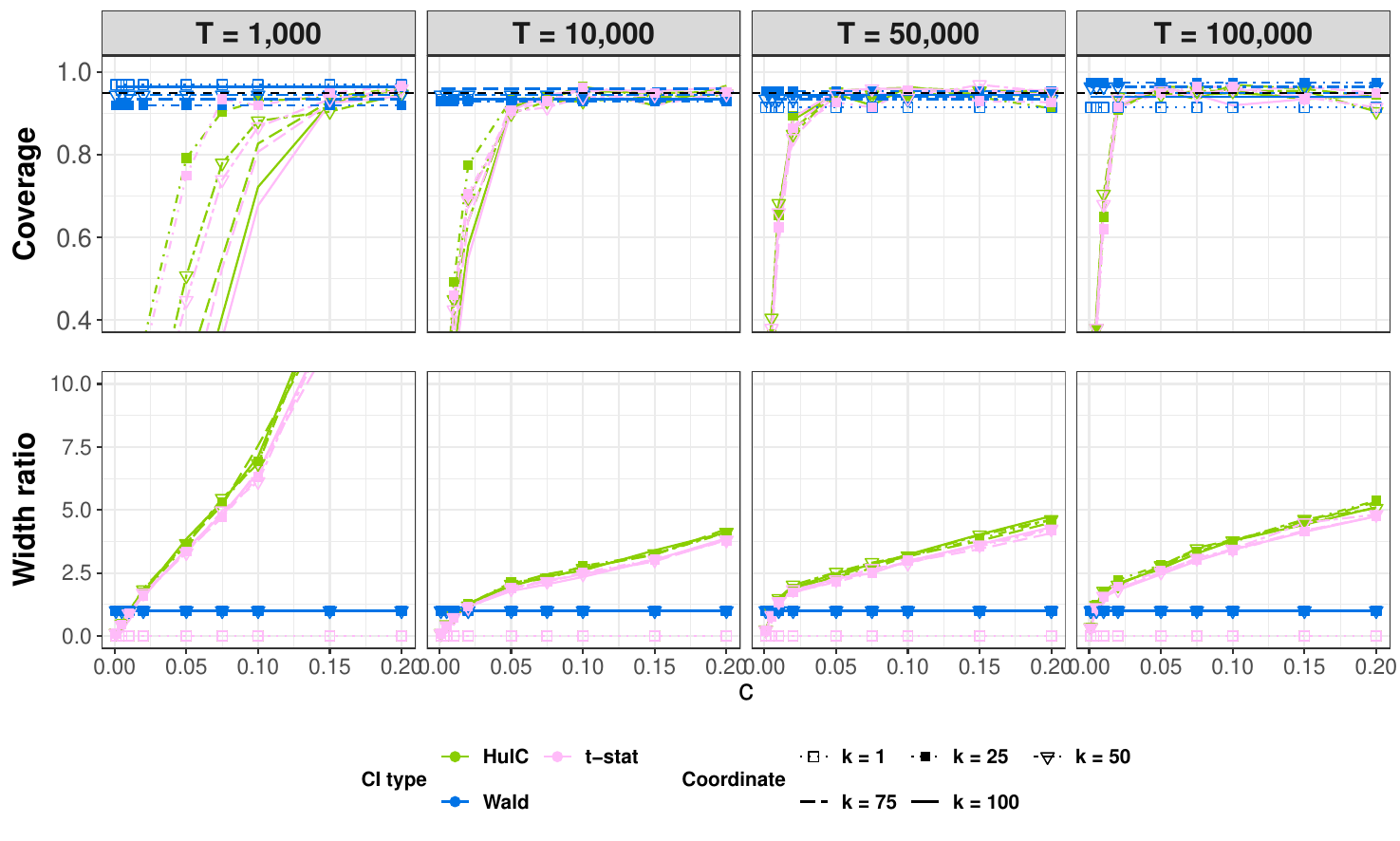}
\caption{Linear regression (Truncated-SGD), Covariance = I, d = 100}
\label{fig:linear_D100_I_cov_wr_truncatedSGD_initTRUE}
\end{figure}

\begin{figure}[H]
\centering
 % \par\medskip
\includegraphics[width=1\textwidth]{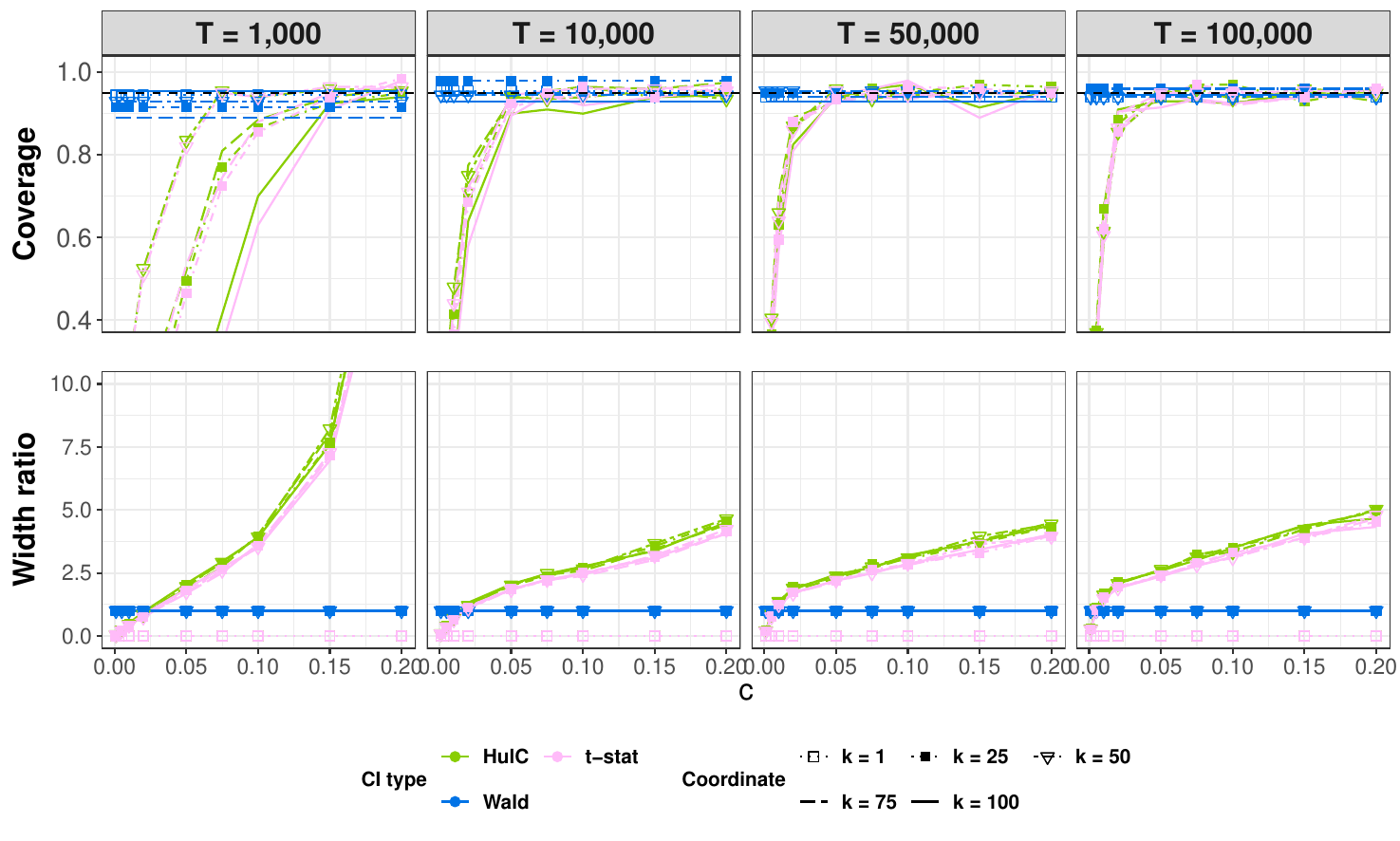}
\caption{Linear regression (Truncated-SGD), Covariance = Equicorrelation, d = 100}
\label{fig:linear_D100_EquiCorr_cov_wr_truncatedSGD_initTRUE}
\end{figure}

\begin{figure}[H]
\centering
 % \par\medskip
\includegraphics[width=1\textwidth]{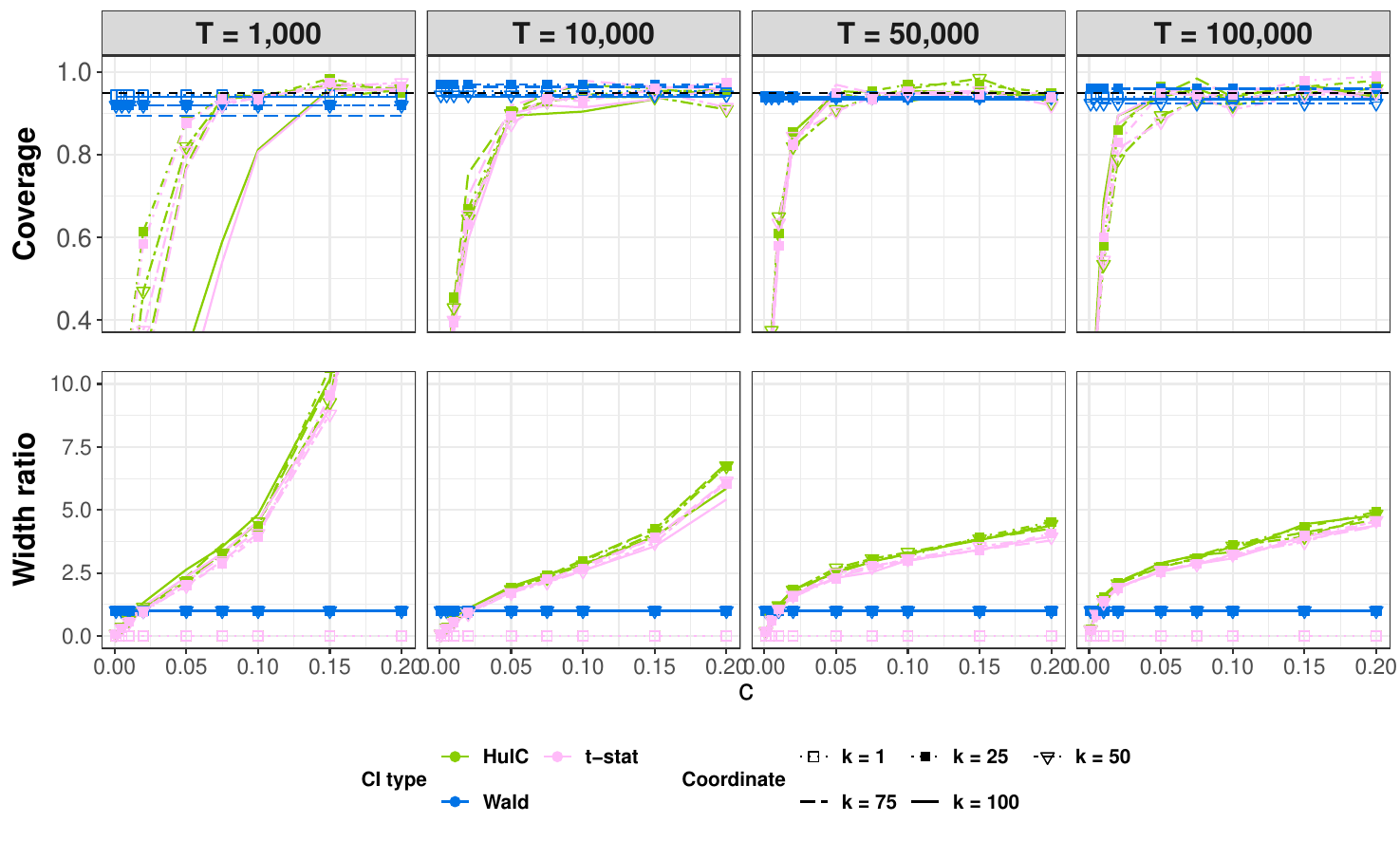}
\caption{Linear regression (Truncated-SGD), Covariance = Toeplitz, d = 100}
\label{fig:linear_D100_Toeplitz_cov_wr_truncatedSGD_initTRUE}
\end{figure}

\subsubsection{Logistic regression}\label{app:Logistic_plots_truncSGD}
\vspace{-20 pt}

\begin{figure}[H]
\centering
 % \par\medskip
\includegraphics[width=1\textwidth]{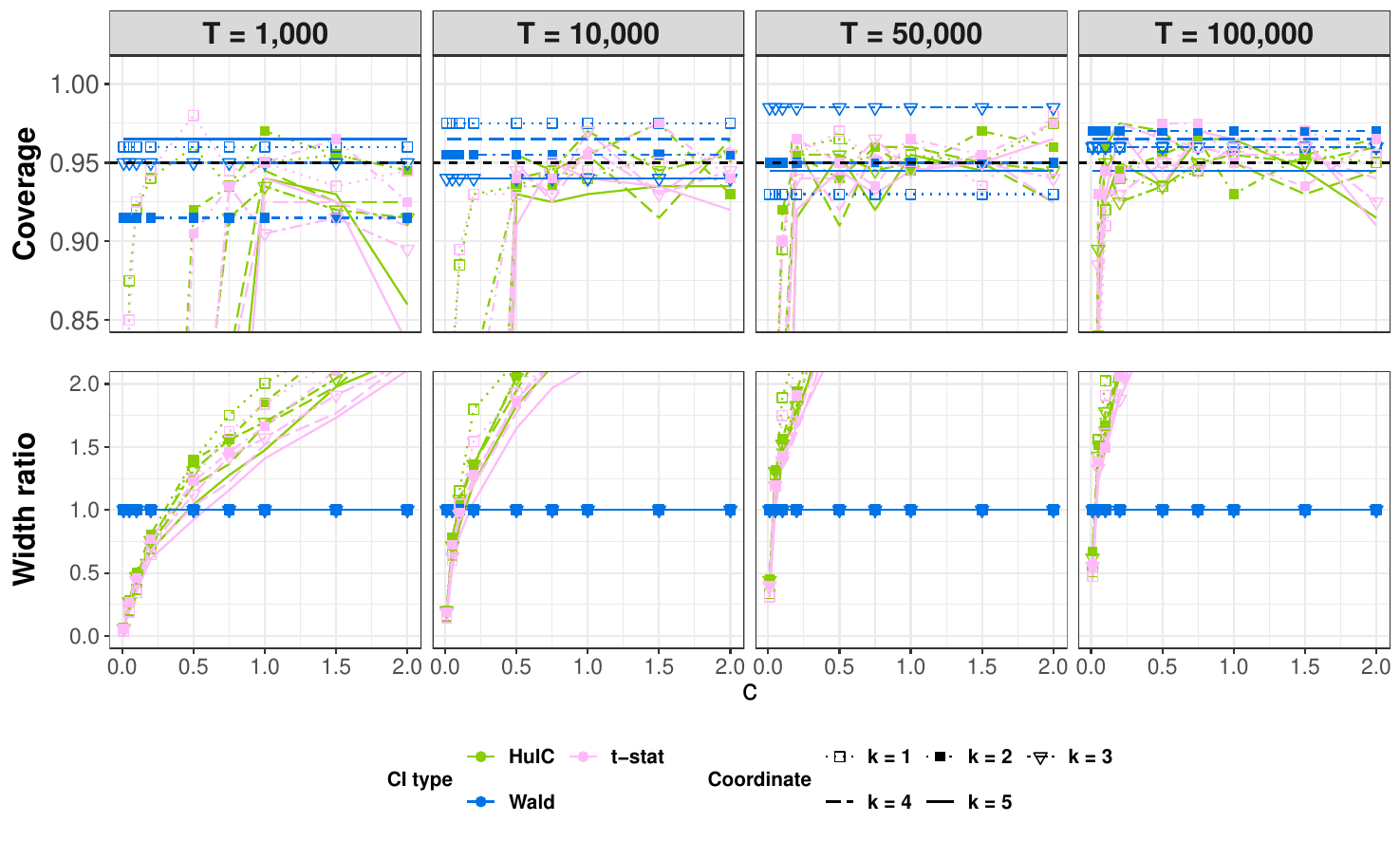}
\caption{Logistic regression (Truncated-SGD), Covariance = I, d = 5}
\label{fig:logistic_D5_I_cov_wr_truncatedSGD_initTRUE}
\end{figure}

\begin{figure}[H]
\centering
 % \par\medskip
\includegraphics[width=1\textwidth]{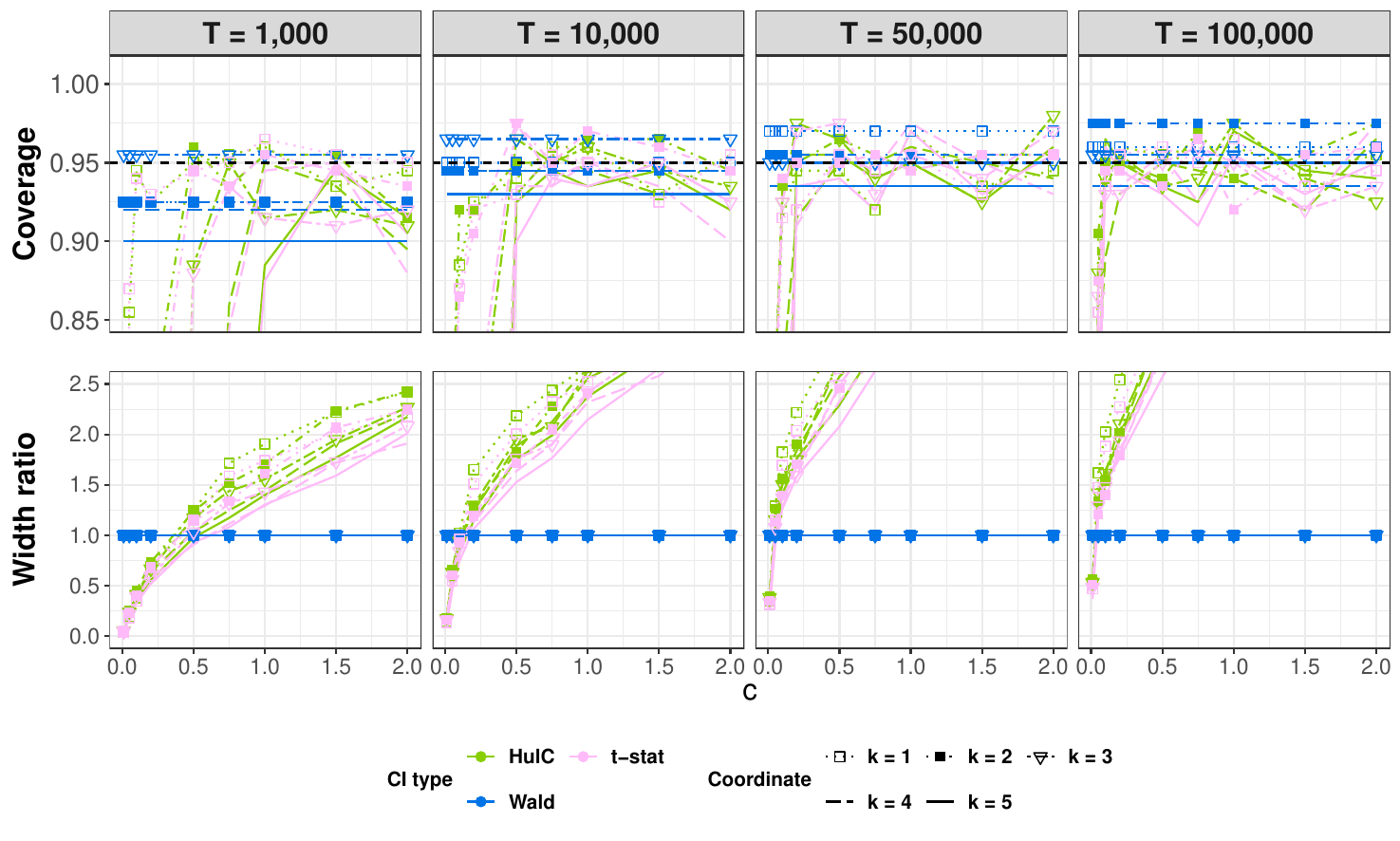}
\caption{Logistic regression (Truncated-SGD), Covariance = Equicorrelation, d = 5}
\label{fig:logistic_D5_EquiCorr_cov_wr_truncatedSGD_initTRUE}
\end{figure}

For logistic regression using truncated-SGD, $d=5$, and Toeplitz covariance, see Figure~\ref{fig:logistic_D5_Toeplitz_cov_wr_truncatedSGD_initTRUE}.

\begin{figure}[H]
\centering
 % \par\medskip
\includegraphics[width=1\textwidth]{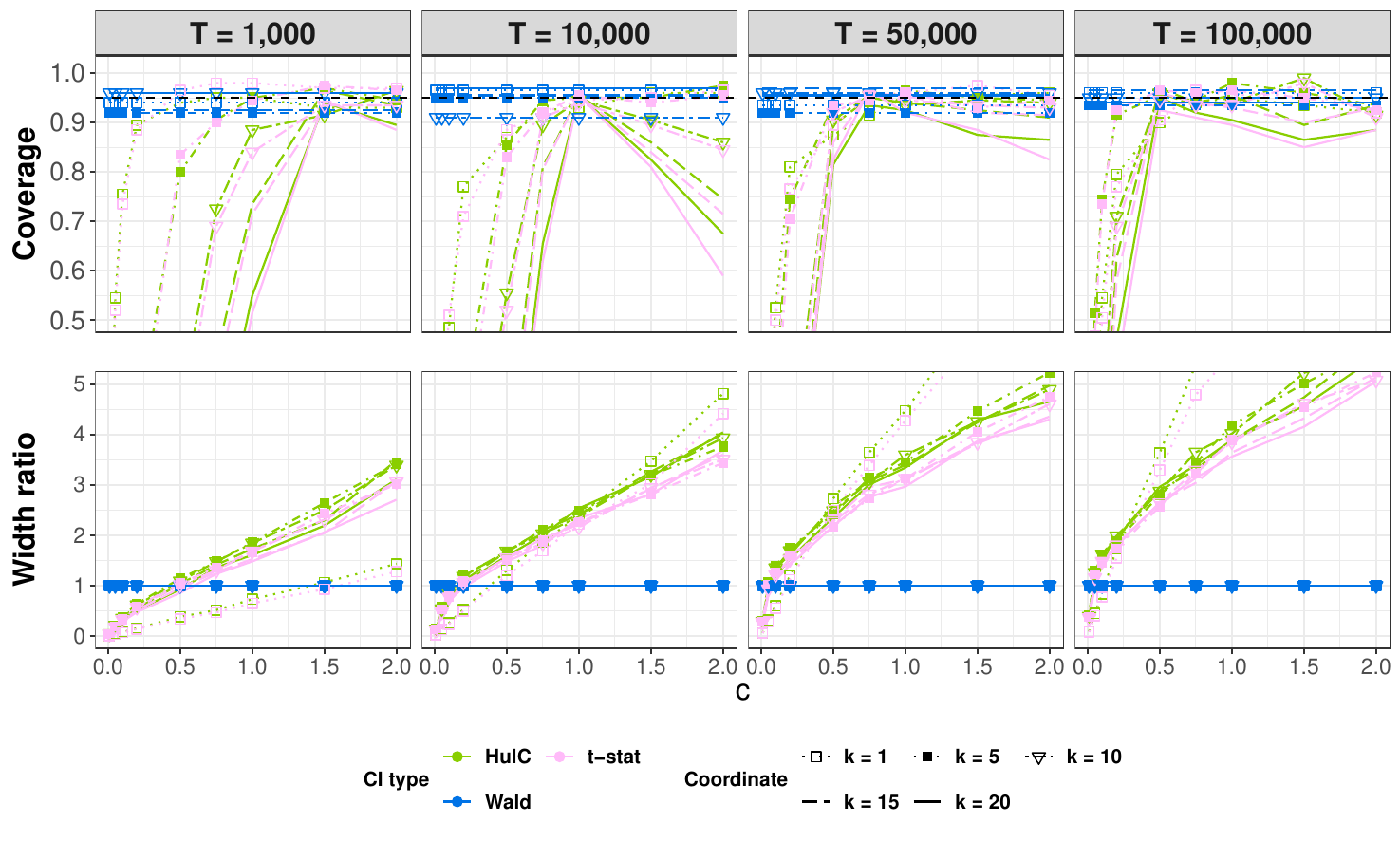}
\caption{Logistic regression (Truncated-SGD), Covariance = I, d = 20}
\label{fig:logistic_D20_I_cov_wr_truncatedSGD_initTRUE}
\end{figure}

\begin{figure}[H]
\centering
 % \par\medskip
\includegraphics[width=1\textwidth]{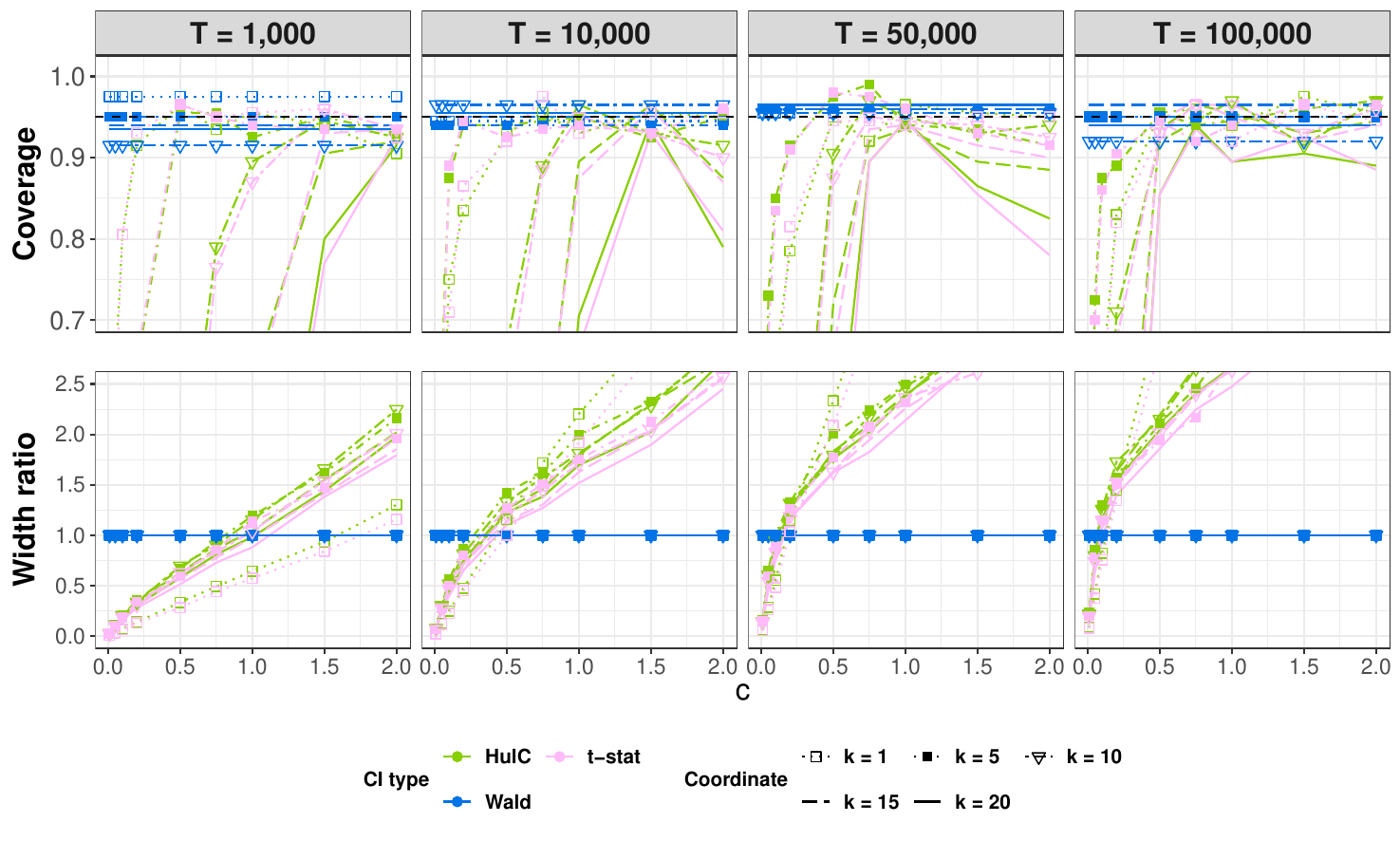}
\caption{Logistic regression (Truncated-SGD), Covariance = Equicorrelation, d = 20}
\label{fig:logistic_D20_EquiCorr_cov_wr_truncatedSGD_initTRUE}
\end{figure}

\begin{figure}[H]
\centering
 % \par\medskip
\includegraphics[width=1\textwidth]{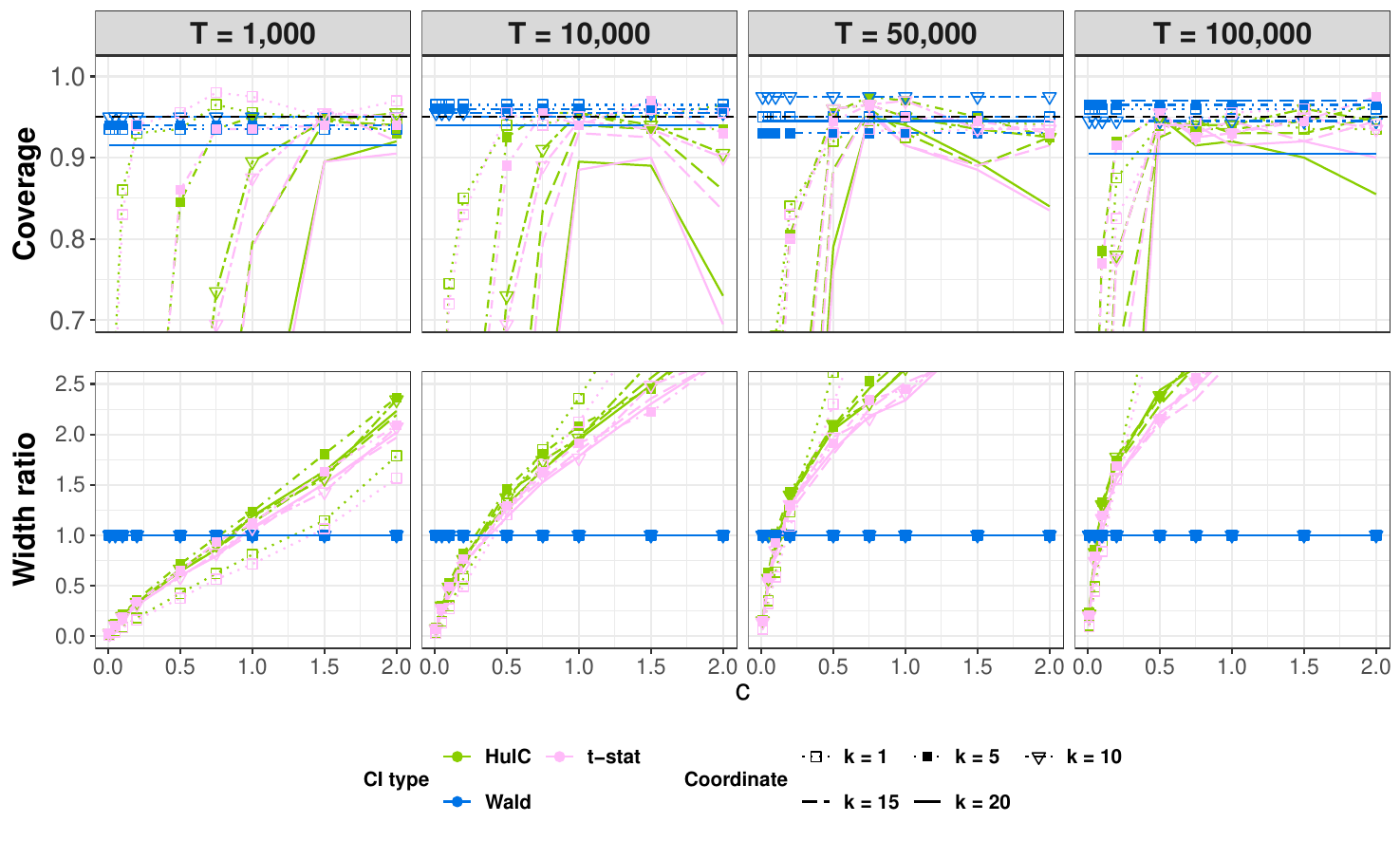}
\caption{Logistic regression (Truncated-SGD), Covariance = Toeplitz, d = 20}
\label{fig:logistic_D20_Toeplitz_cov_wr_truncatedSGD_initTRUE}
\end{figure}

\begin{figure}[H]
\centering
 % \par\medskip
\includegraphics[width=1\textwidth]{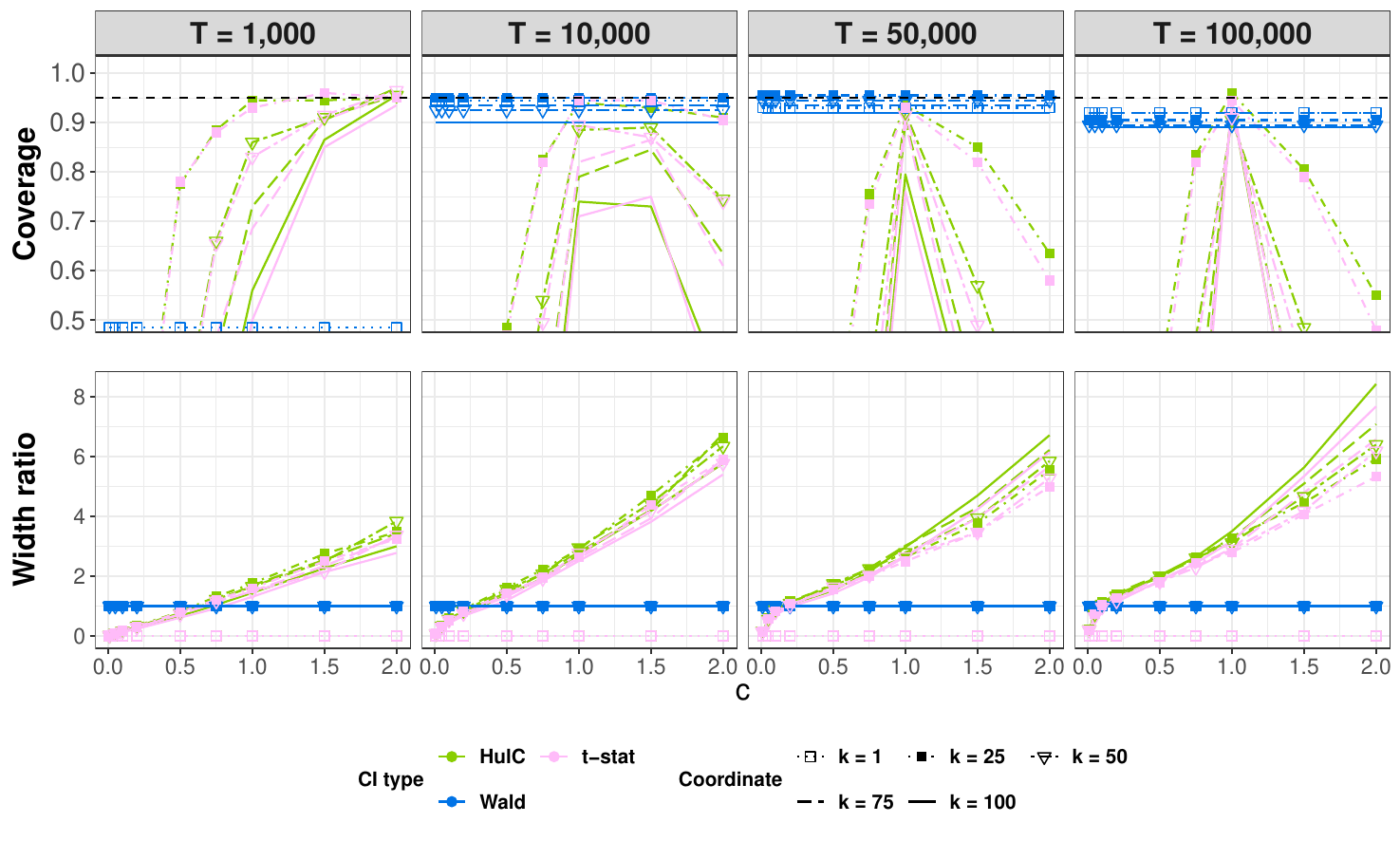}
\caption{Logistic regression (Truncated-SGD), Covariance = I, d = 100}
\label{fig:logistic_D100_I_cov_wr_truncatedSGD_initTRUE}
\end{figure}

Note: Wald estimates were not able to be computed for equicorrelation and Toeplitz covariances in the case of logistic regression using truncated-SGD, $d=100$. To see raw widths, see the online tool.\footnote{\url{https://public.tableau.com/app/profile/selina.carter6629/viz/OnlineinferencesimulationsOLSandlogisticregression/Coverageandwidthratio_paper}}

% \begin{figure}[H]
% \centering
%  % \par\medskip
% \includegraphics[width=1\textwidth]{figures/extra_sims/logistic_D100_EquiCorr_cov_wr_truncatedSGD_initTRUE}
% \caption{Logistic regression (Truncated-SGD), Covariance = Equicorrelation, d = 100}
% \label{fig:logistic_D100_EquiCorr_cov_wr_truncatedSGD_initTRUE}
% \end{figure}

%TODO: update this plot
% \begin{figure}[H]
% \centering
%  % \par\medskip
% \includegraphics[width=1\textwidth]{figures/extra_sims/logistic_D100_Toeplitz_cov_wr_truncatedSGD_initTRUE}
% \caption{Logistic regression (Truncated-SGD), Covariance = Toeplitz, d = 100}
% \label{fig:logistic_D100_Toeplitz_cov_wr_truncatedSGD_initTRUE}
% \end{figure}

\end{document}